# Semantic Similarity
# from Natural Language and Ontology analysis
## *preprint*

Sébastien Harispe, Sylvie Ranwez, Stefan Janaqi, Jacky Montmain
Ecole des mines d'Alès - LGI2P

# Abstract

Artificial Intelligence federates numerous scientific fields in the aim of developing machines able to assist human operators performing complex treatments – most of which demand high cognitive skills (e.g. learning or decision processes). Central to this quest is to give machines the ability to estimate the *likeness* or *similarity* between *things* in the way human beings estimate the similarity between stimuli.

In this context, this book focuses on semantic measures: approaches designed for comparing semantic entities such as units of language, e.g. words, sentences, or concepts and instances defined into knowledge bases. The aim of these measures is to assess the similarity or relatedness of such semantic entities by taking into account their semantics, i.e. their meaning – intuitively, the words *tea* and *coffee*, which both refer to stimulating beverage, will be estimated to be more semantically similar than the words *toffee* (confection) and *coffee*, despite that the last pair has a higher syntactic similarity. The two state-of-the-art approaches for estimating and quantifying semantic similarities/relatedness of semantic entities are presented in detail: the first one relies on corpora analysis and is based on Natural Language Processing techniques and semantic models while the second is based on more or less formal, computer-readable and workable forms of knowledge such as semantic networks, thesaurus or ontologies.

Semantic measures are widely used today to compare units of language, concepts, instances, or even resources indexed by them (e.g., documents, genes). They are central elements of a large variety of Natural Language Processing applications and knowledge-based treatments, and have therefore naturally been subject to intensive and interdisciplinary research efforts during last decades. Beyond a simple inventory and categorization of existing measures, the aim of this monograph is to convey novices as well as researchers of these domains towards a better understanding of semantic similarity estimation and more generally semantic measures. To this end, we propose an in-depth characterisation of existing proposals by discussing their features, the assumptions on which they are based and empirical results regarding their performance in particular applications. By answering these questions and by providing a detailed discussion on the foundations of semantic measures, our aim is to give the reader key knowledge required to: (i) select the more relevant methods according to a particular usage context, (ii) understand the challenges offered to this field of study, (iii) distinguish room of improvements for state-of-the-art approaches and (iv) stimulate creativity towards the development of new approaches. In this aim, several definitions, theoretical and practical details, as well as concrete applications are presented.

**keywords**: Semantic similarity, semantic relatedness, semantic measures, distributional measures, domain ontology, knowledge-based semantic measure.

# Contents

# Preface

In the last decades, numerous researchers from different domains have provided a lot of efforts developing and studying the notion of semantic measure and more specifically the notions of semantic similarity and semantic relatedness. Indeed, from the biomedical domain, where ontologies and conceptual annotations abound – e.g., genes are characterised by concepts from the Gene Ontology, scientific articles are indexed by terms defined into the Medical Subject Heading thesaurus (MeSH) – to Natural Language Processing (NLP) where text mining requires the semantics of units of language to be compared, researchers provided a vast body of research related to semantic measures: algorithms and approaches designed in the aim of comparing concepts, instances characterised by concepts and units of language w.r.t their meaning. Despite the vast literature dedicated to the domain, most of which is related to the definition of new measures, no extensive introduction proposes to highlight the large diversity of contributions which have been proposed so far. In this context, understanding the foundations of these measures, knowing the numerous approaches which have been proposed and distinguishing those to use in particular application contexts is challenging.

This book proposes an extended introduction to semantic measures targeting both students and domain experts. The aim is to provide a general introduction to the diversity of semantic measures in order to distinguish the central notions and the key concepts of the domain. In a second step, we present the two main families of measures to further discuss technical details related to specific implementations. By organising information about measures and by providing references to key research papers, our aim is to improve semantic measure understanding, to facilitate their use and to provide an condensed overview of state-of-the-art contributions related to the domain.

The first chapter introduces the motivations which highlight the importance of studying semantic measures. Starting by presenting various applications that benefit from semantic measures in different usage contexts, it then guides the reader towards a deeper understanding of those measures. Intuitive notions and the vocabulary commonly used in the literature are introduced. We present in particular the central notions of semantic relatedness, semantic similarity, semantic distance... More formal definitions and properties used for studying semantic measures are also proposed. Next, these definitions and properties are used to characterise the broad diversity of measures which have been introduced in the literature. A classification of semantic measures is then proposed; it distinguishes the two main approaches corresponding to corpus-based and knowledge-based semantic measures. These two families of semantic measures are further presented in detail in Chapter 2 and Chapter 3 respectively. The foundations of these measures and several implementations which have been proposed in the literature are discussed – software tools enabling practical use of measures are also presented in appendix. Chapter 4 is dedicated to semantic measures evaluation and selection. It presents several aspects of measures that can be studied for their comparison, as well as state-of-the-art protocols and datasets used for their evaluation. Finally, Chapter 5 concludes by summarising important notions which are introduced in this book, and by highlighting several important research directions which must be studied for improving both semantic measures and their understanding.

By following this progression, we hope that the reader will find a detailed and stimulating introduction to semantic measures. Our aim is to give the reader access to an extensive state-of-the-art of this field, as well as key knowledge required to: (i) select the more relevant methods according to a particular usage context, (ii) understand the challenges offered to this field of study, (iii) distinguish room of improvements for state-of-the-art approaches and (iv) stimulate creativity towards the development of new approaches.

*Sébastien Harispe, Sylvie Ranwez, Stefan Janaqi, and Jacky Montmain*
*Nîmes – France, January 2015*

# Chapter 1

# Introduction to semantic measures

Back in the 60s, the quest for artificial intelligence (AI) had originally been motivated by the assumption that "[...] *every aspect of learning or any other feature of intelligence can in principle be so precisely described that a machine can be made to simulate it* [...]" (McCarthy et al., 2006). Even if this assumption has today proved to be pretentious and perhaps even unattainable, efforts are still made to design intelligent agents which are able to resolve complex problems and to perform elaborated tasks. To this end, AI federates numerous scientific communities to tackle a large diversity of problems in the aim of giving machines the ability to reason, to understand knowledge, to learn, to plan, to manoeuvre, to communicate, and to perceive (Russell and Norvig, 2009). Central to this quest is the ability to estimate some *likeness* between *things* in the way human is able to compare stimuli, e.g. to compare objects, situations. This is a key notion to induce reasoning and therefore to provide machines with intelligence, i.e., the "*ability to acquire and apply knowledge and skills*" (Oxford Dict., 2012). In this context, numerous contributions have focused on designing and studying *semantic measures* for comparing semantic entities such as units of language, e.g. words, sentences, or concepts and instances defined into knowledge bases. The aim of these measures is to assess the similarity or relatedness of such semantic entities by taking into account their semantics, otherwise stated their meaning. These measures are cornerstones of refined processing of texts and ontologies and therefore play an important role in numerous domains.

Semantic measures are widely used today to compare semantic entities such as units of language, concepts or even semantically characterised instances, according to information supporting their meaning. They are based on the analysis of *semantic proxies*, corpora of texts or ontologies[1], from which *semantic evidence* can be extracted. Notice that here, according to the literature related to the field, the notion of semantic measure is not framed in the rigorous mathematical definition of *measure*. It should instead be understood as any theoretical tool, mathematical function, algorithm or approach which enables the comparison of semantic entities according to semantic evidence. Generally speaking, these measures are used to estimate the degree of semantic likeness between semantic entities through a numerical value. Therefore, even if a large diversity of measures exists to estimate the similarity or the distance between specific mathematical objects (e.g., vectors, matrices, graphs, sets, fuzzy sets), data structures (e.g., lists, objects) and data types (e.g., numbers, strings, dates), the main particularity of semantic measures compared to traditional similarity or distance functions relies on two aspects: (i) they are dedicated to the comparison of semantic entities and (ii) they are based on the analysis of semantic proxies from which semantic evidence can be extracted. This semantic evidence is expected to directly or indirectly characterise the meaning of compared elements. As an example, the measures used to compare two words according to their sequences of characters cannot be considered as semantic since only the characters of the words and their ordering is taken into account, not their meaning. Indeed, according to such syntactic measures, the two words *car* and *vehicle* would be regarded as distant despite their closely related semantics. Semantic measures enable to overcome the limitation of such syntactic measures by comparing semantic entities w.r.t their semantics. To this aim, semantic measures rely on the analysis of two broad types of semantic proxies: corpora of texts and ontologies. These semantic proxies are used to extract semantic evidence which will next be used by semantic measures to support the comparison of compared units of language, concepts or instances.

The first type of semantic proxy corresponds to unstructured or semi-structured texts (e.g., plain texts, dictionaries). These texts contain informal evidence of semantic relationships between units of language. Let us consider a simple example. Since it is common to drink *coffee* with *sugar* and nothing particular links *coffee* to *cats*, most will agree that the pair of words (*coffee*,*sugar*) is more semantically *coherent* than the pair of words (*coffee*,*cat*) – otherwise stated that the words (*coffee*,*sugar*) are more semantically related than the words (*coffee*,*cat*). Interestingly, corpus of texts can be used to derive the same conclusion. To this end, a semantic measure will take advantage of the fact that the word *coffee* is more likely to co-occur with the word *sugar* than with the word *cat*. Simply stated, it is possible to use observations regarding the distribution of words in a corpus in order to estimate the strength of the semantic relationship which links two words, e.g., based on the assumption that semantically related words tend to co-occur.

---

[1] Or more generally knowledge bases.

The second type of semantic proxy from which semantic evidence can be extracted encompasses a large range of knowledge models. From structured vocabularies to highly formal ontologies, these proxies are structured and model, in an explicit manner, knowledge about the entities they define. As an example, in an ontology defining the concepts `Coffee` and `Sugar`, a specific relationship will probably explicitly define the link between the two concepts, e.g., that `Coffee canBeDrinkWith Sugar`. Semantic measures based on knowledge base analysis rely on techniques which take advantage of network-based (e.g., thesaurus, taxonomies), or logic-based ontologies to extract semantic evidence on which the comparison will be based.

From gene analysis to recommendation systems, semantic measures have recently been found to cover a broad field of applications and are now essential metrics for leverage data mining, data analysis, classification, knowledge extraction, textual processing or even information retrieval based on text corpora or ontologies. They play an essential role in numerous treatments which require the analysis of the meaning (i.e., semantics) of compared entities. In this context, the study of semantic measures has always been an interdisciplinary effort. Communities of Psychology, Cognitive Sciences, Linguistics, Natural Language Processing, Semantic Web, and Biomedical informatics being among the most active contributors (2014). Due to the interdisciplinary nature of semantic measures, recent decades have been highly prolific in contributions related to the notion of semantic relatedness, semantic similarity and semantic distance, etc. This book provides an organised state of the art of semantic measures and proposes a classification of existing measures. Yet before introducing the technical aspects required to further introduce semantic measures, we will briefly discuss their large diversity of applications.

## 1.1 Semantic measures in action

Semantic measures are used to solve problems in a broad range of applications and domains. They are essential tools for the design of numerous algorithms and treatments in which semantics matters. In this section, we present diverse practical applications involving semantic measures. Three domains of application are considered in particular: (i) Natural Language Processing, (ii) Knowledge Engineering/Semantic Web and Linked Data, and (iii) Biomedical informatics and Bioinformatics. Since they are transversal, additional applications related to information retrieval and clustering are also briefly considered.

The list of applications presented in this section is far from being exhaustive. An extensive classification of contributions related to semantic measures is proposed in the state-of-the-art presented in this manuscript (Harispe et al., 2013b, version 2). This classification underlines the broad range of applications of semantic measures and highlights the large number of communities involved.

### 1.1.1 Natural Language Processing

Linguists have, quite naturally, been among the first to study semantic measures in the aim of comparing units of language (e.g., words, sentences, paragraphs, documents). The estimation of word/concept relatedness plays an important role in detecting paraphrasing, e.g., duplicate content and plagiarism (Fernando and Stevenson, 2008), in generating thesauri or texts (Iordanskaja et al., 1991), in summarising texts (Kozima, 1993), in identifying discourse structure, and in designing question answering systems (Bulskov et al., 2002; Freitas et al., 2011; Wang et al., 2012a) to mention a few. The effectiveness of semantic measures to resolve both syntactic and semantic ambiguities has also been demonstrated on several occasions, e.g., (Sussna, 1993; Resnik, 1999; Patwardhan, 2003).

Several surveys related to the usage of semantic measures and to the techniques used for their design for natural language processing can be found in (Weeds, 2003; Curran, 2004; Sahlgren, 2008; Dinu, 2011; Mohammad and Hirst, 2012a; Panchenko, 2013).

### 1.1.2 Knowledge engineering, Semantic Web and Linked Data

Interest in semantic measures is still growing while several initiatives promote the Semantic Web and Linked Data paradigms to provide *"an extension of the current [Web], in which information is given well-defined meaning, better enabling computers and people to work in cooperation"* (Berners-Lee et al., 2001).

Communities associated to Knowledge Engineering, Semantic Web and Linked Data play an import role in the definition of methodologies and standards to formally express machine-understandable knowledge representations. They extensively study the problematic associated to the expression of structured and controlled vocabularies, as well as ontologies, i.e., formal and explicit specification of a shared conceptualisation defining a set of concepts, their relationships and axioms to model a domain[2] (Gruber, 1993). These models rely on structured knowledge representations in which the semantics of the concepts (classes) and relationships (properties) are rigorously and formally defined in an unambiguous way. These (on-going) efforts have led to the definition of several languages which can be used today to express formal, computer-readable and processable forms of knowledge. Such models are therefore proxies of choice to compare the concepts and the instances of the domain they model. As we will see, the taxonomy of concepts, which is the backbone of ontologies, is particularly useful to estimate the degree of similarity between concepts.

In this field, semantic measures can be used as part of processes aiming to integrate heterogeneous ontologies (refer to ontology alignment and instance matching) (Euzenat and Shvaiko, 2013); they are used to find similar/duplicate entities defined in different ontologies. Applications to provide inexact search capabilities over ontologies or to improve classical information retrieval techniques have also been proposed, e.g., (Hliaoutakis, 2005; Varelas et al., 2005; Hliaoutakis et al., 2006; Kiefer et al., 2007; Sy et al., 2012; Pirró, 2012). In this context, semantic measures have also been successfully applied to learning tasks using Semantic Web technologies (D'Amato, 2007). Their benefits for designing recommendation systems based on the Linked Data paradigm have also been stressed, e.g., (Passant, 2010; Harispe et al., 2013a).

### 1.1.3 Biomedical Informatics & Bioinformatics

A large number of semantic measures have been defined in biomedical informatics and bioinformatics. In these domains, semantic measures are commonly used to study various types of instances which have been semantically characterised using ontologies (genes, proteins, drugs, diseases, phenotypes)[3]. Several surveys related to the usage of semantic measures in the biomedical domain underline the diversity of their applications, e.g. for diagnosis, disease classification, drug design and gene analysis (Pedersen et al., 2007; Pesquita et al., 2009a; Guzzi et al., 2012). As an illustration, here we focus on applications related to studies on the Gene Ontology (GO) (Ashburner et al., 2000).

The GO is the preferred example with which to highlight the large adoption of ontologies in biology; it is extensively used to conceptually annotate gene (products) on the basis of experimental observations or automatic inferences. A gene is classically annotated by a set of concepts structured in the GO. These annotations formally characterise genes regarding their molecular functions, their cellular location and the biological processes in which they are involved. Thanks to semantic measures, these annotations make the automatic comparison of genes possible, not on the basis of particular gene properties (e.g. nucleotidic/proteic sequence, structural similarity, gene expression), but rather on the analysis of biological aspects which are formalised by the GO. Genes can be further analysed by considering their representation in a semantic space expressing our current understanding of particular aspects of biology. In such cases, conceptual annotations bridge the gap between global knowledge of biology defined in the GO (e.g., organisation of molecular functions) and fine-grained understanding of specific instances (e.g., the specific role of a gene at molecular level). In this context, semantic measures enable computers to take advantage of this knowledge to analyse genes and therefore open interesting perspectives to infer new knowledge.

As an example, various studies have highlighted the relevance of semantic measures to assess the functional similarity of genes (Wang et al., 2007; Du et al., 2009), to build gene clusters (Sheehan et al., 2008), to validate and to study protein-protein interactions (Xu et al., 2008), to analyse gene expression (Xu et al., 2009), to evaluate gene set coherence (Diaz-Diaz and Aguilar-Ruiz, 2011) or to recommend gene annotations (Couto et al., 2006), among others. A survey dedicated to semantic measures applied to the GO can be found in (Guzzi et al., 2012).

---

[2] More about ontologies can be found in (Gruber, 1995; Guarino et al., 2009; Fernandez-Lopez and Corcho, 2010).

[3] Biology and biomedicine are heavy users of ontologies, e.g., BioPortal, a portal dedicated to ontologies related to biology and the biomedical domain, references hundreds of ontologies (Whetzel et al., 2011).

### 1.1.4 Other applications

**Information Retrieval** (IR) uses semantic measures to overcome the limitations of techniques based on plain lexicographic term matching, i.e., simple IR models consider that a document is relevant according to a query only if the terms specified in the query occur in the document. Semantic measures enable the meaning of words to be taken into account by going over syntactic search. They can therefore be used to improve classic models, e.g., synonyms will no longer be considered as totally different words. As an example, semantic measures have been successfully used in the design of ontology-based information retrieval systems and for query expansion, e.g., (Hliaoutakis, 2005; Hliaoutakis et al., 2006; Varelas et al., 2005; Baziz et al., 2007; Saruladha et al., 2010b; Sy et al., 2012).

An important aspect is that semantic measures based on ontologies allow for the analysis and querying of non-textual resources and therefore do not restrict IR techniques in text analysis, e.g., genes annotated by concepts can be queried (Sy et al., 2012). In the same vein, the definition of efficient generic indexing solutions based on semantic measures are proposed in (Fiorini et al., 2014).

**GeoInformatics** actively contributes to the study of semantic measures. In this domain, measures have for instance been used to compute the similarity between locations according to semantic characterisations of their geographic features (Janowicz et al., 2011), e.g. estimating the semantic similarity of *tags* defined in the OpenStreetMap Semantic Network[4] (Ballatore et al., 2012). Readers interested in the application of semantic measures in this field may also refer to (Akoka et al., 2005; Andrea Rodríguez and Egenhofer, 2004; Rodríguez et al., 2005; Janowicz, 2006; Formica, 2008; Janowicz et al., 2008).

## 1.2 From similarity towards semantic measures

This section first introduces generalities related to the notion of similarity. We introduce in particular several models which have been proposed by cognitive sciences in order to study the human capacity to evaluate similarity of objects. Presenting these contributions will help us to better understand the models generally used to analyse object comparison. As we will see, they will be of specific interest for studying semantic measures. Indeed, the main models of similarity defined by cognitive sciences play an important role to understand the diversity of approaches which have been proposed to design semantic measures.

The next aim of this section is to introduce the vocabulary which is commonly used to refer to the notion of semantic measures, i.e. semantic similarity, semantic relatedness, semantic distance, etc. Several mathematical definitions and properties related to distance and similarity are also presented. These definitions will be further used to distinguish mathematical properties of interest for the characterization and the study of semantic measures in Section 1.3.

### 1.2.1 Human cognition, similarity and existing models

Human capacity to evaluate the similarity of *things* (e.g., objects, stimuli) has long been studied by cognitive sciences and psychology. It has been characterised as a central element of the human cognitive system and is therefore understood nowadays as a pivotal notion to simulate intelligence (Rissland, 2006). It is indeed a key element to initiating the process of learning in which the capacity to recognise similar situations helps us to build our experience[5], to activate mental traces, to make decisions, to innovate by applying experience gained in solving similar problems[6] (Holyoak and Koh, 1987; Ross, 1987; Novick, 1988; Ross, 1989; Vosniadou and Ortony, 1989; Gentner and Markman, 1997). The importance of the notion of similarity for cognitive processes, and in particular for the process of learning, has also been stressed by the theories of transfer which highlights that new skills are expected to be easier to learn if similar to skills already learned (Markman and Gentner, 1993). In this context, similarity is therefore commonly considered as a central component of memory retrieval, categorisation, pattern recognition, problem solving, reasoning, as well as social judgement, e.g., refer to Markman and Gentner (1993); Hahn et al. (2003); Goldstone and Son (2004) for associated references.

---

[4] http://wiki.openstreetmap.org/wiki/OSM_Semantic_Network
[5] Cognitive models based on categorisation consider that human classify things, e.g., experience of life, according to their similarity to some prototype, abstraction or previous examples (Markman and Gentner, 1993).
[6] Here the similarity is associated to the notion of generalisation and is measured in terms of probability of inter-stimulus-confusion errors (Nosofsky, 1992).

As we have seen, from mathematics to psychology, the notion of similarity is central in numerous fields and is particularly important for human cognition and *intelligent* system design. In this subsection, we provide a brief overview of the psychological theories of similarity by introducing the main models proposed by cognitive sciences to study and explain (human) appreciation of similarity.

Cognitive models of similarity generally aim to study the way humans evaluate the similarity of two mental representations according to some kind of psychological space (Tversky, 2004). They are therefore based on assumptions regarding the mental representation of the compared objects from which the similarity will be estimated. Indeed, as stated by several authors, the notion of similarity, *per se*, can be criticised as a purely artificial notion. In Goodman (1972), the notion of similarity is defined as "*an imposture, a quack*" because objectively, everything is equally similar to everything else. The authors emphasise that, conceptually, two random objects have an infinitive number of properties in common and infinite different properties[7], e.g. a flower and a computer are both smaller than 10m, 9.99m, 9.98m, etc. An important notion to understand, which has been underlined by cognitive sciences, is that different degrees of similarities emerge only when some predicates are selected or weighted more than others. As stated in Hahn (2011), "*this important observation doesn't mean that similarity is not an explanatory notion but rather that the notion of similarity is heavily framed in psychology*". Similarity assessment must therefore not be understood as an attempt to compare object realisations through the evaluation of their properties, but rather as a process aiming to compare objects as they are understood by the agent which estimates the similarity (e.g., a person, an algorithm). The notion of similarity therefore only makes sense according to the consideration of a partial (mental) representation on which the estimation of object similarity is based – this aspect of the notion of similarity will be essential for the rest of this book.

Contrary to real objects, representations of objects do not contain infinitesimal properties. As an example, our mental representations of things only capture a limited number of dimensions of the object which is represented. Therefore, the philosophical worries regarding the soundness of similarity vanish given that similarity aim at comparing partial representations of objects and not objects themselves, e.g., human mental representation of objects (Hahn, 2011). It is important to understand that studying human capacity to assess similarity, the similarity is thus estimated between mental representations – i.e. *representations*, not the real objects. This will also be the case for semantic similarity measures. Considering that these representations are the ones of a human agent, the notion of similarity may thus be understood as how similar objects appear to us. Considering the existential requirement of representations to compare things or objects, much of the history of research on similarity in cognitive sciences focuses on the definition of models of the mental representation of objects, to further consider measures which will be used to compare objects based on their representations.

The central role of cognitive sciences regarding the study of similarity relies on the design of cognitive models of both, mental representations and similarity. These models are used to study how humans store their knowledge and interact with it in order to compare object representations. Cognitive scientists then test these models according to our understanding of human appreciation of similarity. Indeed, evaluations of human appreciation of similarity help us to distinguish constraints/expectations on the properties an accurate model should have. This approach is essential to reject hypotheses and improve the models. As an example, studies have demonstrated that appreciation of similarity is sometimes asymmetric: the similarity between a person and his portrait is commonly expected to be lower than the inverse.[8] Therefore, the expectation of asymmetric estimation of similarity is incompatible with the mathematical properties of a distance, which is symmetric by definition. Models based on distance axioms thus appeared inadequate and have to be revised or to be used with moderation. In this context, the introduction of cognitive models of similarity will be particularly useful to understand the foundations of some approaches adopted for the definition of semantic measures.

Cognitive models of similarity are commonly organised into four different approaches: (i) Spatial models, (ii) Feature models, (iii) Structural models and (iv) Transformational models. We briefly intro-duce these four models – more detailed introductions have been proposed by Goldstone and Son (2004) and Schwering (2008). A captivating talk introducing cognition and similarity, on which this introduction is based, can also be found in (Hahn, 2011).

---

[7] This statement also stands if we restrict the comparison of objects to a finite set of properties. The reader may refer to Andersen's famous story of the Ugly Duckling. Proved by Watanabe and Donovan (1969), the *Ugly Duckling* theorem highlights the intrinsic bias associated to classification, showing that all things are equal and therefore that an ugly duckling is as similar to a swan as two swans are to each other. The important teaching is that biases are required to make a judgement and to classify, i.e., to prefer certain categories over others.

[8] Indeed, Tversky (1977) stresses that *We say "the portrait resembles the person" rather than "the person resembles the portrait"*.

**Spatial models**

The spatial models, also named geometric models, rely on one of the most influencal theories of similarity in cognitive sciences. They are based on the notion of psychological distance and consider objects (here perceptual effects of stimuli or concepts) as points in a multi-dimensional metric space.

Spatial models consider similarity as a function of the distance between the mental representations of the compared objects. These models derive from Shepard's spatial model of similarity. Objects are represented in a multi-dimensional space and their locations are defined by their dimensional differences (Shepard, 1962).

In his seminal work on generalisation, Shepard (1987) provides a statistical technique in the form of Multi-Dimensional Scaling (MDS) to derive locations of objects represented in a multi-dimensional space. MDS can be used to derive some potential spatial representations of objects from proximity data (similarity between pairs of objects). Based on these spatial representations of objects, Shepard derived the *universal law of generalisation* which demonstrates that various kinds of stimuli (e.g., Morse code signals, shapes, sounds) have the same lawful relationship between distance (in an underlined MDS) and perceive similarity measures (in terms of confusability) – the similarity between two stimuli was defined as an exponentially decreasing function of their distance[9].

By demonstrating a negative exponential relationship between similarity and generalisation, Shepard established the first sound model of mental representation on which cognitive sciences will base their studies on similarity. The similarity is in this case assumed to be the inverse of the distance separating the perceptual representations of the compared stimuli (Ashby and Perrin, 1988). Similarity defined as a function of distance is therefore implicitly constrained to the axiomatic properties of distance – these properties will be detailed in the following chapter, Section 1.2.3.

A large number of geometric models have been proposed. They have long been among the most popular in cognitive sciences. However, despite their intuitive nature and large popularity, geometric models have been subject to intense criticism due to the constraints defined by the distance axioms. Indeed, several empirical analyses have questioned and challenged the validity of the geometric framework (i.e., both the model and the notion of psychological distance), by underlying inconsistencies with human appreciation of similarity, e.g., violation of the symmetry, triangle inequality and identity of the indiscernibles, e.g. (Tversky, 1977; Tversky and Itamar, 1978; Tversky and Gati, 1982)[10].

**Feature models**

To respond to the limitation of the geometric models, Tversky (1977) proposes the feature model in which evaluated objects are manipulated through sets of features. A feature "*describes any property, characteristic, or aspect of objects that are relevant to the task under study*" (Tversky and Gati, 1982). Therefore, feature models evaluate the similarity of two stimuli according to a feature-matching function $F$ which makes use of their common and distinct features:

$$sim_F(u,v) = F(U \cap V, U \setminus V, V \setminus U) \qquad (1.1)$$

The function $F$ is expected to be non-decreasing, i.e., the similarity increases when common (distinct) features are added (removed). Feature models are thus based on the assumption that $F$ is monotone and that common and distinct features of compared objects are enough for their comparison. In addition, an important aspect is that the feature-matching process is expressed in terms of a matching function as defined in set theory (i.e., binary evaluation).

The similarity of two objects is further derived as a parametrised function of their common and distinct features. Two models, the *contrast model* ($sim_{CM}$) and the *ratio model* ($sim_{RM}$) were initially proposed by Tversky (1977). They can be used to compare two objects $u$ and $v$ represented through sets of features $U$ and $V$:

$$sim_{CM}(u,v) = \gamma f(U \cap V) - \alpha f(U \setminus V) - \beta f(V \setminus U) \quad (1.2)$$

$$sim_{RM}(u,v) = \frac{f(U \cap V)}{\alpha f(U \setminus V) + \beta f(V \setminus U) + f(U \cap V)} \quad (1.3)$$

---

[9] The similarity between two stimuli is here understood as the probability that a response to one stimulus will be generalised to the other (Shepard, 1987). With $sim(A,B)$ the similarity between two stimuli $A, B$ and $dist(A,B)$ their distance, we obtain the relation $sim(A,B) = e^{-dist(A,B)}$, that is $dist(A,B) = -log\ sim(A,B)$, a form of entropy.

[10] Note that recent contributions propose to answer these inconsistencies by generalising the classical geometric framework through quantum probability (Pothos et al., 2013). Compared objects are represented in a quantum model in which they are not seen as points or distributions of points, but entire subspaces of potentially very high dimensionality, or probability distributions of these spaces.

The symmetry of the measures produced by the two models can be tuned according to the parameters $\alpha$ and $\beta$. This enables the design of asymmetric measures. In addition, one of the major constructs of the feature model is the function $f$ which is used to capture the salience of a (set of) feature(s). The salience of a feature is defined as a notion of specificity: *"the salience of a stimulus includes intensity, frequency, familiarity, good form, and informational content"* (Tversky, 1977). Therefore, the operators $\cap, \cup$ and $\setminus$ are based on feature matching ($F$) and the function $f$ evaluates the contribution of the common or distinct features (distinguished by previous operators) to estimate the similarity[11].

**Structural alignment models**

Structural models are based on the assumption that objects are represented by structured representations. Indeed, a strong criticism of the feature model was that (features of) compared objects are considered to be unstructured, contrary to evidence suggesting that perceptual representations are well characterised by hierarchical systems of relationships, e.g., (Gentner and Markman, 1994; Markman and Gentner, 1993).

Structural alignment models are structure mapping models in which the similarity is estimated using matching functions which will evaluate the correspondence between the compared elements (Gentner and Markman, 1994; Markman and Gentner, 1993). Here, the process of similarity assessment is expected to involve a structural alignment between two mental representations in order to distinguish correspondences. Hence, the greater the number of correspondences, the more similar the objects will be considered. In some cases, the similarity is estimated in an equivalent manner to analogical mapping (Markman and Gentner, 1990) and similarity is expected to involve mapping between both features and relationships.

Another example of a structural model was proposed by Goldstone (1996, 1994a), who proposed to model similarity as an interactive activation and mapping model using connectionism activation networks based on mappings between representations.

**Tranformational models**

Transformational models assume that similarity is defined by the transformational distance between mental representations (Hahn et al., 2003). The similarity is framed in *representational distortion* (Chater and Hahn, 1997) and is expected to be assessed based on the analysis of the modifications required to transform one representation to another. The similarity, which can be explained in terms of the Kolmogorov complexity theory (Li and Vitányi, 2008), is therefore regarded as a decreasing function of transformational complexity (Hahn et al., 2003).

**Unification of cognitive models of similarity**

Several studies highlighted links and deep parallels between the various cognitive models. Tenenbaum and Griffiths (2001) propose a unification of spatial, feature-based and structure-based models through a framework relying on the generalisation of Bayesian inference – see Gentner (2001) for criticisms. Alternatively, Hahn (2011) proposes to introduce the transformational model as a generalisation of the spatial, feature and structure-based models.

In this section, we have briefly presented several cognitive models which have been proposed to explain and study (human) appreciation of similarity. These models are characterised by particular interpretations and assumptions on the way knowledge is mentally represented and processed. This highlights that the notion of object representation is central for comparing objects, and directly impacts the second critical dimension of similarity: the strategy adopted to compare object representations in order to assess the similarity of objects. As we will see, these two components will be of major importance for the definition and characterisation of semantic measures.

---

[11] As an example, the notion of the salience associated to a feature implicitly defines the possibility of designing measures which do not respect the identity of the indiscernibles, i.e. which enable non-maximal self-similarity.

In addition, we have also stressed that the fundamental differences between the models also rely on the conceptual approach used to explain similarity assessment, and more particularly on the mathematical properties of the measure which is used to compare the objects, e.g., symmetry, triangle inequality. These mathematical properties are also of major importance to better understand the models and the approaches which can be used to compare objects – similarly, semantic measures will also be analysed considering these mathematical properties.

We have also introduced that, interestingly and despite the strong differences between the different models presented, several meaningful initiatives have been undertaken to unify the cognitive models. To this end, researchers have proposed to develop frameworks which generalise existing models of similarity – once again this kind of initiative will also be found in studies related to semantic measures.

### 1.2.2 Definitions of semantic measures and related vocabulary

**Generalities**

The goal of semantic measures is easy to understand – they aim to capture the strength of the semantic interaction between semantic elements (e.g., words, concepts) based on their meaning. Are the words *car* and *auto* more semantically related than the words *car* and *mountain*? Most people would agree that they are. This has been proved in multiple experiments, inter-human agreement on semantic similarity ratings is high, e.g. (Rubenstein and Goodenough, 1965; Miller and Charles, 1991; Pakhomov et al., 2010)[12].

Appreciation of similarity is obviously subject to multiple factors. Our personal background is an example of such a factor, e.g., elderly people and teenagers will probably not associate the same score of semantic similarity between the two concepts `Phone` and `Computer`[13]. However, most of the time, a consensus regarding the estimation of the strength of the semantic link between elements, i.e. semantic relatedness, can be reached (Miller and Charles, 1991) – this is what makes the notion of semantic measures intuitive and fascinating[14].

The majority of semantic measures try to mimic the human capacity to assess the degree of relatedness between things according to semantic evidence. However, strictly speaking, semantic measures evaluate the strength of the semantic interactions between semantic entities according to the analysis of semantic proxies (texts, ontologies), nothing more. Thus, not all measures aim at mimicking human appreciation of similarity. In some cases, designers of semantic measures only aim to compare elements according to the information defined in a semantic proxy, no matter if the results produced by the measure correlate with human appreciation of semantic similarity/relatedness. This is, for instance, often the case in the design of semantic measures based on domain-specific ontologies. In these cases, the ontology can be associated to our understanding of the world, or a domain, and the semantic measure can be regarded as our capacity to take advantage of this knowledge to compare things. The aim, therefore, is to be coherent with the knowledge expressed in the considered semantic proxy, without regards to the coherence of the modelled knowledge. As an example, a semantic measure based on an ontology built by animal experts would not consider the two concepts `Sloth` and `Monkey` to be similar, even if most people think sloths are monkeys. Given that semantic measures aim at comparing things according to their meaning captured from semantic evidence, it is difficult to further define the notion of semantic measures without defining the concepts of *Meaning* and *Semantics*.

Though risking the disappointment of the reader, this section will not face the challenge of demystifying the notion of Meaning. As stressed by Sahlgren (2006) "*Some 2000 years of philosophical controversy should warn us to steer well clear of such pursuits*". The reader can refer to the various theories proposed by linguists and philosophers. In this contribution, we only consider that we are dealing with the notion of semantic meaning proposed by linguists: how meaning is conveyed through signs or language. Regarding the notion of semantics, it can be defined as the meaning or interpretation of any lexical units, linguistic expressions or instances which are semantically characterised according to a specific context. We further generally define the notion of semantic measure by the following definition:

> **Definition**: *Semantic measures* are mathematical tools used to estimate the strength of the semantic relationship between units of language, concepts or instances, through a (numerical) description obtained according to the comparison of information supporting their meaning.

---

[12] As an example, considering three benchmarks, Schwartz and Gomez (2011) observed 73% to 89% human inter-agreement between scores of semantic similarity associated to pairs of words.

[13] Given that nowadays smartphones are kinds of computers and very different to the first communication devices.

[14] Despite some hesitations and interrogations regarding the notion of (semantic) similarity, it is commonly admitted that the notions related to similarity make sense. However, there are numerous examples of authors who question their relevance, e.g. "*Similarity, ever ready to solve philosophical problems and overcome obstacles, is a pretender, an impostor, a quack.*"(Goodman, 1972) or "*More studies need to performed with human subjects in order to discover whether semantic distance actually has any meaning independent of a particular person, and how to use semantic distance in a meaningful way*" (Delugach, 1993), refer also to the work of Murphy and Medin (1985); Goldstone (1994b); Hahn and Ramscar (2001).

It is important to stress the diversity of the domain (in a mathematical sense) in which semantic measures can be used. They can be used to drive word-to-word, concept-to-concept, text-to-text or even instance-to-instance comparisons. In this book, when we do not focus on a specific type of measure, we will refer, as much as possible, to any element of the domain of measures through the generic term *element* – which refers to semantic entities or entities semantically characterised. It can therefore be any unit of language (e.g. word, text), a concept/class, an instance which is semantically characterised in an ontology (e.g., gene products, ideas, locations, persons).

We formally define a semantic measure as a function:

$$\sigma_k : E_k \times E_k \to \mathbb{R}^+ \tag{1.4}$$

with $E_k$ the set of elements of type $k \in \mathbb{K}$ and $\mathbb{K}$, the various types of elements which can be compared regarding their semantics, e.g., $\mathbb{K} = \{$words, concepts, sentences, texts, webpages, instances annotated by concepts...$\}$.

This expression can be generalised so as to take into account the comparison of different types of elements. This could be interesting to evaluate entailment of texts or to compare words and concepts, among others. However, here, we restrict our study to the comparison of pairs of elements of the same nature (a domain of study which is already a vast subject of research)[15]. In addition, the co-domain of the function $\sigma_k$ could also be relaxed to consider measures which produce results defined into more complex scales, e.g. discrete or bipolar scales. For convenience we focus on measures which are defined in $\mathbb{R}^+$. We stress that semantic measures must implicitly or explicitly take advantage of semantic evidence. As an example, as we have said, measures comparing words through their syntactical similarity cannot be considered as semantic measures; recall that semantics refers to evidence regarding the meaning of compared elements.

The distinction between approaches that can and cannot be assimilated to semantic measures is sometimes narrow; there is no clear boundary distinguishing non-semantics to semantic-augmented approaches, but rather a range of approaches. Some explanations can be found in the difficulty of clearly characterising the notion of meaning. For instance, someone can say that measures used to evaluate the syntactical distance between words capture semantic evidence related to the meaning of the words. Indeed, the sequence of characters associated to a word derives from its etymology which is sometimes related to its meaning, e.g., words created through morphology derivation such as *subset* from *set*.

Therefore, the notion of semantic measure is sometimes difficult to distinguish from measures used to compare specific data structures. This fine line can also be explained by the fact that some semantic measures compare elements which are represented through canonical forms corresponding to specific data structures for which specific (non-semantic) similarity measures have been defined. As an example, units of language represented as vectors can be compared using vector similarity measures, or pure graph similarity measures can be used to compare entities defined into semantic graphs.

In some cases, the semantics of the measure is therefore not captured by the measure used to compare the canonical forms of the elements. It is rather the process of mapping an element (e.g., word, concept) from a semantic space (text, ontology) to a specific data structure (e.g., vector, set), which semantically enhances the comparison. This, however, is an interesting paradox, the definition of the rigorous semantics of the notion of semantic measure is hard to define.

However, an important aspect to underline is that, over the years, semantic measures have been studied through various notions and not always in rigorous terms, i.e. using a well-defined terminology. Some definitions are even still subject to debate and not all communities agree on the semantics carried by the terminology they use – the notions which are commonly used in the literature to refer to semantic measures are: *semantic similarity*, *semantic relatedness*, *semantic distance*, *taxonomic distance*, *semantic dissimilarity*, *conceptual distance*, etc. These notions can have different meanings depending on the communities and/or the authors which refer to them. This highlights the difficulty to define and reduce these notions into formal mathematical frameworks. We propose to clarify the definitions considered in this book.

---

[15]Note however, that semantic measures for the comparison of units of language of different sizes are also studied in the literature, e.g. to compare the meaning carried by sentences and paragraphs. For more information the reader can refer to the notion of cross-level semantic similarity (Jurgens et al., 2014).

**Semantic relatedness and semantic similarity**

Among the various notions associated to semantic measures, this section defines the two central notions of semantic relatedness and semantic similarity, which are among the most commonly referred to in the literature. Several authors have already distinguished them in different communities, e.g., (Resnik, 1999; Pedersen et al., 2007). Based on these works, we propose the following definitions.

> **Definition** *Semantic relatedness*: the strength of the semantic interactions between two elements with no restrictions on the types of the semantic links considered.

Note that compared to the general definition of semantic measure, the notion of interaction used to define semantic relatedness refers to a positive value, i.e. the more two elements interact the more related they will be considered. As an example, compared to semantic relatedness, semantic distance refers to the degree of repulsion between the two compared elements.

> **Definition** *Semantic similarity*: subset of the notion of semantic relatedness only considering taxonomic relationships in the evaluation of the semantic interaction between two elements.

In other words, semantic similarity measures compare elements regarding the constitutive properties they share and those which are specific to them. The two concepts `Tea` and `Cup` are therefore highly related despite the fact that they are not similar: the concept `Tea` refers to a `Drink` and the concept `Cup` refers to a `Vessel`. Thus, the two concepts share few of their constitutive properties. This highlights a potential interpretation of the notion of similarity, which can be understood in term of substitution, i.e., evaluating the implication to substitute the compared elements: `Tea` by `Coffee` or `Tea` by `Cup`.

In some specific cases, communities such as linguists will consider a more complex definition of the notion of semantic similarity for words. Indeed, word-to-word semantic similarity is sometimes evaluated not only considering (near-)synonymy, or the lexical relations which can be considered as equivalent to the taxonomic relationships for words, e.g., hyponymy and hypernymy or even troponymy for verbs. In some contributions, linguists also consider that the estimation of the semantic similarity of two words must also take into account other lexical relationships, such as antonymy (Mohammad and Hirst, 2012a) – different definitions of the notion of semantic similarity can be found in the literature.

In other cases, the notion of semantic similarity refers to the approach used to compare the elements, not the semantics associated to the results of the measure. As an example, designers of semantic measures relying on ontologies sometimes use the term semantic similarity to denote measures based on a specific type of semantic relatedness which only considers meronymy, e.g., partial ordering of concepts defined by `partWhole` relationships. The semantics associated to the scores of relatedness computed from such restrictions differs from semantic similarity.[16] In this book, for the sake of clarity, we consider that only taxonomic relationships are used to estimate the semantic similarity of compared elements.

Older contributions related to semantic measures do not stress the difference between the notions of similarity and relatedness. The reader must therefore understand that in the literature, authors sometimes introduce semantic similarity measures which estimate semantic relatedness and *vice versa*. In addition, despite the fact that the distinction between the two notions is now commonly admitted by most communities, it is still common to observe improper use of both notions.

Extensive terminology refers to the notion of semantic measures and contributions related to the domain often refer to the notions of semantic distance, closeness, nearness or taxonomic distance, etc. The following subsection proposes to clarify the semantics associated to the terminology which is commonly used in the literature.

---

[16]Nevertheless, as we will see, technically speaking, most approaches defined to compute semantic similarities based on ontologies can be used on any restriction of semantic relatedness considering a type of relationship which is transitive, reflexive and antisymmetric.

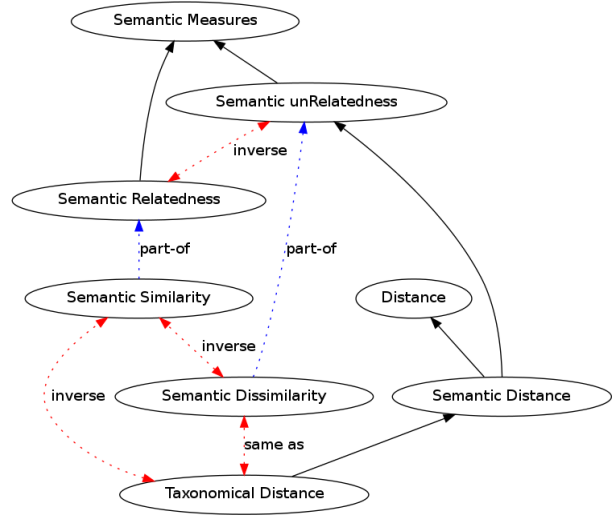

Figure 1.1: Informal semantic graph of the terminology related to semantic measures. It structures various types of semantics which have been associated to semantic measures in the literature. Black (plain) relationships correspond to taxonomic relationships, inverse relationships refer to the semantic interpretation associated to the score of the measure, e.g., semantic similarity and dissimilarity measures have inverse semantic interpretations

**The diversity of types of semantic measures**

We have so far introduced the broad notion of semantic measures and have also distinguished the two central notions of semantic relatedness and semantic similarity. Extensive terminology has been used in the literature to refer to the notion of semantic measure. Thus, we here define the meaning of the terms commonly used (the list may not be exhaustive):

- *Semantic relatedness*, sometimes called *proximity*, *closeness*, *nearness*, or *attributional similarity* (Turney and Pantel, 2010) ; refers to the notion introduced above.

- *Semantic similarity* has also already been defined. In some cases, the term *taxonomic semantic similarity* is used to stress the fact that only taxonomic relationships are used to estimate the similarity.[17]

- *Semantic distance* is generally considered as the inverse of semantic relatedness, and all semantic interactions between the compared elements are considered. These measures respect (for the most part) the mathematical properties of distances which will be introduced later. Semantic distance is also sometimes denoted as *farness*.

- *Semantic dissimilarity* is understood as the inverse of semantic similarity.

- *Taxonomic distance* also corresponds to the semantics associated to the notion of dissimilarity. However, these measures are expected to respect the properties of distances.

Figure 1.1 presents a graph in which the various notions related to semantic measures are (informally) structured through semantic relationships. Most of the time, the notion considered to be the inverse of semantic relatedness is denoted as semantic distance, whether or not the measure respects the mathematical properties characterising a distance. Therefore, for the purpose of organising the different notions, we introduce the term *semantic unrelatedness* to denote the set of measures whose semantics is the inverse to the one carried by semantic relatedness measures, without necessarily respecting the properties of a distance. To our knowledge, this notion has never been used in the literature.

---

[17]Sometimes the notions of attributional or relational similarities are used to refer to semantic similarity (Baroni and Lenci, 2010).

### 1.2.3　From distance and similarities to semantic measures

Are semantic measures mathematical measures? What are the specific properties of a distance or a similarity measure? Do semantic similarity measures correspond to similarity measures in the way mathematicians understand them? As we have seen in Section 1.2.2, contributions related to semantic measures do not rely on formal definitions of the notion of measure or distance. Indeed, contributions related to semantic measures generally rely on the commonly admitted and intuitive expectations regarding these notions, i.e. similarity (resp. distance) must be higher (resp. lower) the more (resp. less) the two compared elements share commonness[18]. However, the notions of measure and distance have been rigorously defined in Mathematics through specific axioms from which particular properties derive. These notions have been expressed for well-defined objects (element domain). Several contributions rely on these axiomatic definitions and interesting results have been demonstrated according to them. This section briefly introduces the mathematical background related to the notions of distance and similarity. It will help us to rigorously define and better characterise semantic measures in mathematical terms; it is a prerequisite to clarify the fuzzy terminology commonly used in studies related to semantic measures.

For more information on the definition of measures, distance and similarity, the reader can refer to: (i) the seminal work of Deza and Deza (2013) – *Encyclopedia of Distances*, (ii) the work of Hagedoorn (2000)– *A theory of similarity measures*, Chapter 2, and (iii) the definitions proposed by D'Amato (2007). Most of the definitions proposed in this section have been formulated based on these contributions. Therefore, for convenience, we will not systematically refer to them. In addition, contrary to most of the definitions presented in these works, here we focus on highlighting the semantics of the various definitions according to the terminology introduced in Section 1.2.2.

**Distance and similarity in Mathematics**

For the definitions presented hereafter, based on D'Amato (2007), we consider a set $D$ which defines the elements of the domain we want to compare and a totally ordered set $(V, \preceq)$. We also consider the element $min_V$ as the element of $V$ such as $\forall v \in V : min_V \preceq v$, $max_V \in V$ such as $\forall v \in V : v \preceq max_V$; and $0_V \in V$ such as $min_V \preceq 0_V \preceq max_V$.[19]

> **Definition** *Distance*: a function $dist : D \times D \to V$ is a distance on $D$ if, $\forall x, y \in D$, the function is:
>
> - Non-negative, $dist(x, y) \succeq 0_V$.
> - Symmetric, $dist(x, y) = dist(y, x)$.
> - Reflexive $dist(x, x) = 0_V$ and $\forall y \in D \land y \neq x : dist(x, x) \prec dist(x, y)$.
>
> To be considered as a distance in a metric space, the distance must additionally respect two properties:
>
> - The *identity of indiscernibles* also known as *strictness* property, *minimality* or *self-identity*, that is $dist(x, y) = 0_V$ iff $x = y$.
> - The *triangle inequality*, when $V \subseteq \mathbb{R}$, the distance between two *points* must be the shortest distance along any path: $dist(x, y) \leq dist(x, z) + dist(z, y)$.

Despite the fact that some formal definitions of similarity have been proposed, e.g., (Hagedoorn, 2000; Deza and Deza, 2013), contrary to the notion of distance, there is no axiomatic definition of similarity that sets the standard; the notion appears in different fields of Mathematics, e.g., figures with the same shape are denoted similar (in geometry), similar matrices are expected to have the same eigenvalues, etc. In this book, we consider the following definition.

> **Definition** *Similarity*: a function $sim : D \times D \to V$ is a similarity on $D$ if, for all $x, y \in D$, the function $sim$ is non-negative ($sim(x, y) \succeq 0_V$), symmetric ($sim(x, y) = sim(y, x)$) and reflexive, i.e., $sim(x, x) = max_V$ and $\forall x, y \in D \land y \neq x : sim(x, x) \succ sim(x, y)$.

---

[18] The works of D'Amato (2007) and Blanchard et al. (2008) are among the exceptions.
[19] E.g. different definitions of $V$ could be $V = \mathbb{R}, V = [0, 1], V = \{\text{very low, low, medium, high, very high}\}$.

> **Definition** *Normalised function*: any function $f$ on $D$ (e.g. similarity, distance) with values in $[0, 1]$.

Notice that a normalised similarity *sim* can be transformed into a distance *dist* considering multiple approaches; inversely, a normalised distance can also be converted into a similarity. Some of the approaches used for such transformations are presented in (Deza and Deza, 2013, Chapter 1).

As we have seen, distance and similarity measures are formally defined in mathematics as functions with specific properties. These properties are extensively used to demonstrate results and to develop proofs. However, the benefits of fulfilling some of these properties, e.g., triangle inequality for distance metric, have been subject to debate among researchers. As an example, Jain et al. (1999) stress that the mutual neighbour distance used in clustering tasks doesn't satisfy the triangle inequality but perform well in practice – to conclude by "*This observation supports the viewpoint that the dissimilarity does not need to be a metric*". Another example is the semantic measure proposed by Resnik for comparing concepts defined in a taxonomy[20] (Resnik, 1995). This measure does not respect the identity of the indiscernibles, the similarity of a concept to itself can even be low when general concepts are evaluated. However, despite the fact that this property may seem counter-intuitive, this measure has proved to perform well in several usage contexts. This highlights the gap which often exists between (i) formal definitions of measures which are based on axiomatic definitions and rigid expectations of well-defined properties and (ii) results provided by empirical evaluations which sometimes challenge the benefits of respecting specific properties characterising measures.

A large number of properties which are not presented in this section have been distinguished to further characterise distance or similarity functions, e.g., see (Deza and Deza, 2013). These properties are important and needed for theoretical proofs. However, as we have seen, the definition of semantic measures proposed in the literature is not framed in the mathematical axiomatic definitions of distance or similarity. In some cases, such a distortion among the terminology creates difficulties in bridging the gap between the various communities involved in the study of semantic measures and similarity/distance. As an example, in the *Encyclopedia of distances*, Deza and Deza (2013) do not distinguish the notions of distance and dissimilarity, which is the case in the literature related to semantic measures (refer to Section 1.2.2). In this context, the following section defines the terminology commonly adopted in the study of semantic measures w.r.t the mathematical properties already introduced.

**Flexibility of semantic measures**

Notice that we have not introduced the precise and technical mathematical definition of a measure proposed by *measure* theory. This can be disturbing considering that this manuscript extensively refers to the notion of semantic measure. The notion of measure we use is indeed not framed in the rigorous mathematical definition of *measure*. Such a definition would exclude many semantic measures defined in the literature. The notion of measure considered in this book therefore refers to the common sense of the term measure, any "*measuring instruments*" which can be used to "*assess the importance, effect, or value of (something)*" (Oxford Dict., 2012) – in our case, any functions answering the definitions of semantic distance/relatedness/similarity/etc. proposed in Section 1.2.2.

Various communities have used the concepts of similarity or distance without considering the rigorous axiomatic definitions proposed in mathematics but rather using their broad intuitive meanings[21]. To be in accordance with most contributions related to the field, and to facilitate the reading of this book, we will not limit ourselves to the mathematical definitions of distance and similarity.

The literature related to semantic measures generally refers to a semantic distance as any (non-negative) function, designed to capture the inverse of the strength of the semantic interactions linking two elements. Such functions must respect that: the higher the strength of the semantic interactions between two elements, the lower their distance. The axiomatic definition of a distance (metric) may not be respected. A semantic distance is, most of the time, what we define as a function estimating semantic unrelatedness. However, to be in accordance with the literature, we will use the term semantic distance to refer to *any* function designed to capture semantic unrelatedness. We will explicitly specify that the function respects (or does not respect) the axiomatic definition of a distance (metric) when required.

---

[20]The measure will be introduced gradually in Chapter 3 – for the interested reader the similarity between two concepts $u, v$ is defined by $sim(u, v) = IC(MICA(u, v))$ with $IC(x)$ the specificity of the concept $x$ which is defined as the inverse of the logarithm of the probability of occurrence of a concept in a corpus of texts, i.e. $-log(p(x))$ and $MICA(u, v)$ the common ancestor of $u$ and $v$ which is the most specific, i.e. the common ancestor with the maximal IC.

[21] As we have seen, researchers in cognitive science have demonstrated that human expectations regarding (semantic) distance challenges the mathematical axiomatic definition of distance. Thus, the communities involved in the definition of semantic measures mainly consider a common vision of these notions without always clearly defining their mathematical properties.

| Properties | Definitions |
|---|---|
| Non-negative | $dist(x,y) \succeq 0_V$ |
| Symmetric | $dist(x,y) = dist(y,x)$ |
| Reflexive | $dist(x,x) = 0_V$ |
| Normalised | $V = [0,1]$ |
| Identity of indiscernibles | $dist(x,y) = 0_V \; iff \; x = y$ |
| Triangle inequality ($V \subseteq \mathbb{R}$) | $dist(x,y) \leq dist(x,z) + dist(z,y)$ |

Table 1.1: Properties which can be used to characterise any function which aims to estimate the notion of distance between two elements. Refer to the notations introduced page 12.

| Properties | Definitions |
|---|---|
| Non-negative | $sim(x,y) \succeq 0_V$ |
| Symmetric | $sim(x,y) = sim(y,x)$ |
| Reflexive | $sim(x,x) = max_V$ |
| Normalised | $V = [0,1]$ |
| Identity of indiscernibles | $sim(x,y) = max_V \; iff \; x = y$ |
| Integrity | $sim(x,y) \preceq sim(x,x)$ |

Table 1.2: Properties which can be used to characterise any function which aims to estimate the notion of similarity/relatedness between two elements. Refer to the notations introduced page 12.

Semantic relatedness measures are functions which are associated to an inverse semantics of the one associated to semantic unrelatedness: the higher the strength of the semantic interactions between two elements, the higher the function will estimate their semantic relatedness.

The terminology we use (distance, relatedness, similarity) refers to the definitions presented in Section 1.2.2. To be clear, the terminology refers to the semantics of the functions, not their mathematical properties. However, we further consider that semantic measures must be characterised through mathematical properties. Table 1.1 and Table 1.2 summarise some of the properties which can be used to formally characterise any function designed in order to capture the intuitive notions of semantic distance and relatedness/similarity. These properties will be used in this book to characterise some of the measures that we will consider. They are essential to further understand the semantics associated to the measures and to distinguish semantic measures which are adapted to specific contexts and usage.

## 1.3 Classification of semantic measures

We have seen that various mathematical properties can be used to characterise technical aspects of semantic measures. This section distinguishes other general aspects which may be interesting to classify semantic measures. They will be used to introduce the large diversity of approaches proposed in the literature. First we present some of the general aspects of semantic measures which can be relevant for their classification, and subsequently introduce two general classes of measures.

### 1.3.1 How to classify semantic measures

The classification of semantic measures can be made according to several aspects; we propose to discuss four of them:

1. The type of elements that the measure aims to compare.

2. The semantic proxies used to extract the semantics required by the measure.

3. The semantic evidence and assumptions considered during the comparison.

4. The canonical form adopted to represent an element and how to handle it.

**Types of elements compared: words, concepts, instances, etc.**

Semantic measures can be used to compare various types of elements:

- Units of language: words, sentences, paragraphs, documents.

- Concepts/Classes, groups of concepts.

- Semantically characterised instances.

Semantic measures can therefore be classified according to the type of elements they aim to compare.

**Semantic proxies from which semantics is distilled**

Semantic measures require a source of information to compare two semantic entities. It will be used to characterise compared elements and to extract the semantics required by the measure.

> **Definition** *Semantic proxy*: any source of information from which indication of the semantics of the compared elements, which will be used by a semantic measure, can be extracted.

Two broad types of semantic proxies can be distinguished:

- Unstructured or semi-structured texts: Text corpora, controlled vocabularies, dictionaries.

- Structured: ontologies, e.g., thesaurus, structured vocabularies, taxonomies.

**Semantic evidence and considered assumptions**

Depending on the semantic proxy used to support the comparison of elements, various types of semantic evidence can be considered. The nature of this evidence conditions the assumptions associated to the measure.

> **Definition** *Semantic evidence*: any clue or indication based on semantic proxy analysis from which, often based on assumptions, a semantic measure will be based.

As an example, considering the measures which rely on text analysis, we have already mentioned that the proximity or relatedness of terms can be assessed considering that pairs of terms which co-occur frequently are more related. In this case, the co-occurrence of words is considered as semantic evidence; its interpretation is governed by the assumption that relatedness of terms is a function of their degree of co-occurrence.

**Canonical forms used to represent compared elements**

The canonical form (representation) chosen to process a specific element can also be used to distinguish the measures defined for comparing a specific type of element. Since a canonical form corresponds to a specific reduction of the element, the degree of granularity with which the element is represented may highly impact the analysis. The selected canonical form is of major importance since it influences the semantics associated to the score produced by a measure, that is to say, how a score must/can be understood/interpreted. This particular aspect is essential when inference must be driven from scores produced by semantic measures.

> It is therefore important to stress that a semantic measure is defined to process a given type of element which is represented through a specific canonical form.

### 1.3.2   A general classification of semantic measures

Figure 1.2 presents a partial overview of the landscape of semantic measures which can be used to compare various types of semantic entities (e.g. words, concepts, instances). It summarizes one of the classifications of semantic measures which can be proposed. As we have seen, measures can first be classified based on the elements they can compare. Based on this aspect, we distinguish two main types of measures:

- Corpus-based measures used to compare units of language, concepts or instances from text analysis, i.e. unstructured semantic proxies. These measures are generally used to compare words or more generally units of language. However, they can also be adapted for comparing concepts or instances by considering that disambiguation techniques have been used to identify concept or instance denotations in texts.

- Knowledge-based measures which are designed for comparing entities defined in ontologies, i.e. structured semantic proxies. Knowledge-based measures can also be used to compare units of language, e.g., sentences or texts, for instance by considering that disambiguation techniques have been used for establishing bridges between texts and ontologies.

Hybrid strategies can also be defined mixing both distributional and knowledge-based measures. Nevertheless, in the literature, measures are generally defined for comparing a specific type of elements: units of language or entities defined in an ontology. Therefore, classifying measures based on the elements they compare and the semantic proxy which is used in the analysis, i.e. texts or ontologies (knowledge representation system), helps to distinguish the general types of measures which have been proposed.

These measures are based on different semantic evidence and assumptions which are used to capture the semantics of compared elements, e.g. the distributional hypothesis, intentional or extensional evidence expressed into ontologies. Based on these evidence and assumptions, a model is defined for comparing two elements – such a model is generally denoted a semantic measure. Various specific types of approaches have been proposed for distributional and knowledge-based measures, Figure 1.2 structures several broad categories. Depending on the strategy which is used for defining the measure and the evidence and assumptions which are considered, the semantics of the measure, i.e. the meaning which can be associated to the scores it produces, may vary.

This chapter has introduced the notion of semantic measures. We have presented their practical usages in different application contexts, general definitions associated to the notion have been proposed, and different semantics which can be associated to them have been distinguished. This latter point helped us to better capture the meaning of semantic measures (results). To this end, we define the terminology classically found in the literature, e.g., semantic similarity/proximity/relatedness/distance, and we proposed an organisation of the notions commonly used, e.g. semantic similarity is a component of semantic relatedness. In a second step, to better understand the characteristics of semantic measures, we distinguished several central aspects of measures which can be used to categorising the large diversity of measure proposals. As a result, a general classification of the variety of semantic measures defined in the literature has been presented. Such a classification highlights the similarities and differences of the numerous measures and approaches which have been proposed in the literature. It can therefore be used to better understand the large diversity of measures and to characterise areas of research which have not been explored for designing measures. Importantly, this overview of semantic measures and the proposed classification also stresses the breadth of this field of study and the difficulty to define the notions on which are based semantic measures, e.g., semantic relatedness and semantic similarity.

The two following chapters are dedicated to an in-depth introduction to both corpus-based and knowledge-based semantic measures.

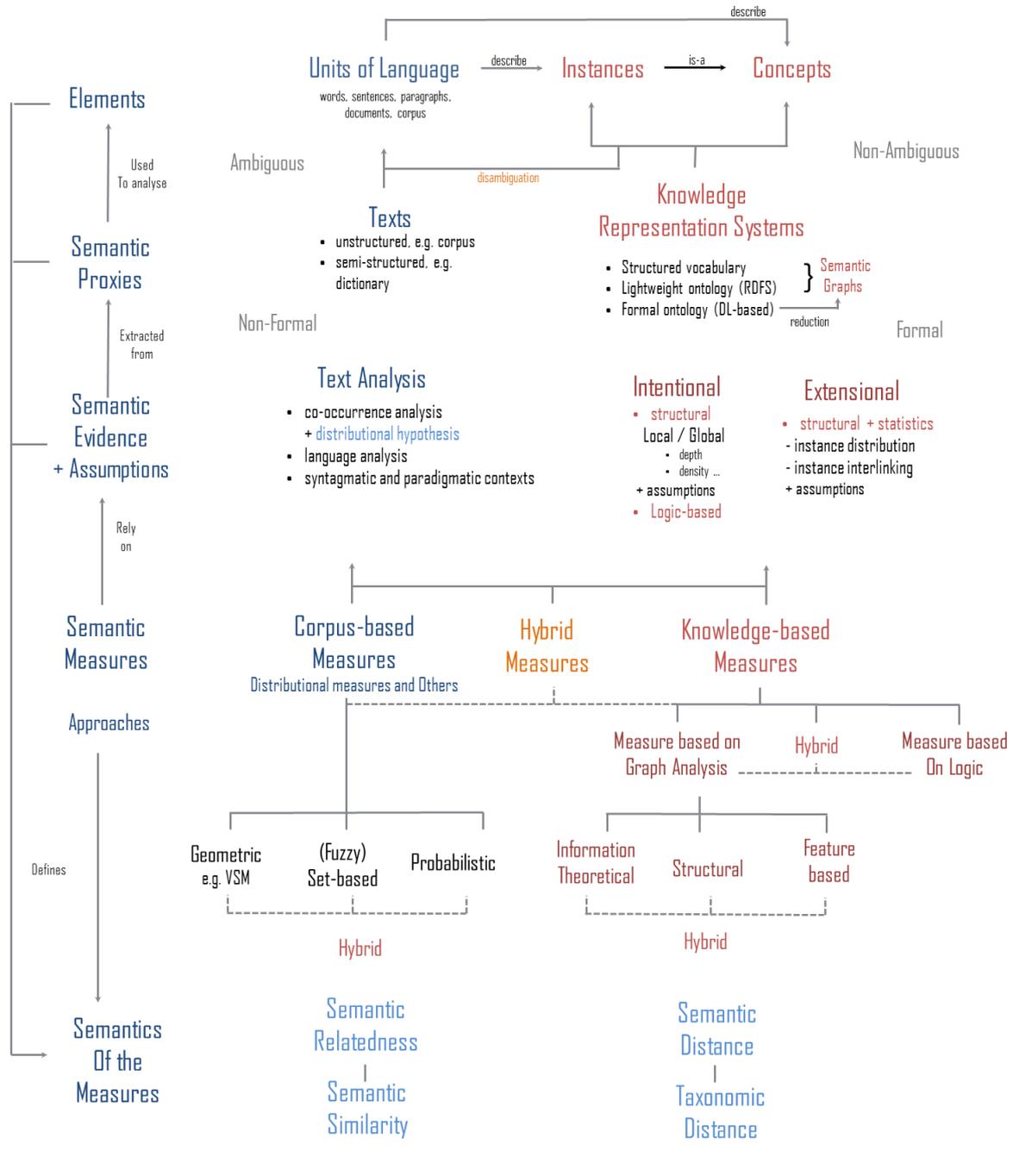

Figure 1.2: Partial overview of the landscape of the types of semantic measures which can be used to compare various types of elements (e.g., words, concepts, instances) (Harispe et al., 2013b)

# Chapter 2

# Corpus-based semantic measures

## 2.1 From text analysis to semantic measures

Corpus-based semantic measures enable the comparison of units of language from the analysis of unstructured or semi-structured texts. They are more generally used to compare words, sentences or texts based on NLP techniques which most often only rely on statistical analysis of word usage in texts, e.g. based on the analysis of word (co-)occurrences and the linguistic contexts in which they occur. As we will see in this chapter, corpus-based measures cannot most of the time be reduced to single mathematical formulae. They rather refer to complex pipelines of treatments which are used (i) to extract the semantics of compared units of language, to further (ii) compare these units by analysing their semantics. To this end, corpus-based semantic measures take advantage of a large variety of NLP and information retrieval algorithms – which makes corpus-based semantic measures a broad field of study at a crossroad between several domains, in particular computational linguistics and information retrieval.

Corpus-based measures are often denoted *distributional measures* in the literature (Mohammad and Hirst, 2012a). This is to stress that most measures are explicitly or implicitly based on the *distributional hypothesis* – a central hypothesis of distributional semantics which states that words occurring in similar contexts convey similar meaning (Harris, 1981). Nevertheless, like Mihalcea et al. (2006) and other authors, we here adopt the more general denotation of corpus-based semantic measures. This is to ease the introduction of the variety of measures which are based on corpus or natural language analysis, including those for which the distributional hypothesis is not considered to be the root of the approach. That is the case for measures which are based on the analysis of results provided by information retrieval systems. To stress this point, different classifications of semantic measures based on text analysis have been proposed in the literature, e.g. (Mihalcea et al., 2006; Panchenko, 2013). The one which has been adopted in this chapter is among the most used. In addition, even if this reflection is out of the scope of this book and will therefore not be discussed hereafter, evidence indicate that all corpus-based semantic measures are somehow implicitly or explicitly defined into the framework of distributional semantics (which will be introduced in this chapter).

Much of the literature related to corpus-based measures focuses on the comparison of a pair of words; extensive surveys have been proposed by Weeds (2003); Curran (2004); Sahlgren (2008); Dinu (2011); Mohammad and Hirst (2012a); Panchenko (2013). Several contributions have also been proposed to compare larger units of language such as pairs of sentences or texts, e.g. (Corley and Mihalcea, 2005; Yu et al., 2006; Hughes and Ramage, 2007; Ramage et al., 2009; Buscaldi et al., 2013). However, most of these latter measures are extensions of measures which have been defined for comparing words, or rely on approaches which are also used to compare words, e.g. topic models such as Latent Semantic Analysis (Lintean et al., 2010) or Latent Dirichlet Allocation (Blei et al., 2003). Therefore, for the sake of clarity and due to space constraint, this chapter mainly introduces semantic measures which have been defined for comparing words. We will not present measures which can be used to compare texts or sentences[1].

Generally speaking, corpus-based semantic measures are based on a strategy which will be used to capture the meaning of a word – this meaning is often regarded as a function of its usage in a semantic space built from a corpus of texts. Depending on the strategy which is adopted to (i) characterise the meaning of a word, and to (ii) represent the semantic space in which this meaning is defined, a specific canonical form will be selected to represent a word. This canonical form corresponds to a data structure which is expected to encompass evidence of the meaning of the word. It can be regarded as the second proxy layer from which the meaning of a word will be processed – the first one being the natural language under study. This representation will be of major importance for defining corpus-based semantic measures. It enables the meaning of words to be processed by algorithms and therefore enables words to be compared based on the comparison of their respective canonical forms.

---

[1] A large literature is dedicated to the subject. As an introduction to the field the reader may refer to the broad overview of measures provided by Achananuparp et al. (2008). We also encourage the reader to take an interest to the central notion of compositionality which is essential in order to adapt models commonly used for assessing the similarity of words in order to process units of language larger than words (Kamp et al., 2014).

Despite the fact that corpus-based semantic measures are most often complex processes composed of multiple algorithms, four main components characterise the definition of a strategy for assessing the semantic similarity of words[2]:

1. A premise on what the meaning of a word is.

2. A set of assumptions which defines the semantic evidence convey by natural language from which the meaning of a word can be captured.

3. A representation of a word such as (most of) its meaning – as defined in 1 and captured by the semantic evidence defined in 2 – can be processed by algorithms. In some cases, this is a two-step process as a representation of the whole semantic space will first be defined to next extract a specific canonical form of a word.

4. An algorithm or mathematical function specifically designed for comparing two word representations.

All corpus-based semantic measures defined in the literature differ regarding the strategies which have been adopted to implement each of these components – mainly components 2, 3 and 4.

The premise which defines the meaning of a word (Component 1) directly impacts the choice of (i) the semantic evidence which will be used to define the representation of a word and (ii) the approach which will be used to compare these representations. Nevertheless, this aspect is not always discussed when measures are defined. As we will see, the meaning of a word is generally considered to be a function of its usage – according to the distributional hypothesis.

Note also that the comparison of two words representations (Component 4) has, technically speaking, no direct relationship with semantic analysis. Indeed, this step mainly refers to the definition or the use of mathematical functions or algorithms for comparing data structures. In some cases, this step corresponds to the definition of *ad hoc* functions used to compare complex representations of words. However, in most cases, words are represented using well-known mathematical objects (e.g. sets, vectors, probability distributions, nodes in graphs). In these cases, state-of-the-art measures defined for comparing these mathematical objects are used, regardless of the semantics carried by these objects.

Figure 2.1 presents a general and conceptual overview of the various steps which can be used for defining corpus-based measures. Most of the approaches proposed in the literature can be break up into some of these steps. Each of these steps will be presented in this chapter and we will show how they can be combined in order to assess the semantic relatedness of words. Below, the five steps presented in Figure 2.1 are introduced:

1. The semantic proxy which is used is a corpus of unstructured or semi-structured texts, for instance retrieved from the Web. This corpus can be preprocessed depending on the strategy which is used for defining the measure. Such a preprocessing can involve, among others: stemming/lemmatization, Part-Of-Speech (POS) tagging, and removal of specific words, e.g. stop words – in some cases the study focuses on specific POS such as nouns or verbs.

2. At this stage, the vocabulary is distinguished, almost always implicitly. It contains all the words for which the measure will be able to compute a score of relatedness. Note that the notion of words may refers to a set of lemma or to complex objects such as a set of pairs ⟨lemma, POS⟩ for instance.

---

[2]Note that these components are found in all the approaches which can be used for comparing units of language (e.g., sentences, texts).

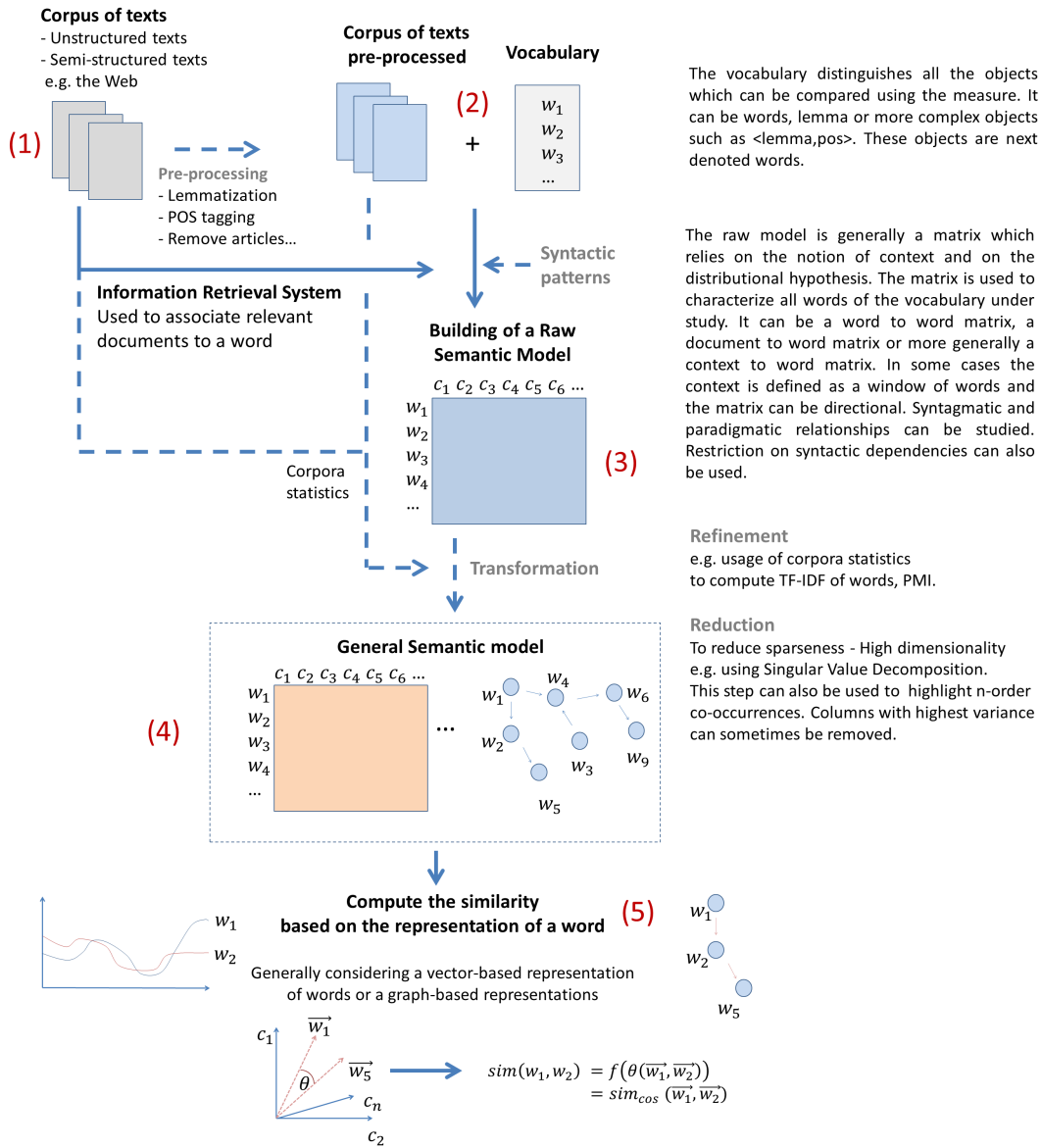

Figure 2.1: General process commonly adopted for the definition of corpus-based semantic measures.

3. Based on the (preprocessed) corpus and the vocabulary considered, a model is next built. It will be used to characterise all the words of the vocabulary under study. This model is generally a matrix which represents each word of the vocabulary using a set of contexts. As we will see, most of the diversity of measures can be explained by the variety of strategies which have been proposed to build such a model. The construction of this model is based on a set of assumptions regarding (i) the meaning of words and (ii) the semantic evidence which can be used for evaluating this meaning. As an example, the model can be a word to word co-occurrence matrix considering that two words co-occur if they appear in a word window of a specified size. In some cases syntactic patterns are used to study word co-occurrences. In other cases, using an information retrieval system, the model can also be a word-document matrix which specifies the relevant documents w.r.t a given word.

At this stage, a raw model is built. Some transformations can be applied on it; we distinguish the phase of refinement and the phase of reduction:

- The refinement refers to the treatments which are used to improve the model by incorporating additional information extracted from the corpus under study. As an example, some of the approaches incorporate corpora statistics to take into account the informativeness of specific

   words, e.g. by taking into account term frequency.

- The reduction refers to the use of specific techniques for reducing the potentially high dimensionality of the matrix. Note that this matrix is generally sparse as most of the words never co-occur in the same contexts. This step generally involves specific matrix processing and reduction techniques such as Singular Value Decomposition (SVD) (Berry et al., 1995; Golub and Van Loan, 2012) – these approaches will be briefly introduced later.

4. The transformation step aims at obtaining the general model, here denoted *semantic model*, which will be analysed for comparing the words. In some cases the semantic model, originally represented as a matrix, will be transformed into a graph representation. Therefore, interestingly, a bridge can be established between corpus-based and knowledge-based measures which will be discussed in detail in Chapter 3. Indeed, some of the strategies represent the semantic space using well-known knowledge representation models. As an example, it has been proposed to build semantic networks or referential networks from dictionaries and thesauruses to further compare words according to the strength of their interconnections in the semantic network, e.g. (Nitta, 1988; Kozima and Furugori, 1993; Niwa and Nitta, 1994; Kozima and Ito, 1997; Blondel, 2002; Ho and Cédrick, 2004; Iosif and Potamianos, 2012). This kind of approaches will be discussed in Chapter 3 which is dedicated to knowledge-based measures.

5. Finally, two words can be compared analysing the semantic model which has been defined. Once again a large variety of measures can be used depending on the representation of a word the model enables. However, despite the diversity of measures proposed in the literature, two general approaches for defining corpus-based measures can be distinguished:

    - Distributional measures: semantic measures based on distributional semantic models (Baroni and Lenci, 2010) and the distributional hypothesis (Harris, 1981).
    - Other types of measures: measures based on approaches which do not explicitly rely on the distributional hypothesis, e.g. approaches based on information retrieval systems.

   These two approaches are presented in this chapter. However, we will mainly focus on distributional measures as they are both the most studied and the most used in the literature. Let us remind that the measures which rely on a graph-based representation of the semantic model will be presented in the chapter dedicated to knowledge-based measures (Chapter 3).

The aim of this chapter is not to introduce the extensive list of corpus-based measures which have been proposed in the literature. It is rather to present the central notions on which corpus-based measures rely. Therefore, a large portion of this chapter is dedicated to the introduction of the key notions of *context* and *semantic model*. Several examples of context and semantic model definitions will be illustrated. We next present several examples of measures which can be used to take advantage of these context and semantic model definitions for designing semantic measures.

The reader must indeed understand that the central aspect of corpus-based semantic measures relies in the definition of the semantic model from which will be extracted the processable representation of a word. Indeed, next, the definition or use of a measure for comparing two word representations is most often only a matter of a technical discussion on mathematical formulae dedicated to the comparison of specific data structures - e.g., vectors, sets.

The remainder of this chapter is organised as follow. After a brief introduction of what it is generally understood by *word meaning*, Section 2.2 presents several types of semantic evidence which can be extracted from natural language for assessing the similarity of words or more generally for assessing the similarity of two units of language. In this section, important concepts related to corpus-based measures will be introduced, e.g. distributional semantics, the distributional hypothesis, the notion of context. Section 2.3 presents several proposals of distributional measures. Section 2.4 briefly introduces the other types of measures which have been proposed in the literature. Section 2.5 discusses the advantage and limits of corpus-based measures. Finally, Section 2.6 concludes the chapter.

## 2.2 Semantic evidence of word similarity in natural language

In this section we first briefly discuss the notion of word meaning to further introduce semantic evidence which is commonly used to compare words from natural language analysis. In particular, we introduce the two structural relationships which can be studied between words, namely paradigmatic and syntagmatic relationships.

### 2.2.1 The meaning of words

The notions of meaning and semantics have been extensively discussed by numerous communities, e.g. (Osgood, 1952; Aitchison, 2012), and this survey does not aim to propose an in-depth analysis of the different theories proposed so far. Nevertheless, as we saw in Section 2.1, the definition of word meaning directly influences the way semantic measures are defined. Indeed, an accurate measurement technique can only be defined if a clear and non-ambiguous definition of what we want to measure has been defined. However, the notion of semantic similarity or semantic relatedness of words – what we want to measure using corpus-based semantic measures – is generally not defined in the contributions related to semantic measures. The authors generally refer to the intuitive notion of word similarity/relatedness w.r.t their meaning by giving examples of the behaviour expected by measures, e.g. the measure must enable to distinguish that the two words (*cup, tea*) are more semantically related than the two words (*cup, wine*)[3]. In this context, considering that measures can be evaluated using benchmarks composed of expected results of similarity, finely defining the notion of word meaning may seem unnecessary. This explains that some authors have questioned the relevance to finely characterise the notion of word meaning for assessing the similarity/relatedness of words, e.g. (Karlgren and Sahlgren, 2001; Sahlgren, 2006). The necessity to define what we consider by word meaning for designing word similarity measure is still an open debate.

An important point to understand is that most contributions related to corpus-based semantic measures consider semantic relatedness as linguistic relatedness. The meaning of a word is assumed to be directly or indirectly explainable by the sole study of language, independently from both (i) the knowledge of language users and (ii) the environment. Therefore, the notion of linguistic relatedness, commonly denoted semantic relatedness, fall within the domain of *Semantics*, i.e. the study of language in isolation, and does not refer to the domain of *Pragmatics* which also incorporates the analysis of the context in which the language is used (Cruse, 2011). As discussed in Sahlgren (2008), this does not mean that linguists do not consider that extralinguistic factors play a role in defining the meaning of language, and therefore that linguistics can for instance be used to compare the meaning of the signified or the referent of a signifier[4] – it only means that it is generally considered that "*For a large class of cases – though not for all – in which we employ the word "meaning" it can be defined thus: the meaning of a word is its use in the language*", i.e. "*meaning is use*" (Wittgenstein, 2010), i.e. "*Not the meanings that are in our heads, and not the meanings that are out there in the world, but the meanings that are in the text*" (Sahlgren, 2008).

Focusing on semantic measures, it rather means that corpus-based measures compare words w.r.t their linguistic meaning, i.e. the meaning of words which can be captured by analysing language in isolation, without regard on the amount of meaning of a word which is explicitly or implicitly conveyed by language. In corpus-based semantic measures words are therefore solely compared w.r.t their usage in texts. Interestingly, the usage of words in texts is assumed to reflect a commonly accepted meaning of words. This central notion relates to some of the early work of Firth: "*The complete meaning of a word is always contextual*" (Firth, 1935). It will be further discussed through the introduction of the distributional hypothesis.

Therefore, in the majority of cases, corpus-based measures are assumed to evaluate the distributional relatedness of words: words are considered to be related if they occur in the same context

### 2.2.2 Structural relationships: Paradigmatic and Syntagmatic

An interesting point to note is that a structural vision of language is adopted in most contributions related to corpus-based semantic measures. Structuralism have been developed based on the work of

---

[3] What we did in this book for introducing the notion of semantic similarity and semantic measures.
[4] Signifier refers to the word, sequence of graphemes (letters), phonetic. The signified refers to the mental representation of a concrete or abstract concept the signifier refer to, and the referent is the concrete object in the real world (Chandler, 2007).

| Relationships | **Syntagmatic** | | |
|---|---|---|---|
| **Paradigmatic** | The | cat | is | eating |
| | A | dog | was | barking |

Table 2.1: Paradigmatic and syntagmatic relationships.

Saussure (1857-1913)[5]. Here the natural language is regarded as a sequence of symbols (such as words) with no *a priori* semantics. These symbols are considered to be structured by two types of relationships:

- *Paradigmatic relationships*. All symbols (e.g. words) are regarded as *paradigms* which are members of a specific class or semantic group. Paradigms are considered to establish a paradigmatic relationship – or associative relation – if they can be substituted, at least without modifying the grammatical coherence of the sentence. Therefore, a paradigm is generally considered as a set of symbols which refer to a specific class, e.g. verbs or nouns are examples of grammatical paradigms. Therefore, the paradigmatic vision considers a sentence as a sequence of disjunction of paradigms $p_0 \vee p_1 \vee p_2 \vee p_3 \ldots \vee p_i$ with $p_i$ a specific paradigm. As an example, in Table 2.1 the word *cat* could be replaced by the word *dog* without implication on the grammatical coherence of the sentence. Note that the notion of class or semantic group which can link the various paradigms is broadly defined in the literature (Booij et al., 2000). It may refer to the lexical category of the syntagm or even to its meaning when classes refer to synonyms, hypo/hypernyms, antonyms.[6]

- *Syntagmatic relationships* correspond to the chain of associations of the symbols which compose a sentence, or a larger lexical unit. The selection of specific paradigms and their combination generates a syntagm which traduces the sequence of lexical units related by syntagmatic relationships. This contributes to define the meaning of a sentence. A syntagm is governed by the grammar of the language in use, i.e. the rules which define the coherency of expressing a specific chain of symbols. According the syntagmatic vision, a sentence is regarded as a sequence of conjunction of paradigms of the form $p_0 \wedge p_1 \wedge p_2 \wedge p_3 \ldots \wedge p_i$ with $p_i$ a specific paradigm, e.g. the two sentences presented in Table 2.1 correspond to specific chains of paradigms.

Paradigmatic analysis refers to the study of specific patterns in texts, in contrast to syntagmatic analysis which will be based on the analysis of a surface representation of the language in which words co-occurrence is a central component. Kozima and Furugori (1993), among others, mention the importance of these two types of structural relationships for discussing the relatedness of words. However, due to the different nature of these relationships, and the different information they convey, some authors have proposed to distinguish the notions of paradigmatic and syntagmatic relatedness (Sahlgren, 2008):

- *Paradigmatic relatedness* refers to the notion of substitutability w.r.t the impact of substituting a word by another, e.g. on the grammatical coherence of a sentence. Indeed, more finely, paradigmatic relatedness of two words can be seen as a function of the impact of substituting two words on the meaning conveyed by a sentence. Paradigmatic relationships between words must therefore be evaluated by means of indirect co-occurrences. As an example, the two synonyms *father* and *dad* will not often co-occur, but will rather tend to occur with the same words – this can be captured by analysing second-order co-occurrences or more generally indirect co-occurrences of words. Paradigmatic relatedness is a broad notion. Indeed, the association of specific paradigms in the same semantic group may be justified by a variety of semantic relationships which can be established between paradigms, e.g. hypo/hypernymy, synonymy, antonymy.

- *Syntagmatic relatedness* is mainly captured inside a specific text region by means of words collocation and direct co-occurrence (Kozima and Furugori, 1993), i.e. first-order co-occurrences.

As we will see, structuralism and more particularly paradigmatic and syntagmatic relationships are central for the definition of corpus-based semantic measures.

### 2.2.3 The notion of context

The notion of context is central in several natural language analysis. It is of major importance for capturing the meaning of a word through the analysis of syntagmatic and paradigmatic relationships.

---
[5]For more information, the reader may refer to the work of Culler (1986); Adedimeji (2007); De Saussure (1989).
[6]Interestingly the association of specific paradigms into a specific class may be motivated by the notion of semantic relatedness (Booij et al., 2000).

|       | $d_0$ | $d_1$ | $d_2$ | $d_3$ |
|-------|-------|-------|-------|-------|
| $w_0$ | 1     | 1     | 0     | 0     |
| $w_1$ | 0     | 1     | 0     | 0     |
| $w_2$ | 1     | 1     | 1     | 0     |
| $w_3$ | 1     | 1     | 0     | 0     |

Table 2.2: Words represented according to the Vector Space Model (VSM).

Indeed, the meaning of a word is generally considered to be understandable only w.r.t a context of use, i.e. according to its usage which is defined through the structural relationships the word establishes. Thus, the notion of context is central to perform structural processing of natural language by analysing paradigmatic and syntagmatic relationships. As an example, first-order co-occurrences of words will be evaluated by studying the syntagmatic relationships between words.

Several approaches have been proposed for defining the context of a word. They differ in their linguistic sophistication, algorithmic complexity, reliability, and information they consider (Curran, 2004). A basic approach is to consider the context of a word as the document of the corpus in which the word occurs. Such a definition of context refers to the semantic model on which is based Vector Space Model (VSM) – a widely known model which have been proposed in Information Retrieval to characterising documents w.r.t. the vocabulary which constitutes the corpus (Salton and McGill, 1983; Turney and Pantel, 2010). In VSM a document is considered as a topical unit. It is represented as a vector which highlights the words it contains. In this aim, the corpus is used to derive a semantic model which corresponds to a document-word matrix. A basic example of such a matrix is presented in Table 2.2 – $w_i$ refers to the word $i$ and $d_j$ to the document $j$ of the corpus.

If the word $w_i$ occurs in the document $d_j$ the cell $(w_i, d_j)$ will be filled by the value 1. Based on this model, when a query must be evaluated in order to find the more relevant documents, this is the vector representation of the query which will be evaluated. The information retrieval system will be based on a similarity function which will assess the similarity between the vector-based representation of the query and the vector representation of each document (i.e. the columns of the matrix) – vector comparison techniques will be introduced in Section 2.3, just consider for now that the similarity of two vectors can easily be computed using specific mathematical formulae.

Therefore, by considering the transpose of the document-word matrix used in VSM, we can consider that each word is also represented according to a vector representation – the rows of the matrix in Table 2.2, e.g. $w_0$ is represented by the vector $[1, 1, 0, 0]$. This vector representation of a word highlights the documents in which the word occurs. Thus, considering that vector similarity can be assessed using specific mathematical functions, the semantic relatedness of two words can simply be defined as a function of the similarity of their vector-based representations.

More generally, as stressed by Sahlgren (2006) and other authors, we can consider that VSM relies on a semantic model which is a specific expression of a more general approach. Such a general approach does not constrain words to be characterised by documents but enables to build semantic models in which words are analysed w.r.t contexts, e.g. paragraph, sentence, word window – in the case of VSM the semantic model refers to a document-word matrix, the context selected is therefore a document[7].

Thus, more generally, a semantic model can be defined as a matrix in which each row $i$ is a vector representation of the word $w_i$, and each dimension of the vector refers to a specific context $c_n$, i.e. the row $i$ provides a vector-based representation of the corresponding word $w_i$ with $\vec{w}_i = (c_1, c_2, \ldots, c_n)$. The notion of context is therefore essential for the definition of semantic measures. It is a core element of the semantic model from which a word will be characterised and represented to further be compared.

In this section we introduce the contexts which can be defined for studying word co-occurrences and to build semantic models. We will next discuss how these semantic models can be used for assessing the relatedness of words. Several alternative approaches have been proposed to define processable definitions of context; this section will introduce the two broad types of contexts which can be analysed but will not cover in detail the large literature dedicated to their tuning and to the numerous variants which have been proposed and evaluated. For more information the reader may refer to (Weeds, 2003; Curran, 2004; Sahlgren, 2006) and to the specific contributions which will be introduced hereafter. The notion of context is generally considered according to the type of structural relationships which is studied. Syntagmatic and paradigmatic contexts are therefore commonly distinguished; both types of contexts

---

[7] In some contributions VSM also refer to other models such as distributional models - and may refer to any model in which dimensions of vectors/matrices/tensors are *derived from event frequencies* (Turney and Pantel, 2010).

are introduced.

**Syntagmatic contexts**

Syntagmatic contexts refer to the study of co-occurrences in sequential ordering of words. Using this approach a specific window size of words will generally be used to define a context, e.g. using a five noun window the word *bananas* will be analysed by considering contexts such as "...[*Monkeys* love *fruits* such as **bananas**, they are a great healthy *food source*] for them...". Considering multiple-words windows enables to not only focus on collocation of words but also to capture words which co-occur considering larger text regions and several words between them. A variety of alternative linguistic constructs can be used to define a syntagmatic context: a sentence, a paragraph, an n-size word window or even a window composed of several characters to mention a few. Different techniques can be used to process a context depending on its definition and on its the tuning:

- Contexts which correspond to relatively large text regions such as documents, paragraphs or sentences will generally be analysed by counting the occurrences of each word in each context. This can be used to build a word-context matrix, e.g. a word-document matrix. Such a semantic model can also be used to compute a word co-occurrence matrix if required, e.g., by considering that two words which occur in the same context co-occur.

- Contexts which are defined considering shorter text regions, e.g. composed of a window of few words, are generally not used to build word-context matrices. In this case the aim is generally to scan the documents by sliding the predefined window using a specific step, e.g. one word. At each step, the window will be processed by considering a focal word which is generally located at the center of the window, e.g. the word banana in the example presented above. Considering this focal word, the co-occurrences between the focal word and the other words composing the window are analysed. Let us note that, intuitively, the size of the window which is considered will define the sparseness of the co-occurrence matrix since the more a window is narrow the less two words will tend to co-occur.

A large variety of approaches and configurations can be adopted to define syntagmatic contexts. Concrete and simple examples of syntagmatic contexts and associated processing are provided in Appendix A. More refined models have also been proposed in the literature. As an example, some authors have proposed to consider directional models based on oriented windows. During the co-occurrence evaluation, these models also take into account the location (left or right) of the words which co-occur with the focal word under study. It has also been proposed to weight co-occurrences according to the distance of the words inside a window. In some cases, the focal word is not located at the center of the window and an asymmetric window is used to evaluate word co-occurrences.

Generally, window-based context extractors and more generally strategies based on syntagmatic context analysis benefit of low algorithmic complexity. They are therefore commonly considered as a solution of choice to process large corpora (Curran, 2004). They generally rely on several user defined parameters but have the advantage to be language independent – except specific pre-processing and particular configurations considering oriented windows.

Adopting another approach, syntagmatic relationships between words can also be studied defining specific lexical patterns. As an example, for studying co-occurrences between two nouns, particular types of lexical relationships can be analysed, e.g. considering two words $w_1$ and $w_2$, we can evaluate their co-occurrences by analysing the following patterns:

- Hyponymy, Hyperonymy: $\langle w_1,\text{is a}, w_2\rangle$, $\langle w_1,\text{such as}, w_2\rangle$

- Synonymy, Antonymy: $\langle w_1,\text{or},w_2\rangle$

- etc.

This patterns can be used to define a three dimensional co-occurrence matrix with dimensions (Word1, Word2, Pattern). Such an approach can be used to define semantic model particularly adapted to the design of parametric semantic measures. Indeed, using such model semantic designer can compare words by controlling the importance which is given to each lexical pattern and therefore finely control the semantics of the scores produced by the measure.

> *Additional references*: Examples of use of syntagmatic contexts in the definition of semantic measures: Schütze and Pedersen (1997); Yoshida et al. (2003).

**Paradigmatic contexts**

Paradigmatic contexts refer to indirect co-occurrences, that is to say situations in which two words occur with the same words but not together. This is often the case for words which establish paradigmatic relationships such as synonymy, hyperonymy, co-hyponymy, troponymy, antonymy or meronymy. Generally, paradigmatic relationships are characterised by analysing words which are surrounded by the same words. Therefore, such contexts are generally defined in terms of grammatical dependency between words. Otherwise stated, word occurrences w.r.t paradigmatic contexts can be used to assess the distributional similarity of words in terms of lexical substituability (Weeds, 2003)[8].

Most of the paradigmatic patterns are of the form $\langle w, \mathbf{t}, \mathbf{x} \rangle$ or $\langle \mathbf{x}, \mathbf{t}, w \rangle$, with $\mathbf{t}$ the dependency under study, $\mathbf{x}$ a word which is part of the pattern and $w$ the word the pattern is used to characterise. As an example, the pattern $\langle w, \text{SUBJ}, \text{GROW} \rangle$ can be defined in order to study words which are subjects of the verb *to grow*. Using such a pattern it is therefore possible to distinguish that the words *plant* and *tree* frequently occur with such a pattern, e.g. in the sentence "Soil provides a base which the roots hold on to as a <u>*plant* **grows**</u> bigger" the word *plant* occurs with this pattern; examples of sentences in which the pattern "*tree grows*" occurs are also numerous. Conversely, the word *botany* will never or significantly less frequently occurs with this specific pattern. This pattern which refers to a specific paradigmatic context can therefore be used to characterise that the words (*plant,tree*) together are probably more semantically related than the word *botany* and *tree* for instance. Paradigmatic contexts are also denoted *pair-pattern* by Turney and Pantel (2010), and refer to the notion of *extended distributional hypothesis* proposed by Lin and Pantel (2001).

In order to analyse words using paradigmatic contexts, a specific size of window and a direction is generally considered. In more advanced approaches specific patterns will be used to characterise slight variation in the contexts, e.g. lexical-syntactic patterns. The reader can refer to the work of Curran (2004) for examples of use of lexical-syntactic patterns. It shows how to characterise words based on complex context extractors, e.g. using Cass or Sextant parsers to mention a few. Panchenko (2013)[9] also presents several references related to the definition of pattern-based approaches and to the use of paradigmatic contexts. He also introduces an approach which can be used to consider multi-word expressions[10] in the definition of paradigmatic contexts.

> *Additional references*: Examples of use of paradigmatic contexts in the definition of semantic measures: Hirschman et al. (1975); Hindle (1990); Hatzivassiloglou and McKeown (1993); Grefenstette (1994); Takenobu et al. (1995); Lin (1998b,a); Dagan et al. (1999); Lee (2001); Weeds (2003); Van Der Plas and Bouma (2004); Heylen et al. (2008); Panchenko (2013).

According to Mohammad and Hirst (2012b), Lin (1998b) highlights that, in the context of paradigmatic context analysis, using more patterns tend to improve the results of semantic measures. In addition, interestingly, McCarthy et al. (2007) showed that syntagmatic contexts and more particularly simple sliding window approaches can be used to obtained results almost as good as those obtained using syntactic patterns. However, as we saw in this section, both syntagmatic and paradigmatic contexts are of particular importance for analysing relationships between words. They will both be used for defining semantic models on which will be based semantic measures. Indeed, as we will see in the next section, syntagmatic and paradigmatic relationships between words are the cornerstones of distributional semantics, a branch of study which have been extremely prolific on the definition of corpus-based semantic measures.

### 2.2.4 Distributional semantics

Distributional semantics is a branch of study which explores how statistical analysis of large corpora, and in particular word distributions and statistical regularities w.r.t linguistic contexts, can be used to model semantics (Lenci, 2008). Distributional semantics proposes a usage-based study of word meaning and extensively relies on one of the main tenets of computational linguistics and statistical semantics: the *distributional hypothesis* (Harris, 1954; Weaver, 1955; Firth, 1957; Sahlgren, 2008)[11].

---

[8] As stressed in Weeds' thesis, several authors only consider paradigmatic contexts to estimate semantic similarity. This is because the notion of semantic similarity is sometimes only evaluated w.r.t synonymy, i.e. what is the impact of substituting a word by another on the meaning of a specific sentence.

[9] Chapter 2.

[10] E.g. "*machine learning*".

The distributional hypothesis, and by extension distributional semantics, is based on the assumption that words occurring in the same contexts tend to be semantically close (Harris, 1981). This hypothesis is of major importance for the definition of corpus-based semantic measures. It was made popular through the idea of (Firth, 1957): *"You shall know a word by the company it keeps"*[12], and is indeed based on the assumption that:

1. The context associated to a word can be characterised by the words surrounding it, i.e. through the definition of syntagmatic and paradigmatic contexts – refers to syntagmatic and paradigmatic structural relationships and to the notion of context introduced in sections 2.2.2 and 2.2.3 respectively.

2. Words occurring in similar contexts, e.g., often surrounded by the same words, are likely to be semantically related; it is assumed that *"similar things are being said about both of them"* (Mohammad and Hirst, 2012b) – otherwise stated the association of their meaning is often used to refer to a specific topic which makes that the two words are *de facto* semantically related.

The relevance of the distributional hypothesis and distributional semantics has been stressed out of linguistics, several contributions in psychology and cognitive sciences use it for studying knowledge acquisition and memory to cite a few. Indeed, the distribution hypothesis has been linked to cognitive processes related to language acquisition, processing and understanding (McDonald and Ramscar, 2001). It has for instance been shown that statistical properties of language, and more particularly statistical relationships between neighbouring speech sounds, plays an important role in word segmentation in infants, and therefore speech acquisition (Saffran et al., 1996). Word distribution analysis has also been identified as an interesting evidence to derive syntactic categories (Redington et al., 1998).

Therefore distributional semantics became an approach of choice to model semantics of words w.r.t their usage contexts. This enthusiasm for distributional semantics relies on the statistical foundation of the approach which generally requires little expensive human supervision and therefore makes distributional semantics particularly adapted to large corpora analysis. In addition, this approach has been proved to be particularly interesting for solving NLP problems. Indeed, based on the distributional hypothesis, several distributional models have been proposed, e.g. topic models such as Latent Semantic Analysis (LSA), or Hypertext Analogue to Language (HAL) – they will be introduced later. As we will see, these distributional models are extensively and successfully used to analyse semantic relatedness of words – they have also proved to be particularly successful to perform a variety of NLP tasks and to design information retrieval systems[13].

Thus, distributional semantics provides a theoretical framework for assessing the semantic similarity or relatedness of words by means of distributional analysis, i.e. by considering an implementation of the distributional hypothesis. Otherwise stated, it is often implicitly considered that distributional similarity of words is equivalent to semantic similarity. However, note that in the study of distributional semantics, some authors distinguish distributional similarity of words to their semantic similarity. This is in particular the case when specific definitions of semantic similarity and distributional similarity are considered. As an example, in (Weeds, 2003, chap 1.), the author considers that (i) the notion of semantic similarity is defined in terms of *inter-substituability*, i.e. regarding the impact to substitute a word by another on the meaning of a sentence, and (ii) the distributional similarity of words is defined only considering the impact of substituting two words on the grammatical coherence of the sentence. Considering this specific definition of semantic similarity, the assumption that distributional similarity equates semantic similarity cannot be true – *"distributional similarity is* [in this case] *a weaker requirement than semantic similarity"* (Weeds, 2003). However, as we will see, in most cases, authors assume that distributional similarity can be considered as a valuable estimator of semantic relatedness.

In addition, even if highly popular and of major importance for computational linguistics, the place of the distributional hypothesis is often subject to debate in the linguistic community. Some researchers, advocates of a *strong* vision of the distributional hypothesis, consider that it can be used to fully characterise the meaning of language, while others, i.e. *weak* distributional hypothesis advocates, only see this hypothesis as a way to study linguistic meaning (Lenci, 2008). Despite the central importance of this debate for studying methods which can be used to manipulate the meaning of units of language through distributional models, we here consider that this debate is out of the scope of this book. In addition, from a practical point of view, distributional semantics and the application of the distributional hypothesis have been proved to be successful in the definition of corpus-based semantic measures – this will be illustrated in the next section which introduces distributional measures for comparing words.

---

[11]The term has been introduced in (McDonald and Ramscar, 2001).

[12]Also implicitly discussed in (Weaver, 1955) originally written in 1949 (source: wiki of the Association for Computational Linguistics http://aclweb.org/aclwiki accessed 09/13).

[13]They have also been proved to be of particular interest in other related domains, e.g. for psychologists interested in semantic representation building and processing for cognitive tasks (Kamp et al., 2014).

In some cases, authors consider the distributional hypothesis as a component of a more general framework denoted the *statistical semantics hypothesis* (Turney and Pantel, 2010). This general hypothesis has been proposed to defend Vector Space Models, a very popular class of models of words and texts that will be presented in this chapter. It states that meaning of texts can be studied by analysing statistical patterns of human word usage. The hypothesis is therefore related to the strong vision of the distributional hypothesis that has been aforementioned. Links with the statistical semantics hypothesis defined by Turney and Pantel (2010) will be provided in the following presentation.

## 2.3 Distributional measures

Distributional measures are the corpus-based semantic measures which have been the more studied in the literature (Weeds, 2003; Curran, 2004; Sahlgren, 2008; Dinu, 2011; Mohammad and Hirst, 2012a; Panchenko, 2013). They rely on the distributional hypothesis introduced in Section 2.2.4 and are therefore based on the assumption that words which occur in the same contexts are semantically related. These measures rely on a *distributional semantic model* which corresponds to a semantic space built from the distributional analysis of corpora (Baroni and Lenci, 2010)[14]. This semantic model, here denoted *distributional model*, is used to manipulate the semantics of specific units of languages, e.g. words, documents. Based on this semantic model, specific canonical representations of words, denoted *distributional profiles* of words, can be built according to the contexts in which they occur[15]. As we will see, an important part of the large literature related to distributional measures is dedicated to the various approaches which have been proposed and tested to instantiate distributional model based on various context definitions. Thus, depending on the distributional model which is selected and depending on the definition of the distributional profile which is used to represent a word, a large diversity of distributional measures have been proposed and evaluated.

### 2.3.1 Implementation of the distributional hypothesis

Studies of distributional measures are tightly related to spatial representations of both the semantic space which characterises a corpus and the words to compare – in the vain of the spatial models proposed in Information Retrieval, e.g. Vector Space Models (VSM) and topic models, and according to the spatial model of similarity which has been widely studied in cognitive sciences (refers to Section 1.2.1). Distributional models are generally defined to capture the meaning of words through distributional profiles of words build from context analysis. These profiles represent words into a multi-dimensional space defined by the distributional model under study. Using this strategy, words are therefore considered as specific points of a highly multi-dimensional space. The various dimensions which characterise a word generally refer to the various words of the vocabulary or to the notion of paradigmatic and syntagmatic contexts which have been introduced in Section 2.2.3[16]

The various distributional measures are therefore mainly distinguished by (i) the strategy which is used to build the distributional model and (ii) the measure which is used to process the distributional profile of words extracted from the model. This section introduces the different strategies which have been proposed to build a distributional model. Next, the approaches which have been proposed to represent words based on these semantic models, and the measures which can be used to compare such word representations will be presented.

The construction of the distributional model mainly depends on the steps presented below, some of them have already been discussed:

---

[14]As stressed by the authors, such semantic models based on distributional semantics have been denoted through a variety of names in the literature, e.g. vector spaces, semantic spaces, word spaces, corpus-based semantic models.

[15]A distributional profile is also denoted a contextual vector in the literature.

[16]Considering these two types of contexts corresponds to the *distributional hypothesis* and the *extended distributional hypothesis* in the general *statistical semantics hypothesis* defined by Turney and Pantel (2010).

- Pre-processing techniques of the corpus (optional), e.g., stop filter, part-of-speech filter. As we have seen, in some cases, this step defines the vocabulary which will be considered and therefore the pairs of words for which a semantic relatedness can be assessed.

- Type of context used to build the distributional model. Depending if syntagmatic or paradigmatic contexts are used, the context which is used to characterise a word may be a document, a paragraph, a sentence, a potentially oriented word windows, a number of letters, etc. The various contexts which can be used to characterise a word by analysing syntagmatic and paradigmatic relationships of words have been introduced in Section 2.2.3.

- Frequency weighting (optional). The function used to transform the raw counts associated to each context in order to incorporate frequency and additional knowledge relative to the informativeness of contexts – this step is used to reduce the impact of frequent words or more generally contexts. A large number of approaches have been proposed in the literature. The most popular approach to weight the importance of words in texts is TF-IDF (Salton and McGill, 1983):

$$tfidf_{i,d} = tf_{i,d} \cdot idf_i \qquad (2.1)$$

  with $tf_{i,d}$ the frequency of the term $t_i$ in a document $d$ (more exactly the number of occurrences of $t_i$ in $d$) and $idf_i$ the inverse document frequency of $t_i$ is defined by:

$$idf_i = log\frac{|D|}{|D_{t_i}|} \qquad (2.2)$$

  with $|D|$ the set of documents of the corpus and $D_{t_i} \subseteq D = \{d_j \in D \mid t_i \in d_j\}$ the set of documents in which the term $t_i$ occurs. The rationale is to consider a word important for a document if it (i) frequently occurs in it (high $tf$) and (ii) is rarely found in the documents of the corpus (high $idf$). Numerous alternative expressions modelling this intuitive idea have been proposed.

  Association measures such as Pointwise Mutual Information (PMI) can also be used to refine raw co-occurrences[17]. More complex weighting schemes commonly used in information retrieval can also be adopted, e.g. *Lnu.tc* (Bellot et al., 2014).

- Dimension reduction technique (optional) used to reduce the distributional model represented as a matrix, generally a word-context matrix. This process is used to reduce the sparseness of the matrix, it can be done by removing the most frequent contexts, i.e. low entropy columns. Matrix factorisation techniques can also be used to reduce the number of dimensions (and therefore the initial amount of information), e.g. using Singular Value Decomposition (SVD) (Berry et al., 1995; Golub and Van Loan, 2012) – alternative approaches for matrix factorization can also be applied, e.g. Principal Component Analysis, Independent Component Analysis to cite a few (Turney and Pantel, 2010). In some cases such treatments have the interesting properties of removing noise by finding the most important axes of variation. By forcing correspondences between words and context in the reduction process this also enables the discovering of latent dimensions. As we will discuss in Section 2.3.3, reduction techniques are also used to highlight high order co-occurrences, i.e. indirect co-occurrences between words. Appendix B introduces the intuitions, interesting properties and limits of SVD.

A large number of distributional models have been proposed and studied in the literature, some of them will be introduced in this chapter (e.g. Latent Semantic Analysis, Explicit Semantic Analysis, Hyperspace Analogue to Language). We invite interested readers to refer to the original contribution for details, and to start by studying contributions related to vector space model, e.g. we recommend the detailed survey provided by Turney and Pantel (2010) as a general introduction to distributional models. Reisinger and Mooney (2010) also propose how to define a VSM that enables context-dependent vector representations of words that tries to solve the problem of lexical ambiguity of words. We also encourage the reader to study the contributions related to the definition of general frameworks dedicated to distributional models unification (Baroni and Lenci, 2010). Finally, since this book focuses on semantic measures, and cannot present the whole literature related to models of units of language, this chapter will only introduce some of the models that have been proposed so far. Other (recent and more complex) models that are of interest for semantic measure design are also briefly presented in Appendix C. We discuss in particular, how language models can be used to derive representations of words and we introduce the reader to compositionality, i.e. how to combine models to build representations of complex units of language such as sentences. We advise the reader to consult this appendix only when traditional distributional models presented herein are fully understood.

---

[17] This approach will be introduced in Equation 2.7 page 34.

Distributional measures differ regarding the type of distributional models they use. They also differ considering the approach used to assess the similarity/distance of words that are characterised w.r.t the distributional model – recall that words are represented by distributional profiles build from the analysis of distributional models. In some cases words will be considered as vectors or (fuzzy) sets to cite a few. In other cases, the matrix will be used to extract statistics on which will be based the measures. Several approaches will be presented in the next section – according to the literature, most of them are particular instantiations of the spatial model of similarity through distributional semantics.

However, an important aspect to understand is that distributional models, e.g. word co-occurrence matrices, are not necessarily tight to the geometrical model. Distributional semantics and implementations of the distributional hypothesis must therefore not only be regarded as a means to obtain a geometrical representation of words through vector representations – even if some authors do not distinguish distributional semantics from the geometrical representation of words which can be made from distributional models (Kamp et al., 2014). Indeed, as we will see, first of all, distributional models contain statistics which are valuable in the aim of assessing the similarity or relatedness of words. However, as we will see, these measures can be defined independently from the geometrical/spatial model commonly considered in the design of distributional measures and can for instance exclusively rely on information theoretical notions.

### 2.3.2 From distributional model to word similarity

We have introduced how, by implementing the distributional hypothesis, distributional models can be used to characterise word by analysing their usage contexts. Based on these models, several distributional measures have been proposed. They differ regarding the conceptual approach on which they are based and on the type of distributional models they have been designed for. Indeed, as we have seen, distributional models can be of various forms, e.g. word-context matrix, word-word matrix or even sometimes three dimensional word-word-context matrices[18]. This section briefly introduces three main types of approaches which have been proposed to assess word similarity from distributional models:

- The geometric/spatial approach which evaluates the relative positions of two words in the semantic space defined by contexts vectors.

- The set-based approach which rely on the analysis of the overlap of the set of contexts in which the words occur.

- The probabilistic approach which is based on probabilistic models and measures proposed by information theory.

Due to space constraint, we will only consider specific examples of measures for each category. For detailed information the reader may refer to the presentations provided in original references. Technical details and several measures have also been introduced by Weeds (2003); Curran (2004); Sahlgren (2006); Mohammad and Hirst (2012b); Panchenko (2013).

**The geometric or spatial approach**

The geometric approach is based on the assumption that compared elements are defined in a semantic space corresponding to the intuitive spatial model of similarity proposed by cognitive sciences (see Section 1.2.1). A word is considered as a point in a multi-dimensional space representing the diversity of the vocabulary in use or more generally the various contexts which are used to characterise a word. Two words are therefore compared regarding their location in this multi-dimensional space. The dimensions which are considered to represent the semantic space are defined by the contexts used to build the distributional model. Words are represented through their corresponding vector in the matrix and are therefore compared through measures used to compare vectors. This approach as already been illustrated when introducing contexts in Section 2.2.3.

---

[18]Refer to the complex syntagmatic context analysis based on lexical patterns which is introduced page 26.

We define **u** and **v** the vector representations of the words $u$ and $v$, with $n$ the size of the vectors, and $\mathbf{u}_k$ the value of **u** in dimension $k$. Among the measures commonly used for comparing vectors, we distinguish:

- Minkowski $L_p$ distance metric:

$$dist_{L_P}(u,v) = \left(\sum_{k=1}^{n} |\mathbf{u}_k - \mathbf{v}_k|^p\right)^{\frac{1}{p}} \qquad (2.3)$$

$$sim_{L_P}(u,v) = \frac{1}{dist_{L_P}(u,v) + 1}$$

    Instances of this measure are $L_1$ Manhattan distance ($p=1$) and the $L_2$ Euclidian distance ($p=2$).

- Cosine similarity, the cosine of the angle between the vectors – the similarity is inversely proportional to the angle. The cosine similarity between $u$ and $v$ is:

$$sim_{cos}(u,v) = \frac{\sum_{k=1}^{n} \mathbf{u}_k \mathbf{v}_k}{\sqrt{\sum_{k=1}^{n} \mathbf{u}_k^2}\sqrt{\sum_{k=1}^{n} \mathbf{v}_k^2}} \qquad (2.4)$$

- Measures of correlation can also be used in some cases (Schütze, 1998; Ganesan et al., 2003). For instance, the similarity of two words can be defined as the coefficient of the Pearson's product-moment correlation between their vector representations (the cosine similarity of re-centered vectors):

$$sim_{Pearson}(u,v) = \frac{\sum_{k=1}^{n}(\mathbf{u}_k - \overline{\mathbf{u}})(\mathbf{v}_k - \overline{\mathbf{v}})}{\sqrt{\sum_{k=1}^{n}(\mathbf{u}_k - \overline{\mathbf{u}})^2}\sqrt{\sum_{k=1}^{n}(\mathbf{v}_k - \overline{\mathbf{v}})^2}} \qquad (2.5)$$

Numerous approaches rely on the cosine similarity which, contrary to the $L_p$ family also incorporates information about contexts in which the words do not co-occur. However, as stressed by Lee (1999), the cosine similarity appears to be less effective in some experiments based on paradigmatic context analysis (Weeds, 2003). Other approaches for comparing vectors are introduced by Mohammad and Hirst (2012b), the reader can also refer to the collection of measures proposed by Deza and Deza (2013).

Numerous approaches based on multidimensional representation of words exist. The most popular approaches based on the spatial model are briefly presented. For each of them we present: (i) the approach used to build the distributional model, (ii) the strategy considered to extract distributional profiles of words and (iii) the measure used to compare these profiles:

- LSA – Latent Semantic Analysis (Deerwester et al., 1990; Landauer and Dumais, 1997; Landauer et al., 1998), also called Latent Semantic Indexing (LSI) in Information Retrieval. This approach is used to represent a word-context matrix, generally a word-document matrix, which can be used to extract a distributional profile of a word. Two word profiles can next be compared using the measures which have been presented above – the cosine similarity measure was used initially (Equation 2.4). In this approach, the sparseness of the matrix is reduced using Singular Value Decomposition which is a linear algebra operation used to reduce the number of contexts considered in the matrix. This operation is used to obtain a low-rank approximation of the matrix and therefore obtain a more compact representation of the word-context matrix which is used to characterise a word. Such a treatment has also the interesting property to put into light latent concepts by highlighting indirect co-occurrences between words – i.e. words occurring within a similar context but not necessarily occurring together.

- ESA – Explicit Semantic Analysis (Gabrilovich and Markovitch, 2007). Compared to LSA which uses latent concepts, ESA is based on explicit description of concepts. The approach relies on the analysis of corpora which are composed of texts describing specific concepts. The approach has originally been developed to be used with Wikipedia assuming that an article refers to a specific concept. Then, a word-concept (i.e. word-article) matrix is built in order to characterise a word. Each cell of the matrix refers to the strength of association between the pairs word/concept – TF-IDF is used (Salton and McGill, 1983). Words are next compared using the cosine similarity of the vector representations of words.

- HAL – Hyperspace Analogue to Language (Lund and Burgess, 1996). This approach is used to generate a word co-occurrence matrix. Co-occurrences are analysed considering syntagmatic context and a ten word sliding window is used by considering a specific model for counting co-occurrences, i.e. the distance between words in the window are taken into account to weight co-occurrences. In addition, each word is characterised by a vector of twice the size of the vocabulary since words co-occurrence counts are oriented, i.e. two word co-occurrence matrices are computed, one for each side. The similarity of word representations was originally computed using the Minkowski distance, Equation 2.3.

- Schutze wordspace (Schütze, 1993). In this approach n-gram co-occurrences are analysed to further build word profiles based on it. In the original approach 5000 frequent four-grams were considered to build a four-gram co-occurrence matrix. A four-gram $n_1$ was considered to co-occur with another four-gram $n_2$ if $n_1$ was located at least at 200 four-grams to the left of $n_2$. The matrix is also reduced using SVD. Next, based on this matrix, the vector representations of a word is built summing the vectors which represent each of the 500 n-gram which occurs to the left and right of each occurrence of the word under study (vector representation of words are also normalised). Finally, the similarity between word vectors are computed using the normalised correlation coefficient, based on Equation 2.5.

- Random indexing (Kanerva et al., 2000). This approach is based on a two step process. To each context used to characterise a word, e.g. a document or a paragraph, is associated an index vector which corresponds to a randomly generated vector. It will be used to build the vector representation of the words. Each time the word is encountered within a context, e.g. paragraph, the vector associated to the corresponding context will be added to the vector representation of the word. This approach can be used to reduce the high-dimensionality of the sparse word-context matrices produces by other topic model approaches, such as LSA, HAL. It can also be used to avoid the computing expensive reduction of the matrix (which must be computed each time the model is updated). The various measures which can be used to compare vector representation of words can be used with this space model. Nevertheless, note that this model to do not generate a statistically meaningful distributional model composed of specific statistics such as co-occurrences.

- COALS – Correlated Occurrence Analogue to Lexical Semantic (Rohde et al., 2006). This approach uses a four word (weighted) sliding window to compute a word-word co-occurrence matrix. A reduction of the number of columns can be performed in order to focus on the most common words, e.g. considering a frequency threshold. A correlation normalisation is next applied to each cell of the matrix by computing the Pearson correlation between the pair of words which is associated to each cell (Equation 2.5). Negative values are discarded and positive values are square rooted. Reduction of the matrix based on SVD can also be applied. Finally, the cosine similarity (Equation 2.4) is used to compute the similarity of word vectors.

We have presented several distributional models from which semantic measures based on the spatial model can be defined. Nevertheless, nothing prevent the use of the statistics provided by most of these distributional model for defining other types of semantic measures.

**The set-based approach**

Words are compared regarding the number of contexts in which they occur which are common and different (Curran, 2004). The comparison can be made using classical set-based measures (e.g., Dice index, Jaccard coefficient). The reader may refer to the several set-based operators which have for instance been used to compare words in (Bollegala, 2007b; Terra and Clarke, 2003). As an example to compare the words $u$ and $v$, with $\mathbf{C}(u)$ the sets of contexts in which the word $u$ occurs, the similarity can be defined using Dice index:

$$sim_{Dice}(u,v) = \frac{2|\mathbf{C}(u) \cap \mathbf{C}(v)|}{|\mathbf{C}(u)| + |\mathbf{C}(v)|} \qquad (2.6)$$

Extensions have also been proposed in order to take into account a weighting scheme through fuzzy sets, e.g. (Grefenstette, 1994). Set-based measures relying on information theory metrics have also been proposed, they are introduced in the following subsection which presents the measures based on probabilistic approaches.

**The probabilistic approach**

The distributional hypothesis enables to express the semantic relatedness of words in term of probability of co-occurrence, i.e. regarding both, the contexts in which compared words appear together and alone. These two evidences can intuitively be used to estimate the strength of association between two words. This strength of association can also be seen as the mutual information of two words which can be expressed regarding the probability the two words occur in the corpus, as well as the probability the two words co-occur in the same context. Once again a large diversity of measures have been proposed in the literature. Only those which are frequently used are presented (Dagan et al., 1999; Mohammad and Hirst, 2012b).

With $p(u)$ the probability that the word $u$ occurs in a context and $p(u,v)$ the probability that the two words $u$ and $v$ occurs in the same context – the notion of context may refer to windows of different sizes or event larger units of language depending on the implementation of the notion of context which is considered (Section 2.2.3). For convenience following formulae are introduced without defining a specific implementation of the notion of context.

- Pointwise Mutual Information (PMI) (Fano, 1961).

$$pmi(u,v) = \log \frac{p(u,v)}{p(u)p(v)} \quad (2.7)$$

The PMI was first adapted for the comparison of words by (Church and Hanks, 1990). It is based on the analysis of the number of co-occurrences and individual occurrences of words (marginal frequencies), e.g., in paragraphs or sentences – examples of use are discussed in (Turney, 2001; Lemaire and Denhière, 2008; Mohammad and Hirst, 2012b). On of the limitation of PMI is its biais towards rare words (Manning and Schütze, 1999): considering that $u$ and $v$ are two words that always co-occur, we will have $p(u,v) = p(u)$, and $pmi(u,v) = log(1/p(u))$. Thus, when $u$ is rare, i.e. $p(u)$ is low, the value of $pmi(u,v)$ will be high.

To overcome the fact that PMI is biased towards infrequent words, various adaptations and correction factors have been proposed (Pantel and Lin, 2002; Mohammad and Hirst, 2012b). For instance, the Normalised PMI (NPMI) is defined as follows:

$$npmi(u,v) = \frac{pmi(u,v)}{-log\ p(u,v)} \quad (2.8)$$

Other adaptations have also been proposed, e.g. (Han et al., 2013) propose $PMI_{max}$, an adaptation of PMI in order to take into account multiple senses of words. Note that (N)PMI functions have also been often used to refine word-context matrices (distributional model) by incorporating information on word association. Another variation of PMI widely used in NLP is the Positive PMI (PPMI) which forces to only consider positive PMI values (Niwa and Nitta, 1994). As an example, Bullinaria and Levy (2007) showed that PPMI outperforms numerous weighting approaches for semantic relatedness computations based on context-co-occurrence analyses. PPMI is defined by :

$$ppmi(u,v) = \max(0, pmi(u,v)) \quad (2.9)$$

- Confusion probability and Maximum likelihood Estimate (MLE) can also be used – refer to the work of Dagan et al. (1999) for both details and examples.

Vector representation of words obtained from most distributional models can also be seen as distribution functions corresponding to distribution profiles[19]. Therefore, the several approaches which have been proposed by information theory to compare probability mass functions[20] can also be used to compare vector representations of words. Note that measures which consider word representations as probability mass functions imply that the vector representation of words are normalised such as the sum of the vector equals one. The functions commonly used in the probabilistic approach are the followings:

---

[19]i.e. word representation. Distributional profiles are introduced in Section 2.3. Notice that these vectors can also be vectors of strength of association if one of the metrics presented above (e.g. PMI) has been used to convert the initial word-context matrix. As an example, considering an initial word-word co-occurrence matrix, this can be done by substituting the value of each cell of the matrix by the corresponding PMI or NPMI values.

[20]The reader may refer to the work of Cover and Thomas (2006) for an introduction to the field of Information Theory.

- Kullback-Leibler divergence (information gain or relative entropy) is a classic measure used to compare two probability distributions. It is often characterised as the loss of information when a probability distribution is approximated by another. The distance between two words $p$ and $q$ can thus be estimated by the relative entropy between their respective distributions $\mathbf{p}$ and $\mathbf{q}$.

$$dist_{KL}(p,q) = \sum_{k=1}^{n} \mathbf{p}_k log \frac{\mathbf{p}_k}{\mathbf{q}_k} \qquad (2.10)$$

  This measure is positive, asymmetric and ensures that $dist_{KL}(p,p) = 0$. However, this distance does not satisfies the triangle inequality. It can also be applied on conditional probabilities (Dagan et al., 1999). Please refer to the details discussed in the contribution of Cover and Thomas (2006) for more information.

- Jensen-Shannon divergence. This function also measures the distance between two probability distributions.

$$dist_{JS}(p,q) = \frac{1}{2} d_{KL}(\mathbf{p},\mathbf{m}) + \frac{1}{2} d_{KL}(\mathbf{q},\mathbf{m}) \qquad (2.11)$$

  with $\mathbf{m} = \frac{\mathbf{p}+\mathbf{q}}{2}$. This measure is based on the Kullback-Leibler divergence with the interesting properties of being symmetric and to be bounded between 0 and 1 – it has also been demonstrated that in specific cases this measure is an approximation of the $\chi^2$ test (Cover and Thomas, 2006).

- $\chi^2$ distance – several variants exists in the literature.

$$dist_{\chi^2}(p,q) = \frac{1}{2} \sum_{i} \frac{(\mathbf{p}_i - \mathbf{q}_i)^2}{\mathbf{p}_i + \mathbf{q}_i} \qquad (2.12)$$

- Other measures such as Skew Divergence or Kendall's $\tau$ (Lee, 1999; Curran, 2004) and the measures presented in Section 2.3.2 for comparing vector representations of words can also be used.

An excerpt of the similarity functions which can be used to compare probability distributions can be found in the work of Pantel and Lin (2002); a comprehensive survey presenting a large collection of measures is also proposed by Cha (2007)[21].

Several combinations of measures can therefore be used to mix both the strength of association (weighting scheme, e.g., PMI) and the measures which can be used to compare the probability functions/vectors. Alternatively, fuzzy metrics can also be considered for comparing words according to their strength of association – the reader may refer to the contribution of Mohammad and Hirst (2012b) for detailed examples.

The probabilistic approach does not only refers to the different measures based on information theory. Several statistical techniques have also been proposed to define distributional models (or topic models); two of them are briefly presented:

- Probabilistic Latent Semantics Analysis (PLSA) is a statistical technique based on mixture decomposition which are derived from latent class model (Hofmann, 1999). Its aim is to propose an approach with a solid foundation in statistics in order to respond to some limitations associated to LSA (e.g. overfitting, word and documents are assumed to be joint by a Gaussian model). The latent variables which are considered in PLSA correspond to topics. The probabilistic model relies on two conditional probabilities: the probability that a word is associated to a given topic and the probability that a document refers to a topic – refer to the work of Cohen and Widdows (2009) for details.

- LDA – Latent Dirichlet Allocation (Blei et al., 2003). This topic model is similar to PLSA but adopt a different approach w.r.t the assumption on the topic distribution in document. The interested reader will refer to the original publication and to the multiple adaptations proposed for defining semantic measures, e.g. (Dinu and Lapata, 2010).

---

[21] An interesting correlation analysis between measures is also provided.

### 2.3.3 Capturing deeper co-occurrences

Most of the measures which have been presented so far are based on distributional models build considering syntagmatic contexts. They can only be used to estimate the similarity of words regarding their first order co-occurrences, i.e., the similarity is mainly assessed by studying the contexts in which the words occur together. However, a strong limitation of first order co-occurrence studies (which only rely on simple syntagmatic context, i.e. those not based on syntagmatic patterns) is that words which do not co-occur in the same context will have a low similarity. However, in some cases, similar or related words never co-occur in the same syntagmatic contexts - in particular if short contexts are considered. As an example, even if syntactic context analyses perform well for synonymy detection in general (Turney, 2001), specific synonyms may not co-occurs syntagmatically. As an example, some studies of large corpus have observed that the words *road* and *street* almost never co-occurs in the same word window, although they can be considered as synonyms in most cases (Lemaire and Denhière, 2008). As we have seen, this specific aspect of word similarity can be studied by analysing paradigmatic contexts, as these words will tend to occur in similar paradigmatic contexts. Approaches have also been proposed to capture this type of word similarity by processing results of syntagmatic context analysis.

Indeed, focusing on syntagmatic contexts, specific techniques have also been proposed to highlight deeper relationships between words, e.g., second order co-occurrences, i.e. co-occurrences which correspond to paradigmatic relationships. These techniques will transform the word-context matrix to enable evidence of deeper co-occurrence to be captured. To this end matrix factorisation techniques can be used (e.g. SVD is presented in Appendix B).

Statistical analysis can be used to distinguish valuable patterns in order to highlight deeper co-occurrences between words. These patterns, which represent the relationships between words, can be identified using several techniques; among them we distinguish:

- LSA (Dumais et al., 1988), HLA (Lund and Burgess, 1996), PLSA (Hofmann, 1999), LDA (Blei et al., 2003). These approaches have been introduced in Section 2.3.2.

- SOC-PMI Second Order Co-occurence PMI (Islam and Inkpen, 2006). This approach can be used to consider related words of the compared words when computing their PMI. Related words were originally obtained using a Web-based search engine; other approaches can also be adopted.

- Syntax or dependency-based model, i.e. models which are based on paradigmatic context analysis – refers to Section 2.2.3.

We recall that Appendix C presents other models that can be used to derive word representations.

## 2.4 Other corpus-based measures

The large majority of corpus-based measures are distributional measures and therefore extensively rely on the distributional hypothesis. However, other approaches have been proposed to assess the semantic similarity or relatedness of words based on corpora analysis. As an example, several authors have proposed to compare words w.r.t results returned by an information retrieval system, e.g. using results provided by Web search engines considering the words to compare (Chen et al., 2006; Sahami and Heilman, 2006; Bollegala, 2007b; Cilibrasi and Vitanyi, 2007; Gracia and Mena, 2008). In this case, even if many information retrieval systems are based on an implementation of the distributional hypothesis, these measures are generally not considered as distributional measures *per se*.

Other approaches are based on specific treatments performed on semi-structured texts. In Lesk (1986) the author proposes to compare words based on the number of words' overlap of their descriptions in dictionaries. In other cases, measures also take into consideration more or less structural information, e.g. information expressed into WordNet, Probase (Wu et al., 2012) or other semantic networks (Nitta, 1988; Kozima and Furugori, 1993; Niwa and Nitta, 1994; Kozima and Ito, 1997; Blondel, 2002; Ho and Cédrick, 2004; Iosif and Potamianos, 2012; Zhila et al., 2013). These approaches are based on concepts which will be introduced in the section dedicated to knowledge-based semantic measures. As an example, (Jarmasz and Szpakowicz, 2003b,a) propose a measure which uses *Roget's Thesaurus* to compute the similarity of words. Words are regarded as if they were structured by a taxonomy and the similarity of two words is defined as a function of the depth of the most specific common ancestor of the two words. In other cases a graph representation of a corpus of texts will be build and used as a semantic proxy. For instance, Muller et al. (2006) analyse a dictionary to generate a graph in which nodes are entries in the dictionary or words which occurs in definitions. An edge is created between an entry and each word in its definition. The similarity between words is next evaluated using a random-walk approach. Numerous measures mixing knowledge-based and corpus-based approaches will be introduced in Section 3.8.

## 2.5 Advantages and limits of corpus-based measures

Here, we list some of the advantages and limits of corpus-based semantic measures and in particular those of distributional measures.

### 2.5.1 Advantages of corpus-based measures

- They are unsupervised and can be used to compare the relatedness of words expressed in corpora without prior knowledge regarding their meaning or usage – in comparison to knowledge-based measures.

- They enable fine-grained analysis of the semantics of measures as tuning of measures can be done by considering syntagmatic and paradigmatic relationships. As we saw, specific measures can also be used to target specific relationships between words, e.g. antonymy.

- They can be used to compare numerous units of language (from words to texts).

### 2.5.2 Limits of corpus-based measures

- The words to compare must occur at least few times.

- They highly depend on the corpus which is used. This specific point can also be considered as an advantage as the measure is context-dependent. Sense-tagged corpora are most of the time not available (Resnik, 1999; Sánchez et al., 2011). The construction of a representative corpus of text can be challenging in some usage contexts, e.g., biomedical studies.

- It is difficult to estimate the relatedness between concepts or instances due to the disambiguation process which is required prior to the comparison. Distributional measures are mainly designed for the comparison of words. However, some pre-processing and disambiguation techniques can be used to enable concepts or instances comparison from text analysis. Interesting words for the consideration of lexical ambiguity in distributional models have also been proposed, e.g. (Reisinger and Mooney, 2010) propose a VSM that enables context-dependent vector representations of words to be build. Nevertheless, the computational complexity of approaches based on disambiguation is most of the time a drawback making such approaches impracticable with large corpora analysis. The comparison of multi-word expressions can also not be performed using most approaches.

- It is difficult to estimate the semantic similarity between two words using these measures. Even if different observations are provided in the literature, it is commonly considered that distributional measures can only be used to capture semantic relatedness. This is due to the large amount of relationships that can be associated to a pair of co-occurring words (Chaffin and Herrmann, 1984; Zhila et al., 2013). And to the fact that co-occurrence is generally, for the most part, only considered as an evidence of relatedness, e.g., (Batet, 2011). Additional considerations have to be made to capture specific relationships. As an example, Mohammad and Hirst (2012b) specifies that similarity can be captured performing specific pre-processing or using specific paradigmatic contexts – Mohammad et al. (2008) show for instance that semantic measures can be used to detect antonyms. In the same vein, Ferret (2010) proposes an interesting study on the use of corpus-based semantic measures for synonymy extraction[22]. The reader may also refer to the recent work of Yih et al. (2012) in which the authors propose a modification of LSA to assess the degree of synonymy and antonymy between words.

- It is generally difficult to explain and to trace the semantics of the relatedness in some cases, e.g. when the approach is based on a general implementation of the distributional hypothesis it is difficult to deeply understand the semantics associated to the co-occurrences and therefore to the scores produced by the measure.

---

[22]As we will see in Chapter 4 which is dedicated to semantic measure evaluation, synonymy detection is frequently used to compare and to analyse measures.

> Software solutions and source code libraries that provide corpus-based semantic measures implementations are presented in Appendix D. Chapter 4 also provides information related to the evaluation protocoles and datasets that can be used to compare these measures.

## 2.6 Conclusion

Due to their central importance for applications based on natural language analysis, corpus-based semantic measures have attracted a lot of interest in the last decades. This has led to the proposal of a large variety of measures which can be used to compare different types of units of language – from words to documents. As an introduction to corpus-based semantic measures and to illustrate the type of semantic evidence which can be extracted from natural language analysis in order to compare units of language, this chapter has focused on presenting measures for comparing words. This type of measures is cornerstone of corpus-based semantic measures. Indeed, these measures are based on important notions and models which are used to design measures for comparing larger lexical units.

As we saw, most of the measures which have been proposed in the literature are based on distributional semantics and rely on distributional semantics through the definition or extension of topic models, e.g., LSA, HAL, LDA. These measures are based on syntagmatic and paradigmatic analyses of language, i.e. according to the distributional hypothesis, words are compared w.r.t the contexts in which they occur. An important aspect of this approach is therefore that its roots are based on a quantitative analysis of word occurrences w.r.t. the central notion of context detailed in this chapter.

We have also introduced alternative approaches to pure-distributional measures. They take advantage of external algorithms which are used to distinguish relevant resources considering specific words. Words will therefore be compared by analysing the set of resources which are associated to them. It's worth to note that most of the time these resources are retrieved based on algorithms which also take into account the distributional hypothesis, sometimes implicitly, e.g. approaches based on information retrieval algorithms. This challenges the ability to differentiate corpus-based measures defined as purely distributional, i.e. distributional measures, and other measures.

An interesting and important aspect of corpus-based measures is therefore the large diversity of lexical units they can compare and the diversity of approaches and adaptations defined in the literature. In this context, important research efforts focus on defining methods for adapting corpus-based measures dedicated to the comparison of words, and more particularly distributional measures, in order to compare larger lexical units such as sentences, paragraphs or documents. To this end, a central aspect to study is semantic compositionality w.r.t semantic similarity: how to evaluate larger units of language than words using approaches commonly used to define corpus-based measures, e.g., using the distributional hypothesis and by adapting distributional models (Kamp et al., 2014). Challenges related to this field of study also aim at incorporating the consideration of linguistic expressions, e.g. anaphora or even negation (to cite a few) into existing models. It is indeed commonly considered that current models and in particular distributional semantics provide quantitative evidence of semantic similarity and therefore only give access to a rough approximation of the meaning which is conveyed by natural language (Baroni and Lenci, 2010). This highlights the narrow link which exists between the study of semantic measures and the study of computable models of natural language such as distributional models.

Another interesting and important topic refers to the unification of corpus-based measures in order to define models which are generic enough to be used for comparing lexical units of different granularities, i.e. in order that the same approach/measure can be used to define corpus-based semantic measures for comparing words or paragraphs (Pilehvar et al., 2013).

# Chapter 3

# Knowledge-based semantic measures

As we have seen, two main families of semantic measures can be distinguished: corpus-based measures, which take advantage of unstructured or semi-structured texts, and knowledge-based measures which rely on ontologies.

Corpus-based measures are essential for comparing units of languages such as words, or even concepts when there is no formal expression of knowledge available to drive the comparison. On the contrary, knowledge-based semantic measures rely on more or less formal expressions of knowledge explicitly defining how the compared entities, i.e. concepts or instances, must be understood. Thus, they are not constrained to the comparison of units of language and can be used to drive the comparison of any formally described pieces of knowledge, which encompasses a large diversity of elements, e.g., concepts, genes, person, music bands, etc.

This chapter focuses on knowledge-based measures and we will more particularly introduce measures which rely on ontologies processed as semantic graphs or semantic networks. These measures are generally used to compare terms structured through unambiguous semantic relationships or concepts defined in taxonomies and knowledge organisation systems. It also encompasses measures commonly used to compare terms or senses defined into lexical databases such as WordNet (Miller, 1998; Fellbaum, 2010).

The main limitation of knowledge-based measures is their strong dependence on the availability of an ontology – an expression of knowledge which can be difficult to obtain and may therefore not be available for all fields of studies. However, in recent decades, we have observed, both in numerous scientific communities and industrial fields, the growing adoption of knowledge-enhanced approaches based on ontologies. As an example the Open Biological and Biomedical Ontology (OBO) foundry gives access to hundreds of ontologies related to biology and biomedicine. Moreover, thanks to the large efforts made to standardise the technology stack which can be used to define and to take advantage of ontologies (e.g., RDF(S), OWL, SPARQL – triple stores implementations) and thanks to the increasing adoption of the Linked Data and Semantic Web paradigms, a large number of initiatives give access to ontologies in numerous domains (e.g., biology, geography, cooking, sports).

In the introduction, we also point out that several large corporations adopt ontologies to support large-scale worldwide systems. The most significant example over recent years is the adoption of the Knowledge Graph by Google, a graph built from a large collection of billions of non-ambiguous subject-predicate-object statements used to formally describe general or domain-specific pieces of knowledge (Singhal, 2012). This ontology is used to enhance their search engine capabilities and millions of users benefit from it on a daily basis. Several examples of such large ontologies are now available: DBpedia, Freebase, Wikidata, Yago.

Another significant fact about the increasing adoption of ontologies is the joint effort made by the major search engines companies, e.g., Bing (Microsoft), Google, Yahoo! and Yandex, to design Schema.org[1], a set of structured schemas defining a vocabulary which can be used by publishers to define meta-data with the aim of characterising the content of their web pages in an unambiguous manner.

Another interesting aspect of the last few years is the growing adoption of graph databases (e.g., Neo4J[2], OrientDB[3], Titan[4]). These databases rely on a graph model to describe information in a NoSQL fashion. They actively contribute to the growing adoption of the graph property model – thinking in terms of connected entities (Robinson et al., 2013).

---

[1] http://schema.org
[2] http://www.neo4j.org/
[3] http://www.orientechnologies.com/orientdb/
[4] http://thinkaurelius.github.io/titan/

In this context, a lot of attention has been given to ontologies, which in numerous cases merely correspond to semantic graphs or semantic networks – characterised elements (concepts, instances and relationships) are defined in an unambiguous manner without using complex logical constructs. Such semantic graphs have the interesting properties of being easily expressed and maintained while ensuring a good ratio between semantic expressiveness and effectiveness (in term of computational complexity). This justifies the large number of contributions related to the design of semantic measures dedicated to semantic graphs – a diversity of measures to which this chapter is dedicated.

Due to space constraint we will not introduce knowledge-based measures which refer to logic-based measures, i.e. measures used to compare knowledge base entities defined using complex logical constructs. These measures are used to process knowledge bases, generally defined using description logics, which cannot most of the time be considered as simple graphs – even if a partial representation of the knowledge they define can be reduced as a graph. An introduction to logic-based measures can be found in (D'Amato, 2007), several references are also provided in Section 3.6.1 and in (Harispe et al., 2013b, version 2). In addition, this chapter will not detail the semantic measures which have been proposed for comparing concepts taking into account multiple ontologies, they are however introduced in Section 3.6.2.

This chapter is structured as follow. Section 3.1 provides information regarding the processing of ontologies through semantic graph representations. It also defines the formal notations which will be used to present the measures. Section 3.2 distinguishes different types of measures w.r.t the properties of the semantic graph considered. Section 3.3 presents the foundations of numerous knowledge-based measures by introducing semantic evidence which can be extracted from semantic graph analysis (mainly taxonomies). Section 3.5 presents numerous measures which have been proposed for assessing the semantic similarity of concepts defined in a taxonomy. Section 3.6 briefly introduces knowledge-based measures which are not detailed in this chapter, i.e. logic-based measures and measures based on the analysis of multiple ontologies – several references are provided. Section 3.7 discusses the advantages and limits of knowledge-based measures. Finally, Section 3.9 concludes this chapter.

## 3.1 Ontologies as graphs and formal notations

This section presents the type of ontologies and knowledge organisation system considered in this book. It also provides important information regarding ontology processing through graph analysis – information which is generally not provided in contribution related to semantic measures. We next introduce the notations which will be used in this chapter to refer to particular constitutive elements of a semantic graph.

### 3.1.1 Ontologies as graphs

We briefly introduces the reader to ontologies which can be processed as graphs. However, this section does not aim at: (i) presenting the field of Knowledge Representation, (ii) discussing the broad diversity of ontologies which have been proposed in the literature, and (iii) introducing the language and specifications which can be used to express ontologies, e.g., RDF(S), OWL. Here, we assume that the reader is already familiar with knowledge modelling and to the associated terminology. An introduction to this field of study is provided by Gruber (1993); Baader et al. (2010); Hitzler et al. (2011).

Numerous ontologies can be expressed as graphs. In addition, more complex ontologies can be reduced or used to generate knowledge represented as a graph. This section discusses specific aspects of ontologies related to graph representations. We first introduce simple ontologies which can be represented as graphs (e.g., taxonomies) to further discuss the case of more complex ontologies.

**Taxonomies and partially ordered sets**

Taxonomies are used to structure elements which have similar characteristics into ordered classes. They were originally used in biology to define *taxa* (classes), by categorising organisms sharing common properties. A taxonomy is a function of a taxonomic scheme which defines the properties considered to distinguish classes. Depending on this scheme, the number of classes and their ordering may vary.

The semantics carried by a taxonomy is non-ambiguous as the interpretation of the taxonomic relationship is formally expressed through particular properties/axioms. Indeed, considering a set of elements $C$ (e.g. concepts), a taxonomy is a non-strict partial order (*poset*) of $C$. It can be defined by $\preceq_C$, a binary relation $\preceq$ over $C$ which is[5] : (i) reflexive $\forall c \in C : c \preceq c$, (ii) antisymmetric $\forall u, v \in C : (u \preceq v \wedge v \preceq u) \Rightarrow u = v$ and (iii) transitive $\forall u, v, w \in C : (u \preceq v \wedge v \preceq w) \Rightarrow u \preceq w$.

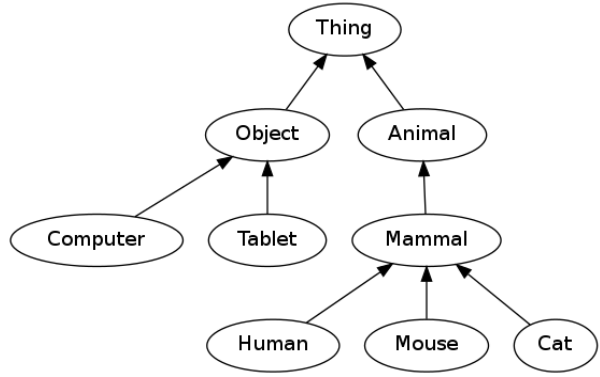

Figure 3.1: Taxonomy of concepts represented as a graph

Note that in some rare cases taxonomies are totally ordered, but generally, they are only partially ordered, i.e., $\exists (u, v) \in C : u \npreceq v \wedge v \npreceq u$. Given that they generally contain a root element denoted $\top$ which subsumes all other elements, i.e., $\forall c \in C, c \preceq \top$, they can be represented as a connected, Rooted and Directed Acyclic Graph (RDAG).

A taxonomy of concepts $\preceq_C$ can therefore be formally defined as a semantic graph $O : \langle C, R, E, A^O \rangle$ with $C$ the set of concepts, $R$ a singleton defining the unique predicate which can be used to order the concepts, i.e. $R = \{\texttt{subClassOf}\}$ and $E \subseteq C \times R \times C$ the set of oriented relationships (edges) which defines the ordering of $C$.

Only considering $O : \langle C, R, E \rangle$ leads to a labelled graph structuring elements of $C$ through labelled oriented edges. Nevertheless, by defining the sets of axioms associated to the taxonomic predicate defined in $R$, e.g., associated relationships are considered reflexive, antisymmetric and transitive, $A^O$ explicitly and formally states that $O$ is a taxonomy *per se* and not a simple graph data structure. These axioms can be used to define inference techniques and more generally to ensure the coherence of specific algorithms w.r.t the knowledge defined in the representations. As an example, Figure 3.1 denotes an example of taxonomy represented by a graph structure.

Although simple, taxonomies are knowledge organisation systems which are used in numerous processes; they are also the backbones of more refined ontologies and are therefore considered as essential components of knowledge modelling. Most of the measures detailed in this chapter rely on these simple ontologies. The notion of taxonomy has been detailed here through a taxonomy of concepts, we also consider $\preceq_R$ the taxonomic of predicates in which $\texttt{subPredicateOf}$ refers to the taxonomic relationship defining that one predicate inherits from another[6].

**General discussion on ontologies as graphs**

Although some ontologies cannot be reduced to simple graphs, a large part of the knowledge they model can generally be expressed as a graph. Therefore, an important aspect to understand is that ontologies, even if they are not explicitly defined as graphs, can be reduced into graphs. Indeed, in all cases, a partial representation of the knowledge defined in expressive ontologies can be manipulated as a graph. The example of the taxonomy has been underlined but this is also the case for the knowledge which links instances to classes (also obtained by a common reasoning procedure). In this case, the ontology can be reduced as a graph in which instances are represented as nodes and linked to their class(es) by simple subject–predicate–object ($\texttt{spo}$) statements. Therefore, any complex ontology, in which sets of concepts and instances have been defined, can be represented as a connected graph in which nodes denote concepts or instances.

In this chapter we consider a graph-based formalism frequently used to manipulate ontologies. It can be used to express numerous network-based ontologies and, sometimes through reductions, ontologies which rely on complex logic constructs. It corresponds to an extension of the structure $O : \langle C, R, E, A^O \rangle$ which has been presented to introduce taxonomies as graphs. Extensions have been made to take instances, data values and multiple predicates into consideration. The next subsection presents this formalism in detail.

---

[5]Note that we adopt the notation used in the literature related to poset instead of the notation commonly used in description logics ($\sqsubseteq$).
[6]Generally named $\texttt{subPropertyOf}$, e.g., in RDFS.

**Types of ontologies considered in this chapter**

Regardless of the particularities of some domain-specific ontologies and regardless of the language considered for the modelling, all approaches used to represent knowledge share common components:

- *Concepts* (Classes), set of things sharing common properties, e.g., `Human`.

- *Instances*, i.e., members of classes, e.g., `alan` (an instance of the class `Human`).

- *Predicates*, the types of relationships defining the semantic relationships which can be established between instances or classes, e.g., `subClassOf`.

- *Relationships*, concrete links between classes and instances which carry a specific semantics, e.g., `alan isA Human` – `alan worksAt BletchleyPark`. Relationships form `spo` statements.

- *Attributes*, properties of instances, e.g., `Alan hasName` *Turing*.

- *Axioms*, for instance defined through properties of the predicates, e.g. *taxonomic relationships are transitive*, the definition of the *domain* and the *range* (co-domain) of predicates, or constraints on predicate and attributes, e.g., *Any Human has exactly 2 legs*.

In practice, numerous ontologies do not rely on complex logical constructs or complex concept/predicate definitions but rather correspond to a formal semantic network, here denoted *semantic graphs*. In addition, we have stressed the fact that complex ontologies can also be regarded as semantic graphs (sometimes considering partial reductions).

A semantic graph, in which instances of classes and data values of specific datatypes are considered, can formally be defined by $O : \langle C, R, I, V, E, A^O \rangle$, with:

- $C$ the set of concepts.
- $R$ the set of predicates.
- $I$ the set of instances.
- $V$ the set of data values.
- $E$ the set of oriented relationships of a specific predicate $r \in R$:
  $E \subseteq E_{CC} \cup E_{RR} \cup E_{II} \cup E_{IC} \cup E_{CI} \cup E_{CV} \cup E_{RV} \cup E_{IV}$ with:

    - $E_{CC} \subseteq C \times R \times C$
    - $E_{RR} \subseteq R \times R \times R$
    - $E_{II} \subseteq I \times R \times I$
    - $E_{IC} \subseteq I \times R \times C$
    - $E_{CI} \subseteq C \times R \times I$
    - $E_{CV} \subseteq C \times R \times V$
    - $E_{RV} \subseteq R \times R \times V$
    - $E_{IV} \subseteq I \times R \times V$

- $A^O$ the set of axioms defining the interpretations of classes and predicates.

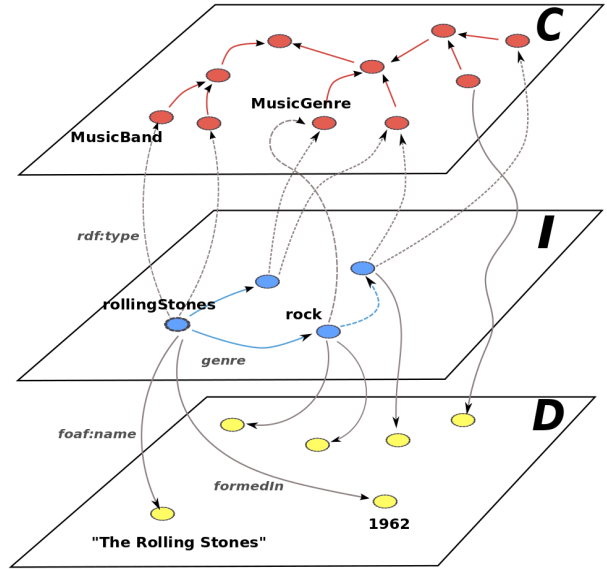

Figure 3.2: Example of a semantic graph related to the music domain. Concepts, instances and data values are represented (Harispe et al., 2013a).

The sets of concepts ($C$), predicates ($R$), instances ($I$), values ($V$) are expected to be mutually disjoint[7]. We consider that each instance is a member of at least one concept and that the taxonomies of concepts $\preceq_C$, and predicates $\preceq_R$ (if any), correspond to connected and rooted directed acyclic graphs (RDAGs). In this manuscript we will mainly manipulate such an ontology without considering predicate taxonomies.

In the following, we consider that a lexical reference – *didactical device* (Guarino and Giaretta, 1995) – is used to refer, in an unambiguous manner, to any node which refers to a concept/predicate/instance. Although we will use a literal in this manuscript, in practice, this unique identifier is a URI or a domain-specific identifier – except data values (V) which are (typed) literals.

Figure 3.2 presents an example of a semantic graph related to the music domain which involves related concepts, predicates, instances and data values[8].

In this example, concepts are taxonomically structured in the layer $C$, e.g. `MusicBand`, `MusicGenre`. Several types of instances are also defined in layer $I$, e.g. `rollingStones`, `rock`. These instances can be characterised according to specific concepts, e.g. the statement `rollingStones isA MusicBand` defines that `rollingStones` is a member of the class `MusicBand`. In addition, instances can be interconnected through specific predicates, e.g., `rollingStones hasGenre rock`. Specific data values (layer $D$) can also be used to specify information relative to both concepts and instances, e.g., `rollingStones haveBeenFormedIn "1962-01-01"^^xsd:date`. All relationships which link the various nodes of the graph are directed and semantically characterised, i.e., they carry an unambiguous and controlled semantics. Notice that extra information are not represented in this figure, e.g., the taxonomy of predicates, axiomatic definitions of predicate properties.

We denote $G_T$, the taxonomic reduction of the ontology, i.e. the layer $C$ in Figure 3.2. $G_T$ corresponds to the taxonomy $\preceq_C$, and therefore only contains concepts. As we will see, this reduction is widely used to compute the semantic similarity between concepts; it will be extensively used in this chapter.

We will alos denote $G_{TI}$ ($T$ stands for Taxonomic and $I$ for `isA` relationship) the graph composed of the layers $C$ and $I$ in Figure 3.2 (only considering edges in $E_{CC}, E_{II}$ and $E_{IC}$[9]).

Knowledge modelling is a vast domain and a large diversity of ontologies have been proposed to express knowledge in a machine understandable form. This section has briefly introduced several ontologies which can be processed as graphs. We have also introduced the formalism adopted in this manuscript to represent such ontologies and to refer to some of their essential components.

---

[7] Note that a set of data types ($D$) can easily be added.
[8] Representation of a subgraph extracted from DBpedia (Auer et al., 2007).
[9] Triplets are rarely defined in $E_{CI}$.

### 3.1.2 Relationships

The relationships of a semantic graph are distinguished according to their predicate and to the pair of elements they link. They can also be denoted statements or triplets in some contributions. As an example, the triplet $(u, t, v)$ corresponds to the unique relationship of type $t \in R$ which links the elements $u, v$: $u$ is named the subject, $t$ the predicate and $v$ the object. Relationships are central elements of semantic graphs and will be used to define algorithms and to characterise paths in the graph.

Since the relationships are oriented, we denote $t^-$ the type of relationship carrying the inverse semantic of $t$. We therefore consider that any relationship $(u, t, v)$ implicitly implies $(v, t^-, u)$, even if the type of relationship $t^-$ and the relationship $(v, t^-, u)$ are not explicitly defined in the graph. As an example, the relationship `Human subClassOf Mammal` implies the inverse relationship `Mammal subClassOf` $^-$ `Human` (even if the ontology defines that `superClassOf` is the inverse of `subClassOf`$^-$, i.e. `subClassOf`$^- \equiv$ `superClassOf`). The notion of inverse predicate will be considered to discuss detailed paths. In some ontology languages, inverse relationships between predicates are explicitly defined by specific construct, e.g., `owl:inverseOf` in OWL.

### 3.1.3 Graph traversals

Graph traversals are often represented through paths in a graph, i.e., sequence of relationships linking two nodes. To express such graph paths, we adopt the following notations[10].

- **Path**: Sequence of relationships $[(c_{i-1}, t_i, c_i), (c_i, t_{i+1}, c_{i+1}), \ldots]$. To lighten the formalism, if a single predicate is used the path is denoted $[c_{i-1}, c_i, c_{i+1}, \ldots]^t$.

- **Path Pattern**: We denote $\pi = \langle t_1, \ldots, t_n \rangle$ with $t_n \in R$, a path pattern which corresponds to a list of predicates[11]. Therefore, any path which is a sequence of relationships is an instance of a specific path pattern $\pi$.

We extend the use of the path pattern notation to express concise expressions of paths:

- $\langle t_* \rangle$ corresponds to the set of paths of any length composed only of relationships having for predicate $t$.

- $\langle t_*^* \rangle$ corresponds to the set of paths of any length composed of relationships associated to the predicate $t$ or $t^-$.

As an example, $\{\texttt{Human}, \langle \texttt{subClassOf}_* \rangle, \texttt{Animal}\}$ refers to all paths which link concepts `Human` and `Animal` and which are only composed of relationships `subClassOf` (they do not contain relationships of type `subClassOf`$^-$).

We also mix the notations to characterise set of paths between specific elements. As an example, $\{u, \langle t, \texttt{subClassOf}_* \rangle, v\}$ represents the set of paths which (i) link the elements $u$ and $v$, (ii) start by a relationship of predicate $t$, and (iii) end by a (possibly empty) path of `subClassOf` relationships. As an example the concept membership function $\mathcal{I}$ which characterises instances of a specific concept can formally be redefined by:

$$\mathcal{I}(\texttt{X}) = \{i | \{i, \langle \texttt{isA}, \texttt{subClassOf}_* \rangle, \texttt{X}\} \neq \emptyset \} \quad (3.1)$$

To lighten the formalism, we consider that the set of paths $\{u, \langle p_* \rangle, v\}$ can be shortened by $\{u, p, v\}$, e.g. $\{\texttt{Human}, \langle \texttt{subClassOf}_* \rangle, \texttt{Animal}\} = \{\texttt{Human}, \texttt{subClassOf}, \texttt{Animal}\}$ and $\{\texttt{Human}, \langle \texttt{subClassOf}_*^* \rangle, \texttt{Animal}\} = \{\texttt{Human}, \texttt{subClassOf}^*, \texttt{Animal}\}$

### 3.1.4 Notations for taxonomies

The taxonomy $G_T$ is the semantic graph associated to the non-strict partial order defined over the set of concepts $C$. We introduce the notations used to characterise $G_T$ as well as its concepts; some of them have already been introduced and are repeated for clarity:

- $C(G_T)$ shortened by $C$ refers to the set of concepts defined in $G_T$.

---

[10]These notations are based on an adaptation of the notations used by Lao (2012).
[11]In SPARQL 1.1, such paths are denoted using path properties $t_1/t_2/t_3$.

- $E(G_T)$ shortened by $E_T$ refers to the set of relationships defined in $G_T$ with:

$$E_T \subseteq C \times \{\texttt{subClassOf}\} \times C \subseteq E_T \subseteq E_{CC}{}^{12}$$

- A concept $v$ subsumes another concept $u$ if $u \preceq v$, i.e., $\{u, \texttt{subClassOf}, v\} \neq \emptyset$. Several additional denominations will be used; it is commonly said that $v$ is an ancestor of $u$, that $u$ is subsumed by $v$ and that $u$ is a descendant of $v$.

- $C^+(u) \subseteq C$, with $u \in C$, the set of concepts such as:

$$C^+(u) = \{c | (u, \texttt{subClassOf}, c) \in E_T\}$$

- $C^-(u) \subseteq C$, with $u \in C$, the set of concepts such as:

$$C^-(u) = \{c | (c, \texttt{subClassOf}, u) \in E_T\}$$

- $C(u) \subseteq C$, with $u \in C$, the set of neighbours of concepts such as:

$$C(u) = C^+(u) \cup C^-(u)$$

- $A(u)$ the set of concepts which subsumes $u$, also named the ancestors of $u$, i.e., $A(u) = \{c|\{u, \texttt{subClassOf}, c\} \neq \emptyset\} \cup \{u\}$. We also denote $A^-(u) = A(u) \setminus \{u\}$ the exclusive set of ancestors of $u$.

- $parents(u)$ the minimal subset of $A^-(u)$ from which $A^-(u)$ can be inferred according to the taxonomy $G_T$, i.e., if $G_T$ doesn't contain taxonomic redundancies[13] we obtain: $parents(u) = C^+(u)$.

- $D(u)$ the set of concepts which are subsumed by $u$, also named the descendants of $u$, i.e., $D(u) = \{c|\{c, \texttt{subClassOf}, u\} \neq \emptyset\} \cup \{u\}$. We also denote $D^-(u) = D(u) \setminus \{u\}$ the exclusive set of descendants of $u$.

- $children(u)$ the minimal subset of $D^-(u)$ from which $D^-(u)$ can be inferred according to the taxonomy $G_T$, i.e., if $G_T$ doesn't contain taxonomic redundancies we obtain: $children(u) = C^-(u)$.

- $roots(G_T)$, shortened by $roots$, the set of concepts $\{c|A(c) = \{c\}\}$. We call the $root$, denoted as $\top$, the unique concept (if any) which subsumes all concepts, i.e., $\forall c \in C, c \preceq \top$.

- $leaves(G_T)$, shortened by $leaves$, the set of concepts without descendants, i.e. $leaves = \{c|D(c) = \{c\}\}$. We also note $leaves(u)$ the set of leaves subsumed by a concept (inclusive if $u$ is a leaf), i.e., $leaves(u) = D(u) \cap leaves$.

- $depth(u)$, the length of the longest path in $\{u, \texttt{subClassOf}, \top\}$, for convenience we also consider $depth(G_T) = \underset{c \in C}{\operatorname{argmax}}\ depth(c)$.

- $G_T^+(u)$ the graph composed of $A(u)$ and the set of relationships which link two concepts in $A(u)$.

- $G_T^-(u)$ the graph composed of $D(u)$ and the set of relationships which link two concepts in $D(u)$.

- $G_T(u) = G_T^+(u) \cup G_T^-(u)$ the graph induced by $A(u) \cup D(u)$.

- $\Omega(u, v)$, the set of Non Comparable Common Ancestors (NCCAs) of the concepts $u, v$. $\Omega(u, v)$ is formally defined by: $\forall (x, y) \in \Omega(u, v), (x, y) \in \{A(u) \cap A(v)\} \times \{A(u) \cap A(v)\} \wedge x \notin A(y) \wedge y \notin A(x)$. NCCAs are also called the Disjoint Common Ancestors (DCAs) in some contributions, e.g. (Couto et al., 2005).

---

[12]$E_{CC}$ were used to introduce semantic graphs

[13]Taxonomic redundancies refer to taxonomic relationships which can be inferred analysing other relationships. As an example if it is define that $u \prec v$ and $v \prec w$, a relationship which explicitly defines that $u \prec w$ is considered redundant.

- A *taxonomic tree* is defined as a special case of taxonomy in which: $\forall c \in C : |parents(c)| < 2$.

Despite the fact that these notations are used to characterise the taxonomy of concepts $G_T$ and that specific semantics is associated to the notations (e.g., parents, children), they can be used to characterise any poset.

Using these notations, the next section introduces semantic evidence which are commonly used to assess the semantic similarity or relatedness of concepts or instances defined into semantic graphs.

## 3.2  Types of semantic measures and graph properties

Two main groups of measures can be distinguished depending on the properties of semantic graphs they are adapted to:

- Measures adapted to semantic graphs composed of (multiple) predicate(s) which potentially induce cycles.

- Measures adapted to taxonomies, i.e., acyclic semantic graphs composed of a unique predicate inducing transitivity.

The two types are introduced in the following sections.

### 3.2.1  Semantic measures on cyclic semantic graphs

Considering all predicates defined in a semantic graph potentially leads to a cyclic graph. Nevertheless, only few semantic measures framed in the relational setting have been designed to deal with cycles. Since these measures take advantage of all predicates, they are generally used to evaluate semantic relatedness (and not semantic similarity). Notice that they can be used to compare concepts and instances. Two types of measures can be further distinguished:

- *Measures based on graph traversal*, i.e., pure graph-based measures. These measures have initially been proposed to study node interactions in a graph and essentially derive from graph theory contributions. They can be used to estimate the relatedness of nodes considering that greater the (direct or indirect) interconnection between two nodes, the more related they are. These measures are not semantic measures *per se* but rather graph measures used to compare nodes. However, they can be used on semantic graphs and can also be adapted in order to take into account evidence of semantics defined in the graph (e.g. strength of connotation).

- *Measures based on the graph property model*. These measures consider concepts or instances as sets of properties distinguished from the graph.

**Semantic measures based on graph traversals**

Measures based on graph traversals can be used to compare any pair of concepts or instances represented as nodes. These measures rely on algorithms designed for graph analysis which are generally used in a straightforward manner. Nevertheless, some adaptations have been proposed in order to take into account the semantics defined in the graph. Among the large diversity of measures and metrics which can be used to estimate the relatedness (distance, interconnection, etc.) of two nodes in a graph, we distinguish:

- Shortest path approaches.

- Random-walk approaches.

- Other interconnection measures.

The main advantage of these measures is their unsupervised nature. Their main drawback is the absence of extensive control over the semantics which are taken into account; this generates difficulties in justifying, explaining, and therefore analysing the resulting scores. However, in some cases, these drawbacks are reduced by enabling fine-grain control over the predicates considered during the comparison. This is done by tuning the contribution of each relationship or predicate.

**Shortest path approaches**

The shortest path problem is one of oldest problems of graph theory. It can be applied to compare both pairs of instances and concepts considering their relatedness as a function of the distance between their respective nodes. More generally, the relatedness is estimated as a function of the weight of the shortest path linking them. Classical algorithms proposed by graph theory can be used. The algorithm to use depends on specific properties of the graph, e.g., Do the constraints applied to the shortest path (really) induce cycles? Are there non-negative weights associated to relationships? Is the graph considered to be oriented?

Rada et al. (1989) were among the first to use the shortest path technique to compare two concepts defined in a semantic graph (initially a taxonomy). This approach is sometimes denoted as the *edge-counting strategy* in the literature (edge refers to relationship). As the shortest path may contain relationships of any predicate we call it unconstrained shortest path (*usp*).

One of the drawbacks of the *usp* in the design of semantic measures lies in the fact that the meaning of the relationships from where the relatedness derives is not taken into account. In fact, complex semantic paths which involve multiple predicates and only those composed of taxonomic relationships are considered equally. Therefore, propositions to penalise any *usp* reflecting complex semantic relationships have been proposed (Hirst and St-Onge, 1998; Bulskov et al., 2002). Approaches for considering particular predicates in a specific manner have also been described. To this end, a *weighting scheme* can be considered in order to tune the contribution of each relationship or predicate in the computation of the final score – this weighting scheme can be derived from the notion of strength of connotation (Section 3.3.3).

**Random walk approaches**

These approaches are based on a Markov chain model of random walks (Spitzer, 1964). The random walk is defined through a transition probability associated to each relationship. The walker can therefore walk from node to node – each node represents a state of the Markov chain. Several measures can be used to compare two nodes $u$ and $v$ based on this technique; a selection of measures introduced by Fouss et al. (2007) is listed:

- The average first-passage time, hitting time, i.e., the average number of steps needed by the walker to go from $u$ to $v$.

- The average commute time, Euclidean commute time distance.

- The average first passage cost.

- The pseudo inverse of the Laplacian matrix.

These approaches are closely related to spectral clustering and spectral embedding techniques (Saerens et al., 2004). Examples of measures based on random walk techniques are defined and discussed in (Muller et al., 2006; Hughes and Ramage, 2007; Ramage et al., 2009; Fouss et al., 2007; Alvarez and Yan, 2011; Garla and Brandt, 2012).

As an example, the hitting time $H(u,v)$ of two nodes $u, v$ is defined as the expected number of steps needed by a random walker to go from $u$ to $v$. The hitting time can recursively be defined by:

$$H(u,v) = 1 + \sum_{k \in N^+(u)} p(u,k) H(k,v) \quad (3.2)$$

With $N^+(u)$ the set of nodes which are linked to $u$ by an outgoing relationship starting from $u$ and $p(u,k)$ the transition probability of the Markov Chain:

$$p(u,k) = \frac{w(u,k)}{\sum_{i \in N^+(u)} w(u,i)}$$

With $w(u,k)$ the weight of the relationship between $u$ and $k$.

The commute time $C(u,v) = H(u,v) + H(v,u)$ corresponds to the expected time needed for a random walker to travel from $u$ to $v$ and back to $u$. Intuitively, the more paths that connect $u$ and $v$, the smaller their commute distance becomes. Several technical criticisms of classical approaches used to evaluate hitting and commute times, as well as associated extensions, have been formulated in the literature, e.g., (Sarkar et al., 2008; von Luxburg et al., 2011).

In a similar vein, approaches based on graph-kernel can also be used to estimate the relatedness of two nodes in a graph (Kondor and Lafferty, 2002); they have already been applied to the design of semantic measures by Guo et al. (2006).

Note that these measures take advantage of second-order information which is generally hard to interpret (in terms of semantics).

**Other measures based on interaction analysis**

Several approaches exploiting graph structure analysis can be used to estimate the relatedness of two nodes through their interconnections. Chebotarev and Shamis (2006b,a) proposed the use of indirect paths linking two nodes by means of the matrix-forest theorem. $sim_{Rank}$, proposed by Jeh and Widom (2002), is an example of such a measure. Considering $N$ as the set of nodes of the graph, $N^-(n)$ as the nodes linked to the node $n$ by a single relationship ending with $n$ (i.e., in-neighbours), $sim_{Rank}$ similarity is defined by:

$$sim_{Rank}(u,v) = \frac{|N|}{|N^-(u)||N^-(v)|} \sum_{x \in N^-(u)} \sum_{y \in N^-(v)} sim_{Rank}(x,y) \quad (3.3)$$

Note that $sim_{Rank}$ is a normalised function. Olsson et al. (2011) propose an adaptation of the measure for semantic graphs built from Linked Data.

**Semantic measures for the graph property model**

The second type of measures which can be used to compare a pair of instances/concepts defined in a (potentially) cyclic semantic graph relies on the graph property model. Here the graph is not only considered as a data structure which highlights the interactions between the different elements it defines. It is considered as a data model in which concepts and instances are describes through sets of properties. The properties may sometimes refer to specific data types. Therefore, the nodes of the graph may refer to data values, concepts, instances or even predicates – the semantic graphs generally correspond to RDF graphs or labelled graphs.

In this case the measures take advantage of semantic graphs by encompassing expressive definitions of concepts/instances through properties. The measures rely on the comparison of the different properties which characterise the concepts or instances being compared. Therefore the study of these measures inherits from early work related to both the comparison of objects defined into knowledge base and the comparison of entities defined in a subset of the first order logic (Bisson, 1992, 1995). As an example, these measures have been extensively studied for comparing objects analysing their different properties. They are based on the aggregation of specific measures enabling the comparison of each of the properties characterising compared objects (Valtchev and Euzenat, 1997; Valtchev, 1999b,a). Considering the domain of knowledge representation, these contributions have formed the basis of several frameworks which are used for comparing instances or concepts in the field of ontology alignment or instance matching, e.g., OWL Lite Alignment (OLA) method has been proposed to compare ontologies based on aggregations of several measures (Euzenat and Valtchev, 2004; Euzenat et al., 2004).

In this presentation, we do not introduce the expressive formalisms which have been introduced in earlier contributions (Bisson, 1992, 1995; Euzenat and Valtchev, 2004; Euzenat et al., 2004; Harispe et al., 2013a), e.g. for comparing objects defined in a knowledge base (Valtchev and Euzenat, 1997; Valtchev, 1999b,a; Harispe et al., 2013a). We rather distinguish two general approaches which have been proposed and which are commonly used to compare concepts or instances.

**Elements represented as a list of direct property**

An element can be evaluated by studying its direct properties, i.e., the set of values associated to the element according to a specific predicate. As an example, focusing on relationships related to instances, two types of relationships can be distinguished:

- *Taxonomic relationships* (`isA`) – relationships which link instances to concepts.

- *Non-taxonomic relationships*:

- Which link two instances (*object properties* in OWL).
- Which link instances to data values (*datatype properties* in OWL).

Two elements will be compared w.r.t values associated to each property considered. To this end, for each property considered, a specific measure will be used to compare associated values (concepts, data values, instances).

Properties which link two instances associate a set of instances to the instance which is characterised. Considering Figure 3.2 page 43, the property `genre` can be used to characterise the instance `rollingStones` through a set of instances $\{i|\exists(\texttt{rollingStones}, \texttt{genre}, i)\}$, i.e., $\{\texttt{rock}, \ldots\}$. Such properties therefore refer to sets, they are often compared using simple set-based measures – they will for example evaluate the cardinality of the intersection (e.g., the number of music genres that two bands have in common).

Taxonomic properties are evaluated using semantic measures adapted to concept comparison. These measures will be presented in Section 3.4.

Properties associated to data values can be compared using measures adapted to the type of data considered, e.g., a measure for comparing dates if the corresponding property refers to a date.

Finally, the scores produced by the various measures (associated to the various properties) are aggregated in order to obtain a global score of relatedness of the two elements (Euzenat and Shvaiko, 2013). Such a representation has been formalised in the framework proposed by Ehrig et al. (2004). This is a strategy which is commonly adopted in ontology alignment, instance matching or link discovery between instances; SemMF (Oldakowski and Bizer, 2005), SERIMI (Araujo et al., 2011) and SILK (Volz et al., 2009) are all based on this approach. The reader can also refers to the extensive survey presented in (Euzenat and Shvaiko, 2013).

**Consideration of indirect properties of elements**

Several contributions underline the relevance of indirect properties in comparing entities represented through graphs, especially in object models (Bisson, 1995). Referring to Figure 3.2 (page 43), indirect properties might be used to consider properties of music genres (e.g., `rock`, `rockNroll`) to compare two music bands (e.g., `rollingStones` - `doors`).

This approach relies on a representation of the compared elements which is an extension of the canonical form used to represent an element as a list of properties. This approach can be implemented to take into account the indirect properties of compared elements, e.g., properties induced by the elements associated to the element that we want to characterise.

Albertoni and De Martino (2006) extended the formal framework proposed in Ehrig et al. (2004) to allow for the consideration of some indirect properties. This framework is dedicated to instance comparison. It formally defines an indirect property of an instance along a path in the graph. The indirect properties to be taken into account depend on the context of use of the framework, e.g., application context.

From a different perspective, Andrejko and Bieliková (2013) suggested an unsupervised approach to compare two instances by considering their indirect properties. Each direct property which is shared between the compared instances plays a role in computing the global relatedness. When the property links two instances, a recursive process is applied to take into account properties of associated instances with the instances being processed. Lastly, the measure aggregates the scores obtained during the recursive process. The authors have also proposed to weigh the contribution of the various properties in the aggregation so as to define a personalised information retrieval approach.

In (Harispe et al., 2013a) the authors also proposed an approach enabling more expressive formulations of measures to compare entities based on the analysis of direct and indirect properties. This approach enables to define aggregations of properties in order to consider new properties during the comparison of entities, e.g. it is impossible to evaluate an instance of a class `Person` whose `weight` and `size` have been specified through his *body mass index* (which can be computed from the `weight` and `size` alone).

All the measures which can be used on the whole semantic graph $G$ can also be used for any reduction $G_{R'} \subseteq G$; the reduction $G_{R'}$ is the subgraph of $G$ built only considering the relationships of $G$ that are of specific types defined into $R' \subset R$, with $R$ the set of types of relationships of $G$. As an example, the taxonomy $G_T$ is the reduction $G_{R'} \subseteq G$ obtained defining $R' = \{\texttt{subClassOf}\}$.[14] Depending on the types of relationships that are considered, i.e. $R'$, a large number of reductions can be obtained. Some of them have the interesting property to be acyclic. Depending on the topological properties of the reduction, two cases can be distinguished:

1. The reduction $G_{R'}$ leads to a cyclic graph. Measures presented for cyclic graphs can be used.

2. $G_{R'}$ is acyclic – particular techniques and algorithms can be used. Most semantic measures defined for acyclic graphs focus on taxonomic relationships defined in $G_{R'}$ and consider the reduction to be the taxonomy of concepts $G_T$. However, some measures consider a specific subset of $R'$, e.g., $R' = \{\texttt{isA}, \texttt{partOf}\}$, which also produces an acyclic graph (Wang et al., 2007). The measures which can be used in this case are usually a generalisation of semantic similarity measures designed for $G_T$.

### 3.2.2 Semantic measures on acyclic graphs

Semantic measures applied to graph-based ontologies were originally designed for taxonomies. Since most ontologies are usually composed mainly of taxonomic relationships or represent poset structures, substantial literature is dedicated to semantic similarity measures[15]. In particular, a large diversity of semantic measures focus on $G_T$ and have been defined for the comparison of pairs of concepts. These measures are presented in details in Section 3.4.

## 3.3 Semantic evidence in semantic graphs and their interpretations

A semantic graph carries explicit semantics, e.g. through the taxonomy defining concepts partial ordering. It also contains implicit semantic evidence. According to Section 1.3.1 page (15), we consider semantic evidence as any information on which interpretations can be based according to the meaning carried by the ontology or the elements it defines (concepts, instances, relationships).

Semantic evidence derives from the study of specific factors (e.g., number of concepts, depth of a concepts, average number of relationships associated to a concept) which can be used to discuss particular properties of the semantic graph (e.g., coverage, expressiveness) or particular properties of its elements (e.g. specificity of concepts). The acquisition of semantic evidence is generally made using the following process. Based on the analysis of specific factors using particular metrics, some properties of both the semantic graph and the elements it defines can be obtained. Based on these properties, and either based on high assumptions or theoretically justified by the core semantics on which relies the ontology, semantic evidence can be obtained. As an example of semantic evidence, the number of concepts described in a taxonomy can be interpreted as a clue on the degree of coverage of the ontology. One can also consider that the deeper a concept w.r.t the depth of $G_T$, the more specific the concept.

As we will see, several properties are used to consider extra semantics from semantic graphs; they are especially important for the design of semantic measures. Indeed, semantic evidence is core elements of measures; it has been used for instance to: (i) normalise measures, (ii) estimate the specificity of concepts and to (iii) weigh the relationships defined in the graph, i.e., to estimate the *strength of connotation* between concept/instances. It is therefore central for both designers and users of semantic measures to know: (i) the properties which can be used to derive semantic evidence, (ii) how it is computed, and (iii) the assumptions on which its interpretation relies.

Most of the properties used to derive semantic evidence are well-known graph properties defined by graph theory. In this section, we only introduce the main properties which are based on the study of a taxonomy of concepts ($G_T$). We go on to introduce two applications of these properties: the estimation of the specificity of concepts and the estimation of the strength of connotation between concepts.

### 3.3.1 Semantic evidence in taxonomies

Semantic evidence used to process more general semantic graphs are adaptations of the graph properties commonly used to assess the similarity of nodes of a taxonomy. The interested reader can refer to the numerous contributions proposed by the graph theory community. In this section we mainly focus on semantic evidence commonly exploited in taxonomies. Two kinds of semantic evidence can be distinguished:

- *Intentional evidence* which can also be called *intrinsic evidence*, which is based on the analysis of properties associated to the topology of $G_T$.

- *Extensional evidence* which is based on the analysis of both the topology of $G_T$ and the distribution of concepts' usage, i.e., the number of instances associated to concepts.

---

[14]In the Figure 3.2, page 43, subClassOf relationships are in red color and the taxonomic reduction $G_T = G_{R'} \subseteq G$ obtained defining $R' = \{\texttt{subClassOf}\}$ thus corresponds the graph shown onto the layer $C$.

[15]According to the literature we consider that semantic measures on $G_T$ are necessarily semantic similarity measure.

Notice that we don't consider semantic evidence *purely* extensional, i.e., only based on concepts' usage, without taking the taxonomy into account. Indeed, in most cases, the distribution of concepts' usage must be evaluated considering the transitivity of the taxonomic relationship. If this is not the case, incoherent results could be obtained, As an example, if the transitivity of the taxonomic relationship is not considered to propagate the usage of concepts (instance membership), the distribution of instances can be incoherent w.r.t the partial order defined by the taxonomy, i.e., a concept can have more instances than one of its ancestors.

We further distinguish the evidence which is based on *global properties* (i.e., derived from the full taxonomy), from that based on *local properties* of concepts.

**Intentional evidence**

Global properties

- Depth of the taxonomy – maximal number of ancestors of a concept

The depth of the taxonomy corresponds to the maximal depth of a concept in $G_T$. It informs on the degree of expressiveness/granularity of the taxonomy. As an example, the deeper $G_T$, the more detailed the taxonomy is expected to be.

The maximal number of ancestors of a concept is also used as an estimator of the upper bound of the degree of expressiveness of a concept. Inversely, the number of concepts defined in $G_T$, i.e., $|D(\top)|$ if $\top$ exists, can also be used as an upper bound of the degree of generality of a concept.

- Diameter – width of the taxonomy

The width of the taxonomy corresponds to the length of the longest shortest path which links two concepts in $G_T$.[16] It also informs on the degree of coverage of the taxonomy. $G_T$ is generally assumed to better cover a domain the bigger its diameter. The rational of this assumption is that bigger is the diameter, the higher the maximal number of concepts separating two concepts and therefore the higher is the likelihood that meaningful concepts of the domain have been expressed into the ontology.

Local properties

- Local density

It can be considered that relationships in dense parts of a taxonomy represent smaller taxonomic distances. Metrics such as compactness can be used to characterise local density (Botafogo et al., 1992)[17]. Other metrics such as the (in/out)-branching factor of a concept ($|C^+(u)|$, $|C^-(u)|$), the number of neighbours of a given concept ($|C(u)|$), can also be used (Sussna, 1993). It is generally assumed that the higher the number of neighbours of a concept, the more general it is.

- Number of ancestors – depth – number of descendants – number of subsumed leaves – distance to leaves.

---

[16] Backtracks, loops or detours excluded, ref: http://mathworld.wolfram.com/GraphDiameter.html
[17] Author also introduces interesting factors for graph-based analysis; the depth of a node is also introduced.

The number of ancestors of a concept is often considered to be directly proportional to its degree of expressiveness. The more a concept is subsumed, the more detailed/restrictive the concept is expected to be. The number of ancestors can also be interpreted w.r.t the maximal number of ancestors a concept of the taxonomy can have. The depth of a concept is also expected to be directly proportional to its degree of expressiveness. The deeper the concept (according to the maximal depth), the more detailed/restrictive the concept is regarded[18]. A local depth of a concept can also be evaluated according to the depth of the branch in which it is defined.

In a similar fashion, in some cases the distance of a concept to the leaves it subsumes, or the number of leaves it subsumes, will be considered as an estimator of expressiveness: the greater the distance/number the less expressive the concept is considered.

**Extensional evidence**

Global properties

- Distribution of instances among the concepts.

The distribution of instances among concepts, i.e., concept usage, can be used to design local correction factors, e.g., to correct estimations of the expressiveness of a concept. This is generally made by evaluating the balance of the distribution.

Local properties

- Number of instances associated to a concept

The number of instances of a concept is expected to be inversely proportional to its expressiveness, the less instances a concept has, the more specific it is expected to be.

---

This semantic evidence and its interpretations have been used to characterise notions extensively used by semantic measures. They are indeed used to estimate the specificity of concepts as well as the strength of connotations between concepts. These two notions are introduced in detail in the following subsections.

### 3.3.2 Concept specificity

Not all concepts have the same degree of specificity. Indeed, most people will agree that `Dog` is a more specific description of a `LivingBeing` than `Animal`. The notion of specificity can be associated to the concept of *salience* which has been defined by Tversky (1977) to characterise a stimulus according to its *"intensity, frequency, familiarity, good form, and informational content"*. In Bell et al. (1988), it is also specified that *"salience is a joint function of intensity and what Tversky calls diagnosticity, which is related to the variability of a feature in a particular set* [i.e., universe, collection of instances]*"*. The idea is to capture the amount of information carried by a concept – this amount is expected to be directly proportional to its degree of specificity and generality.

The notion of specificity of a concept is not totally artificial and can be explained by the roots of taxonomies. Indeed, the transitivity of the taxonomic relationship specifies that not all concepts have the same degree of specificity or detail. In knowledge modelling, the ordering of two concepts $u \prec v$ defines that $u$ must be considered as more abstract (less specific) than $v$. In fact, the taxonomy explicitly defines that all instances of $u$ are also instances of $v$. This expression is illustrated by Figure 3.3; we can see that the more a concept is subsumed by numerous concepts: (A) the number of properties which characterise the concept increases (intentional interpretation), and (B) its number of instances decreases (extensional interpretation).

Therefore, another way of comparing the specificity of concepts defined in a total order[19] is to study their usage, analysing their respective number of instances. The concept which contains the highest number of instances will be the least specific (its universe of interpretation is larger). In this case, it is therefore possible to assess the specificity of ordered concepts either by studying the topology of the graph, or the set of instances associated to them.

---

[18]Note that the depth of a concept as an estimator of its degree of expressiveness can be seen as an inverse function of the notion of *status* introduced by Harary et al. (1965) for organisation study.
[19]For any pair of concepts $u, v$ either $u \preceq v$ or $v \preceq u$.

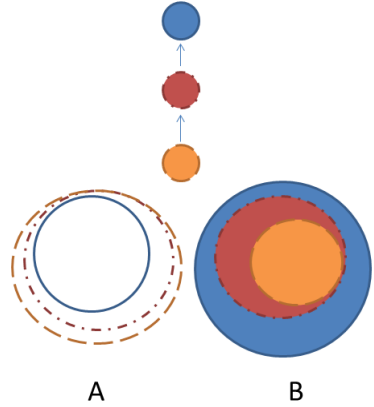

Figure 3.3: Set-based representations of ordered concepts according to (A) their intentional expressions in term of properties characterising the concepts, and (B), in term of their extensional expressions, i.e., the set of instances which compose the concept. Figure based on Blanchard (2008).

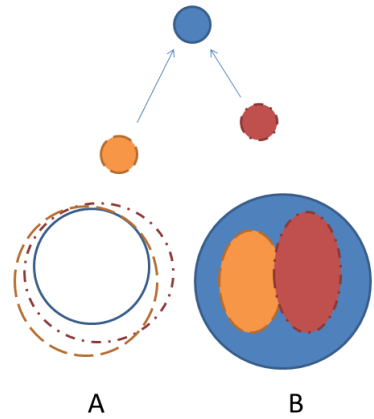

Figure 3.4: Potential set-based representations of non-ordered concepts according to (A) their intentional expressions in term of properties characterising the concepts, and (B), in term of their extensional expressions, i.e., the set of instances associated to the concepts. Figure based on Blanchard (2008).

Nevertheless, in taxonomies, concepts are generally only partially ordered. This implies that presented evidence used to compare the specificity of two ordered concepts cannot be used without assumptions, i.e., concepts which are not ordered are in some sense not comparable. This aspect is underlined in Figure 3.4. It is impossible to compare, in an exact manner, the specificity of two non-ordered concepts. This is due to the fact that the amount of shared and distinct properties between these concepts can only be estimated w.r.t the properties which characterise the common concepts they derive from, i.e., their Non Comparable Common Ancestors (NCCAs). However, this estimation can only be a lower bound of their commonality since extra properties shared by the two concepts may not be carried by such NCCAs.

As we will see, the estimation of the degree of specificity of concepts is of major importance in the design of semantic measures. Therefore, given that discrete levels of concept specificity are not explicitly expressed in a taxonomy, various approaches and functions have been explored to evaluate concept specificity. We denote such a function as $\theta$:

$$\theta : C \to \mathbb{R}_+ \qquad (3.4)$$

The function $\theta$ may rely on the intrinsic and/or extrinsic properties presented above. It must be in agreement with the taxonomic representation which defines that concepts are always semantically broader than their specialisations[20]. Thus, $\theta$ must monotonically decrease from the leaves (concepts without descendants) to the root(s) of the taxonomy:

$$x \preceq y \Rightarrow \theta(x) \geq \theta(y) \tag{3.5}$$

We present examples of $\theta$ functions which have been defined in the literature.

**Basic intrinsic estimators of concept specificity**

The specificity of concepts can be estimated considering the location of its corresponding node in the graph. A naive approach will define the specificity of the concept $c$, $\theta(c)$, as a function of some simple properties related to $c$, e.g., $\theta(c) = f(depth(c))$, $\theta(c) = f(A(c))$ or $\theta(c) = f(D(c))$ with $A(c)$ and $D(c)$ the ancestors and descendants of $c$.

The main drawback of simple specificity estimators is that concepts with a similar depth or an equal number of ancestors/descendants will have similar specificities, which is a heavy assumption. In fact, two concepts can be described with various degrees of detail, independently of their depth, e.g., (Yu et al., 2007a). More refined $\theta$ functions have been proposed to address this limitation.

**Extrinsic information content**

Another strategy explored by designers of semantic measures has been to characterise the specificity of concepts according to Shannon's Information Theory. The specificity of a concept will further be regarded as the amount of information the concept conveys, its Information Content (IC). The IC of a concept can for example be estimated as a function of the size of the universe of interpretations associated to it (i.e. instances). The IC is a common expression of $\theta$ and was originally defined by Resnik (1995) to assess the informativeness of concepts from a corpus of texts.

The IC of the concept $c$ is defined as inversely proportional to $p(c)$, the probability to encounter an instance of $c$ in a collection of instances (negative entropy). The original IC definition was based on the number of occurrences of a concept in a corpus of texts.

We denote $eIC$ any IC which relies on extrinsic information, i.e., information not provided by the ontology[21]. They are based on the analysis of concept usage in a corpus of texts or on the analysis of a collection of instances for which associated concepts are known[22]. We consider the formulation of $eIC$ originally defined by Resnik (1995):

$$p(c) = \frac{|\mathcal{I}(c)|}{|I|}$$

with $\mathcal{I}(c)$ the set of instances of $c$, e.g., occurrences of $c$ in a corpus, instances in an ontology $\{i|\{i, \langle \texttt{isA}, \texttt{subClassOf}_*\rangle, c\} \neq \emptyset\}$.

$$\begin{aligned} eIC_{Resnik}(c) &= -log(p(c)) \\ &= log(|I|) - log(|\mathcal{I}(c)|) \end{aligned} \tag{3.6}$$

The suitability of the log function can be supported by the work of Shepard (1987)[23]. Notice also the link with Inverse Document Frequency (IDF) which is commonly used in information retrieval (Jones, 1972):

$$\begin{aligned} IDF(c) &= log(\frac{|I|}{|\mathcal{I}(c)|}) \\ &= log(|I|) - log(|\mathcal{I}(c)|) \\ &= IC(c) \end{aligned} \tag{3.7}$$

---

[20]This explains that the specificity of concepts cannot be estimated only considering extrinsic information.
[21]Note that if the instances are represented in the graph, some $eIC$ are indeed $iIC$.
[22]As an example, usage of concepts defined in the Gene Ontology can be known studying gene annotations which provide genes and associated Gene Ontology concepts, e.g. refer to UniprotKB.
[23]Shepard derived his universal law of stimulus generalisation based on the consideration that logarithm functions are suited to approximate semantic distance (Al-Mubaid and Nguyen, 2006).

The main drawback of $\theta$ functions based on extrinsic information lies in their tight dependence on concepts usage: they will automatically reflect its bias[24]. Nevertheless, in some cases, the consideration of such bias is desired as all concepts which are highly represented will be considered less informative, even the concepts which seem specific w.r.t intrinsic factors (e.g., depth of concepts). However, in some cases, bias in concept usage can badly affect IC estimation and may not be adapted. In addition, IC computation based on text analysis can be both time consuming and challenging given that, in order to be accurate, complex disambiguation techniques have to be used to detect which concept occurs in texts or knowledge bases.

**Intrinsic information content**

In order to avoid the dependency of $eIC$ calculus to concept usage, various intrinsic IC formulations ($iIC$) have been proposed. They can be used to define $\theta$ functions by only considering structural information extracted from the ontology, e.g., the intrinsic factors presented in Section 3.3.1. $iIC$ formulations extend basic specificity estimators presented above.

Multiple topological characteristics can be used to express $iIC$, e.g., number of descendants, ancestors, depths, etc. (Seco et al., 2004; Schickel-Zuber and Faltings, 2007; Zhou et al., 2008; Sánchez et al., 2011). As an example, the formulation proposed by Zhou et al. (2008) enables to consider the contribution of both the depth and the number of descendants of a concept to compute its specificity:

$$iIC_{Zhou}(c) = k \left(1 - \frac{log(|D(c)|)}{log(|C|)}\right) + (1-k)\frac{log(depth(c))}{log(depth(G_T))} \quad (3.8)$$

with $|C|$ the number of concepts defined in the taxonomy, $D(c)$ the descendants of $c$, $depth(c)$ its depth, $depth(G_T)$ the maximal depth of a concept in $G_T$ and $k \in [0,1]$ a parameter used to set the contribution of both components (originally set to 0.6).

In (Sánchez et al., 2011), the $iIC$ incorporates additional semantic evidence in the aim of better distinguishing the concepts with the same numbers of descendants but different degrees of *concreteness* – here captured as a function of the number of ancestors a concept has.

$$iIC_{Sanchez}(c) = -log\left(\frac{\frac{|leaves^-(c)|}{|A(c)|} + 1}{|leaves| + 1}\right) \quad (3.9)$$

We denote $leaves^-(c)$ the exclusive set of leaves of the concept $c$, i.e., if $c$ is a leaf $leaves^-(c) = \emptyset$. Note that $iIC_{Sanchez}$ will set the same $iIC$ for each leaf. To avoid this, we propose the following modification:

$$iIC_{Sanchez'}(c) = -log\left(\frac{\frac{|leaves(c)|}{|A(c)|} + 1}{|leaves| + 1}\right) \quad (3.10)$$

$iICs$ are of particular interest as only the topology of the taxonomy is considered. They prevent errors related to bias on concept usage. However, the relevance of $iIC$ relies on the assumption that $G_T$ expresses enough knowledge to rigorously evaluate the specificities of concepts. Therefore, as a counterpart, $iICs$ are sensitive to structural bias in the taxonomy and are therefore sensitive to *unbalanced taxonomy, degrees of completeness, homogeneity and coverage* of the taxonomy (Batet et al., 2010a).

**Non-taxonomic information content**

Both introduced $iIC$ and $eIC$ only take taxonomic relationships into account. In order to take advantage of all predicates and semantic relationships, Pirró and Euzenat (2010a) proposed the *extended IC* ($extIC$).

$$extIC(c) = \alpha EIC(c) + \beta IC(c) \quad (3.11)$$

$$EIC(c) = \sum_{r \in R} \frac{\sum_{u \in C_r(u)} iIC(u)}{|C_r(u)|}$$

With $C_r(u)$ the set of concepts linked to the concept $c$ by any relationship of type $r \in R$ (i.e., generalisation of $C(u)$). In this formula, the contribution of the various relationships of the same predicate is averaged. However, the more a concept establishes relationships of different predicates, the higher its $EIC$ will be. We thus propose to average $EIC$ by $|R|$, or to weigh the contribution of the different predicates.

---

[24] As an example, this can be problematic for GO-based studies as some genes are studied and annotated more than others (e.g., drug related genes) and annotation distribution patterns among species reflect abnormal distortions, e.g. human – mouse (Thomas et al., 2012).

**List of functions defined to estimate concept specificity**

We have presented various strategies which can be used to estimate the specificities of concepts defined in a partially ordered set ($\theta$ functions). It is important to understand that these estimators are based on assumptions regarding ontologies. Table 3.1 lists some of the properties of some of the $\theta$ functions proposed in the literature – proposals are ordered by date. Ex inf. refers to the use of extensional information.

| Names – References | Ex inf | Co-domain | Comments |
|---|---|---|---|
| Depth | No | [0, 1] | Normalised depth or max depth can be used. In a graph, considering the minimal depth of a concept doesn't ensure that the specificity increases according to the partial ordering (due to multi-inheritance). |
| IC Resnik (Resnik, 1995) | Yes | $[0, +\infty[$, $[0, 1]$ | Depend on concept usage. $IC(c) = -log(|\mathcal{I}(c)|/|I|)$. Normalised versions have also been proposed, e.g., (Sheehan et al., 2008). |
| IC Resnik intrinsic (Resnik, 1995) | No | [0, 1] | Resnik's IC with $\forall c \in C$ the number of instances associated to $c$ (without taxonomic inferences) set to one. |
| IC Seco (Seco, 2005) | No | [0, 1] | IC estimated from the number of descendants. |
| Depth non-linear (Seco, 2005) | No | [0, 1] | Use log to introduce non-linear estimation of depth. |
| TAM (Yu et al., 2007a) | Yes | $[0, +\infty[$ | The probability $p(c)$ associated to a concept is computed as the number of pairs of instances which are members of $c$ divided by total the number of pairs. |
| IDF (Chabalier et al., 2007) | Yes | $[0, +\infty[$ | Inverse Document Frequency (IDF) obtained by dividing the number of instances by the number of instances of the concept (Jones, 1972), i.e. $IDF(c) = log(|I|/|\mathcal{I}(c)|)$. As we saw, in Section 3.3.2, this formulation is similar to IC proposed by Resnik (1995). |
| APS (Schickel-Zuber and Faltings, 2007) | No | [0,1/2] | $iIC$ based on the number of descendants of a concept. |
| IC Zhou (Zhou et al., 2008) | No | [0,1] | Parametric hybrid $iIC$ mixing Seco's IC and nonlinear depth. |
| extIC (Pirró and Euzenat, 2010a) | No | [0, 1] | $iIC$ based on all predicates. |
| IC Sanchez et al (A) (Sánchez et al., 2011) | No | $[0, +\infty[$ | Consider the number of leaves contained in $D(c)$, the higher it is, the less specific $c$ is considered. |
| IC Sanchez et al. (B) (Sánchez et al., 2011) | No | $[0, +\infty[$ | Refined version A (see above) exploiting $D(c)$. |

Table 3.1: Selection of $\theta$ functions which can be used to estimate the specificity of a concept defined in a taxonomy.

### 3.3.3 Strength of connotations between concepts

A notion strongly linked to concept specificity is the strength of connotation between a pair of concepts/instances, i.e., the strength of the relationship(s) which links two concepts/instances. Otherwise stated, this notion can be used to assess the strength of interaction associated to a specific relationship.

Considering taxonomic relationships, it is generally considered that the strength of connotation between concepts is stronger the deeper two concepts are in the taxonomy. As an example, the taxonomic relationship linking concepts `SiberianTiger` and `Tiger` will generally be considered to be stronger than the one linking the concepts `Animal` and `LivingBeing`. Such a notion is quite intuitive and has for instance been studied by Quillian and Collins in the early studies of semantic networks (Collins and Quillian, 1969) – hierarchical network models were built according to response time to questions, i.e., mental activations evaluated w.r.t the time people took to correctly answer questions related to two concepts, e.g., a `Canary` is an `Animal` – a `Canary` is a `Bird` – a `Canary` is a `Canary`. Based on the variation of times taken to correctly answer questions involving two ordered concepts (e.g., `Canary` – `Animal`), the authors highlighted human sensibility to non-uniform strength of connotation and its link to concept specificity.

It is worth noting that the estimation of the strength of connotation of two linked concepts is in some sort a measure of the semantic similarity or taxonomic distance between two directly ordered concepts. The models used to estimate the strength of connotation between two concepts are generally based on the assumption that the taxonomic distance associated to a taxonomic relationship *shrinks* with the depth of the two concepts it links (Richardson et al., 1994). Given that the strength of connotation between concepts is not explicitly expressed in a taxonomy, it has been suggested that several intrinsic factors need considering in order to refine its estimation, e.g., (Richardson et al., 1994; Young Whan and Kim, 1990; Sussna, 1993).

A taxonomy only explicitly defines the partial ordering of its concepts, which means that if a concept $v$ subsumes another concept $u$, all the instances of $u$ are also instances of $v$, i.e., $u \preceq v \Rightarrow \mathcal{I}(u) \subseteq \mathcal{I}(v)$. Nevertheless, non-uniform strength of connotation aim to consider that all taxonomic relationships do not convey the same semantics.

Strictly speaking, taxonomic relationships only define concept ordering and concept inclusion. Therefore, according to the extensional interpretation which can be made of a taxonomy, the size of the universe of interpretation of a concept, i.e., the size of the set of its possible instances w.r.t the whole set of instances, must reduce the more a concept is specialised[25]. This reduction of the universe of eligible interpretations associated to a concept (i.e. instances), corresponds to a specific understanding of the semantics of non-uniform strengths of connotation. Alternative explanations which convey the same semantics can also be expressed according to the insights of the various cognitive models which have been introduced in Section 1.2.1:

- Spatial/Geometric model: it states that the distance between concepts is a non-linear function which must take salience of concept into account.

- Feature model (which represents a concept as a set of properties): It can be seen as the difficulty to further distinguish a concept which is relevant to characterise the set of instances of a domain.

- Alignment and Transformational models: the effort of specialisation which must be done to extend a concept increases the more a concept has been specialised.

All these interpretations state the same central notion – the strength of connotation which links two concepts is a function of two factors: (i) the specificities of the linked concepts, and (ii) the variation of these specificities. The semantic evidence introduced in the previous section, as well as the notion of IC, can be used to assess the strength of connotation of two concepts.

As an example, the strength of connotation $w$ which characterises a taxonomic relationship linking two concepts $u, v$, with $u \preceq v$, can be defined as a function of the ICs of $u$ and $v$ (Jiang and Conrath, 1997): $w(u,v) = IC(u) - IC(v)$.

It is important to stress that estimations of the strength of connotations based on the density of concepts, the branching factor, the maximal depth or the width of the taxonomy, are based on assumptions regarding the definition of the ontology.

We have presented various semantic evidence which can be used to extract knowledge from an ontology represented as a semantic graph. We have also presented two applications of such semantic evidence for assessing the specificity of a concept defined in a taxonomy and the strength of interaction between two elements defined in a semantic graph. As we will see, semantic evidence are central for the definition of semantic measures. We will now introduce semantic measures which can used to compare pairs of concepts and pairs of groups of concepts.

---

[25]We here consider a finite universe.

## 3.4 Semantic similarity between a pair of concepts

The majority of semantic measures framed in the relational setting have been proposed to assess the semantic similarity or taxonomic distance of a pair of concepts defined in a taxonomy. Given that they are designed to compare two concepts, these measures are denoted as *pairwise measures* in some communities, e.g., bioinformatics (Pesquita et al., 2009a). As we will see, extensive literature is dedicated to these measures – they can be used to compare any pairs of nodes expressed in a graph which defines a (partial) ordering, that is to say, any graph structured by relationships which are transitive, reflexive and antisymmetric (e.g., `isA`, `partOf`).

The main approaches used to compare concepts defined in a taxonomy are:

- **Measures based on graph structure analysis**. They estimate the similarity as a function of the degree of interconnection between concepts. They are generally regarded as measures which are framed in the spatial model – the similarity of two concepts is estimated as a function of their distance in the graph, e.g., based on the analysis of the lengths of the paths which link the concepts. These measures can also be considered as being framed in the transformational model by considering them as functions which estimate the similarity of two concepts regarding the difficulty to transform one concept into another.

- **Measures based on concept features analysis**. This approach extracts features of concepts from the graph. These features will be subsequently analysed to estimate the similarity as a function of shared and distinct features of the compared concepts. This approach is conceptually framed in the feature model. The diversity of feature-based measures relies on the diversity of strategies which have been proposed to characterise concept features, and to take advantage of them in order to assess the similarity.

- **Measures based on Information Theory**. Based on a function used to estimate the amount of information carried by a concept, i.e., its Information Content (IC), these measures assess the similarity w.r.t the amount of information which is shared and distinct between compared concepts. This approach is framed in information theory. However, in some cases, it can be seen as a derivative of the feature-based approach in which features are not compared using a boolean feature-matching evaluation (shared/not shared), but also incorporate their saliency, i.e. their degree of informativeness.

- **Hybrid measures**. Measures which are based on multiple paradigms.

The broad classification of measures that we propose is interesting as an introduction to basic approaches defined to assess the similarity of two concepts – and to put them in perspective with the models of similarity proposed by cognitive sciences. It is however challenging to constrain the diversity of measures to this broad classification. It is important to understand that these four main approaches are highly interlinked and cannot be seen as disjoint categories. As an example, all measures rely in some sense on the analysis of the structure of the taxonomy, i.e., they all take advantage of the partial ordering defined by the (structure of the) taxonomy. These categories must be seen as devices used by designers of semantic measures to introduce approaches and to highlight relationships between several proposals. Indeed, as we will see, numerous approaches can be regarded as hybrid measures which take advantage of techniques and paradigms used to characterise measures of a specific approach. Therefore, the affiliation of a specific measure to a particular category is often subject to debate, e.g., as it is exposed in (Batet, 2011). This can be partially explained by the fact that several measures can be redefined or approximated using reformulations, in a way that further challenge the classification. Indeed, the more you analyse semantic measures, the harder it is to restrict them to specific boxes; the analogy can be made with the relationship between cognitive models of similarity[26].

---

[26] Refer to dedicated Section 1.2.1 and more particularly to efforts made for the unification of the various models.

Several classifications of measures have been proposed. The most common one is to distinguish measures according to the elements of the graph that they take into account (Pesquita et al., 2009a). This classification distinguishes three approaches: (i) *edge-based* – measures focusing on relationship analysis, (ii) *node-based* – measures based on node analysis, and (iii) *hybrid measures* – measures which mix both approaches. In the literature, edge-based measures often refer to structural measures, node-based measures refer to measures framed in the feature-model and those based on information theory. Hybrid measures are those which implicitly or explicitly mix several paradigms.

Another interesting way to classify measures is to study whether they are (i) *intentional*, i.e., based on the explicit definition of the concepts expressed by the taxonomy, (ii) *extensional*, i.e., based on the analysis of the realisations of the concepts (i.e., instances), or (iii) *hybrid*, measures which mix both intentional and extensional information about concepts. Refer to (Gandon et al., 2005; Aimé, 2011)[27] for examples of such classifications.

In some cases, authors will mix several types of classifications to present measures. In this section, we will introduce the measures according to the four approaches presented above: (i) structural, (ii) feature-based, (iii) framed in information theory, and (iv) hybrid. We will also specify the extensional, intentional, or hybrid nature of the measures.

Numerous concept-to-concept measures have been defined for trees, i.e. special taxonomies which correspond to graphs without multiple inheritances. In the literature, these measures are generally considered to be applied *as it is* on taxonomies (which are DAGs). However, in DAGs, some adaptations deserve to be made and several components of measures generally need to be redefined in order to avoid ambiguity, e.g., to be implemented on computer software. For the sake of clarity, we first highlight the diversity of proposals by introducing the most representative measures defined according to the different approaches. In most cases, measures will be presented according to their original definitions. When the measures have been defined for trees, we will not necessarily stress the modifications which must be taken into account for them to be used on DAGs. These modifications will be discussed after the introduction of the diversity of measures. For convenience, `subClassOf` relationships will be denoted *isa* – there is no ambiguity with `isA` since $G_T$ only contains concepts.

### 3.4.1 Structural approach

Structural measures rely on graph-traversal approaches presented in Section 3.2.1 (e.g., shortest path techniques, random walk approaches). They focus on the analysis of the interconnection between concepts to estimate their similarity. However, most of the time, they consider specific tuning in order to take into account specific properties and interpretations induced by the transitivity of the taxonomic relationships. In this context, some authors, e.g., (Hliaoutakis, 2005), have linked this approach to the spreading activation theory (Collins and Loftus, 1975). The similarity is in this case seen as a function of propagation between concepts through the graph.

Back in the eighties, Rada et al. (1989) expressed the taxonomic distance of two concepts defined in a taxonomic tree as a function of the shortest path linking them[28]. We denote $sp(u, isa^*, v)$ the shortest path between two concepts $u$ and $v$, i.e., the path of minimal length in $\{u, isa^*, v\}$. Remember that the length of a path has been defined as the sum of the weights associated to the edges which compose the path. When the edges are not weighted we refer to the edge-counting strategy – the length of the shortest path is the number of edges it contains. The taxonomic distance is therefore defined by[29]:

$$dist_{Rada}(u,v) = sp(u, isa^*, v) \qquad (3.12)$$

Distance-to-similarity conversions can also be applied to express a similarity from a distance. A semantic similarity can therefore be defined in a straightforward manner:

$$sim_{Rada}(u,v) = \frac{1}{dist_{Rada}(u,v) + 1} \qquad (3.13)$$

Notice the importance of considering the transitive reduction of the tree/graph to obtain coherent results using measures based on the shortest path. In the following presentation, we consider that the taxonomy $G_T$ doesn't contain redundant relationships (here redundancies refer to relationships which can be inferred due to the transitivity of taxonomic relationships).

---

[27] In french.

[28] It is worth noting that they didn't invent the notion of shortest path in a graph. In addition, in Foo et al. (1992), the authors refer to a measure proposed by Gardner et al. (1987) to compare concepts defined in a conceptual graph using the shortest path technique.

[29] In this chapter, equations named *dist* refer to taxonomic distances.

In a tree, the shortest path $sp(u, isa^*, v)$ contains a unique common ancestor of $u$ and $v$. This common ancestor is the Least Common Ancestor (LCA)[30] of the two concepts according to any function $\theta$ (since the $\theta$ function is monotonically decreasing)[31]. Therefore, in trees, we obtain $dist_{Rada}(u,v) = sp(u, isa, LCA(u,v)) + sp(v, isa, LCA(u,v))$.

Several issues with the shortest path techniques have been formulated. The edge-counting strategy, or more generally any shortest path approach with uniform edge weight, has been criticised for the fact that the distance represented by an edge linking two concepts does not take concept specificities/salience into account[32]. Several modifications have therefore been proposed to break this constraining uniform appreciation of edges. Implicit or explicit models defining non-uniform strength of connotation between concepts have therefore been introduced e.g., (Young Whan and Kim, 1990; Sussna, 1993; Richardson et al., 1994).

One of the main challenges of designers of semantic measures over the years has therefore been to refine measures by taking advantage of semantic evidence related to concept specificity and the strength of connotation between concepts. The different strategies and factors used to appreciate concept specificity as well as strength of connotations have already been introduced in Section 3.3. Another use of the various semantic evidence which can be extracted from $G_T$ has been to normalise the measures. As an example, Resnik (1995) suggested considering the maximal depth of the taxonomy to bound the edge-counting strategy:

$$sim_{Resnik-eb}(u,v) = 2 \cdot depth(G_T) - sp(u, isa, LCA(u,v)) - sp(v, isa, LCA(u,v)) \qquad (3.14)$$

To simulate non uniform edge weighing, Leacock and Chodorow (1998)[33] introduced a logarithmic transformation of the edge counting strategy:

$$sim_{LC}(u,v) = -log\left(\frac{N}{2 \cdot depth(G_T)}\right) = log(2 \cdot depth(G_T)) - log(N) \qquad (3.15)$$

with $N$ the cardinality of the union of the sets of nodes involved in the shortest paths $sp(u, isa, LCA(u,v))$ and $sp(v, isa, LCA(u,v))$.

Authors have also proposed taking into account the specificity of compared concepts, e.g., (Mao and Chu, 2002), sometimes as a function of the depth of their LCA, e.g.,(Wu and Palmer, 1994; Pekar and Staab, 2002; Wang et al., 2012b). As an example, Wu and Palmer (1994) proposed expressing the similarity of two concepts as a ratio taking into account the shortest path linking the concepts as well as the depth of their LCA.

$$sim_{WP}(u,v) = \frac{2 \cdot depth(LCA(u,v))}{2 \cdot depth(LCA(u,v)) + sp(u, isa, LCA(u,v)) + sp(v, isa, LCA(u,v))} \qquad (3.16)$$

This function is of the form:

$$f(x,y,z) = \frac{x}{(x + (y+z)/2)}$$

with $x$ the depth of the LCA of the two concepts $u,v$ and $y+z$ the length of the shortest path linking $u,v$. It is easy to see that for any given non-null length of the shortest path, this function increases with $x$; otherwise stated, to a given shortest path length, $sim_{WP}(u,v)$ increases with the depth of $LCA(u,v)$. In addition, as expected, for a given depth of the $LCA$, the longer the shortest path which links $u,v$, less similar they will be considered.

Based on a specific expression of the notion of depth, a parametrised expression of $sim_{WP}$ has been proposed in Wang and Hirst (2011). A variation was also proposed by Pekar and Staab (2002):

$$sim_{PS}(u,v) = \frac{depth(LCA(u,v))}{sp(u, isa, LCA(u,v)) + sp(v, isa, LCA(u,v)) + depth(LCA(u,v))} \qquad (3.17)$$

---

[30] The Least Common Ancestor is also denoted as the Last Common Ancestor (LCA), the Most Specific Common Ancestor (MSCA), the Least Common Subsumer/Superconcept (LCS) or Lowest SUPER-ordinate (LSuper) in the literature.

[31] Here relies the importance of applying the transitive reduction of the taxonomic graph/tree, redundant taxonomic relationships can challenge this statement and therefore heavily impact the semantics of the results.

[32] As an example, Foo et al. (1992) quotes remarks made in Sowa personal communication.

[33] Note that according to Resnik (1995), this approach was already proposed in an 1994 unpublished paper by the same authors (Leacock and Chodorow, 1994).

Zhong et al. (2002) also proposed comparing concepts taking into account the notion of depth:

$$dist_{Zhong}(u,v) = 2 \cdot \frac{1}{2k^{depth(LCA(u,v))}} - \frac{1}{2k^{depth(u)}} - \frac{1}{2k^{depth(v)}} \tag{3.18}$$

with $k > 1$ a factor defining the contribution of the depth.

In a similar fashion, Li et al. (2003, 2006) defined a parametric function in which both the length of the shortest path and the depth of the LCA are taken into account:

$$sim_{LB}(u,v) = e^{-\alpha dist_{Rada}(u,v)} \times df(u,v) \tag{3.19}$$

with,

$$df(u,v) = \frac{e^{\beta h} - e^{-\beta h}}{e^{\beta h} + e^{-\beta h}}$$

The parameter $h$ corresponds to the depth of the LCA of the compared concepts, i.e. $h = depth(LCA(u,v))$. The parameter $\beta > 0$ is used to tune the depth factor ($df$) and to set the importance given to the degree of specificity of concepts. The function used to express $df$ corresponds to the hyperbolic tangent which is normalised between 0 and 1. It defines the degree of non-linearity to associate to the depth of the LCA. In addition, $\alpha \geq 0$ controls the importance of the taxonomic distance expressed as a function of the length of the shortest path linking the two concepts.

Approaches have also been proposed to modify existing measures in order to obtain particular properties. As an example, Slimani et al. (2006) proposed an adaptation of the measure proposed by Wu and Palmer (1994) (Equation 3.16) in order to avoid the fact that, in some cases, neighbour concepts can be estimated as more similar than ordered concepts. To this end, the authors introduced $sim_{tbk}$ which is based on a factor used to penalise concepts defined in the neighbourhood:

$$sim_{tbk}(u,v) = sim_{WP}(u,v) \times pf(u,v) \tag{3.20}$$

with,

$$pf(u,v) = (1-\lambda)(min(depth(u), depth(v)) - depth(G_T)) + \lambda(depth(u) + depth(v) + 1)^{-1}$$

In the same vein (Ganesan et al., 2012; Shenoy et al., 2012) recently proposed alternative measures answering the same problem. The approach proposed by Shenoy et al. (2012) is presented[34]:

$$sim_{Shenoy}(u,v) = \frac{2 \cdot depth(G_T) \cdot e^{-\lambda L/depth(G_T)}}{depth(u) + depth(v)} \tag{3.21}$$

with $L$ the weight of the shortest path computed by penalising paths with multiple changes of type of relationships, e.g. a path following the pattern $\langle isa, isa^-, isa, \ldots \rangle$. Note that the penalisation of paths inducing complex semantics, e.g., which involves multiple types of relationships, was already introduced in (Hirst and St-Onge, 1998; Bulskov et al., 2002).

Several approaches have also been proposed to consider density of concepts, e.g., through analysis of cluster of concepts (Al-Mubaid and Nguyen, 2006). Other adaptations also proposed taking into account concepts' distance to leaves (Wu et al., 2006), and variable strengths of connotation considering particular strategies (Lee et al., 1993; Zhong et al., 2002), e.g., using IC variability among two linked concepts or multiple topological criteria (Jiang and Conrath, 1997; Alvarez et al., 2011).

In terms of the spreading activation theory, measures have also been defined as a function of transfer between the compared concepts (Schickel-Zuber and Faltings, 2007). Wang et al. (2007) use a similar approach based on a specific definition of the strength of connotation. Finally, pure graph-based approaches defined for the comparison of nodes can also be used to compare concepts defined in a taxonomy (refer to Section 3.2.1). As an example, Garla and Brandt (2012) and Yang et al. (2012) define semantic similarity measures using random walk techniques such as the personalised page rank approach.

As we have seen, most structural semantic similarity measures are extensions or refinements of the intuitive shortest path distance considering intrinsic factors to consider both the specificity of concepts and variable strengths of connotations. Nevertheless, the algorithmic complexity of the shortest path algorithms hampers the suitability of these measures for large semantic graphs[35]. To remedy this problem, we have seen that shortest path computation can be substituted by approximation based on the depth of the LCA of the compared concepts[36], and that several measures proposed by graph theory can be used instead.

---

[34]Note that we assume that the paper contains an error in the equation defining the measure. The formula is considered to be $X/(Y+Z)$, not $X/Y+Z$ as written in the paper.

[35]A linear algorithm in $O(C+E)$ exists for DAGs; nevertheless search for $sp(u, isa^*, v)$ requires the consideration of cyclic graphs for which algorithms, such as Dijkstra's, are in $O(C^2)$ or $O(E+C \cdot logC)$ using sophisticated implementation.

**Towards other estimators of semantic similarity**

Most criticisms related to the initial edge-counting approach were linked to the uniform consideration of edge weights. As we have seen, to remedy this, several authors proposed considering a great deal of semantic evidence to differentiate strengths of connotation between concepts.

One of the central findings conveyed by early developments in structure-based measures is that the similarity function can be broken down into several components, in particular those distinguished by the feature model: commonality and difference. Indeed, the shortest path between two concepts can be seen as the difference between the two concepts (considering that all specialisation add properties to a concept). More particularly, in trees, or under specific constraints in graphs, we have seen that the shortest path linking two concepts contains their LCA. It can therefore be broken down into two parts corresponding to the shortest paths which link compared concepts to their LCA: in most cases, $sp(u, isa^*, v) = sp(u, isa, LCA(u,v)) + sp(v, isa, LCA(u,v))$. Therefore, the LCA can be seen as a *proxy* which partially summarises the commonality of compared concepts[37]. Distances between compared concepts and their LCA can therefore be used to estimate their differences.

The fact that measures can be broken down into specific components evaluating commonalities and differences is central in the design of the approaches which will further be introduced: the feature-based strategy and the information theoretical strategy. As we will see, they mainly define alternative strategies to characterise compared concepts in order to express semantic measures as a function of their commonalities and differences.

---

[36] The algorithmic complexity of the LCA computation is significantly lower than the computation of the shortest path: constant after linear preprocessing (Harel and Tarjan, 1984).

[37] The LCA only partially summarises commonality. Indeed, it can only be considered as an upper-bound of the commonality since highly similar concepts (`Man`, `Women`) may have a general concept for LCA (`LivingBeing`). This LCA will only encompass a partial amount of their commonalities. Please refer to Section 3.3.2. In addition, notice that in some cases the set of NCCAs contains other concepts than the LCA.

### 3.4.2 Feature-based approach

The feature-based approach generally refers to measures which rely on a taxonomic interpretation of the feature model proposed by Tversky (1977) (introduced in Section 1.2.1). However, as we will see, contrary to the original definition of the feature model, this approach is not necessarily framed in set theory[38].

The main idea is to represent concepts as collections of features, i.e., characteristics describing the concepts, to further express measures based on the analysis of their common and distinct features. The score of the measures will only be influenced by the strategy adopted to characterise concept features[39], and the strategy adopted for their comparison.

As we will see, the reduction of concepts to collections of features makes it possible to set the semantic similarity estimation back in the context of classical binary similarity or distance measures (e.g., set-based measures).

An approach commonly used to represent the features of a concept is to consider its ancestors as features[40]. We denote $A(u)$ the set of ancestors of the concept $u$. Since the Jaccard index that was proposed 100 years ago, numerous binary measures have been defined in various fields. A survey of these measures distinguishes 76 of them in Choi et al. (2010). Considering that the features of a concept $u$ are defined by $A(u)$, an example of a semantic similarity measure expressed from the Jaccard index was proposed in Maedche and Staab (2001)[41]:

$$sim_{CMatch}(u,v) = \frac{|A(u) \cap A(v)|}{|A(u) \cup A(v)|} \quad (3.22)$$

Another example of a set-based expression of the feature-based approach is proposed in Bulskov et al. (2002):

$$sim_{Bulskov}(u,v) = \alpha \frac{|A(u) \cup A(v)|}{|A(u)|} + (1-\alpha) \frac{|A(u) \cup A(v)|}{|A(v)|} \quad (3.23)$$

with $\alpha \in [0,1]$ a parameter used to tune the symmetry of the measure.

Rodríguez and Egenhofer (2003) also proposed a formulation derived from the *ratio model* defined by Tversky (introduced in Section 1.2.1):

$$sim_{RE}(u,v) = \frac{|A(u) \cap A(v)|}{\gamma|A(u) \setminus A(v)| + (1-\gamma)|A(v) \setminus A(u)| + |A(u) \cap A(v)|} \quad (3.24)$$

with $\gamma \in [0,1]$, a parameter that enables the tuning of the symmetry of the measure.

Sánchez et al. (2012) define the taxonomic distance of two concepts as a function of the ratio between their distinct and shared features:

$$dist_{Sanchez}(u,v) = log_2 \left(1 + \frac{|A(u) \setminus A(v)| + |A(v) \setminus A(u)|}{|A(u) \setminus A(v)| + |A(v) \setminus A(u)| + |A(u) \cap A(v)|}\right) \quad (3.25)$$

Various refinements of these measures have been proposed, e.g., to enrich concept features by taking their descendants into account (Ranwez et al., 2006).

The feature-based measures may not be intentional, i.e., they are not expected to solely rely on the knowledge defined in the taxonomy. When instances of the concepts are known, the feature of a concept can also be seen by extension and be defined on the basis of instances associated to concepts. As an example, the Jaccard index can be used to compare two ordered concepts according to their shared and distinct features, here characterised by extension:

$$sim_{JacExt}(u,v) = \frac{|\mathcal{I}(u) \cap \mathcal{I}(v)|}{|\mathcal{I}(u) \cup \mathcal{I}(v)|} \quad (3.26)$$

with $\mathcal{I}(u) \subseteq I$ the set of instances of the concept $u$. Note that this approach makes no sense if the desire is to compare concepts which are not ordered – the set $\mathcal{I}(u) \cap \mathcal{I}(v)$ will tend to be empty.

---

[38]You will recall that the feature matching function on which the feature model is based, relies on binary evaluations of the features *"In the present theory, the assessment of similarity is described as a feature-matching process. It is formulated, therefore, in terms of the set-theoretical notion of a matching function rather than in terms of the geometric concept of distance"* (Tversky and Itamar, 1978).

[39]As stressed in Schickel-Zuber and Faltings (2007), there is a narrow link with the multi-attribute utility theory (Keeney, 1993) in which the utility of an item is a function of the preference on the attributes of the item.

[40]Its implicit senses if the concept refers to a synset.

[41]This is actually a component of a more refined measure.

D'Amato et al. (2008) also define an extensional measures considering:

$$sim_{D'Amato}(u,v) = \frac{min(|\mathcal{I}(u)|, |\mathcal{I}(v)|)}{|\mathcal{I}(LCA(u,v))|} \left(1 - \frac{|\mathcal{I}(LCA(u,v))|}{|I|}\right) \left(1 - \frac{min(|\mathcal{I}(u)|, |\mathcal{I}(v)|)}{|\mathcal{I}(LCA(u,v))|}\right) \quad (3.27)$$

Classical feature-based measures summarise the features of a concept through a set representation which generally corresponds to a set of concepts or instances. However, alternative approaches can also be explored. Therefore, even if, to our knowledge, such approaches have not been defined, the features of a concept could also be represented as a set of relationships, as a subgraph, etc.

In addition, regardless of the strategy adopted to characterise the features of a concept (other concepts, relationships, instances), the comparison of the features is not necessarily driven by a set-based measure. Indeed, the collections of features can also be seen as vectors. As an example, a concept $u$ can be represented by a vector $U$ in a chosen real space of dimension $|C|$, e.g., considering that each dimension associated to an ancestor of $u$ is set to 1. Vector-based measures will evaluate the distance of two concepts by studying the coordinates of their respective projections.

In this vein, Bodenreider et al. (2005) proposed the comparison of two concepts according to their representation through the Vector Space Model. Considering a concept-to-instance matrix, a weight corresponding to the IC[42] of the concept $u$ is associated to the cell $(u, i)$ of the matrix if the instance $i \in \mathcal{I}(u)$. The vectors representing two concepts are then compared using the classical dot product of the vectors presented in Section 2.3.2.

### 3.4.3 Information theoretical approach

The information theoretical approach relies on Shannon's information theory (Shannon, 1948). As with the feature-based strategy, these measures rely on the comparison of two concepts according to their commonalities and differences, here defined in terms of information. This approach formally introduces the notion of salience of concepts through the definition of their informativeness Information Content (IC) – Section 3.3.2 introduces the notion of IC.

Resnik (1995) defines the similarity of a couple of concepts as a function of the IC of their common ancestor which maximises an IC function (originally $eIC$), i.e., their Most Informative Common Ancestor (MICA).

$$sim_{Resnik}(u,v) = IC(MICA(u,v)) \quad (3.28)$$

Resnik's measure doesn't explicitly capture the specificities of compared concepts. Indeed, pairs of concepts with an equivalent MICA will have the same semantic similarity, whatever their respective ICs. To correct this limitation, several authors refined the measure proposed by Resnik to incorporate specificities of compared concepts. We here present the measures proposed by Lin (1998a)[43] – $sim_{Lin}$, (Jiang and Conrath, 1997) – $dist_{JC}$, (Mazandu and Mulder, 2013) – $sim_{Nunivers}$, (Pirró and Seco, 2008; Pirró, 2009) – $sim_{PSec}$ and (Pirró and Euzenat, 2010b) – $sim_{Faith}$:

$$sim_{Lin}(u,v) = \frac{2 \cdot IC(MICA(u,v))}{IC(u) + IC(v)} \quad (3.29)$$

$$dist_{JC}(u,v) = IC(u) + IC(v) - 2 \cdot IC(MICA(u,v)) \quad (3.30)$$

$$sim_{Nunivers}(u,v) = \frac{IC(MICA(u,v))}{max(IC(u), IC(v))} \quad (3.31)$$

$$sim_{PSec}(u,v) = 3 \cdot IC(MICA(u,v)) - IC(u) - IC(v) \quad (3.32)$$

$$sim_{Faith}(u,v) = \frac{IC(MICA(u,v))}{IC(u) + IC(v) - IC(MICA(u,v))} \quad (3.33)$$

Taking into account specificities of compared concepts can lead to high similarities (low distances) when comparing general concepts. As an example, when comparing general concepts using $sim_{Lin}$, the maximal similarity will be obtained comparing a (general) concept to itself. In fact, the identity of the indiscernibles is generally ensured (except for the root which generally has an IC equal to 0). However,

---

[42] Originally the authors used the IDF but we saw that both the IC and the IDF are similar (Section 3.3.2).
[43] Originally defined as: $sim_{Lin}(u,v) = \frac{2 \times log(MICA(u,v))}{log(u) + log(v)}$

some treatments require this property not to be respected. Authors have therefore proposed to lower the similarity of two concepts according to the specificity of their MICA, e.g. (Schlicker et al., 2006; Li et al., 2010). The measure proposed by Schlicker et al. (2006) is presented:

$$sim_{Rel}(u,v) = sim_{Lin}(u,v) \times (1 - p(MICA(u,v))) \quad (3.34)$$

with $p(MICA(u,v))$ the probability of occurrence of the MICA. An alternative approach proposed by Li et al. (2010) relies on the IC of the MICA and can therefore be used without extensional information on concepts, i.e., using an intrinsic expression of the IC.

Authors have also proposed to characterise the information carried by a concept by summing the ICs of its ancestors (Mazandu and Mulder, 2011; Cross and Yu, 2011):

$$sim_{Mazandu}(u,v) = \frac{2 \cdot \sum_{c \in A(u) \cap A(v)} IC(c)}{\sum_{c \in A(u)} IC(c) + \sum_{c \in A(v)} IC(c)} \quad (3.35)$$

$$sim_{JacAnc}(u,v) = \frac{\sum_{c \in A(u) \cap A(v)} IC(c)}{\sum_{c \in A(u) \cup A(v)} IC(c)} \quad (3.36)$$

These measures can also be considered as hybrid strategies between the feature-based and information theory approaches. One can consider that these measures rely on a redefinition of the way to characterise the information conveyed by a concept (by summing the IC of the ancestors). Other interpretations can simply consider that features are weighted. Thus, following the set-based representations of features, authors have also studied these measures as fuzzy measures (Cross, 2004; Popescu et al., 2006; Cross, 2006; Cross and Sun, 2007; Cross and Yu, 2010, 2011), e.g., defining the membership function of a feature corresponding to a concept as a function of its IC.

Finally, other measures based on information theory have also been proposed, e.g., (Maguitman and Menczer, 2005; Maguitman et al., 2006; Cazzanti and Gupta, 2006). As an example, in Maguitman and Menczer (2005) the similarity is estimated as a function of prior and posterior probability regarding instances and concept membership.

### 3.4.4 Hybrid approach

Other techniques take advantage of the various aforementioned paradigms. Among the numerous proposals, (Jiang and Conrath, 1997; Bin et al., 2009) defined measures in which density, depth, strength of connotation and ICs of concepts are taken into account. We present the measure proposed by Jiang and Conrath (1997)[44]. The strength of association $w(u,v)$ between two concepts $u$, $v$ is defined as follows:

$$w(u,v) = (\beta + (1-\beta)) \frac{\overline{dens}}{|children(v)|} \times \left( \frac{depth(v)+1}{depth(v)} \right)^\alpha \times (IC(u) - IC(v)) \times T(u,v)$$

The factor $\overline{dens}$ refers to the average density of the whole taxonomy, see Jiang and Conrath (1997) for details. The factors $\alpha \geq 0$ and $\beta \in [0,1]$ control the importance of the density factor and the depth respectively. $T(u,v)$ defines weights associated to predicates. Finally, the similarity is defined by the weight of the shortest path which links compared concepts and which contains their LCA:

$$dist_{JC-Hybrid}(u,v) = \sum_{(s,p,o) \in sp(u,isa,LCA(u,v)) \cup sp(v,isa,LCA(u,v))} w(s,o)$$

Defining $\alpha = 0$, $\beta = 1$ and $T(u,v) = 1$, we obtain the information theoretical measure proposed by the same authors, i.e., $dist_{JC}(u,v) = IC(u) + IC(v) - 2 \cdot IC(MICA(u,v))$ (Equation 3.30).

Singh et al. (2013) proposed a mixing strategy based on (Jiang and Conrath, 1997) IC-based measure $dist_{JC}$. They consider transition probabilities between concepts relying on a depth-based estimation of the strength of connotation.

Rodríguez and Egenhofer (2003) also proposed mixing a feature-based approach considering structural properties such as the concepts' depth. Finally, Paul et al. (2012) defined multiple measures based on an aggregation of several existing measures.

---

[44]This measure is a parametric distance. Couto et al. (2003) discuss the implementation, Othman et al. (2008) propose a genetic algorithm which can be used to tune the parameters and Wang and Hirst (2011) propose a redefinition of the notion of depth and density initially proposed.

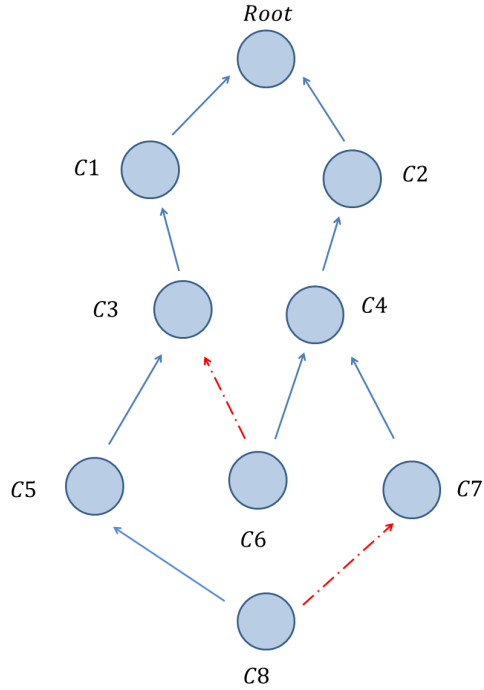

Figure 3.5: The graph composed of the plain (blue) edges is a taxonomic tree, i.e., it doesn't contain concepts with multiple parents. If the (red) dotted relationships are also considered, the graph is a directed acyclic graph (e.g., a taxonomic graph)

### 3.4.5 Considerations when comparing concepts in semantic graphs

Several measures introduced in the previous sections were initially defined to compare concepts expressed in a tree. However, despite the fact that this subject is almost never discussed in the literature, several considerations must be taken into account in order to estimate the similarity of concepts defined in a semantic graph in which multiple inheritances can exist – please refer to notations introduced in Section 3.1.

**Shortest path**

A tree is a specific type of graph in which multiple inheritances cannot be encountered, i.e. $\forall c \in C, |parents(c)| < 2$. This implies that two concepts $u, v$ which are not ordered will have no common descendants, i.e., $G_T^-(u) \cap G_T^-(v) = \emptyset$. Therefore, if there is no redundant taxonomic relationship, the shortest path which links $u, v$ always contains a single common ancestor of $u, v$: $LCA(u, v)$. However, in a graph, since two non-ordered concepts $u, v$ can have common descendants, i.e., $G_T^-(u) \cap G_T^-(v) \neq \emptyset$, the shortest path which links $u, v$ can in some cases not contain one of their common ancestors. Figure 3.5 illustrates the modifications induced by multiple inheritances.

In Figure 3.5, the shortest path linking the two non-ordered concepts $C5$ and $C7$ in the tree (i.e. without considering red dotted edges) is $[C5 - C3 - C1 - Root - C2 - C4 - C7]$. However, if we consider multiple inheritances (red dotted edges), it is possible to link $C5$ and $C7$ through paths which do not contain one of their common ancestors, e.g., $[C5 - C3 - C6 - C4 - C7]$ or even $[C5 - C8 - C7]$. The shortest path which contains a common ancestor of the compared concepts is defined in the search space corresponding to the graph $G_T^+(u) \cup G_T^+(v)$. In practice, despite the fact that in most graphs $G_T^-(u) \cap G_T^-(v) \neq \emptyset$ (for two non-ordered concepts), it is commonly admitted that the shortest path must contain a single ancestor of the two compared concepts. Given this constraint, the edge-counting taxonomic distance of $u$ and $v$ in $G_T^+(u) \cup G_T^+(u)$ is generally (implicitly[45]) defined by: $dist_{SP}(u, v) =$

---

[45] Generalisation of measures defined from trees to graphs is poorly documented in the literature.

$sp(u, isa, LCA(u,v)) + sp(v, isa, LCA(u,v))$.

Note that when non comparable common ancestors (NCCAs) are shared between compared concepts, the ancestor which maximises the similarity is expected to be considered. Depending on the $\theta$ function which is used, the shortest path doesn't necessarily involve the concept of the NCCAs which maximise $\theta$, e.g. the deeper in the taxonomy. As an example, in order to distinguish which NCCA to consider, Schickel-Zuber and Faltings (2007) took into account a mix between depth and reinforcement (number of different paths leading from one concept to another).

Nevertheless, the shortest path techniques can also be relaxed to consider paths which do not involve common ancestors or which involve multiple common ancestors:

$$sim_{SP-R}(u,v) = \frac{1}{sp(u, isa^*, v) + 1}$$

**Notion of depth**

The definition of the notion of depth must also be reconsidered when the taxonomy is not a tree. Remember that, in a tree without redundancies, the depth of a concept has been defined as the length of the shortest path linking the concept to the root. The depth of a concept is a simple example of specificity estimator. In a tree, this estimator makes perfect sense since the depth of a concept is directly correlated to its number of ancestors since $depth(c) = |A(c)| - 1$.

In a graph, or in a tree with redundant taxonomic relationships, we must ensure that the depth is monotonically decreasing according to the ordering of concepts. As an example, to apply depth-based measures to graphs, we must ensure that $depth(LCA(u,v))$ is lower or equal to both $depth(u)$ and $depth(v)$. To this end, the maximal depth of a concept must be used, i.e., the length of the longest path in $\{u, isa, \top\}$, denoted $lp(u, isa, \top)$. As an example, the measure proposed by Pekar and Staab (2002) – Equation 3.17 – is therefore implicitly generalised to:

$$sim_{PS-G}(u,v) = \frac{lp(LCA(u,v), isa, \top)}{lp(u, isa, LCA(u,v)) + lp(v, isa, LCA(u,v)) + lp(LCA(u,v), isa, \top)}$$

**Notion of least common ancestors**

Most measures which have been presented take advantage of the notions of LCA or MICA. However, in graphs, these measures do not consider the whole set of NCCAs – denoted $\Omega(u,v)$ for the concepts $u$ and $v$. To remedy this, several authors have proposed adaptations of existing measures. As an example, Couto et al. (2005); Couto and Silva (2011) proposed *GraSM* and *DiShIn* strategies.

In (Couto et al., 2005) the authors proposed the modification of information theoretical measures based on the notion of MICA. The authors recommended substituting the IC of the MICA by the average of the ICs of the concepts which compose the set of NCCAs. A redefinition of the measure proposed by (Lin, 1998a) – Equation 3.29 – is presented:

$$sim_{Lin-GraSM}(u,v) = \frac{2 \cdot \frac{\sum_{c \in \Omega(u,v)} IC(c)}{|\Omega(u,v)|}}{IC(u) + IC(v)} \tag{3.37}$$

$$= \frac{2 \cdot \sum_{c \in \Omega(u,v)} IC(c)}{|\Omega(u,v)| \times (IC(u) + IC(v))}$$

Wang et al. (2012b) also proposed averaging the similarity between the concepts according to their multiple NCCAs:

$$sim_{Wang}(u,v) = \frac{\sum_{a \in \Omega(u,v)} \frac{2 \cdot depth(a)^2}{d_a(\top,u) \times d_a(\top,v)}}{|\Omega(u,v)|}$$

With $d_a(\top, u)$ the average length of the set of paths which contain the concept $a$ and which link the concept $u$ to the root of the taxonomy ($\top$).

As we have underlined, numerous approaches have been defined to compare pairs of concepts defined in a taxonomy, these measures can be used to compare any pair of nodes defined in a poset. Table 3.2 to Table 3.5 present some properties of a selection of measures defined to compare pairs of concepts.

### 3.4.6 List of pairwise semantic similarity measures

Several semantic measures which can be used to compare concepts defined in a taxonomy or any pair of elements defined in a poset. Measures are ordered according to their date of publication. Other contributions studying some properties of pairwise measures can be found in Yu (2010); Slimani (2013). IOI: Identify of the Indiscernibles. Some of the values associated to specific measures have not been complete yet. This is generally because the reference associated to the measure was not available or because the properties of the measure are still under study.

| | Structural Measures | | | | |
|---|---|---|---|---|---|
| Name | Type | Const. | Range | IOI | Comment |
| Shortest Path strategy | Sim / Rel | None | $\mathbb{R}_+$ | Yes | Measures are defined as a function of the weight of the shortest path linking compared concepts. Several modifications can be considered depending on the strategy adopted, e.g., weighing of the relationships (according to their predicate), constraints on the inclusion of a common ancestor of the compared concepts, etc. |
| Rada et al. (1989) | Dist (ISA) | DAG | $\mathbb{R}_+$ | Yes | Specific shortest path strategy with uniform edge weight. The shortest path is constrained to containing the LCA of the compared concepts. |
| Young Whan and Kim (1990) | x | x | x | x | x |
| Lee et al. (1993) | x | x | x | x | x |
| Sussna (1993) | Dist | RDAG | $\mathbb{R}_+$ | Yes | Originally defined as a parametric semantic relatedness. Under specific constraints, this measure can be used as a semantic similarity. Shortest path technique which take into account non-uniform strengths of connotation. This latter is tuned according to the depth of the compared concepts and to the weights associated to predicates. |
| Richardson et al. (1994) | x | x | x | x | The authors propose several intrinsic metrics (e.g., depth, density) to weigh the relationships and define hybrid measures by mixing the structural and information theoretical approach. No measure explicitly is defined. |
| Wu and Palmer (1994) | Sim | RDAG | [0,1] | Yes | The similarity is assessed as a function of the depth of the compared concepts and the depth of their LCA. |

| Reference | Type | Graph | Range | Norm. | Description |
|---|---|---|---|---|---|
| Leacock and Chodorow (1994, 1998) | Sim | RDAG | $\mathbb{R}_+$ | Yes | Rada et al. (1989) formulation penalising long shortest path between the compared concepts according to the depth of the taxonomy. |
| Resnik (1995) | Sim | RDAG | $[0, 2D]$ | Yes | Similarity based on the shortest path technique which has been bound by (twice) the depth of the taxonomy (D). |
| Hirst and St-Onge (1998) | Sim / Rel | None | $\mathbb{R}_+$ | Yes | Shortest path penalising multiple changes of predicate. Can be used as a similarity or relatedness measure depending on the relationships which are considered. |
| Zhong et al. (2002) | Dist | RDAG | $[0, max[$ | Yes | Taxonomic distance taking into account the depth of the compared concepts. With $max$ defined as $max = 1/k^{depth(LCA(u,v))}$ with $k$ a given constant. |
| Pekar and Staab (2002) | Sim | RDAG | $[0,1]$ | Yes | Shortest path technique which takes into account the depth of the LCA of the compared concepts. |
| Mao and Chu (2002) | Sim | DAG | $\mathbb{R}_+$ | No | Modification of Rada et al. (1989) formulation taking into account concept specificity as a non-linear function of the number of descendants a concept has. |
| Li et al. (2003) | x | x | x | x | x |
| Li et al. (2006) | Sim | RDAG | $\mathbb{R}_+$ | No | Measure considering both the length of the shortest path linking the compared concepts and their depth. |
| Ganesan et al. (2003) | Sim | x | x | x | Refer to the function named *leafsim* in the publication. |
| Yu et al. (2005) | Sim | RDAG | $[0,1]$ | Yes | Measure allowing non-null similarity only to ordered pairs of concepts. |
| Wu et al. (2006) | Sim | RDAG | [0,1] | No | Take into account compared concepts (i) commonality (length of the longest shared path from the concepts to the root), (ii) specificity (defined as a function of the shortest path from the concept to the leaves it subsumes) and (iii) local distance (Rada et al. (1989) distance). |
| Slimani et al. (2006) | Sim | RDAG | [0,1] | Yes | Modification of Wu and Palmer (1994) measure to avoid cases in which neighbours of a concept might have higher similarity values than concepts which are ordered with it. |
| Blanchard et al. (2006) | Sim | RDAG | [0,1] | No | Measure which compare two concepts w.r.t their depth and the depth of their LCA. Originally defined for trees and extended for DAG in (Blanchard, 2008) |
| Nagar and Al-Mubaid (2008) | Sim | DAG | $\mathbb{R}_+$ | Yes | Use a modification of shortest path constrained by the LCA. |
| Cho et al. (2003) | Sim | DAG | $\mathbb{R}_+$ | No | Multiple factors are considered to take into account the specificity of the compared concepts. |
| Alvarez and Yan (2011) (SSA) | Sim / Rel | RDAG | [0,1] | No | This measure relies on three components evaluating (i) the shortest path linking the compared concepts (a weighing scheme is applied to the graph), (ii) their LCA, and (iii) their literal definitions. |
| Wang et al. (2012b) | Sim | RDAG | [0,1] | Yes | Approach taking into account the depth of the compared concepts, as well as the depth of all their DCAs. |
| Shenoy et al. (2012) | Sim | RDAG | x | x | Modification of Wu and Palmer (1994) measure to avoid cases in which neighbours of a concept might have higher similarity values than concepts which are ordered with it. |

| | Sim | RDAG | | |
|---|---|---|---|---|
| Ganesan et al. (2012) | | x | x | Modification of Wu and Palmer (1994) measure to avoid cases in which neighbours of a concept might have higher similarity values than concepts which are ordered with it. |

Table 3.2: Semantic similarity measures or taxonomic distances defined using a structural approach. These measures can be used to compare a pair of concepts defined in a taxonomy or any pair of elements defined in a partially ordered set

| Information theoretical Measures | | | | | |
|---|---|---|---|---|---|
| Name | Type | Const. | Range | IOI | Comment |
| Resnik (1995) | Sim | DAG | $[0,1], [0,+\infty[$ | No | The similarity is defined as the IC of MICA. The range of the measure depends on the IC. |
| Jiang and Conrath (1997) | Dist | DAG | $[0,1]$ | Yes | The taxonomic distance computed as a function of the IC of the compared classes and their MICA. |
| Lin (1998a) | Sim | DAG | $[0,1]$ | Yes | The similarity is computed as a ratio between the IC of the MICA of compared classes and the sum of their respective ICs. |
| Schlicker et al. (2006) | Sim | DAG | $[0,1]$ | No | Modification of Lin (1998a) in order to take into account the specificity of the MICA, i.e., to avoid high score of similarity comparing two general classes (due to the ratio). |
| Couto et al. (2007) | Sim | DAG | $[0,1]$ | No | Derivative of Lin (1998a) measure in which all the DCAs of the compared classes are taken into account. |
| Yu et al. (2007b) | Sim | DAG | $[0,+\infty[$ | No | Total Ancestry Measure (TAM) – measure based on a specific definition of the LCA. |
| Pirró and Seco (2008); Pirró (2009) | Sim | DAG | $[0,x]$ | No | With $x$ the maximal IC value. Formulation similar to Jiang and Conrath (1997) but which gives more importance to the informativeness of the MICA. |
| Li et al. (2010) | Sim | DAG | $[0,1]$ | No | Lin (1998a) measure modified to take specificity into account, i.e. to avoid high score of similarity comparing two general classes. |
| Pirró and Euzenat (2010b) | Sim | DAG | $[0,1]$ | Yes | Ratio formulation similar to the measure proposed by Lin (1998a) but which gives more importance to the difference between the compared concepts. |
| Mazandu and Mulder (2011) simDIC | Sim | DAG | [0,1] | Yes | Measure similar to Lin (1998a) but which uses a new approach to characterise the IC of a concept. |
| Mazandu and Mulder (2013) Nunivers | Sim | DAG | [0,1] | Yes | IC of the MICA of the compared classes divided by the maximal IC of the compared classes. |

Table 3.3: Semantic similarity measures or taxonomic distances defined using an information theoretical approach. These measures can be used to compare a pair of concepts defined in a taxonomy or any pair of elements defined in a partially ordered set

| Feature-based Measures | | | | | |
|---|---|---|---|---|---|
| Name | Type | Const. | Range | IOI | Comment |
| Maedche and Staab (2001) | Sim | DAG | [0,1] | Yes | Feature-based expression relying on the Jaccard index. |
| Bodenreider et al. (2005) | Sim | DAG | [0,1] | x | Cosine similarity on a vector-based representation of the classes. The vector representation is built according to the set of instances of the classes. |
| Ranwez et al. (2006) | Dist | DAG | $\mathbb{R}_+$ | Yes | The distance is defined as a function of the number of descendants of the LCA of the compared concepts. This measure respects distance axioms (i.e., positivity, symmetry, triangle inequality). |
| Jain and Bader (2010) | Sim | DAG | [0,1] | No | Build a meta-graph reducing the original ontology into cluster of related concepts. Similarity is assessed through a specific function evaluating LCA information content. |
| Batet et al. (2010b) | Sim | DAG | $\mathbb{R}_+$ | Yes | Comparison of the classes according to their ancestors. Formulation expressed as a distance converted to a similarity using negative log. |

Table 3.4: Semantic similarity measures or taxonomic distances designed using a feature-based approach. These measures can be used to compare a pair of concepts defined in a taxonomy or any pair of elements defined in a partially ordered set

| | | Hybrid Measures | | | |
|---|---|---|---|---|---|
| Name | Type | Const. | Range | IOI | Comment |
| Jiang and Conrath (1997); Couto et al. (2003); Othman et al. (2008) | Sim / Dist | RDAG | [0,1] | Var. | Strategy based on the shortest path constrained by the LCA of the compared classes. Relationships are weighted according to the difference of IC of the classes they link. |
| Al-Mubaid and Nguyen (2006) | Sim | DAG | $[0,+\infty[$ | Yes | Assigns cluster(s) to classes. The similarity is computed considering multiple metrics. |
| Wang et al. (2007) | Sim/Rel | RDAG | [0,1] | Yes | This measure as originally been defined as a semantic relatedness. It can also be used to compute semantic similarity. It relies on a non-linear approach to characterise the strength of connotation and a specific approach to characterise the informativeness of a concepts. |
| Alvarez and Yan (2011) | Sim / Rel | RDAG | [0,1] | No | Exploits three components evaluating concepts, their shortest path (a weighting scheme is applied to the graph), their LCA, and their literal definitions. |
| Paul et al. (2012) | Sim | x | x | x | Multiple approaches are mixed |

Table 3.5: Semantic similarity measures or taxonomic distances designed using an hybrid approach. These measures can be used to compare a pair of concepts defined in a taxonomy or any pair of elements defined in a partially ordered set

## 3.5　Semantic similarity between groups of concepts

Two main approaches are commonly distinguished to introduce semantic similarity measures designed for the comparison of two sets of concepts, i.e., *groupwise measures*:

- *Direct approach*, the measures which can be used to directly compare the sets of concepts according to information characterising the sets w.r.t the information defined in the taxonomy.
- *Indirect approach*, the measures which assess the similarity of two sets of concepts using one or several pairwise measures, i.e. measures designed for the comparison of a pair of concepts. They are generally simple aggregations of the scores of similarities associated to the pairs of concepts defined in the Cartesian product of the two compared sets.

Note that the sets are generally expected to not contain semantically redundant concepts, i.e., they do not contain any pair of ordered concepts – $\forall (u,v) \in X, u \npreceq v \land v \npreceq u$.

Once again, a large diversity of measures have been proposed, some of which are presented in the next subsections.

### 3.5.1　Direct approach

The direct approach corresponds to a generalisation of the approaches defined for the comparison of pairs of concepts in order to compare two sets of concepts. It is worth noting that classical set-based approaches can be used. The sets can also be compared through their vector representations, e.g., using the cosine similarity measure. Nevertheless, these measures are in most cases not relevant to be used considering the semantics they convey – they do not take into account the similarity of the elements composing compared sets[46], e.g., $sim(\{\texttt{Man},\texttt{Girl}\}, \{\texttt{Women},\texttt{Boy}\}) = 0$.

**Structural approach**

Considering $G_T^+(X)$ as the graph induced by the union of the ancestors of the concepts which compose the set $X$, Gentleman (2007) defined the similarity of two sets of concepts $(U,V)$ according to the length of the longest $sp(c, isa, \top)$ which links the concept $c \in G_T^+(U) \cap G_T^+(V)$ to the root ($\top$).

**Feature-based approach**

The feature-based measures are characterised by the approach adopted to express the features of a set of concepts.

Several measures have been proposed from set-based measures. We introduce $sim_{UI}$ (Gentleman, 2007)[47], and the Normalised Term Overlap measure $sim_{NTO}$ (Mistry and Pavlidis, 2008). For convenience, we consider $C_T^+(X)$ as the set of concepts contained in $G_T^+(X)$:

$$sim_{UI}(U,V) = \frac{|C_T^+(U) \cap C_T^+(V)|}{|C_T^+(U) \cup C_T^+(V)|} \tag{3.38}$$

$$sim_{NTO}(U,V) = \frac{|C_T^+(U) \cap C_T^+(V)|}{min(|C_T^+(U)|, |C_T^+(V)|)} \tag{3.39}$$

**Information theoretical measures**

Among others, Pesquita et al. (2007) proposed considering the information content of the concepts (originally an *eIC* expression):

$$sim_{GIC}(U,V) = \frac{\sum_{c \in C_T^+(U) \cap C_T^+(V)} IC(c)}{\sum_{c \in C_T^+(U) \cup C_T^+(V)} IC(c)} \tag{3.40}$$

### 3.5.2　Indirect approach

In Section 3.4, we introduced numerous measures for comparing a pair of concepts (pairwise measures). They can be used to drive the comparison of sets of concepts.

---

[46]These simple approaches are generally used when the compared sets contain semantically redundant concepts.
[47]Also published through the name *Term Overlap* (TO) in Mistry and Pavlidis (2008).

**Improvements of direct measures using concept similarity**

One of the main drawbacks of basic vector-based measures is that they consider dimensions as mutually orthogonal and do not exploit concept relationships. In order to remedy this, vector-based measures have been formulated to:

- Weigh dimensions considering concept specificity evaluations (e.g., IC) (Huang et al., 2007; Chabalier et al., 2007; Benabderrahmane et al., 2010b).

- Exploit an existing pairwise measure to perform vector products (Ganesan et al., 2003; Benabderrahmane et al., 2010b).

Therefore, pairwise measures can be used to refine the measures proposed to compare sets of concepts using a direct approach.

**Aggregation strategies**

A two-step indirect strategy can also be adopted in order to take advantage of pairwise measures to compare sets of concepts:

1. The similarity of pairs of concepts obtained from the Cartesian product of the two compared sets has to be computed.

2. Pairwise scores are then summarised using an aggregation strategy, also called mixing strategy in the literature.

Classic aggregation strategies can be applied (e.g. max, min, average); more refined strategies have also been proposed. Among the most commonly used we present: Max average (Max-Avg), Best Match Max – BMM (Schlicker et al., 2006) and Best Match Average – BMA (Pesquita et al., 2008):

$$sim_{Avg}(U,V) = \frac{\sum_{u \in U} \sum_{v \in V} sim(u,v)}{|U| \times |V|} \tag{3.41}$$

$$sim_{Max-Avg}(U,V) = \frac{1}{|U|} \sum_{u \in U} max_{v \in V} sim(u,v) \tag{3.42}$$

$$sim_{BMM}(U,V) = max(sim_{Max-Avg}(U,V), sim_{Max-Avg}(V,U)) \tag{3.43}$$

$$sim_{BMA}(U,V) = \frac{sim_{Max-Avg}(U,V) + sim_{Max-Avg}(V,U)}{2} \tag{3.44}$$

### 3.5.3 List of groupwise semantic similarity measures

| | | Direct Groupwise Measures | | | | |
|---|---|---|---|---|---|---|
| Name | Type | Approach | Const. | Range | IOI | Comment |
| (Ganesan et al., 2003) Optimistic Genealogy Measure | Sim | Hybrid | RDAG | [0,1] | Yes | Feature-based approach taking into consideration structural properties during the comparison. |
| (Popescu et al., 2006) A | Sim | Feature-based | DAG | [0,1] | yes | Weighted Jaccard |
| (Popescu et al., 2006) B | Sim | Feature-based | DAG | [0,1] | x | Fuzzy Measure |
| (Chabalier et al., 2007) | Sim | Feature-based (Vector) | RDAG | [0,1] | Yes | Groups of concepts are represented using the Vector Space Model and compared using the cosine similarity. |
| (Gentleman, 2007) SimLP | Sim | Structural | RDAG | [0,1] | Yes | Similarity as a function of the longest common path found in the graph induced by the compared groups of concepts. |
| (Cho et al., 2007) | Sim | Feature-based | RDAG | $\mathbb{R}_+$ | No | Feature-based measure taking into account the specificity of the compared concepts. |
| (Pesquita et al., 2008) SimGIC | Sim | Feature-based | DAG | [0,1] | Yes | Jaccard measure in which a set of concepts is represented by the concepts contained in the graph it induces. |
| (Sheehan et al., 2008) SSA | Sim | Feature-based | RDAG | [0,1] | Yes | Extends the notion of MICA to pair of groups of concepts then redefine the Dice coefficient. |
| (Ali and Deane, 2009) | Sim | Feature-based | DAG | [0,1] | No | Commonality is assessed considering shared nodes in the graph induced by the ancestors of the compared sets of concepts. |
| (Jain and Bader, 2010) TCSS | Sim | Feature-based | RDAG | [0,1] | No | Max Strategy considering a specific pairwise measure |
| (Diaz-Diaz and Aguilar-Ruiz, 2011) | Sim | Structural | RDAG | [0,1] | Yes | Distance taking into account the shortest path between the concepts and the depths of the compared concepts. |
| (Alvarez and Yan, 2011) | Sim/Rel | Structural | None | $\mathbb{R}_+$ | Yes | Measure based on the analysis of structural properties of the graph. |
| (Alvarez et al., 2011) SPGK | Sim | Structural | | None | Yes | The set of concepts is represented by the graph induced by the concepts it subsumes. A similarity measure is used to compare the two graphs. |
| (Teng et al., 2013) | Sim | x | | x | x | x |

Table 3.6: Semantic similarity measures or taxonomic distances designed using a direct approach. These measures can be used to compare a pair of groups of concepts defined in a taxonomy or any pair of group of elements defined in a partially ordered set

| Indirect Groupwise Measures based on a direct approach | | | | | | |
|---|---|---|---|---|---|---|
| Name | Type | Approach | Const. | Range | IOI | Comment |
| (Ganesan et al., 2003) GCSM | Sim | Feature-based | RDAG | [0,1] | Yes | GCSM: Generalised Cosine-Similarity Measure. Groups of concepts are represented using the Vector Space Model. Dimensions are not considered independent, i.e. the similarity of two dimensions is computed using an approach similar to the one proposed by Wu and Palmer (1994) measure. The similarity between the vector representations of two groups of concepts is estimated using to the cosine similarity. |
| (Huang et al., 2007) | x | x | RDAG | [0,1] | Yes | x |
| (Benabderrahmane et al., 2010b) Intelligo | Sim | Feature-based (Vector) | RDAG | [0,1] | Yes | Groups of classes are represented using the Vector Space Model - Also consider (Benabderrahmane et al., 2010a). The dimensions are not considered to be independent. |

Table 3.7: Semantic similarity measures or taxonomic distances designed using an indirect approach. These measures can be used to compare a pair of groups of concepts defined in a taxonomy or any pair of group of elements defined in a partially ordered set

| Indirect Groupwise Measures (Mixing strategy) | | |
|---|---|---|
| Mixing strategies | Range | IOI |
| Classic approaches Max/Min/AVG, etc. | depends | depends |
| Best Match Max (BMM) | | |
| Best Match Average (Azuaje et al., 2005) | | |

Table 3.8: Semantic similarity measures or taxonomic distances designed using an indirect approach (mixing strategy). These measures can be used to compare a pair of groups of concepts defined in a taxonomy or any pair of group of elements defined in a partially ordered set

## 3.6 Other knowledge-based measures

### 3.6.1 Semantic measures based on logic-based semantics

Semantic measures based on the relational setting cannot be used to directly compare complex descriptions of classes or instances which rely on logic-based semantics, e.g. description logics (DLs). To this end, semantic measures which are capable of taking into account logic-based semantics have been proposed. They are for instance used to compare complex concept definitions expressed in OWL.

Among the diversity of proposals, measures based on simple DLs, e.g., only allowing concept conjunction (logic $\mathcal{A}$), were initially proposed through extensions of semantic measures based on graph analysis (Borgida et al., 2005). More refined semantic measures have since been designed to exploit high expressiveness of DLs, e.g. $\mathcal{ALC}$, $\mathcal{ALN}$, $\mathcal{SHI}$, $\mathcal{ELH}$ description logics (D'Amato et al., 2005a; Fanizzi and D'Amato, 2006; Janowicz, 2006; Hall, 2006; Araújo and Pinto, 2007; D'Amato et al., 2008, 2005b; Janowicz and Wilkes, 2009; Stuckenschmidt, 2009; Lehmann and Turhan, 2012).

As an example D'Amato et al. (2005a) proposed to compare complex concept descriptions by aggregating functions which consider various components of their $\mathcal{ALC}$ normal forms[48]. These measures rely mostly on extensions of the feature model proposed by Tversky. They have been extensively covered in the thesis of D'Amato (2007).

### 3.6.2 Semantic measures for multiple ontologies

Several approaches have been designed to estimate the relatedness of concepts or instances using multiple ontologies. These approaches are sometimes named cross-ontology semantic similarity/relatedness measures in the literature, e.g., (Petrakis et al., 2006). Their aim is twofold:

- To enable the comparison of elements which have not been defined in the same ontology (the ontologies must model a subset of equivalent elements).

- To refine the comparison of elements by incorporating a larger amount of information during the process.

These measures are in some senses related to those commonly used for the task of ontology alignment/mapping and instance matching (Euzenat and Shvaiko, 2013). Therefore, prior to their introduction we will first highlight the relationship between these measures and those designed for the aforementioned processes.

**Comparison with ontology alignment/mapping and instance matching**

The task of ontology mapping aims at finding links between the classes and predicates defined in a collection of ontologies. These mappings are further used to build an alignment between ontologies. Instance matching focuses on finding similar instances defined in a collection of ontologies. These approaches generally rely on multiple matchers which will be aggregated for evaluating the similarity of the compared elements (Euzenat and Shvaiko, 2013; Shvaiko and Euzenat, 2013). The commonly distinguished matchers are:

- *Terminological* – based on string comparison of the labels or definitions.

- *Structural* – mainly based on the structuration of classes and predicates.

- *Extensional* – based on instance analysis.

---

[48]*Primitives* and restrictions (both existential and universal) are considered.

- *Logic-based* – rely on logical constructs used to define the elements of the ontologies.

The score produced by these matchers is generally aggregated; a threshold is used to estimate if two (groups of) elements are similar enough to define a mapping between them. In some cases, the mapping will be defined between an element and a set of elements, e.g., depending on the difference of granularity of the compared ontologies, a concept can be mapped to a set of concepts. The problem of ontology alignment/mapping and instance matching is a field of study in itself. The techniques used for this purpose involve semantic similarity measures for the design of structural, extensional and logic-based matchers (terminological matchers are not semantic). However, the measures used in this context aim to find exact matches and are therefore generally not suited for the comparison of non-equivalent elements defined in different ontologies. Indeed, techniques used for ontology alignment are for instance not suited to answering questions such as: to which degree are the two concepts `Coffee` and `Cup` related?

In every instance, technically speaking, nothing prevents the use of matching techniques to estimate the similarity between elements defined in different ontologies. Indeed, the problem of knowing if two elements must be considered as equivalent can be reformulated as a function of their degree of semantic similarity. Nevertheless, a clear distinction of the problem of ontology alignment and semantic measure design exists in the literature. This can be partially explained by the fact that, in practice, compared to approaches used for ontology alignment and instance matching, semantic measures based on multiple ontologies:

- Can be used to estimate the semantic relatedness and not only the semantic similarity of compared elements.

- Sometimes rely on strong assumptions and approximations which cannot be considered to derive alignments, e.g., measures based on shortest path techniques.

- Focus on the design of techniques for the comparison of elements defined in different ontologies which generally consider a set of existing mappings between ontologies.

In short, ontology alignment and instance matching are complex processes which use specific types of (semantic) similarity measures and which can be used to support the design of semantic measures involving multiple ontologies. We briefly present the main approaches which have been proposed for the definition of semantic measures based on multiple ontologies.

**Main approaches for the definition of semantic measures using multiple ontologies**

The design of semantic measures for the comparison of elements defined in different ontologies have attracted less attention than classical semantic measures designed for single ontologies. They have been successfully used to support data integration (Rodríguez and Egenhofer, 2003; Lange et al., 2007), clustering (Batet et al., 2010b), or information retrieval tasks (Xiao and Cruz, 2005), to cite a few. In this context, several contributions have focused on the design of semantic measures based on multiple ontologies without focusing on specific application contexts.

The measures proposed in the literature can be distinguished according to the approach they adopt – we consider the same classification used for semantic measures defined for a single ontology (the list of references may not be exhaustive):

- Structural approach: (Al-Mubaid and Nguyen, 2009).

- Feature-based approach: (Petrakis et al., 2006; Batet et al., 2010b, 2013; Solé-Ribalta et al., 2014; Sánchez and Batet, 2013).

- Information Theoretical approach: (Saruladha, 2011; Saruladha and Aghila, 2011; Saruladha et al., 2010a; Sánchez and Batet, 2013; Batet et al., 2014).

- Hybrid approach: (Rodríguez and Egenhofer, 2003).

## 3.7 Advantages and limits of knowledge-based measures

**Advantages**

- They can be used to compare all types of elements defined in an ontology, i.e., terms, concepts, instances. These measures can therefore be used to compare entities which cannot be compared using text analysis. Indeed, knowledge-based measures can be used to compare any entities which is defined into an ontology through their semantic representations. As an example, knowledge-based semantic measures can be used to compare gene products according to conceptual annotations corresponding to their molecular functions, the biological processes in which they are involved or their cellular location.

- They give access to fine control on the semantic relationships taken into account to compare the elements. This aspect is important to understand the semantics associated to a score of semantic measures, e.g., semantic similarity/relatedness.

- Generally easier and less complex to compute than corpus-based measures measures. Indeed, knowledge-based semantic measures do not require complex and time-consuming preprocessing, such as the semantic model building in corpus-based measures. In addition, several efficient measures have been proposed and efficient implementation enable the use of some knowledge-based measures in computational intensive applications. Some problems can however be encountered in using measures which take into account complex logic constructors (e.g. some measures introduced in Section 3.6.1).

**Limits**

- Require an ontology describing the elements to compare. This is a strong limitation if no ontology is available for the domain to consider. Nevertheless, we stress that a large body of literature is dedicated to knowledge base generation/enrichment from text analysis (e.g., refer to the field of ontology learning and Information extraction). Several knowledge base semantic measures could therefore be applied on knowledge base generated by aforementioned text analysis techniques.

- The use of logic-based measures can be challenging to compare elements defined in large ontologies (high computational complexity).

- Measures based on graph analysis generally require the knowledge to be modelled in a specific manner in the graph and are not designed to take non-binary relationships into account. Such relationships are used in specific ontologies and play an important role in defining specific properties to relationships/statements. This can be an issue when reification techniques are used to express such knowledge[49]. However, most measures based on graph analysis are not adapted to this case. This aspect is relative to the mapping of an ontology to a semantic graph.

## 3.8 Mixing knowledge-based and corpus-based approaches

As stated in Chapter 2, corpus-based semantic measures are of particular interest for comparing units of language by taking into account (quantitative) evidence of semantic similarity/relatedness encompassed in texts. In addition to these measures, this chapter has introduced knowledge-based measures that can be used to compare resources characterized by knowledge bases (concepts, instances, annotated objects). A spectrum of solutions may be envisaged to deal with either heterogeneous and unstructured texts on one hand and well-structured and annotated resources on the other hand. This section presents measures that propose to combine the two approaches presented so far, i.e. corpus-based and knowledge-based, into hybrid solutions. Section 3.8.1 introduces generalities about hybrid measures and briefly presents the different approaches that can be used for their definition. In Section 3.8.2, we will in particular focus on Wikipedia-based measures; by relying on the well-known Wikipedia encyclopedia they have the interesting property to take advantage of both a rich corpus of texts and a conceptual organization of categories.

### 3.8.1 Generalities

Hybrid measures have been proposed to take advantage of both corpus-based and knowledge-based semantic measures to compare units of language and entities defined into ontologies. They will not be presented in detail, only references are provided. Most of the time hybrid measures combine several single semantic similarity (Panchenko and Morozova, 2012). Among the various mixing strategies, two broad types of approaches can be distinguished: *Pure-hybrid measures* and *Aggregated measures*.

---

[49]This is done by defining a ternary relationship, i.e., the (binary) relationship is expressed by a node of the graph.

- *Pure-hybrid measures* correspond to measures which are not based on the aggregation of several measures; they are designed by defining a strategy which takes advantage of both corpus and ontology analysis. First and most common examples of pure-hybrid measures are semantic measures based on the information theoretical approach. As an example, (Resnik, 1995) proposed to estimate the amount of information carried by a concept as the inverse of the probability of the concept occurring in texts (refer to Section 3.3.2 for more information). The information content is the cornerstone of information theoretical measures, it can therefore be used to take advantage of several knowledge-based measures by considering corpus-based information. Other authors have also proposed to mix text analysis and structure-based (knowledge-based) measures. The extended gloss overlap measure introduced by (Banerjee and Pedersen, 2002), and the two measures based on context vectors proposed by (Patwardhan, 2003) are good examples. Another interesting approach is proposed in (Mohammad and Hirst, 2006). The authors propose a framework to exploit a thesaurus in order to derive word-concept distributional models that have the interesting property to be very compressed while finely characterising words. In (Alvarez and Lim, 2007), the authors propose to use WordNet and to adopt an hybrid approach in order to build a semantic model represented as a graph. The approach mixes gloss analysis as well as the analysis of WordNet structure. Interested readers may also consider (Li et al., 2003; Patwardhan et al., 2003; Banerjee and Pedersen, 2003; Patwardhan and Pedersen, 2006; Muller et al., 2006).

- *Aggregated measures* derive from the aggregation combining corpus-based, knowledge-based and even hybrid semantic measures[50]. Scores of selected measures are aggregated according to the average, min, max, median or any aggregation function which can be designed to aggregate matrix of scores[51].

Several studies have demonstrated the gain of performance mixing knowledge-based and corpus-based approaches (Panchenko and Morozova, 2012) – see also the work of (Petrakis et al., 2006).

### 3.8.2 Wikipedia-based measure: how to benefit from structured encyclopaedia knowledge

Started in 2001, the Wikipedia[52] initiative rapidly aroused great interest and became a reference encyclopaedia for all Internet users, including (self-proclaimed) experts that do not hesitate to share and organise valuable knowledge covering a variety of subjects. Then, it is not surprising that, in the middle of the 2000's, many efforts have been made to exploit this organised and free knowledge source in order to achieve information retrieval, classification, mining, or even business intelligence to mention a few. Indeed with more than 4.5 million articles[53] providing free-access to textual definitions in many languages, that are linked together and associated with structured categories[54], Wikipedia constitutes a stimulating playground for scientists involved in Computational Linguistics.

Several Wikipedia-based initiatives have been proposed in order to assess the semantic relatedness between words or Wikipedia topics (sometimes denoted concepts). These measures take advantage of the various facets of Wikipedia. They generally jointly rely on corpus-based measures exploiting textual information and approaches that analyse Wikipedia topics organisations or structured representations extracted from Wikipedia, e.g. structures defined by the hyperlinks between the articles. This diversity prevents precisely positioning Wikipedia-based approaches all together at a unique place the measure classification provided in Figure 1.2 (page 17). In addition, as we will see, a particularity of Wikipedia-based measures is that a large majority of them exploit hyperlink relationships between topics – a kind of proxy that, even if structured and informative, cannot be regarded as knowledge models (i.e. ontologies)[55].

---

[50] Pure-hybrid measures can also be part of the aggregation.
[51] Several aggregations will be discussed in the introduction of semantic similarity measures which can be used to compare groups of concepts – Section 3.5.2
[52] https://www.wikipedia.org
[53] 4 675 000 articles on 2015, January.
[54] Since May 2004.
[55] Contrary to texts and knowledge bases that have been denoted semantic proxies in Section 1.3.1, this proxy do not always convey semantics.

As far as the definition of semantic measures based on Wikipedia covers a lot of scientific works, for reading convenience, this section is organised with respect to the various techniques that have been proposed and the background knowledge they exploit:

1. The graph structure that rely on hyperlink relationships defined between articles,
2. The text content of these articles
3. The underlying structured categories that are associated to each article.

It should be noted that all of these approaches aim at quantifying semantic relatedness between words, texts or Wikipedia topics (as we said, topics, also denoted articles, are sometimes considered as concept definitions).

**Measuring semantic relatedness by exploiting Wikipedia's hyperlink relationships**

Each hyperlink of an article pointing towards another article indicates a relation between them. The semantics of this relationship is unknown but the target article is often assumed to help understanding the source article. This graph of hyperlinks between topics thus represents a substantial amount of (human) knowledge that is embodied into Wikipedia (Yazdani and Popescu-Belis, 2013)[56]. Several approaches have been proposed to extract information analysing this graph; some works consider the original orientation of links while other consider that they only traduce an association between topics and therefore consider the graph to be undirected.

According to (Milne and Witten, 2008), leaving aside textual and hierarchical contexts to focus on hyperlinks (more than 90 million links), leads to intermediate solutions, faster than text-based ones and having quite good accuracy compared to knowledge-based measures. Wikipedia Link-based Measure (WLM[57]), the solution proposed by Milne and Witten, only takes into account hyperlinks among articles to assess the relatedness of two words. Using anchors found in the body of Wikipedia articles, they identify candidate article/word relations in the aim of characterising a word through a set of articles. Each article is represented by a list of incoming and outgoing links. Then, by comparing links, they deduce the semantic relatedness between articles. To achieve this, they defined the following measure:

$$sr(a,b) = \frac{log(max(|A|,|B|)) - log(|A \cap B|)}{log(|W|) - log(min(|A|,|B|))} \quad (3.45)$$

Where $a$ and $b$ are the two articles of interest, $A$ and $B$ are the sets of articles linked to $a$ and $b$ respectively, and $W$ is the entire set of Wikipedia articles. The relatedness between two words is then computed using $sr$ to compare the sets of articles they are associated to.

The approach of (Turdakov and Velikhov, 2008) also proposes to compute relatedness[58] between articles based on hyperlink analysis. The measure relies on the Dice coefficient; it also uses heuristics and statistical properties on the links included in the articles. The semantic relatedness between two articles is assessed according to their shared and distinct hyperlinks (i.e. linked articles):

$$dice(a,b) = \frac{2 \times |A \cap B|}{|A| + |B|} \quad (3.46)$$

Where $A$ (resp. $B$) is the set of articles linked to article $a$ (resp. $b$) considering both incoming and outgoing links. In this approach the links are weighted according to some characteristics (e.g. symmetry) and some weights are related to the category structure associated with the articles (the category structure of Wikipedia will be discussed in Section 3.8.2). The heuristics they use proposes to reduce the search space according to the number of (incoming/outgoing) links an article has – this is done to avoid comparing all articles if the more similar have to be found. This relatedness measure leads to good results when applied to word sense disambiguation[59]. To this end, they proposed the following approach. For each word, the set of articles that contain the word in their title is identified (including disambiguation pages). The semantic relatedness between two words is then assessed according to the links their corresponding articles share – only articles with the highest relatedness are taken into account. Despite the good results of these methods, it should be noticed that a previous phase including text analysis is needed to ensure word disambiguation. This may sometimes leads to high computation time due to the large number of articles to process.

---

[56] Note that some of these links are automatically generated based on disambiguation techniques.
[57] This measure is one of the more cited in the literature.
[58] The authors also use the term similarity to describe their work but even if the distinction between similarity and relatedness is not clearly made in their contribution and may be discussed, we here consider that their approach assesses relatedness.
[59] Good enough for (Rui-Qin, 2012) to simply copy this approach for words relatedness estimation.

For (Yeh et al., 2009), Wikipedia encyclopaedia contains articles in a variety of topics wide enough to allow some association between words that are apparently not related only considering a simple text analysis. They go further by claiming that the graph structure of Wikipedia provides some relatedness information not present in the text of the articles. Contrary to previous cited works that only consider the links included in the articles that are compared, they consider the whole graph composed of all the links within Wikipedia to assess the relatedness between texts. This graph is extracted using a random walk strategy. Vertices refer to articles while edges are defined by the hyperlinks among articles. Three types of links are distinguished:

1. Infobox links – articles often contains infoboxes that enumerate attributes and characteristics for a given topic,

2. Categorical links – links that refer to the category associated to a topic/article and thus provide hyponymic and meronymic information, this will be discussed in Section 3.8.2 below.

3. Content links – all the other links.

This measure also relies on a generality attribute that encodes how much an article is more general than another. This attribute is computed by comparing the number of their incoming links – an article that is more specific than another one is assumed to have less incoming links. The semantic relatedness is then computed using personalised PageRank (random walk) algorithm (Hughes and Ramage, 2007), see Section 3.4. To apply personalized PageRank, it is necessary to construct a custom teleport vector representing the initial mass distribution over the article nodes. Two strategies are proposed to build this vector. The first one is based on a dictionary built either by using the article titles or extracted anchors given by WMiner (Lokeshkumar and Sengottuvelan, 2014). This dictionary is then filtered in different ways for their tests. They observed that smaller graphs might lead to information loss: by pruning the dictionary, some entities are isolated and will not be taken into account during the relatedness calculus. The second strategy proposes to initialise the random walk by using Explicit Semantic Analysis (ESA) (Gabrilovich and Markovitch, 2009) and to analyse the whole set of articles. This latter strategy gives the best results and slightly improves the ESA method (that will be discussed in the next section).

In the meantime, Wubben and Van den Bosch argue that the graph extracted from Wikipedia is well adapted to relatedness assessment. They propose FLP (Free Link Path), a semantic relatedness metric based on the notion of shortest-path between two Wikipedia articles – the graph is composed[60] of 2 million nodes (articles) and 55 million edges (internal links - links outside Wikipedia are ignored) (Wubben, 2008; Wubben and van den Bosch, 2009). All the articles are parsed to extract their titles and the outgoing links they contain. In the same time, an inverted index is built to associate to each article the incoming links and thus to extract the whole hyperlink graph. Using a breadth-first search over this graph, the shortest-path between articles is computed and the relatedness is derived from it. The results they obtained are interesting but are outperformed by vector-based approaches that will be presented in the next section.

A similar approach is adopted by Yazdani et al. to compare texts using a random walk strategy on a generated graph[61] (Yazdani and Popescu-Belis, 2011, 2013). In this graph the nodes are Wikipedia articles[62], they are intended to represent concepts; links are either hyperlinks between articles or derived from similarity of content. For the latter, a word co-occurrence approach is used to represent text articles by vectors, a cosine similarity is then used to assess vector similarity. Each article is linked with the $k$ most similar articles, with a weight that reflects this similarity score ($k$ is set to 10 in their evaluation). The semantic relatedness of two concepts is then assessed by analysing their distance into the graph. To do this, an adaptation of the random walk method denoted Visiting Probability is used. Therefore, when two texts are compared, they are projected onto a set of nodes (Wikipedia articles) by using aforementioned vector-based similarity. The semantic relatedness of these texts is then deduced based on the distances of their corresponding nodes. This method has been applied for text clustering with success – this, using several sets of parameters (e.g. weights attributed to the types of links). That makes the author underline the relevance of hyperlinks within this process. They also claim that, in the context of document clustering, random walk methods clearly outperform other methods, namely cosine similarity between the TF-IDF vectors of documents. However by mixing various strategies (vector-based measure, random walks and probabilistic adjustments) their method requires a lot of memory space and computational time. This approach has been improved, as exposed in (Yazdani and Popescu-Belis, 2013), in order to address this problem, among other things. Two truncation methods are proposed to make the algorithm tractable. In addition they propose to improve the Visiting Probability approach by integrating additional factors, such as the density of connections associated to articles. They also enlarge the scope of their evaluation and applied their work to word similarity, text similarity, document clustering/classification and information retrieval. By this way they demonstrate the generality of the knowledge resource associated to their approach and that this method provides a unified and robust answer to measuring semantic relatedness.

---

[60]In 2008.

[61]A distinction may be done here between what (Yazdani and Popescu-Belis, 2013) call concept network extracted from Wikipedia (each node is an article associated to a concepts and the links are the hyperlinks content in them) and a formal conceptual model (formalized taxonomy or ontology) that may be associated to Wikipedia but that will be discussed in Section 3.8.2.

[62]As many other Wikipedia-based approaches, some pruning is done to keep only articles that correspond to proper concepts. Here all articles from the following categories are removed: Talk, Image, Template, Category, Portal and List, and also disambiguation pages.

Recently, (West et al., 2009; Singer et al., 2013) studied an original strategy that does not use the links themselves but the human navigational path on them. The authors argue that while many hyperlinks correspond to semantic links in Wikipedia, many other do not. *"Links are often added based on the inclination of the author, rather than because the concepts are related"* (West et al., 2009). Therefore, other approaches only capture semantics from a limited set of people (Wikipedia editors) and neglect pragmatics, i.e. how Wikipedia is used (Singer et al., 2013). For the authors: *"Humans tend to find intuitive paths instead of necessarily short paths, while contrary an automatic algorithm would try to find a shortest path between two concepts that may not be as semantically rich and intuitive as navigational path conducted by a human"*. Their second argument lies in the nature of compared entities. Using such a method allows, for example, assessing semantic relatedness between an image and a textual page. Their experiments are based on the way people navigate on Wikipedia network. Observations have been gathered from online games played on Wikipedia; players have to reach an article from another unrelated article, only by clicking links in the articles that are encountered – the games that have been used are Wikispeedia[63] for (West et al., 2009) or from Wikigame[64] for (Singer et al., 2013). The measure proposed by West et al. is based on information theory after calculating a probability distribution over out-links of a current page. This method has some benefits, in particular its asymmetry may be interesting for some treatments. But it has also a major limit: only the distance between nodes (articles) that belong to a path that has been encountered during a game may be calculated. The authors argue that it is incremental but it is however limited to a small subset of Wikipedia. To overcome this limitation, (Singer et al., 2013) proposes a new way of calculating semantic relatedness between two concepts (articles) by using the similarity between their corresponding co-occurrence vectors. The underlying assumption is that word are semantically related if they share similar neighbours. Henceforth, two concepts may be compared even if they do not appear in the same path. The semantic relatedness between any Wikipedia concepts may hence be calculated; moreover, they have identified characteristics of navigational path that are most useful for its computing. An interesting thought about semantics of navigational path is described in (Singer et al., 2013). We will not report it here since it is a bit far from our purpose but we encourage the reader to refer to the original contribution for more details.

In a general manner, considering complexity, time calculus and correlation with human expert judgments, link-based Wikipedia semantic measures seem to be very efficient solutions. However they suffer from a noise problem. Indeed, as underlined by (Yeh et al., 2009), similar methods based, for example, on WordNet does not need pruning (i.e. cleaning) to compete with text-based methods. Yet, this pruning is necessary when using Wikipedia. It is due to the fact that links in Wikipedia may convey to marginal information with regard to the subject they are related to. A previous step of *ad-hoc* pruning is thus always needed before applying link-based methods. Therefore, (Yeh et al., 2009) argue that texts of Wikipedia provide a stronger signal than their link structure, which is in line with (Yazdani and Popescu-Belis, 2011) that recommends using both hyperlinks and lexical similarity links in order to take into account linguistic as well as extra-linguistic dimensions of texts.

---

[63] http://cs.mcgill.ca/~rwest/wikispeedia/
[64] http://www.thewikigame.com

**Text analysis of Wikipedia content and vector-based approaches**

As explained in Chapter 2, corpus-based approaches estimate semantic relatedness using statistical analysis of collection of texts that must be large enough to ensure a correct characterisation of words. Wikipedia offers such a large corpus.

Inspired by Vector Space Models used in information retrieval, (Gabrilovich and Markovitch, 2007) proposed Explicit Semantic Analysis (ESA)[65] to measure the semantic relatedness between words and texts (e.g. queries and documents) – this approach has briefly been presented in Section 2.3.2. In ESA, text meaning is represented into a high-dimensional space of concepts. Those concepts are derived from Wikipedia, once again each article of the encyclopaedia is considered to match a concept. Texts are processed to be represented as a vectors of Wikipedia concepts called interpretation vectors. To achieve this, ESA exploits a weighted inverted index extracted from Wikipedia articles, where weighted concepts are attached to words (the weights are function of TF-IDF measures). This inverted index allows, when parsing an input text, to associate to it a weighted vector of Wikipedia concepts by merging the concepts associated to each of the words it contains. The semantic relatedness between two texts is then assessed thanks to vector similarity measures applied to the interpretation vectors of the two texts. For two words $a$ and $b$ and their corresponding ESA vectors $\vec{a}$ and $\vec{b}$, the relatedness is for instance given by their cosine similarity. This semantic interpretation of texts leads to disambiguate word senses since it takes into account the context of the neighbourhood of words. However, even if the results are very close to human judgements, when processing some datasets, the main drawback of this method is time calculus. Indeed, since all the articles have to be compared, the computation complexity is really high. This method has been extended in (Radinsky et al., 2011) where the authors present a new semantic relatedness model, Temporal Semantic Analysis (TSA). This method captures temporal information in addition to knowledge extracted from Wikipedia. While ESA represents word semantics as a vector of concepts, TSA uses a more refined representation. Each concept is no longer a scalar, but is instead represented as time series over a corpus of temporally-ordered documents. This attempt to incorporate temporal evidence into models of semantic relatedness is quite innovative and their evaluation shows that TSA provides consistent improvements over state-of-the-art results of ESA on multiple benchmarks.

In line with ESA, Zesch et al. suggested to apply concept vector based measure to assess semantic relatedness (Zesch et al., 2008). In their proposal, several popular resources are compared: Wiktionary, Wikipedia, English and German WordNets. Several approaches are studied, one relying on a path-based approach, another one generalising the vector-based approach described above. We will focus here on the latter since it is the one used by the authors of (Zesch et al., 2008) to deal with Wikipedia content. They propose to capture the meaning of a word $w$ using a high dimensional concept vector $\vec{v} = (v_1, v_2, \ldots, v_{|W|})$, where $|W|$ is the number of Wikipedia documents. The value $v_i$ depends on the number of occurrences of the word $w$ in the article numbered $i$ (e.g. using TF-IDF score). Each word being represented in this concept vector space, vector measures may be used to assess the relatedness of two words (e.g. cosine). The authors claim that this approach "*may be applied to any lexical semantic resource that offers a textual representation of a concept*". Using Wikipedia, they only take into account the first paragraph of the article considering that it contains shorter and more focussed information. From their comparative works, vector-based measures have proved to better perform than path-length ones when operating on collaboratively constructed resources. The choice of the resource is also discussed – it is showed that Wiktionary outperforms Wikipedia when ranking word pairs, even if it leads to lower performances on a German dataset.

To take into account the semantic wealth offered by the multiple languages spoken around the word would be of particular interest in many domains. In information retrieval, for example, one can search for images related to a specific subject and characterised in different language (e.g. a given dermatologic manifestation in the medical domain). The growing need for cross-lingual solutions for information retrieval, text classification or annotation, to cite a few, has been underlined in (Hassan and Mihalcea, 2009). The authors propose to address this challenge by exploiting the interlanguage links contained into Wikipedia (250 language versions exist). In order to achieve this, they propose an extension of the ESA approach (in which each article is considered to match a concept). To calculate the cross-lingual relatedness of two words, they measure the closeness of their concept vector representations, which are built from Wikipedia using an extension of ESA. Three major changes of ESA have been proposed. The first one concerns the relatedness metric. Instead of the cosine calculus between the vectors, they chose to apply a Lesk-like metric (briefly discussed in Section 2.4). This choice has been made to take into account the possible asymmetry between languages. With $a$ and $b$ two words and $\vec{A}$ and $\vec{B}$ their ESA concept vectors, we denote $A$ and $B$ the sets of concepts with a non-zero weight encountered in $\vec{A}$ and $\vec{B}$ respectively. The coverage of $\vec{A}$ by $\vec{B}$ is defined by:

$$G(\vec{A}|\vec{B}) = \sum_{t \in B} \omega_{c_i}(a) \tag{3.47}$$

---

[65]This work is also one of the more cited in the Wikipedia-based measure literature. Due to the high quality of its results, it is often used as a way to derive a semantic representation of articles or words(Singer et al., 2013).

With $\omega_{c_i}(a)$ the weight associated to the concept $c_i$ in vector $\vec{A}$. The relatedness between two words $a$ and $b$ is defined by:

$$rel_{Hassan}(a,b) = \frac{G(\vec{B}|\vec{A}) + G(\vec{A}|\vec{B})}{2} \tag{3.48}$$

The second change they suggest is to modify the weight calculus of ESA in order to take into account the length of the articles that are associated to a concept. Instead of giving as weight the TF-IDF value associated with a concept $c_i$ they propose to use:

$$\omega_{c_i}(a) = tf_i(a) \times log(\frac{M}{|c_i|}) \tag{3.49}$$

With $tf_i(a)$ the term frequency of the word $a$ in the concept $c_i$ (i.e. in the related article), $M$ a constant representing the maximum vocabulary size on Wikipedia concepts and $|c_i|$ the size of the vocabulary used in the description of the concept $c_i$.

The last change concerns the use of Wikipedia category graph (see following section). The weight is scaled by the inverse of the distance $d_i$ of the concept category (category associated with the concept $c_i$) to the root one.

$$\omega_{c_i}(a) = \frac{tf_i(a) \times log(\frac{M}{|c_i|})}{d_i} \tag{3.50}$$

The cross-lingual relatedness is then computed as follows. Given $C_x$ and $C_y$ the sets of all Wikipedia concepts in languages $x$ and $y$. If $tr_{xy} : C_x \to C_y$ is a translation function that maps a concept $c_i \in C_x$ to a concept $c'_i \in C_y$ via the interlanguage links. The projection of the ESA vector $\vec{t}$ from the language $x$ to the language $y$ can be written: $tr_{xy}^{vec}(\vec{t}) = \{\omega_{tr_{xy}(c_1)}, \ldots, \omega_{tr_{xy}(c_n)}\}$. The relatedness between two words $a_x$ and $b_x$ in given languages $x$, $y$ is then defined by:

$$rel_{cross_lingual}(a_x, b_y) = \frac{G(tr_{xy}^{vec}(\vec{B})|\vec{A}) + G(\vec{A}|tr_{xy}^{vec}(\vec{B}))}{2} \tag{3.51}$$

If the relation described by the interlanguage links is assumed to be reflexive, in practice it is not always the case since users are accredited with the responsibility of maintaining these links. A pre-treatment is applied to detect the missing links and enforces the reflexivity property. Even if the results vary from a language to another according to the coverage conveyed by Wikipedia in these languages (the number of pages and the number of interlanguage links), the correlations obtained by this method on well-known benchmarks slightly outperforms results obtained by monolingual measures.

A recent work bridges the two research streams presented in both previous sections. It consists in using frequency occurrences and link probability to assess relatedness between words (and by extension between texts) (Jabeen et al., 2013). This CPRel method (Context Profile based Relatedness) relies on a context profile extracted from Wikipedia and that is associated to each word. Wikipedia articles are filtered to keep only relevant words and links that point towards another Wikipedia article. The match between a word and article titles is done using Link Probability (LP) [Mihalcea et al. 2007]. Once a set of articles is associated to a word, they are weighted according to their term frequency (based on TF calculus) and link probability. These weights define a vector representation of the word into the space of articles. The relatedness between two words is then computed using cosine similarity. On some benchmark their results are comparable with other Wikipedia-based approaches but ESA performs best overall.

**Wikipedia relies on a category organisation that may be used to measure semantic relatedness**

As we saw, in addition to text articles, Wikipedia stores a great deal of information about the relationships between the articles in the form of hyperlinks, info boxes and category pages (Yeh et al., 2009). In the contributions presented in the previous sections, authors often assimilate Wikipedia articles to concepts since articles are considered to define a particular entity. However, the structure offered by the hyperlink relationships between them cannot be regarded as knowledge models (i.e. ontologies). In this section, we are really talking about a knowledge organisation that is composed of Wikipedia categories structured through conceptual relationships (e.g. hyponymy, meronymy). Wikipedia's categories have been collaboratively developed and used to tag Wikipedia articles. They are assigned to pages in order to group together those discussing similar subjects. They are next used to help readers to find an article or to navigate on related ones. Wikipedia's categories have been standardised and organised by unambiguous relationships – the structure organizing the categories can indeed be regarded as a *light* ontology. These categories are therefore of particular interest for semantic similarity calculus since they provide a way to compare topic articles not only using corpus-based measures, but also using semantic measures based on knowledge organization (in particular those introduced in Section 3.4).

The first authors that have used the structure of Wikipedia's categories to assess semantic similarity or relatedness of words were (Strube and Ponzetto, 2006). In their WikiRelate! method, words are first mapped to articles; this is done by analysing the titles and the hyperlinks of articles. Then words are compared considering two features of the texts they are associated to: (i) their content overlap (using an adaptation of aforementioned Lesk's measure) and (ii) the relatedness of their Wikipedia categories – several measures have been tested (Rada, Leacock and Chodorow, Wu and Palmer, Resnik measures, please refer to Section 3.4 for an introduction to these measures). They obtain good results on datasets of human judgements, which make them stress the benefits of using the categories organisation for designing Wikipedia-based semantic measures.

Within their BabelRelate! method, (Navigli and Ponzetto, 2012) further explore the hybrid approach of semantic relatedness between two words by using BabelNet, a very large multilingual lexical knowledge resource that integrates Wikipedia and WordNet (Navigli and Ponzetto, 2010). This solution combines the analysis of the graph of word senses by using a graph-based algorithm (a kind of node counting strategy). By using the whole graph containing word translations, they are able to prune parts of the graph that are the result of ambiguity and polysemy in the input language (infrequent senses, noisy relations). This strategy helps to rapidly characterise the subgraph that represents the *core* semantic of words, i.e. the most frequent translations, senses, in all languages. The relatedness is then assessed by analysing these graphs. Interestingly, the authors argue that, contrary to the method defined by (Hassan and Mihalcea, 2009), their proposal does not suffer from unbalanced performance across languages since it is based on a unique and common resource (BabelNet). They also observed that the more languages are used, the better the results.

Finally, in (Taieb et al., 2013), a system combining all semantic information in the different components of Wikipedia (articles, graph of hyperlinks, category organisation) is presented to compute the semantic relatedness between words. A pre-processing step provides for each Wikipedia category a semantic description vector (named CSD for Category Semantic Depiction). It is computed by considering the weights of stems extracted from the articles assigned to the target category. Then, a vector representation of a word is derived analysing the CSDs of the set of all categories it is associated to. Finally, the semantic relatedness of two words is assessed according to the similarity of their respective vectors (e.g. using Dice, overlap and cosine measures). This approach performs well and even sometimes outperforms ESA.

---

With an increasing quality and coverage, Wikipedia is an undeniable worldwide success. As we saw in this section, by providing a free multi-language encyclopaedia in which topics are interlinked, organised by structured categories and finely characterised (e.g., using infoboxes), the several facets of Wikipedia are of particular interest and have proved to be particularly helpful in the design of accurate semantic measures. Wikipedia has always been a catalyser of new semantic measure design and will, for sure, continue to be. Research tracks related to this unique resource are numerous. As an example, through the definition of DBpedia (Auer et al., 2007), recent contributions in Knowledge Representation and Information Extraction have proved that Wikipedia can also be used to generate large knowledge bases. This will help us to finely structure and characterise Wikipedia content, and to better capture the semantic interactions between the underlying concepts associated to articles. Interestingly, by obtaining such a large knowledge base and its associated natural language counterpart, a new kind of hybrid semantic proxy will arise and will probably open both interesting and promising perspectives for semantic measures.

> Software solutions and source code libraries that provide knowledge-based semantic measures implementations are presented in Appendix D. Information about evaluation protocols and datasets that can be used to compare these measures are also provided in Chapter 4.

## 3.9 Conclusion

This chapter introduced the reader to the diversity of knowledge-based measures which can be used to compare concepts or instances defined into ontologies. We focused in particular on the measures which are based on the analysis of a graph representation of the ontology, i.e. the strategy commonly adopted to define knowledge-based measures. We proposed an in-depth analysis of the measures which can be used to assess the similarity of concepts defined in a taxonomy. By analysing these measures we distinguished the semantic evidence which can be extracted from ontologies. We next presented in detail the three main types of measures, i.e., structure-based, feature-based and those based on information-theoretical approaches. This helped us to highlight the foundation of knowledge-based semantic measures and to discuss the core elements of these measures. Next, we briefly introduced the measures which rely on complex logic-based construct analysis and those which rely on the analysis of several ontologies. Finally, we defined the advantages and limits of knowledge-based measures and we introduced some hybrid-measures which have been proposed to mix corpus-based and knowledge-based measures.

# Chapter 4

# Methods and datasets for the evaluation of semantic measures

This chapter is dedicated to semantic measure evaluation and discusses in particular two important topics: (i) how to objectively evaluate measures and (ii) how to guide their selection with regard to specific needs. To tackle these central questions, we propose technical discussions on both methodological and practical aspects related to semantic measure evaluation. This will help us to underline, among others, the properties of measures that must be considered for their analysis. An overview of the underlying mathematical frameworks that can support measure evaluation and a detailed discussion of existing evaluation protocols are also provided. Concrete examples of datasets used to evaluate semantic measures in the literature are next introduced. The various topics discussed in this chapter are suited for both research and practical purposes, e.g. they are adapted to measure designers willing to evaluate new proposals, as well as users seeking for existing solutions adapted to their needs.

Despite its central importance – considering the large diversity and the widespread use of semantic measures – the topic of measure evaluation is far from being extensively discussed in the literature. To overcome this substantial lack, this chapter proposes to aggregate a large body of literature in order to provide an overview of state-of-the-art approaches and resources available. However, considering the width of this topic and the diversity of approaches that have been explored to evaluate measures, this chapter will not propose an in-depth analysis of all relevant aspects of measures that can be discussed for their evaluations. In addition, even if some references to interesting state-of-the-art comparisons are provided, we will not analyse, cross and aggregate existing results to summarize measure performances in different evaluation settings.

The first section of this chapter discusses the problems of semantic measure evaluation and selection in a general manner. Section 2 presents several criteria of measures that can be considered for their evaluations. Section 3 introduces the protocols and datasets that are commonly used to evaluate measure accuracy. Finally, Section 4 concludes this chapter by highlighting its important teaching and by discussing, among others, the open challenges related to this important topic.

## 4.1 A general introduction to semantic measure evaluation

In general terms, any evaluation aims to distinguish the benefits and drawbacks of the compared alternatives according to specific criteria. Such comparisons are most of the time used to rank the relevance of using an alternative in a specific usage context regarding a set of criteria – in our case the alternatives are the measures. Semantic measures can be evaluated with regard to theoretical or empirical properties. Based on these evaluations, comparison will be made possible considering a way to (i) compare the values of considered properties and (ii) to aggregate these comparisons. This underlines the strong dependency that exists between the criteria that are considered for the evaluation and the conclusions that can be obtained. Indeed, as we will see in this chapter, an important notion to understand is that there is no best semantic measure *in absolute*. There are only measures that outperform others in specific conditions. Even if this does not prevent the fact that specific measures may outperform other measures in most of (experimental) conditions, it makes clear that conclusions about measure performance will be difficult to generalise. Indeed the set of criteria to consider and the way to interpret the values taken for each of these criteria strongly depend on the purpose of the comparison, e.g. comparing measures in order to distinguish the one which is the best adapted to a specific use case.

Considering that measure evaluation can only be made considering specific aspects of them, prior to compare a set of alternatives, a user which is searching for a measure must pay attention to carefully analyse his needs. It will for instance not be possible to distinguish the measure a user must use *"to compare two concepts defined into an ontology"*, or *"to compare two terms"* without defining: the constraint the measure must respect (e.g. symmetry), the information which is available (e.g., corpora, knowledge bases), and what the elusive objective *to compare two concepts/terms* means for him, i.e. are we talking about semantic similarity / relatedness, etc. Users must understand that answering these questions may be critical to distinguish a measure that is adapted to their needs.

The comparison of measures is therefore only possible considering a set of criteria and a way to compare and to aggregate these criteria. The latter are defined by the use case that motivates the comparison. Considering this intuitive but important remark, three important questions arise in the aim of comparing semantic measures:

1. What are the criteria that can be used?

2. How to evaluate the relevance of a measure regarding a specific set of criteria?

3. Which criteria must be considered to evaluate measures in a specific context?

Even if a complete answer to these three questions cannot be provided only considering existing research contributions –, and could alone justify a complete textbook –, this section proposes substantial material and reflections in the aim of contributing and better understanding semantic measure evaluation and comparison.

## 4.2 Criteria for semantic measure evaluation

Several criteria can be used to analyse semantic measures. Some can be studied theoretically while others require empirical analyses. Among the criteria that are the most frequently considered evaluating semantic measures, we distinguish their:

- Accuracy, precision and robustness.
- Computational complexity, e.g., algorithmic complexity.
- Mathematical properties.
- Semantics.
- Characterisation regarding technical details.

These criteria can be used to discuss numerous facets of semantic measures. They are detailed in corresponding subsections.

### 4.2.1 Accuracy, Precision and Robustness

The accuracy of a measure can only be discussed according to predefined expectations regarding the results produced by the measure. Indeed, as defined in the field of metrology, the science of measurement, the accuracy of a measurement must be understood as the closeness of the measurement of a quantity regarding the (actual) true value of that quantity (BIPM et al., 2012). As we have seen when defining semantic measures, the quantities estimated by semantic (similarity, relatedness, etc.) measures, even if they may be intuitive, are today only weakly characterised by abstract terms – in comparison to quantities commonly measured in other fields, e.g. inertial mass in Physics. In addition, as stressed when semantic measures have been introduced in Chapter 1, formal definitions may only be considered when a system of formal definitions will be accepted and adopted by the various researchers studying this broad topic. The possibility to reach such an objective is still an open research topic. As we will see, this intuitive and volatile nature of semantic measures has deep implications for their evaluation.

Indeed, it is important to stress that evaluating accuracy of semantic measures is made difficult by the fact that the *true value* of a semantic measure (similarity, relatedness, etc.) is unknown; and, more importantly, may not exist. The notion of similarity of terms or concepts is by nature a subjective notion. As an example, there is no, *per se*, true value associated to the semantic relatedness between the two concepts *Communism* and *Freedom*. Disregarding as much as possible philosophical considerations, it can be said that concepts are in our mind and that there are as many true values of semantic measures as concept representations of pairs of compared concepts (i.e. individuals).

Therefore, the accuracy of semantic measures is generally evaluated by considering averaged expectations regarding their values, that is to say, by considering consensual values of the quantities they try to capture. This is generally made by analysing human expectations of semantic similarity/relatedness/etc., and by considering averaged values of these expectations. As an example, measures will be compared to scores of semantic relatedness of word pairs that have been assessed by humans into a specific scale, e.g. 0-4 – several datasets of this kind will be introduced in Section 4.3.2 . Thus, according to the implicit considerations made in the literature, the accuracy of a semantic measure is often considered as the closeness of its measurement with the averaged human expectations. Correlations are preferred to distances between measurements and average experts' assessments because they only take into account the way the two signals behave instead of too arbitrary absolute values. In this case, the accuracy is generally defined as the Pearson's correlation coefficient $r$. Considering that expected and measured values are contained into two vectors $x$ and $y$ of size $n$, the Pearson correlation $r$ is defined by:

$$r(x,y) = \frac{\sum_{i=1}^{n}(x_i - \overline{x})(y_i - \overline{y})}{\sqrt{\sum_{i=1}^{n}(x_i - \overline{x})^2}\sqrt{\sum_{i=1}^{n}(y_i - \overline{y})^2}} \quad (4.1)$$

Pearson's correlation is a linear correlation; less frequently, non-linear correlations are studied, e.g. in (Shen et al., 2010).

Sometimes, the accuracy of a measure is only evaluated by analysing the ordering of the pairs that is induced by their scores of similarity/relatedness. In this case evaluations are made using Spearman's correlation, e.g. in (Huynh et al., 2014; Maguitman and Menczer, 2005; Pedersen et al., 2007; Pesaranghader et al., 2014). Considering two vectors $x$ and $y$ of size $n$ that specify respectively the expected and estimated rank of each pair, Spearman's correlation $\rho$ is defined by:

$$\rho(x,y) = 1 - \frac{6\sum_{i=1}^{n}(x_i - y_i)^2}{n(n^2 - 1)} \quad (4.2)$$

Adaptations of this definition are also often considered in order to evaluate measure accuracy indirectly, that is to say, using datasets that do not directly refer to human expectations regarding measured notions. In this case, the accuracy of a measure is often indirectly evaluated by evaluating (the accuracy of) a system that depends on it. As an example, the aim of such a system may be to resolve a classification problem – e.g. by considering (i) that the degree of membership of an item to a class is defined as a function of its similarity with the gravity centre of the class, and (ii) that the class to which an item will be affected to is the class for which the item has the highest membership degree. Using this approach, systems are generally evaluated studying their accuracy, using the traditional formula of accuracy:

$$accuracy = \frac{\#(true\_positive) + \#(true\_negative)}{\#(true\_positive, true\_negative, false\_positive, false\_negative)} \quad (4.3)$$

With $\#(X)$ the number of elements assigned to class $X$, e.g. $\#(true\_positive)$ is the number of elements which have been correctly classified by the system.

In all cases, the notion of accuracy of a measure is *per se* defined according to a context, e.g., the *true values* considered, the semantic proxy (specific corpus, ontology, etc.), the tuning of the measure parameters (if any). Indeed, there is no guarantee that a measure that has been proved accurate in a specific evaluation setting will be accurate in other settings. The accuracy of a particular semantic measure tuning can therefore only be discussed with regard to a specific usage context, without absolute guarantee that the results that are obtained can be generalized to other usage contexts – even if empirical analyses have shown that specific measures tend to outperform others in many contexts.

The precision of a measure (or more generally any system of measurement) corresponds to the degree of reproducibility or repeatability of the score produced by the measure under unchanged conditions. Since most semantic measures are based on deterministic algorithms, and therefore produce the same result given a specific input, evaluating the precision of a measure generally makes no sense. Evaluations of semantic measures therefore focus most of the time on the notion of accuracy. We will further discuss the precision of a measure as a mathematical property. Nevertheless, when a measure is evaluated analysing the precision of a system depending on it, e.g. a classification system, the precision will be defined by:

$$precision = \frac{\#(true\_positive)}{\#(true\_positive, false\_positive)} \quad (4.4)$$

Accuracy and precision can be used to estimate the performance of a measure according to expected results. As we said, these results are only valid in a very specific evaluation setting – which depends on the dataset, measure parameters and resources (text corpora, knowledge base, etc.) that are considered. Nevertheless, these criteria alone may not be sufficient to analyse measure performance. For instance, most of the time, nobody will be able to ensure that expected results that are considered into a specific dataset are not impacted by uncertainty – and that they will be the same if the benchmark was obtained using other participants. Therefore, intuitively, rather than only considering the measure which best performs according to specific expected results (provided by humans), most system designers will prefer to use a simply satisfying measure which, in revenge, would still ensure a good performance even if human assessments were slightly different. Imprecision and variability in human evaluations, as well as any disturbances that may affect the human assessment process support this cautious behaviour. Put another way, evaluation protocols must also take into account the capacity for a measure to produce robust scores considering the uncertainty related to the way the human expected similarity values have been obtained, or disturbances of the semantic proxies on which relies the measure (modification of the ontologies, corpora).

In this context, it is important to know how sensible/robust measure performances are to perturbation of evaluation settings. This aspect of a measure can be evaluated by analysing its robustness. It has been studied in (Janaqi et al., 2014); the authors propose a framework that can be used to evaluate semantic measure robustness, e.g. by disturbing expected results by an uncertainty model – technical details are not provided herein. To date, the robustness of semantic measures has only faintly been analysed and only few comparisons have been proposed.

### 4.2.2 Computational complexity

The computational complexity of semantic measures is of major importance in most applications. Indeed, considering the growing volumes of datasets processed in semantic analysis (large corpus of texts and knowledge bases), this aspect is most often critical for concrete usages of semantic measures.

As an example, considering alternatives with equivalent performances in a specific evaluation setting, most system designers will prefer to make moderate concessions on measure accuracy for a significant reduction of computational time. Indeed, reducing the expectation on measure accuracy may lower the final performance of a system, although using a measure with a strong computational complexity may simply prevent its use, e.g. information retrieval systems often depend on semantic measure computations that are made *on-the-fly*.

However, the literature relative to the computational complexity of semantic measures is very limited. In particular, the algorithmic complexity of semantic measures, i.e. the amount of resources required (e.g. time, storage), is most of the time never analysed – the paper of Turney and Pantel (2010) is among the exceptions. It is therefore difficult to discuss algorithmic implications of current proposals. This hampers non-empirical evaluations of measures and burdens measure selection. It is nevertheless difficult to blame semantic measure designers for not providing detailed algorithmic analyses of their proposal. First, these extensive analyses are both technical and difficult to make. Second, they are only possible for specific cases as they depend on technical considerations that exceed semantic measure definitions, e.g., the type of data structures used to represent the semantic proxy on which rely the measures, the algorithm used in specific treatments (e.g. graph traversals). This later point may create a gap between theoretical capabilities of measures and computational complexity of their practical implementations into software.

The reader must therefore understand that, to date, and despite the major importance of the computational complexity of measures, evaluation and comparison of measures regarding this criterion is difficult and most of the time not possible. Nowadays, comparisons of measures regarding this specific aspect most often rely on empirical evaluation of specific measure implementations. For an example of such a comparison in a specific setting you can refer to https://github.com/sharispe/sm-tools-evaluation.

### 4.2.3 Mathematical properties

Several mathematical properties that are of interest for characterising semantic measures have been distinguished in Section 1.2.3, e.g., symmetry, identity of the indiscernibles[1], precision (for non-deterministic measures), and normalization. These properties enable to deeply characterize semantic measures and to select proposals adapted to specific usage contexts. Indeed, specific properties may be required to ensure the coherency of treatments depending on semantic measures. As an example, considering that a measure is not symmetric or does not respect the identity of the indiscernibles can lead to undesired results and therefore may be inappropriate for some applications.[2] Some of these mathematical properties are also essential to apply specific optimization techniques in order to reduce the computational complexity of measures while ensuring valid results. In addition, and this is a critical point that will be discussed hereafter, these properties play an important role to finely characterise the semantics carried by measures, i.e., the meaning of the results the measures produced.

---

[1] i.e. does not return the maximal similarity when comparing a term/concept to itself.

### 4.2.4 Semantics

The meaning (semantics) of the results obtained by a measure must be carefully considered when selecting a measure. This semantics is defined by the assumptions on which relies the algorithmic design of the measure. Some of these assumptions can be understood through the mathematical properties of the measure. The others are defined by the cognitive model on which the measure is based, the semantic proxy in use and the semantic evidence analysed. As saw in Section 1.3.1, the semantic evidence taken into account by the measure defines its type and therefore its general semantics (e.g., the measure evaluates semantic similarity, relatedness, etc.) – it therefore largely impacts the results.

It is however difficult to finally compare measures regarding the semantics they carry. Nevertheless, it is essential for end-users to understand that measure selection may in some case strongly impact the conclusions that can be supported by the measurement (e.g. semantic similarity, relatedness, etc.). As an example, designing a recommendation system that will return items considering the focal concept *Coffee* may return completely different results depending on the semantic of the measure; if it is a semantic similarity the system will favour results that can (partially) substitute *Coffee*, e.g. *Arabica*, *Tea*, *Hot Chocolate*, while using a semantic proximity it will not only return beverages but also close concepts, e.g. *Cup*, *Coffee bean*.

### 4.2.5 Technical details

What we denote technical details are the several extra-parameters that have to be considered when choosing semantic measures for a specific usage context. Some of them are not relevant for comparing measure proposals but may play an important role when a measure has to be used in specific applications. They may therefore be of interest for designers of systems that rely on semantic measures. Among the numerous technical details that have to be considered we distinguish:

- The availability of supported implementations of the measure and the license associated to these implementations. Some measures require technical and substantial work in order to be implemented. When comparing measures for a specific application, it may therefore be important to reduce the set of alternatives that are considered according to this practical aspect. Appendix D presents numerous software tools that provide implementations of state-of-the-art-measures.

- The dependency of the measure to specific resources (knowledge base, corpora, training datasets). Are these resources freely available in the domain of interest? In addition, the end-user must consider the sensibility of measure accuracy on these resources (if reported).

- The availability of several evaluations that support the performance of the measure in multiple evaluation settings.

---

Some aspects of measures that have been discussed in this section (e.g. accuracy) may be evaluated empirically using specific datasets. The following section details protocols that are generally adopted to evaluate semantic measures. Datasets that are commonly used are next presented.

---

[2]It is however the case using particular measures in specific contexts. As an example using Resnik's measure to compare concepts defined into a taxonomy (c.f. Section 3.4.3), the similarity of a general concept (near to the root) to itself will be low.

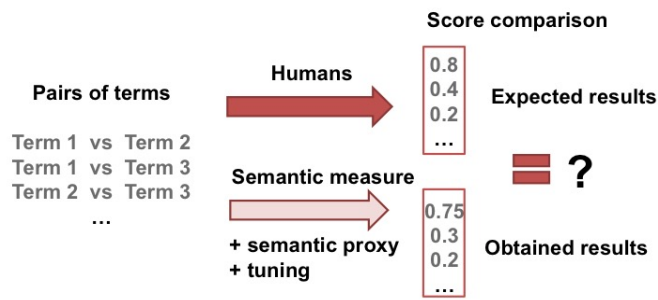

Figure 4.1: Example of a direct evaluation in which a semantic measure is evaluated by comparing the scores obtained by the measure to human expectations (e.g. average of human scores).

## 4.3 Existing protocols and datasets

In the literature, accuracy of semantic measures is generally considered as the *de-facto* metric to evaluate and compare measure performance. It can be evaluated using a direct or an indirect approach depending on the expected scores (*true values*) that are considered. In this section, we detail the protocols commonly used to evaluate measures using the two approaches. Then, we present numerous datasets that can be used to compare measures.

### 4.3.1 Protocols used to compare measures

In most cases, accuracy of measures is evaluated using a direct approach, i.e., based on expected scores of measurement (e.g., similarity, relatedness) of pairs of terms/concepts. In other cases, measures are evaluated indirectly, by analysing results of treatments which depend on semantic measures. In all cases, the evaluation of measure accuracy is performed regarding specific expectations/assumptions of expected results:

- Direct evaluation: based on the correlation of semantic measures with expected scores. Measures are generally evaluated regarding their capacity to mimic human ratings of semantic similarity/relatedness. In this case, the accuracy of measures is discussed based on their correlations with gold standard benchmarks composed of pairs of terms/concepts associated to expected ratings. Results are commonly evaluated using Pearson's and Spearman's correlations. Figure 4.1 illustrates the direct evaluation approach.

- Indirect evaluation: This evaluation highly depends on the domain of study, e.g. NLP, Bioinformatics. Figure 4.2 illustrates the approach. Two strategies can be applied:

  1. The first one evaluates measure accuracy analysing the performance of applications or algorithms which depend on semantic measures, e.g., accuracy of term disambiguation techniques, performance of a classifier or a clustering approach relying on semantic measures to mention a few. As an example, considering that the measure is evaluated through the analysis of disambiguation algorithm, the object of study would be the terms in specific (sentential) contexts; the expected results would be the mappings between the terms and their disambiguated forms and the results obtained would be produced by a disambiguation system based on the evaluated measure. In this scenario, the performance of the measure will be discussed with regard to the performance of the disambiguation process, i.e. the semantic measure that permits to obtain the best results (fixing the other parameters) will be considered to be the best in this context. Therefore, the method used to compare expected and obtained results depends on the type of the results considered.

  2. The second strategy works by analysing the correlation of measures with domain-specific metrics that are expected to behave like semantic measures in specific contexts. As an example, in Bioinformatics, semantic measures are designed to compare gene products according to their conceptual annotations. Evaluations have been made by comparing correlations between scores of measures and the similarity of gene evaluated according to their DNA sequences (Lord et al., 2003).

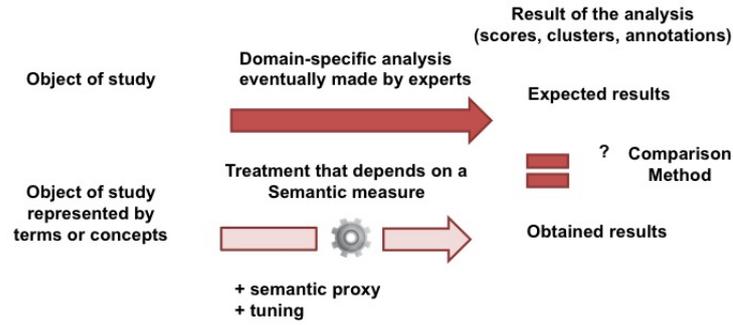

Figure 4.2: General illustration of an indirect evaluation approach.

In the following section different datasets that can be used to compare semantic measures are presented in a syntactic manner. Then each one is detailed with regard to specific properties.

### 4.3.2 Datasets

Most of datasets are based on human-ratings and are composed of pairs of terms/concepts for which humans have been asked to assign scores of semantic similarity or relatedness. The instructions provided to the participants vary among the datasets. These variations may impact the notion of similarity or relatedness considered by the participants. Generally, the quality of a dataset is assessed by analysing inter-agreement of participants, i.e. how scores of participants correlate. This inter-agreement also defines the level of accuracy measure designers may go after. These datasets can be used to evaluate measures according to a direct approach, i.e. regarding the (Pearson's or Spearman's) correlations of measures with averaged scores provided by humans. In some cases, cleaning techniques are applied to exclude abnormal ratings (outliers) from the datasets prior to the evaluation.

Several datasets are listed in Table 4.1 – they are organised in chronological order. The large majority of them are in English. Some have been manually or automatically translated into other languages or mapped to knowledge bases (e.g. WordNet). In this case, word pairs are (manually) mapped to unambiguous pairs of concepts in order to be used to evaluate knowledge-based semantic measures[3]. Some datasets are also dedicated to specific domains (e.g., medicine) and other focus on specific type of words (e.g. verbs, rare words).

---

[3] In most cases, the concepts associated to the terms are not communicated in contributions related to knowledge-based semantic measures. In some cases, words are mapped to multiple concepts and the best score is considered for the evaluation, in accordance with the fact that annotators seem to consider the closest sense pair when evaluating the similarities (Mohammad and Hirst, 2012b). Nevertheless, these particular cases are poorly documented in the literature.

| References | Type of evaluation | Size | Inter-agreement Pearson ($r$), Spearman ($\rho$) |
|---|---|---|---|
| (Rubenstein and Goodenough, 1965) | Semantic similarity of pairs of nouns | 65 | $r = 0.99$ on avg. values of two groups. Intra-agreement $r = 0.85$ |
| (Miller and Charles, 1991) | Semantic similarity of pairs of nouns | 30 | $r = 0.8848$ |
| TOEFL (Landauer and Dumais, 1997) | Semantic Similarity of words using synonymy questions | 80 | 52.7 |
| WordSim353 (Finkelstein et al., 2002) | Semantic relatedness of pairs of nouns. | Two sets 153/200 | $\rho = 0.661$ |
| ESL (Turney, 2001) | Semantic similarity of word pairs based on synonymy questions | 50 | Not found |
| (Turney et al., 2003) SAT analogy questions | Semantic similarity of words based on analogy questions | 374 | Not found |
| (Jarmasz and Szpakowicz, 2003a) | Semantic similarity based on synonym detection problems | 300 | Not found |
| (Boyd-Graber et al., 2006) | Semantic relatedness of pairs of WordNet Synsets | 100000 | Cf. paper[4] |
| Gur65 (Gurevych, 2005) | German version of Rubenstein & Goodenough benchmark | 65 | $r = 0.81$ |
| Gur350 dataset[5] | Semantic relatedness of German word pairs | 222 | $r = 0.69$ |
| ZG222 dataset (Zesch and Gurevych, 2006) | Semantic relatedness of German word pairs | 350 | $r = 0.49$ |
| (Pedersen et al., 2007) | Semantic relatedness between medical words[6] | 29 | For 101 pairs $r = 0.51$ For 29 pairs $r = 0.68$(physicians) $r = 0.78$ (coders) |
| (Pakhomov et al., 2010) | Semantic similarity and relatedness of pairs of UMLS concepts (medical domain) | 566 (sim) 587 (rel) | $r = 0.50$ (sim) $r = 0.47$ (rel) |

---

[4]Specific details have to be considered.
[5]Described in https://www.ukp.tu-darmstadt.de/data/semantic-relatedness/german-relatedness-datasets
[6]A subset of 29 pairs with higher inter-agreement is generally considered among the 101 pairs.

| | | | |
|---|---|---|---|
| ConceptSim (Schwartz and Gomez, 2011) | Semantic similarity of pairs of WordNet synsets. Disambiguate the pairs of nouns of Rubenstein & Goodenough (RG), Miller & Charles (MC) and WordSim353 (WS) | 28 (MC) 65 (RG) 97 (WS) | $r = 0.93$ (RG) $r = 0.89$ (MC) $r = 0.86$ (WS) |
| (Radinsky et al., 2011) | Semantic relatedness of pairs of nouns | 280 | Not found |
| Mturk-771 (Halawi et al., 2012) | Semantic relatedness of pairs of nouns | 771 | $r = 0.89$ |
| (Ziegler et al., 2012) | Semantic relatedness of pairs of nouns which are disambiguated by DBpedia URIs[7] | 25 and 30 | $r = 0.71$ and $r = 0.70$ |
| Stanford's Contextual Word Similarities (SCWS) (Huang et al., 2012) | Semantic relatedness of pairs of words in context | 2003 | Not found |
| The Stanford Rare Word (RW) Similarity Dataset (Luong et al., 2013) | Semantic relatedness of pairs of rare words | 2034 | Not found |
| MEN test Collection (Bruni et al., 2014) | Semantic relatedness of pairs of words | 3000 | $\rho = 0.84$ |
| SimLex-999 (Hill et al., 2014) | Semantic similarity of pairs of words | 999 | $\rho = 0.67$ |
| (Baker et al., 2014) | Semantic relatedness of pairs of verbs | 143 | Not found |

Table 4.1: Summary of datasets that are commonly used for comparing semantic measures. Inter-agreement: refer to the details provided in the description of the dataset and/or to the original publications to understand how these values have been computed. Each result has specificities that must be considered for further analyses – comparison between values is not always directly possible.

---

[7]Uniform Resource Identifiers are used to refer to concept without ambiguity.

| Noun 1 | Noun 2 | Average similarity $[0, 4]$ |
|---|---|---|
| cord | smile | 0.02 |
| noon | string | 0.04 |
| food | fruit | 2.69 |
| forest | woodland | 3.65 |
| automobile | car | 3.92 |

Table 4.2: Example of entries of the Rubenstein and Goodenough dataset.

The reader can also consider these interesting websites about datasets (that may be updated in the future):

- Manaal Faruqi[8] website maintains a list of various datasets that can be used to evaluate semantic measures.

- The ACL website[9] and the Semantic Measure Library website[10] also provide information about datasets.

In the following, we detail several aspects of each aforementioned datasets. We discuss in particular the specific aims of the datasets. We also introduce the protocols and settings that have been used during acquisition. This will help us to underline the semantics which is associated to the expected scores that have been provided by the participants.

### (Rubenstein and Goodenough, 1965) – noun similarity

The procedure used to obtain this dataset is well documented in the associated paper. The dataset is composed of 65 pairs of nouns (ordinary English nouns), e.g. *cord/smile* – pairs were introduced as *theme pairs* in the experiment. Each pair was written into a card in order to obtain a shuffled deck of 65 cards for each participant. Next, participants were asked to order the cards according to the similarity of the pairs of nouns written on them. Finally, participants were asked to evaluate the (semantic) similarity of each pair using a 0-4 scale – the higher the number associated to the card, the greater the *"similiarity of meaning"*. This experiment was designed to evaluate semantic similarity, which was defined as the *"amount of similarity of meaning"* (i.e. degree of synonymy) to the participants. Participants were 51 paid college undergraduates. Two groups of 15 and 36 subjects were considered – respectively called group 1 and 2.

The intra-subject reliability was computed using group 1 on 36 pairs of nouns for which participants were asked to assign the similarity twice, two weeks apart. The intra-subject reliability was computed using Pearson's correlation: a score of $r = 0.85$ was obtained. Inter-subject correlation is not communicated in this experiment but the reported correlation of mean judgements of the two different groups was impressively high ($r = 0.99$). This encouraged the authors to merge the results of the two groups to finally only consider a single group of 51 participants. For each pair of nouns, only the average similarity of the scores provided by all the participants is available. Examples of results are provided in Table 4.2.

The benchmark of Rubenstein and Goodenough is largely used in the evaluation of semantic similarity measures. It has also been translated into German by Gurevych (2005).

### (Miller and Charles, 1991) – noun similarity

Miller & Charles benchmark is a subset of Rubenstein and Goodenough's benchmark composed of a selection of 30 pairs. Three sets of 10 pairs of nouns with high, intermediate and low levels of similarity were chosen; for each set of pairs of nouns, the scores in the original Rubenstein and Goodenough's benchmark were respectively 3 or 4, between 1 and 3, and 0 or 1. The similarities of the 30 pairs of nouns were then assessed by 38 participants. These participants received the instructions provided by Rubenstein and Goodenough. Inter-subject correlation is 0.8848. Interestingly, human rating obtained by Miller and Charles, and Rubenstein and Goodenough are highly correlated, with a Pearson's correlation of 0.97, e.g. reported by Bollegala (2007a). Note that, as stressed by Budanitsky and Hirst (2006), due to a typographical error, the pair cord/smile was changed to *chord/smile*. This seems to have no impact on the overall evaluation since both pairs have low levels of similarity. Only the averaged similarities are provided in the dataset. This benchmark has also been translated into Arabic, Romanian and Spanish (Hassan and Mihalcea, 2009).

---
[8] http://www.cs.cmu.edu/~mfaruqui/suite.html
[9] http://www.aclweb.org/aclwiki
[10] http://www.semantic-measures-library.org

**TOEFL (Landauer and Dumais, 1997) – word similarity**

This dataset is designed to evaluate semantic similarity of words. It evaluates the degree of synonymy of pairs of nouns, verbs, and adjectives. It is composed of 80 multiple-choice synonymy questions that have been selected from the Test of English as a Foreign Language (TOEFL). Each question provides a problem word and 4 choices of synonyms with a single expected answer – an example of question is provided below. For each question, participants were asked to select the synonym with the "*most similar meaning*" with the problem word associated to the question. Averaged result obtained from a large sample of applicants to U.S colleges from non-English countries is 51.6 correct answers (64.5%) – reduced to 52.7% when these scores are corrected by penalizing errors to lower the impact of correct answers that could have been obtained by guessing.

Additional information about this dataset can be found at: `http://lsa.colorado.edu/papers/plato/plato.annote.html#evaluate`

Example of question[11]: considering the focal word *levied* and the following choices (a) *imposed*, (b) *believed*, (c) *requested*, (d) *correlated*, the expected solution is (a) *imposed*.

**WordSim353 (Finkelstein et al., 2002) – word relatedness**

Two sets of English pairs of words along with human ratings of semantic relatedness. The first set contains 153 pairs including the 30 nouns pairs contained in (Miller and Charles, 1991) dataset. 13 participants evaluated it. The second set contains 200 pairs evaluated by 16 participants. Participants were asked to assess the "*relatedness*" of words in a 0-10 discrete scale associated to the semantics "*totally unrelated words*" (0) to "*very much related or identical words*" (10). A correlation of 0.95 is reported between the scores proposed by the participants of WordSim353 and those assessed in Miller and Charles experiment. Row results, as well as mean scores are provided for the two sets. A concatenation of the sets composed of 353 pairs with averaged scores is also provided. (Hill et al., 2014) reports an inter-agreement of $\rho = 0.661$ using Spearman's correlation. The benchmark is available at `http://www.cs.technion.ac.il/~gabr/resources/data/wordsim353`.

WordSim353 have also been translated into French (Joubarne and Inkpen, 2011), Arabic, Romanian and Spanish (Hassan and Mihalcea, 2009).

**ESL (Turney, 2001) – word similarity**

ESL, English as a Second Language, is similar to the TOEFL dataset (Landauer and Dumais, 1997). It proposes to evaluate the semantic similarity of pairs of nouns, verbs or adjectives by providing 50 multiple-choice synonym questions (4 choices by question). These questions have been selected from a collection of questions for students of ESL. Inter-agreement between student results is not provided. This dataset is available on request from Peter Turney.

Additional information about this dataset can be found at: `http://a4esl.org/q/j/dt/mc-2000-01syn.html`

Providing the definition "*A **rusty** nail is not as strong as a clean, new one.*" and the following choices (a) *corroded*, (b) *black*, (c) *dirty*, (d) *painted*, the expected solution is (a) *corroded*.[12]

**SAT Analogy questions (Turney and Littman, 2003) – similarity of word relationships**

This dataset is composed of 374 multiple-choice analogy questions collected by Michael Littman from the Scholastic Aptitude Test (SAT). It has been used to evaluate contextual semantic similarity of semantic relationship between words (Turney, 2006). This notion is evaluated studying the analogy between the relationships that link words. Each question is composed of a problem pair of words with a single valid answer among 5 choices of pair of words. As an example, considering the problem pair *cat:meow* and the 5 choices *mouse:scamper*, *bird:peck*, *dog:bark*, *horse:groom*, *lion:scratch*, the expected answer is *dog:bark* since the semantic relationships which link these words and the words of the problem pair (*cat:meow*) are the same "*the name of the sound made by the animal*". Another example of question is provided below. This benchmark is available on request from Peter Turney. Human performances are not provided.

---

[11]Example from `http://www.aclweb.org/aclwiki/index.php?title=TOEFL_Synonym_Questions_(State_of_the_art)`
[12]Example from `http://www.aclweb.org/aclwiki/index.php?title=ESL_Synonym_Questions_(State_of_the_art)`

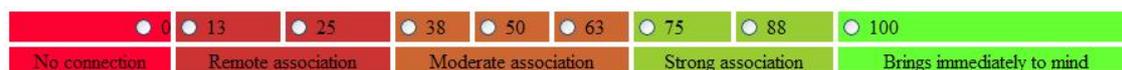

Figure 4.3: Screenshot of the interface and scale used to obtain the score of evocation for a pair of synsets. From (Nikolova et al., 2012).

Providing the stem *mason:stone* and the following choices (a) *teacher:chalk*, (b) *carpenter:wood*, (c) *soldier:gun*, (d) *photograph:camera*, (e) *book:word*, the expected solution is (b) *carpenter:wood*.[13]

### 300 RDWP – (Jarmasz and Szpakowicz, 2003a) word similarity

This dataset is composed of 300 synonym detection problems. They have been selected for the Word Power game of the Canadian edition proposed at Reader's Digest Word (2000, 2001) – refer to the work of Jarmasz and Szpakowicz (2003a) for details about the original dataset. Each question is composed of a problem word, and 4 candidate answers. Participants have been asked to "*Check the word or phrase you believe is nearest in meaning*". The following example is provided by Jarmasz and Szpakowicz (2003a).

Providing the problem word: *ode* and the following choices (a) *heavy debt*, (b) *poem*, (c) *sweet smell*, (d) *surprise*, the expected solution is (b) *poem*.

### (Boyd-Graber et al., 2006) – WordNet synset relatedness

This benchmark provides human appreciation of semantic relatedness for a large number of pairs of WordNet synsets. It is composed of two datasets of about 100K randomly selected pairs of synsets. Trained undergraduates were used to build the first dataset. The second dataset was made from participants selected on the Amazon Mechanical Turk platform. It is reported to contain noise – refer to the original documentation for more details. Figure 4.3 presents the scale proposed to the participants.

Additional information about this dataset can be found at: http://wordnet.cs.princeton.edu/downloads.html

### (Yang and Powers, 2006) – verb relatedness

This dataset provides semantic similarities for 144 pairs of verbs. According to the authors this is the first benchmark dedicated to the evaluation of verb semantic similarity. The verbs have been selected from TOEFL (Test of English as a Foreign Language) and ESL (English as a second Language) tests – details on how the pairs of verbs have been obtained from these tests are provided in the contribution of Yang and Powers (2006). 2 academic staffs and 4 postgraduate students have assessed the semantic similarities – 4 were native Australian English speakers and 2 were *near-native* speakers. A 0 (*not at all related*) to 4 (*inseparably related*) discrete scale has been used. A Pearson's correlation of $r = 0.866$ is reported among the participants.

### (Pedersen et al., 2007) – concept relatedness

This dataset is composed of pairs of medical concepts represented by pairs of non-ambiguous words. Similarly to the definition of Rubenstein and Goodenough, 120 pairs of terms equally divided into four classes of degree of relatedness from *practically synonymous* to *unrelated* were chosen. Participants were 3 physicians and 9 medical coders. They annotated each pair using the following discrete scale: *practically synonymous* (4.0), *related* (3.0), *marginally related* (2.0) and *unrelated* (1.0). A low correlation of 0.51 between the participant scores was obtained. Therefore, a subset of this set composed of 30 pairs with higher inter-agreement is generally considered – one pair was later deleted since no correspondence with a concept was found into the SNOMED-CT (structured terminology). Using the subset of 29 pairs, the average correlation among physicians and medical coders is 0.68 and 0.78 respectively. Considering average scores of each member of each group, the correlation between physicians and medical coders is 0.85. 10 medical coders also assessed the similarity of the pairs of words of the Rubenstein and Goodenough and Miller and Charles benchmarks. They obtained an inter-agreement of 0.84 and 0.88 with the values obtained in the original experiments. Only averaged similarities of both physicians and medical coders are available. Table 4.3 presents some of the entries that compose the dataset.

---

[13]Example from http://www.aclweb.org/aclwiki/index.php?title=SAT_Analogy_Questions_(State_of_the_art)

| Concept 1 | Concept 2 | Physician Avg. | Coder Avg. |
|---|---|---|---|
| Renal failure | Kidney failure | 4 | 4 |
| Heart | Myocardium | 3.3 | 3 |
| Stroke | Infarct | 3 | 2.8 |
| Delusion | Schizophrenia | 3 | 2.2 |

Table 4.3: Example of entries of Pedersen et al. dataset. Values are in the interval $[0, 4]$.

This dataset has also been used to compare concepts defined into MeSH or SNOMED-CT – correspondences between labels and concept identifiers are provided by Batet et al. (2014) and Harispe et al. (2013c).

**WS Sim (Agirre et al., 2009) – cross-lingual word similarity**

This dataset is used to evaluate Spanish/English cross-lingual semantic similarity and relatedness. It is composed of the set of pairs of words of Rubenstein & Goodenough and WordSim 353 datasets, in which the second word of each pair has been translated into Spanish. The two translators agreed on translating 72% and 84% of Rubenstein & Goodenough and WordSim 353 pairs respectively. Information about this benchmark is available at `http://alfonseca.org/eng/research/wordsim353.html`.

**(Pakhomov et al., 2010) – word similarity/relatedness**

This dataset provides scores of semantic similarity and relatedness between pairs of medical terms – terms refer to UMLS concepts. Two sets of concept pairs are studied. The first contains 566 pairs and is dedicated to semantic similarity. The second is composed of 587 pairs rated for semantic relatedness. Scores were obtained from 8 medical of the University of Minnesota Medical School, 4 participated to the similarity task and 4 to the relatedness task. This work can also be used to compare concepts defined into medical knowledge-base, e.g. MeSH or SNOMED-CT (Batet et al., 2014).

**ConceptSim (Schwartz and Gomez, 2011) – concept similarity**

ConceptSim disambiguates WordSim 353, Rubenstein & Goodenough (and therefore Miller & Charles) datasets by mapping each word to a unique synset of WordNet 3.0. Two annotators with an inter-agreement that ranged from 86 to 93% made the disambiguation. The final version of the benchmark contains the mapped pairs after annotator agreement. Original mapping of the two annotators are provided. Even if differences could be observed between the similarity of the ambiguous and disambiguated pairs, the similarities of the original experiments are generally considered when evaluating measures using this dataset. The benchmark is available to download at `http://www.seas.upenn.edu/~hansens/conceptSim`.

**(Radinsky et al., 2011) – word relatedness**

This dataset is composed of 280 pairs of words with associated semantic relatedness scores. Participants were Amazon Mechanical Turk workers. Each participant evaluated 50 pairs and an average of 23 rating is reported for each pair. Outliers were removed using correlations with ten pairs of WordSim353 datasets. A version of this dataset is available at: `http://tx.technion.ac.il/~kirar/Datasets.html`

| Concept 1 | Concept 2 | Relatedness Avg. $[1,5]$ |
|---|---|---|
| artillery | gun | 3.542 |
| cement | glue | 3 |
| pyramid | speculation | 1.7 |

Table 4.4: Example of entries of Halawi et al. dataset.

| Word 1 | Word 2 | Relatedness Avg. $[1,5]$ | Standard deviation |
|---|---|---|---|
| EasyJet | Cheap Flight | 4.294 | 0.749 |
| Eminem | Music | 4.137 | 0.971 |
| Microsoft | Internet Explorer | 4.314 | 0.727 |
| Periodic Table | Toyota | 1.176 | 0.55 |

Table 4.5: Example of entries of Ziegler et al. dataset.

**Mturk-771 (Halawi et al., 2012) – word relatedness**

This dataset is composed of 771 pairs of English words along with their semantic relatedness. Each pair of words has been evaluated by at least 20 participants from the Amazon Mechanical Turk platform. Inter-correlation of the results was assessed to 0.89 with small variance. Participants were asked to assign the relatedness of batch of 50 pairs of words using a 1-5 discrete scale ranging to "*not related*" to "*highly related*". In order to assess the quality and reliability of the evaluation, each batch of 50 word pairs contained 10 pairs known to have extreme relatedness values. This was used to control dataset quality. Evaluations which contained more than one error on the control quality pairs were not considered for building the final benchmark. Both raw and mean scores are available for download at http://www2.mta.ac.il/~gideon/mturk771.html[14]. Examples of entries of this dataset are provided in Table 4.4.

**(Ziegler et al., 2012) – concept relatedness**

Two sets of concept/instance pairs denoted by English words. Contrary to most of existing benchmarks, the compared word refers to concepts or instances such as name of artists, brand names, qualified concepts. The first set contains 25 pairs, the second 30. Participants were asked to assign the relatedness of pairs of concepts using a 1-5 discrete scale ranging to the semantics "*no proximity*" to "*synonymy*". The scores of relatedness were assessed based on an online survey. Most participants were not native people (Germans, Italians, Turkish). Inter-subject correlations based on Pearson's correlation are about 0.70 for the two sets – which is good considering the facts that the benchmark is composed of names of brands, artists, making the comparison more difficult, e.g. the label has to be understood by the participants. In addition, contrary to most experiments the majority of the participants were not native people. Both average similarities and associated standard deviations are provided. Examples of entries are given in Table 4.5.

**SCWS (Huang et al., 2012) – word similarity in context**

Stanford's Contextual Word Similarities (SCWS) provides the semantic similarity of 2003 pairs of words in sentential contexts to ensure that the meaning of words is not ambiguous. Each pair has been evaluated by 10 participants taken on the Amazon Mechanical Turk platform. Details about the dataset are provided in (Huang et al., 2012). It can be downloaded at: http://www-nlp.stanford.edu/~ehhuang/SCWS.zip.

Example of results that have been obtained for the comparison of the words *activity* (n) and *inaction* (n) considering the following contexts:

---

[14] Contrary to the details provided in the paper, the documentation specified that each pair was evaluated by ten people.

| Word 1 | Word 2 | Relatedness Avg. [0, 10] |
|---|---|---|
| omnipotence | state | 0.86 |
| calcify | change | 4.83 |
| dwarfish | fish | 8.38 |
| prescriptions | medicine | 9.11 |

Table 4.6: Example of entries of Luong et al. dataset.

| Word 1 | Word 2 | Relatedness Avg. [0, 50] |
|---|---|---|
| Sun | Sunlight | 50 |
| River | Water | 49 |
| Green | Grey | 36 |
| Garage | Garder | 18 |

Table 4.7: Example of entries of Bruni et al. dataset.

*activity (n)*: "tons of poultry and 61 million tons of eggs were produced worldwide. Chickens account for much of human poultry consumption, though turkeys, ducks, and geese are also relatively common. Many species of birds are also hunted for meat. Bird hunting is primarily a recreational **activity** except in extremely undeveloped areas. The most important birds hunted in North and South America are waterfowl ; other widely hunted birds include pheasants, wild turkeys, quail, doves , partridge , grouse , snipe , and woodcock . Muttonbirding is also popular in Australia and".

*inaction (n)*: "respect which always marked his communications with the court. It has been insinuated both by contemporary and by later critics that being disappointed at his loss of popularity, and convinced of the impossibility of co-operating with his colleagues, he exaggerated his malady as a pretext for the **inaction** that was forced upon him by circumstances. But there is no sufficient reason to doubt that he was really, as his friends represented, in a state that utterly unfitted him for business. He seems to have been freed for a time from the pangs of".

The average relatedness is 6. Details of the 10 scores are: 10, 9, 8 ($\times 3$), 7, 5, 3, 2, 0.

**RW (Luong et al., 2013) – Rare word relatedness**

This benchmark provides the semantic similarity for 2034 pairs of rare words. Details about the dataset and its construction are provided in (Luong et al., 2013). Participants were recruited from the Amazon Mechanical Turk platform. For each pair, 10 ratings have been obtained using a 0-10 discrete scale. The benchmark can be downloaded at: `http://www-nlp.stanford.edu/~lmthang/morphoNLM`. Examples of entries are provided in Table 4.6.

**MEN Test (Bruni et al., 2014) – word relatedness**

The MEN Test Collection provides scores of semantic relatedness for 3000 pairs of words. Scores were obtained from English native speaker available from the Amazon Mechanical Turk platform. Contrary to other experiments, participants did not rate the semantic relatedness of pairs of terms using a discrete scale but rather used comparative tests. They were asked to distinguish the most related pair of words among two candidate word pairs. For each pair, 50 binary scores (more or less related than the other pair) have been obtained and used to compute the score of relatedness in a 0-50 discrete scale. Since candidate pairs were randomly generated no score of inter-agreement between participants has been reported. However, as a control, a $\rho = 0.68$ Spearman correlation between the scores of relatedness provided by two of the authors in a 0-7 discrete scale is reported. In addition, a Spearman correlation of $\rho = 0.84$ between the average values of these scores and those obtained using participants is reported. Details and downloads are provided at `http://clic.cimec.unitn.it/~elia.bruni/MEN.html`. Examples of entries are provided in Table 4.7.

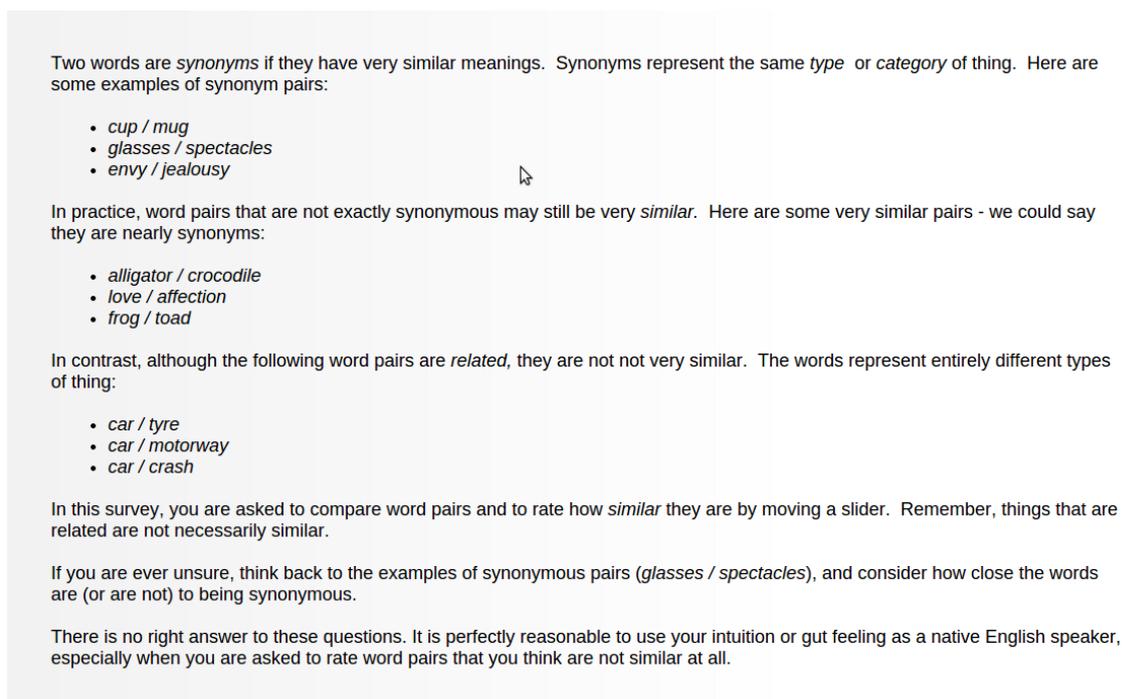

Figure 4.4: Instructions provided to participants of Sim-Lex-9999 experiments. From (Hill et al., 2014).

| **Verb 1** | **Verb 2** | **Relatedness Avg.** $[0, 10]$ | **Standard deviation** |
|---|---|---|---|
| organising | developed | 3.6 | 0.35 |
| form | employ | 1.9 | 0.19 |
| showing | showed | 7.8 | 0.7 |

Table 4.8: Example of entries of Baker et al. dataset.

**Sim-Lex-999 (Hill et al., 2014) – word similarity**

Sim-Lex-999 is dedicated to the evaluation of the semantic similarity between words. It provides semantic similarity scores for 999 pairs of words that have been assessed by participants selected from the Amazon Mechanical Turk platform. Words refer to concepts selected from the University of South Florida (USF) Norms dataset – a dataset that provides ratings for the concreteness of several concepts (Nelson et al., 2004). However, contrary to (Huang et al., 2012) which disambiguates words by providing a sentential context, in this case only labels are provided to the participants – instructions are presented in Figure 4.4.

Each participant had to provide similarity ratings in a 0-7 discrete scale for 20 pairs of words. Each pair was evaluated by an average of 50 participants. The Spearman correlation between participants is 0.67. Refer to the work of Hill et al. (2014) for details about the protocol used to obtain this benchmark. Sim-Lex-999 can be downloaded at: `http://www.cl.cam.ac.uk/~fh295/simlex.html`.

**(Baker et al., 2014) – verb similarity**

This dataset provides score of semantic relatedness for 143 pairs of verbs. For each pair, 10 participants have a relatedness score in a 0-10 discrete scale. The dataset can be downloaded from `http://ie.technion.ac.il/~roiri/`. Examples of entries are provided in Table 4.8.

**The Semantic Textual Similarity campaign and related datasets**

The Semantic Textual Similarity (STS) campaign provides state-of-the-art resources and methods for comparing measures and systems dedicated to sentence semantic similarity evaluations – defined as the

| Sentence 1 | Sentence 2 | Similarity [1,5] |
|---|---|---|
| The bird is bathing in the sink | Birdie is washing itself in the water basin | 5 – The two sentences are completely equivalent, as they mean the same thing. |
| The woman is playing violin | The young lady enjoys listening to the guitar | 1 – The two sentences are not equivalent, but are on the same topic. |

Table 4.9: Pairs of sentences from the STS dataset.

"*degree of semantic equivalence*" between two sentences. Detailed information about STS, the resources it proposes and the community it federates can be found at `http://ixa2.si.ehu.es/stswiki/index.php`.

The STS campaign is proposed since 2012 at SemEval – a series of evaluations dedicated to computer systems related to semantics (Agirre et al., 2012). The datasets used in this campaign are composed of pairs of sentences from specific or existing paraphrase and machine translation datasets. The similarity between the sentences of each pair is defined by participants into a 0 to 5 discrete scale – considering that 5 means "*The two sentences are completely equivalent, as they mean the same thing*" while 0 means "*The two sentences are on different topics*". Details are provided in the associated documentation. The challenge is to design a system that will assign a score of relatedness with an optional score of confidence when comparing two sentences. Example of expected results are presented in Table 4.9.

Some datasets provided by STS are also available in Spanish. Specific evaluations have also been proposed. As an example, in 2014, a Cross-Level Semantic Similarity Task proposed to evaluate measures able to compare semantic similarity across different sizes of texts (paragraph to sentence, phrase to word...).

Among the benchmarks used to evaluate text relatedness, users can also consider:

- (Li et al., 2006) – sentence similarity benchmark which is based on the definitions of the terms that compose (Rubenstein and Goodenough, 1965) dataset. 32 participants assessed similarity using a 0 to 4 scale (*unrelated* to *alike*). A subset of 30 pairs of definitions is generally considered.

- (Lee et al., 2005) – text to text similarity which is based on pairs of documents built from a collection of 50 short documents presenting news (source: Australian Broad-casting Corporation's news mail service). For each pairs, ten participants have proposed relatedness scores into a 1 to 5 discrete scale (*unrelated – alike*). The final benchmark contains similarity for all the 2500 document pairs.

- Microsoft research paraphrase benchmark[15] (MSR Paraphrase) can also be used to compare the similarity of texts. This manually created benchmark contains next to 6K pairs of sentences from diverse Web sources – Inter-agreement was evaluated between 82 and 84%. In the context of STS evaluation, this benchmark has been enriched by specifying how related are the sentences which are not considered to be paraphrased.

- Microsoft research Video Paraphrase Corpus. Built using Amazon Mechanical Turk to obtain a sentence description of the video – 120k descriptions have been collected for 2000 videos. Based on this dataset 1500 pairs of sentences have been generated considering sentences describing the same video or different videos – details are provided in the work of Agirre et al. (2012).

- Other datasets can be adapted to evaluate semantic similarity or relatedness, e.g. benchmarks used to evaluate distributional semantic models (Baroni and Lenci, 2011) or domain-specific datasets (Hassan et al., 2012).

**Other evaluations not based on human ratings**

We have presented numerous datasets that can be used to evaluate semantic measures with regard to expected scores of similarity or relatedness of compared elements (terms, concepts, sentences). These datasets can be used to assess measure accuracy by directly analysing their ability to mimic human appreciation of semantic similarity or relatedness. Using indirect evaluations, the accuracy of measures can also be evaluated by analysing the impact of a specific choice of measure on the performance of systems. Accuracy and precision measures introduced in Section 4.2.1 are generally used in this case. As an example of indirect evaluation, McInnes and Pedersen (2013) evaluated measures by analysing the performance of a disambiguation system. Particular strategies are defined for specific domains. In Bioinformatics, evaluated systems can be used to obtain expected groups of genes or proteins, e.g. protein families, genes involved in similar metabolic pathways, or to distinguish pairs of proteins that will interact. In other cases, semantic measures will be evaluated by analysing the correlation between the semantic similarity of gene conceptual annotations (groups of concepts defined into an ontology) and the score of similarity of gene DNA sequences. Please refer to the work of Guzzi et al. (2012) and Pesquita et al. (2009a,b) for details related to evaluations in Bioinformatics.

---

[15]`http://research.microsoft.com/en-us/downloads/607D14D9-20CD-47E3-85BC-A2F65CD28042/default.aspx`

## 4.4 Discussions

In this chapter, we have introduced several important aspects related to the evaluation and the selection of semantic measures. First, we have discussed multiple characteristics of measures that can be discussed in order to select the measures that are adapted to a specific usage context: accuracy, precision, robustness, computational complexity, mathematical properties, semantics, as well as other technical details such as the availability of free concrete implementations. Next, we have presented state-of-the-art methodologies and datasets commonly used to empirically evaluate measure accuracy. Using this information, users will be able (i) to deeply analyse a specific choice of measure and (ii) to better understand the implication of this choice on the system he develops.

In the process of selecting a specific measure, end-users must first characterise as much as possible the properties his ideal measure must respect, the resources which are available and other constraints that will be useful to orient the selection (e.g. computational complexity, computer resources, and implementation availability). These points are dependant on the usage context and therefore prevent the definition of a generic turnkey solution. However, once this work has been performed, we have introduced and identified a variety of datasets that can be used to empirically evaluate measures with regard to particular applications. Existing comparisons can be analysed from the literature or publicly available results published onto the Web[16]. An important point to stress – that has voluntary not been discussed in this chapter – is that for most datasets measure proposals equalling human inter-agreement have been proposed. This stresses that efficient solutions have already been proposed and that, in most cases, *ad-hoc* solutions are not required. Using existing implementations of measures and tools dedicated to evaluation presented in Appendix D, as well as publicly available benchmarks, specific empirical analysis can also be done to perform fine-grained analysis of measures and eventually tune studied alternatives. Finally, if the datasets that have been presented are not representative of a specific use case of interest, the reader can refer to the benchmarks descriptions provided in this chapter to design new ones.

Despite the numerous datasets and evaluation approaches presented in this chapter, evaluating semantic measures is both conceptually and practically an open challenge. Many biases are yet to be excluded from future datasets to improve evaluation process. For instance we stress that:

- Uncertainty in expected values could be better managed if individual assessments were provided instead of average values as a summary of the collected evaluations.

- Similarity expected scores are imprecise values and should be modelled as such whereas they are currently defined on a linearly ordered finite scale, e.g. 0-4.

- Providing similarity scores for individual pairs of words is probably less natural for participants than providing pairwise comparison of the similarity of word pairs.

- The difficulty of evaluation depends on the pairs of concepts to be compared; hence, all the required scores should not contribute equally to the overall assessment of the measure accuracy.

In this context, research on how to better evaluate semantic measures must be made. Initiatives defining systematic and reproducible approaches for comparing measures, such as the Semantic Textual Similarity campaign (Agirre et al., 2012), must also be encouraged. Indeed, nowadays, this field of study suffers from the lack of reproducible results and shared evaluation platform – they could greatly ease the comparison of existing proposals, and help us to better understand both benefits and drawbacks of existing proposals – initiatives such as (Faruqui and Dyer, 2014; Pesquita et al., 2009b) must also be encouraged, supported, improved and adopted. This is necessary to enable the emergence of general knowledge about semantic measures from existing comparison results.

---

[16] As an example ACL website provides provides several results for numerous approaches. Consult: http://www.aclweb.org/aclwiki

More analysis of existing benchmarks must also be performed in order to criticise their relevance and to underline their strengths and limits. Differences between contextualised semantic similarity and semantic similarity of ambiguous words must deeply be studied. To date, by only providing pairs of words, most datasets delegate sense selection to the participants, which necessarily impact the quality of the evaluation (Budanitsky, 2001), and probably the whole meaning of the evaluation. Difference between the semantic similarity/relatedness of different types of words, e.g. nouns, verbs, adjectives must also be better understood. Better understanding expected scores of semantic similarity or relatedness will, without doubt, help us to improve and better define semantic measures.

A large number of state-of-the-art evaluation results related to semantic measures are based on reduced datasets that only contain a few dozens of pairs of elements, e.g. Miller and Charles dataset. This highly reduces and even questions the conclusions that can be derived from them, e.g. are these results biased? Additional evaluations must be performed using the various larger datasets recently proposed.

All these contributions will help us to better compare semantic measures and to precise more finely the *quantities* estimated by these measures. This will be required to characterise the various dimensions today encompassed into the notions of semantic similarity, proximity or distance.

# Chapter 5

# Conclusion and research directions

The capacity of assessing the similarity of objects or stimuli has long been characterised as a central component for establishing numerous cognitive processes. It is therefore not surprising that measures of similarity or distance play an important role in a large variety of treatments and algorithms, and are of particular interest for the development of Artificial Intelligence. In this context, this book focused on the complex notion of semantic similarity and more generally on semantic measures: how to compare units of languages (words, paragraphs, texts) or conceptualised entities, i.e. concepts or instances defined in ontologies, by taking into account their semantics and therefore their meaning.

We first introduced numerous applications in which approaches for the estimation of semantic similarity, relatedness or distance play an important role, as well as the numerous communities involved in their study. We also introduced important contributions related to the notion of similarity: the various models of similarity proposed and studied by Psychology and the formal axiomatic definitions of distance and similarity considered in Mathematics. This helped us identifying the depth of this interdisciplinary area of research, a depth which has been in particular illustrated by highlighting and organizing the large and often poorly defined vocabulary introduced in the literature, i.e. semantic similarity, semantic relatedness, semantic distance. Next, we introduced a general classification of semantic measures and the two broad families of semantic measures: corpus-based measures used to compare units of languages from natural language analysis and knowledge-based measures used to compare concepts and instances defined into ontologies. The foundations of these two approaches have been detailed and several examples of approaches have been introduced.

We hope this introduction helped the reader to understand the concept of semantic measure and to discover the variety of approaches proposed so far. Indeed, even if this book only propose an introduction to semantic measures, and do not discuss in detail all types of measures proposed in the literature, as well as important topics such as the evaluation of measures or the selection of context-specific measures, we are convinced it provides enough material and references to give the reader access to a deep understanding of this topic. We also hope that this overview of the field will help researchers identifying research directions to galvanise this topic. To conclude, we propose to discuss some of the challenges we have identified based on the analysis of a large body of literature dedicated to semantic measures:

- Better characterise semantic measures and their semantics;

- Provide theoretical and software tools for the study of semantic measures;

- Standardise ontology handling;

- Improve models for compositionality;

- Study current models of semantic measures w.r.t language specificities;

- Promote interdisciplinary;

- Study the algorithmic complexity of measures.

- Support context-specific selection of semantic measures.

# Research directions

### Better characterise semantic measures and their semantics

Throughout the introduction of semantic measures, we have stressed the importance of controlling their semantics, i.e., the meaning of the scores they produce. This particular aspect is of major importance since the semantics of measures must explicitly be understood by end-users: it conditions the relevance to use a specific measure in a particular context.

Nevertheless, the semantics of semantic measures is generally not discussed in proposals (except some broad distinction between the notion of semantic similarity and relatedness). However, semantic similarity based on taxonomies can have different meanings depending on the assumptions on which they rely. In this introduction, we have underlined that the semantics associated to semantic measures can only be understood w.r.t: (i) the semantic proxy used to support the comparison, (ii) the mathematical properties associated to the measures, and (iii) the semantic evidence and assumptions on which the measures are based.

The semantics of the measures can therefore only be captured if a deep characterisation of semantic measures is provided. In recent decades, researchers have mainly focused on the design of semantic measures, and despite the central role of the semantics of semantic measures, few contributions have focused on this specific aspect. This can be partially explained by the fact that numerous semantic measures have been designed in order to mimic human appreciation of semantic similarity/relatedness. In this case, the semantics to be carried by the measures is expected to be implicitly constrained by the benchmarks used to evaluate the accuracy of measures. Nevertheless, despite evaluation protocols based on *ad hoc* benchmarks being relevant to compare semantic measures in specific contexts of use, they do not give access to a deep understanding of measures. Indeed, even if their usefulness is not to discuss, they often do not provide enough general knowledge on semantic measures, e.g. to predict the behaviour of measures in other usage contexts.

There are numerous implications involved in a better characterisation of semantic measures. We have already stressed its importance for the selection of semantic measures in specific contexts of use. Such a characterisation could also benefit cognitive sciences. Indeed, as we saw in Section 1.2.1, cognitive models aiming to explain human appreciation of similarity have been supported by the study of properties expected by the measures. As an example, remember that spatial models have been challenged according to the fact that human appreciation of similarity has proven not to be in accordance with axioms of distance. Therefore, characterising (i) which semantic measures best performed according to human expectations of semantic similarity/relatedness and (ii) the properties satisfied by these measures could help cognitive scientists to improve existing models of similarity or to derive more accurate ones.

In Chapter 1, we have presented an overview of the various semantic measures which have been proposed to compare units of language, concepts or instances which are semantically characterised. In particular, in Section 1.3, we distinguished various aspects of semantic measures which must be taken into account for their broad classification: (i) The types of elements which can be compared, (ii) The semantic proxies used to extract semantic evidence on which the measures will be based and (iii) The canonical form adopted to represent the compared elements and therefore enable the design of algorithms for their comparison.

In Section 1.2.2, based on several notions introduced in the literature, we proposed a characterisation of the general semantics which can be associated to semantic measures (e.g., similarity, relatedness, distance, taxonomic distance). In Section 1.2.3, we recalled some of the mathematical properties which can be used to further characterise semantic measures. Finally, throughout this introduction, and particularly in Section 2.2 and Section 3.3, we distinguished extensive semantic evidence on which corpus-based and knowledge-based semantic measures can be defined. We also underlined the assumptions associated to their consideration.

We encourage designer of semantic measures to provide an in-depth characterisation of measures they propose. To this end, they can use, among others, the various aspects and properties of the measures distinguished in this book and briefly summarised above.

We also encourage the communities involved in the study of semantic measures to better define what a good semantic measure is and to define exactly what makes one measure better than another. Within this goal, the study of the role of usage context seems to be of major importance. Indeed, the accuracy of measures can only be discussed w.r.t specific expectations of measures. Several other properties of measures could also be taken into account and further investigated:

- Algorithmic complexity, which is of major importance for most of concrete use cases, but so far only poorly discussed in the literature.

- Degree of control on the semantics of the scores produced by the measures. A measure could be evaluated according to the understanding of its semantics. It could also be interesting to consider the degree of control a end-user can have on the semantics of the results. Indeed, this aspect is important in several applications based on semantic measures, e.g. synonymy detection.

- The trust which can be associated to a score. Is it possible to evaluate the trust we can have on a specific similarity or relatedness assessment, e.g. based on prior analysis of the semantic proxy which is used.

- The robustness of a measure, i.e., the capacity for a measure to produce robust scores considering the uncertainty associated to expected scores, or disturbances of the semantic proxies on which the measure relies (modification of the ontologies, corpus modifications). As an example, we proposed an approach for studying the robustness of a semantic measure in (Janaqi et al., 2014).

- The discriminative power of the measure, i.e., the distribution of the scores produced by a measure.

## Provide tools for the study of semantic measures

The communities studying and using semantic measures require software solutions, benchmarks, and theoretical tools to compute, compare and analyse semantic measures.

### Develop datasets

There are a host of benchmarks for evaluating semantic similarity and relatedness (Rubenstein and Goodenough, 1965; Miller and Charles, 1991; Finkelstein et al., 2002; Pedersen et al., 2007; Pakhomov et al., 2010; Halawi et al., 2012; Ziegler et al., 2012) – several references and additional information are provided in Chapter 4. Most of them aim at evaluating the accuracy of semantic measures according to human appreciation of similarity/relatedness. For the most part, they are composed of a reduced number of entries, e.g., pairs of words/concepts, and have been computed using a reduced pool of subjects.

Initiatives for the development of benchmarks must be encouraged in order to obtain larger benchmarks in various domains of study. Word-to-word benchmarks must be conceptualised (as much as possible)[1] in order for them to be used to evaluate knowledge-based semantic measures. It is also important to propose benchmarks which are not based on human appreciation of similarity, i.e., benchmarks relying on an indirect evaluation strategy – evaluations based on the analysis of the performance of processes which rely on semantic measures.

Nevertheless, we underline that important efforts have recently been made to propose valuable evaluation campaigns related to semantic measures. Thus, at SemEval conferences[2], SEM tasks have been organised in 2012, 2013, 2014 and 2015 in order to compare systems in several task related to semantic similarity, e.g. text similarity, cross-level semantic similarity[3](Agirre et al., 2012).

### Develop generic open-source software

An overview of the main software solutions dedicated to semantic measures is provided in Appendix D. They are of major importance to: (i) ease the use of the theoretical contributions related to semantic measures, (ii) support large scale comparisons of measures and therefore (iii) better understand the measures and (iv) develop new proposals.

Software solutions dedicated to distributional measures are generally developed without being restricted to a specific corpus of texts. They can therefore be used in a large diversity of contexts of use, as long as the semantic proxy considered corresponds to a corpus of texts, e.g. DKPro (Bär et al., 2013), Semilar (Rus et al., 2013), Disco (Kolb, 2008) among others (Harispe et al., 2013b).

Conversely, software solutions dedicated to knowledge-based semantic measures are generally developed for a specific domain, e.g., WordNet (Pedersen et al., 2004), UMLS (McInnes et al., 2009) – you can also refer to the large number of solutions developed for the Gene Ontology alone (Harispe et al., 2013b). Such a diversity of software is limiting for designers of semantic measures since implementations made for a specific ontology cannot be reused in applications relying on others ontologies. In addition, it hampers the reproducibility of results since some of our experiments have shown that specific implementations tend to produce different results[4]. In this context, we encourage the development of generic open-source software solutions which are not restricted to specific ontologies. This is challenging since the formalism used to express ontologies is not always the same and specificities of particular ontologies sometimes deserve to be taken into account in order to develop semantic measures. However, there are several cases in which generic software can be developed. As an example, numerous knowledge-based semantic measures rely on data structures corresponding to poset or more generally semantic graphs. Other measures are designed to take advantage of ontologies expressed in standardised languages such as RDF(S), OWL. Generic software solutions can be developed to encompass these cases. Reaching such a goal could open interesting perspectives. Indeed, based on such generic and robust software supported by several communities, domain specific tools and various programming language interfaces can subsequently be developed to support specific use cases and ontologies.

---

[1] E.g. using DBpedia URIs.
[2] http://en.wikipedia.org/wiki/SemEval
[3] When language units of different sizes are compared.

In this context, we initiated the Semantic Measures Library project which aims to develop open source software solutions dedicated to semantic measures, e.g. the Semantic Measures Library (SML) and Toolkit (Harispe et al., 2014b). These software solutions can be used for large-scale computation and analysis of semantic similarities, proximities or distances between terms or concepts defined in ontologies, e.g., structured vocabularies, taxonomies, RDF graphs. The comparison of instances (e.g., documents, patient records, genes) annotated by concepts is also supported. An important aspect of these software solutions is that they are generic and are therefore not tailored to a specific application context. They can thus be used with various controlled vocabularies and knowledge representation languages (e.g. OBO, RDF, OWL). The project targets both designers and practitioners of semantic measures providing a Java source code library, as well as a command-line toolkit which can be used on personal computers or computer clusters. More information about this project can be found in the Appendix D and at http://www.semantic-measures-library.org.

The diversity of software solutions is also beneficial as it generally stimulates the development of robust solutions. Therefore, another interesting initiative, complementary to the development of generic software solutions dedicated to semantic measures, could be to provide generic and domain specific tests to facilitate both the development and the evaluation of software solutions. Such tests could for instance inventory expected scores of semantic measures for a reduced example of a corpus/ontology. This specific aspect is important in order to standardise software solutions dedicated to semantic measures and to ensure the users of specific solutions that the score produced by measure implementations are in accordance with the original definitions of the measures.

As discussed in (Harispe et al., 2013b) and other contributions, the evaluation of semantic measures is mainly governed by empirical studies used to assess their accuracy according to expected scores/behaviours of the measures. Therefore, the lack of open-source software solutions implementing a large diversity of measures hampers their study. It explains, for instance, that evaluations of measures available in the literature only involve the comparison of a subset of measures which is not representative of the diversity of semantic measures available today. Initiatives aiming at developing robust open-source software solutions which give access to a large catalogue of measures must therefore be encouraged. It is worth noting the importance of these solutions being open-source. Our communities also lack open-source software dedicated to the evaluation of semantic measures. Indeed, despite some initiatives such as DKPro Similar and CESSM (Collaborative Evaluation of Semantic Similarity Measures) - please refer to Appendix D -, evaluations are not made through a common framework as it is done in most communities, e.g. information retrieval (Voorhees and Harman, 2005; NIST, 2012), ontology alignment (Grau et al., 2013; Euzenat and Shvaiko, 2013). Large efforts have therefore to be made to ease the systematic use of the evaluation protocols presented in Chapter 4. This is mandatory to finely compare and evaluate semantic measures in a large scale fashion. Complementary to evaluation campaign such as SemEval, the development of such tools must ensure fair comparison of results, as well as experiment reproducibility - one of the aim of three recent tools: DKPro Similarity, Semilar and Semantic Measures Library.

**Develop theoretical tools**

It is currently difficult to study the overwhelming amount of proposed semantic measures, e.g., deriving the interesting properties of measures requires the analysis of each measure independently, which is limiting and hampers both theoretical and empirical analyses of measures. However, several initiatives have proposed theoretical tools to ease the characterisation of measures, for instance by means of measure unification (Cross et al., 2013; Cross, 2006; Cross and Yu, 2010; Pirró and Euzenat, 2010a; Sánchez and Batet, 2011; Mazandu and Mulder, 2013; Cross et al., 2013; Harispe et al., 2013c), and by means of semantic model unification in distributional semantics, e.g. (Baroni and Lenci, 2010).

---

[4]You can refer to the evaluation proposed at https://github.com/sharispe/sm-tools-evaluation.

As an example, in (Harispe et al., 2013c), we show how several knowledge-based measures can be unified through a limited number of abstract core elements – this highlights that several measure expressions which have been proposed independently in the literature are only particular expressions of more general measures. Therefore, based on this theoretical characterisation of measures, we further illustrate how thousands of specific semantic measure expressions can be analysed in detail. In particular, we demonstrate how such an approach, based on a theoretical framework and implemented using SML open-source library, can be used to analyse the core elements of semantic measures which best impact measure accuracy in a specific context of use.

More efforts must be done to use existing theoretical tools and to propose new ones and to analyse semantic measures. Indeed, these contributions open interesting perspectives on studying groups of measures and enable to obtain more thoughtful results on semantic measures. They are also essential to better understand limitations of existing measures and the benefits of new proposals. Finally, they are central to distinguish the main components on which measures rely, and to improve our understanding of semantic measures based on detailed characterisation of family of measures.

### Standardise ontology handling

Several knowledge-based semantic measures are based on the transformation of an ontology into a semantic graph. However, this process of transformation is currently overly subject to interpretations and deserves to be carefully discussed and formalised. Indeed, we stress that numerous measures consider ontologies as semantic graphs despite the fact that the formalism on which some ontologies rely cannot be mapped to semantic graphs without reductions – this is the case for some expressive logic-based ontologies (e.g. numerous ontologies which are expressed into OWL). The impact of such a reduction of ontologies is of major importance since it can highly affect semantic measure results[5]. Therefore, the treatment performed to map an ontology into a semantic graph is generally not documented, which explains some of the difficulties encountered to reproduce the results of some experiments[6]. In this context, we think that ontology handling must be carefully discussed and standardized if possible. This is important to ensure that theoretical contributions such as measure definitions can indeed be compared without the comparison being dependent on the implementation of the measure.

### Compositionality for comparing large units of language

One of the most important challenges offered to semantic measure designer is to integrate and to adapt existing approaches in order to compare large units of language such as sentences, paragraphs or documents. To this end, important research efforts are made and are required to study semantic compositionality w.r.t semantic similarity: (i) how to evaluate the semantics of larger units of language than words or concepts, (ii) how to capture this semantics into semantic models and (iii) how to design measures adapted to these models.

As we saw in Chapter 2 and more particularly in C, several attempts have been proposed to adapt distributional models and new original approaches are designed (Kamp et al., 2014; Baroni et al., 2014) – challenges related to this field of study aim to take advantage of more refined language analysis to improve existing models, e.g. by considering linguistic expressions, e.g., anaphora, negation or named entity recognition to cite a few.

### Corpus-based semantic measures and language specificities

It is important to analyse semantic measures in order to ensure that accurate models of semantic similarity and semantic relatedness are available to process resources which are not expressed in English. Among others, open problems are related to: (i) the evaluation of the use of exiting models in different languages, (ii) the definition of measures which are accurate for multiple languages and (iii) the definition and study of language-specific processes which can be used to improve measure accuracy.

---

[5] Consider, for instance, a taxonomy in which redundant relationships have been defined – redundancies highly impact shortest path computation. Should they be considered?

[6] Refer to the evaluation proposed at: https://github.com/sharispe/sm-tools-evaluation

**Promote interdisciplinarity**

From cognitive sciences to biomedical informatics, the study of semantic measures involves numerous communities. Efforts have to be made to promote interdisciplinary studies and to federate the contributions made in the various fields. We briefly provide a non-exhaustive list of the main communities involved in semantic measure study and the communities/fields of study which must be relevant to solicit to further analyse semantic measures. The list is alphabetically ordered and may not be exhaustive:

- *Biomedical Informatics* and *Bioinformatics* are very active in the definition and study of semantic measures; these communities are also active users of semantic measures.

- *Cognitive Sciences* propose cognitive models of similarity and mental representations which can be used to (i) improve the design of semantic measures and (ii) better understand human expectations w.r.t similarity/relatedness. This community can also use empirical evaluation studies of semantic measures to discuss the cognitive models it proposes.

- *Complexity Theory* is an important field of study which is essential to analyse complexity of semantic measures.

- *Geoinformatics* defines and studies numerous semantic measures. Members of this community are also active users of semantic measures.

- *Graph Theory* proposed several major contributions relative to graph processing. Such theoretical works are essential for the optimisation of measures relying on network-based ontologies. This community will probably play an important role on knowledge-based semantic measures in the near future, since large semantic graphs composed of billions of relationships are now available – processing such graphs requires the development of optimised techniques.

- *Information Retrieval* defines and studies semantic measures taking advantage of corpus of texts or ontologies. They also actively contribute to the development of topic models which are central for the design of corpus-based distributional measures.

- *Information Theory* plays an important role in better understanding the notion of information and in defining metrics which can be used to capture the amount of information which is conveyed, shared and distinct between compared elements, e.g., notion of information content.

- *Knowledge Engineering* studies and defines ontologies which will further be used in applications involving semantic measure calculus. It could, for instance, play an important role in characterising the assumptions made by semantic measures.

- *Linguistics and Natural Language Processing* are actively involved in the definition of distributional measures. They propose models to characterise corpus-based semantic proxies and to define measures for the comparison of units of language.

- *Logic* defines formal methods to express and take advantage of knowledge. This community can play an important role in characterising the complexity of knowledge-based semantic measures, for instance.

- *Machine Learning* plays an important role in the definition of techniques and parameterised functions which can be used for the definition and tuning of semantic measures.

- *Measure Theory* defines a mathematical framework to study and define the notion of measure. It is essential for deriving properties of measures, better characterising semantic measures and taking advantage of theoretical contributions proposed by this community.

- *Metrology* studies both theoretical and practical aspects of measurements.

- *Optimisation area* may lead to important contributions in order to optimise measures, to study their complexity and to improve their tuning.

- *Philosophy* also plays an important role in the definition of essential concepts on which semantic measures rely, e.g., definition of the notions of *Meaning*, *Context*.

- *Semantic Web* and *Linked Data* define standards (e.g., languages, protocols) and processes to take advantage of ontologies. The problem of ontology alignment and instance matching are actively involved in the definition of (semantic) measures based on ontologies.

- *Statistics* and *Data Mining* propose several data analysis techniques which can be used to better characterise and understand semantic measures.

## Study the algorithmic complexity of semantic measures

Most contributions have focused on the definition of semantic measures. However, their algorithmic complexity is *near inexistent* despite the fact that this aspect is essential for practical applications to be efficient and tractable. Therefore, to date, no comparative studies can be made to discuss the benefits of using computationally expensive measures. However, these aspects are essential for comparing semantic measures. Indeed, in most application contexts, users will prefer to reduce measure accuracy for a significant reduction of the computational time and resources required to use a measure. To this end, designers of semantic measures must as much as possible provide the algorithmic complexity of their proposals. In addition, as the theoretical complexity and the practical efficiency of an implementation may differ, developers of software tools must provide metrics to discuss and compare the performance of measure implementations.

## Support context-specific selection of semantic measures

Both theoretical and software tools must be proposed to orient end-users of semantic measures in the selection of measures according to the needs defined by their application contexts. Indeed, despite the fact that most people only (*blindly*) consider benchmark results in order to select a measure, efforts have to be made in order to orient end-users in the selection of the best suited approaches according to their usage context – understanding the implications (if any) of using one approach compared to another. To this end, the numerous properties of the measures presented in this book can be used to guide the selection of semantic measures. However, numerous large-scale comparative studies have to be performed in order to better understand the benefits of selecting a specific semantic measure in a particular context of use.

# Appendix A

# Examples of syntagmatic contexts

To illustrate examples of syntagmatic context we will consider the raw text extracted from the Wikipedia article which refers to the topic Tree[1] – sentences are numbered:

"*(1) In botany, a tree is a perennial plant with an elongated stem, or trunk, supporting branches and leaves in most species. (2) Trees play a significant role in reducing erosion and moderating the climate. (3) They remove carbon dioxide from the atmosphere and store large quantities of carbon in their tissues. (4) Trees and forests provide a habitat for many species of animals and plants.*"

For convenience, in this example only nouns are considered; we obtain the following surface representation of the text:

*(1) botany tree plant stem trunk branch leaf species. (2) tree role erosion climate. (3) carbon dioxide atmosphere quantity carbon tissue. (4) tree forest habitat species animal plant.*

We will now present three examples of syntagmatic context definitions and examples of processing which can be done to represent words w.r.t their syntagmatic contexts. To reduce the size of the matrices which are used to characterise the words, not all the nouns which compose the surface representation presented above will be represented in matrices. To each context definition we present:

- The corresponding word-context matrix which characterises the words w.r.t the context definition which has been defined.

- A Figure showing the similarity of the vector representation of each word according to a two dimensional projection – the aim is to intuitively appreciate the impact that the notion of context can have on the definition of words representation and therefore on the estimation of word similarity. Technically speaking the figure is generated using a MultiDimensional Scaling (MDS) algorithms (Borg and Groenen, 2005) on a similarity matrix between the word vector representations[2]. The results will not be discussed in detail but we invite the reader to compare the results which have been obtained using those reduced examples.

Three examples are presented:

1. The context is defined as a sentence, $s_j$ refers to Sentence $j$. The semantic model is a word-sentence matrix and a word is therefore represented by a vector in which each dimension refers to a sentence – therefore, if the word $w_i$ composes the sentence $s_j$ the cell $(w_i, s_j)$ will be set to 1.

2. The context is defined as a sentence – the semantic model is a word-word matrix which highlights the number of time two words co-occured in sentences. The co-occurrences are computed counting all pairs of words which can be made from each sentence. Note that this matrix can be computed from the matrix which have been obtained in Example 1.

3. The semantic model is a word-word matrix. The context is defined as a five word window, i.e. two words to the left and the right of the focal word. For instance, considering Sentence 1, the window associated to the focal word *stem* would be: *botany tree plant* **stem** *trunk branches leaves species*. The sliding step of the window is defined as one word. Finally for each iteration, the current window is processed and the co-occurrences associated to the focal word are updated, i.e. by counting the pairs of words which can be built with the focal word and the words which compose the window.

---

[1] http://en.wikipedia.org/wiki/Tree (accessed 26/06/14)
[2] The similarity between the vectors is computed using the cosine similarity – presented in Section 2.3.2, Equation 2.4.

|  |  | $s_1$ | $s_2$ | $s_3$ | $s_4$ |
|---|---|---|---|---|---|
| atmosphere | $w_1$ | 0 | 0 | 1 | 0 |
| botany | $w_2$ | 1 | 0 | 0 | 0 |
| branch | $w_3$ | 1 | 0 | 0 | 0 |
| climate | $w_4$ | 0 | 1 | 0 | 0 |
| forest | $w_5$ | 0 | 0 | 0 | 1 |
| habitat | $w_6$ | 0 | 0 | 0 | 1 |
| leaf | $w_7$ | 1 | 0 | 0 | 0 |
| plant | $w_8$ | 1 | 0 | 0 | 1 |
| species | $w_9$ | 1 | 0 | 0 | 1 |
| stem | $w_{10}$ | 1 | 0 | 0 | 0 |
| tree | $w_{11}$ | 1 | 1 | 0 | 1 |
| trunk | $w_{12}$ | 1 | 0 | 0 | 0 |

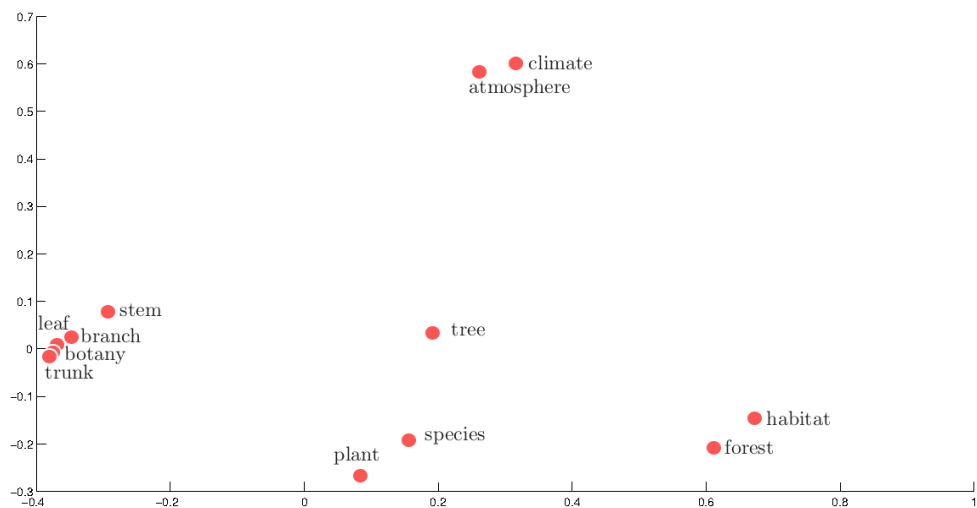

Figure A.1: MDS of a similarity matrix computed from the vector representation of words extracted from a word-sentence matrix.

|  |  | $w_1$ | $w_2$ | $w_3$ | $w_4$ | $w_5$ | $w_6$ | $w_7$ | $w_8$ | $w_9$ | $w_{10}$ | $w_{11}$ | $w_{12}$ |
|---|---|---|---|---|---|---|---|---|---|---|---|---|---|
| atmosphere | $w_1$ | 1 | 0 | 0 | 0 | 0 | 0 | 0 | 0 | 0 | 0 | 0 | 0 |
| botany | $w_2$ | 0 | 1 | 1 | 0 | 0 | 0 | 1 | 1 | 1 | 1 | 1 | 1 |
| branch | $w_3$ | 0 | 1 | 1 | 0 | 0 | 0 | 1 | 1 | 1 | 1 | 1 | 1 |
| climate | $w_4$ | 0 | 0 | 0 | 1 | 0 | 0 | 0 | 0 | 0 | 0 | 1 | 0 |
| forest | $w_5$ | 0 | 0 | 0 | 0 | 1 | 1 | 0 | 1 | 1 | 0 | 1 | 0 |
| habitat | $w_6$ | 0 | 0 | 0 | 0 | 1 | 1 | 0 | 1 | 1 | 0 | 1 | 0 |
| leaf | $w_7$ | 0 | 1 | 1 | 0 | 0 | 0 | 1 | 1 | 1 | 1 | 1 | 1 |
| plant | $w_8$ | 0 | 1 | 1 | 0 | 1 | 1 | 1 | 2 | 2 | 1 | 2 | 1 |
| species | $w_9$ | 0 | 1 | 1 | 0 | 1 | 1 | 1 | 2 | 2 | 1 | 2 | 1 |
| stem | $w_{10}$ | 0 | 1 | 1 | 0 | 0 | 0 | 1 | 1 | 1 | 1 | 1 | 1 |
| tree | $w_{11}$ | 0 | 1 | 1 | 1 | 1 | 1 | 1 | 2 | 2 | 1 | 3 | 1 |
| trunk | $w_{12}$ | 0 | 1 | 1 | 0 | 0 | 0 | 1 | 1 | 1 | 1 | 1 | 1 |

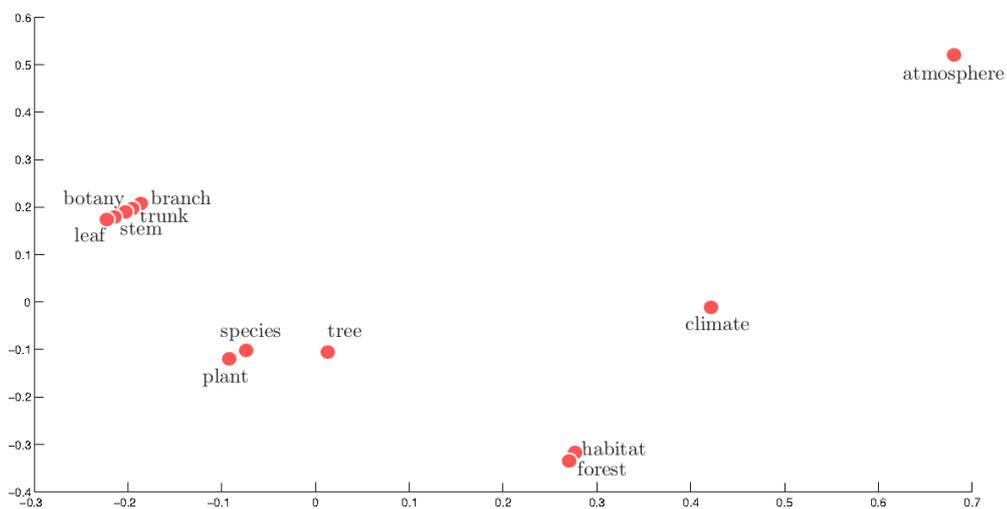

Figure A.2: MDS of a similarity matrix computed from the vector representation of words extracted from a word-word matrix.

|  |  | $w_1$ | $w_2$ | $w_3$ | $w_4$ | $w_5$ | $w_6$ | $w_7$ | $w_8$ | $w_9$ | $w_{10}$ | $w_{11}$ | $w_{12}$ |
|---|---|---|---|---|---|---|---|---|---|---|---|---|---|
| atmosphere | $w_1$ | 1 | 0 | 0 | 0 | 0 | 0 | 0 | 0 | 0 | 0 | 0 | 0 |
| botany | $w_2$ | 0 | 1 | 0 | 0 | 0 | 0 | 0 | 2 | 0 | 0 | 2 | 0 |
| branch | $w_3$ | 0 | 0 | 1 | 0 | 0 | 0 | 2 | 0 | 2 | 2 | 0 | 2 |
| climate | $w_4$ | 0 | 0 | 0 | 1 | 0 | 0 | 0 | 0 | 0 | 0 | 0 | 0 |
| forest | $w_5$ | 0 | 0 | 0 | 0 | 1 | 2 | 0 | 0 | 2 | 0 | 2 | 0 |
| habitat | $w_6$ | 0 | 0 | 0 | 0 | 2 | 1 | 0 | 0 | 2 | 0 | 2 | 0 |
| leaf | $w_7$ | 0 | 0 | 2 | 0 | 0 | 0 | 1 | 0 | 2 | 0 | 2 | 2 |
| plant | $w_8$ | 0 | 2 | 0 | 0 | 0 | 0 | 0 | 2 | 2 | 2 | 2 | 2 |
| species | $w_9$ | 0 | 0 | 2 | 0 | 2 | 2 | 2 | 2 | 2 | 0 | 2 | 0 |
| stem | $w_{10}$ | 0 | 0 | 2 | 0 | 0 | 0 | 0 | 2 | 0 | 1 | 2 | 2 |
| tree | $w_{11}$ | 0 | 2 | 0 | 0 | 2 | 2 | 2 | 2 | 2 | 2 | 3 | 0 |
| trunk | $w_{12}$ | 0 | 0 | 2 | 0 | 0 | 0 | 2 | 2 | 0 | 2 | 0 | 1 |

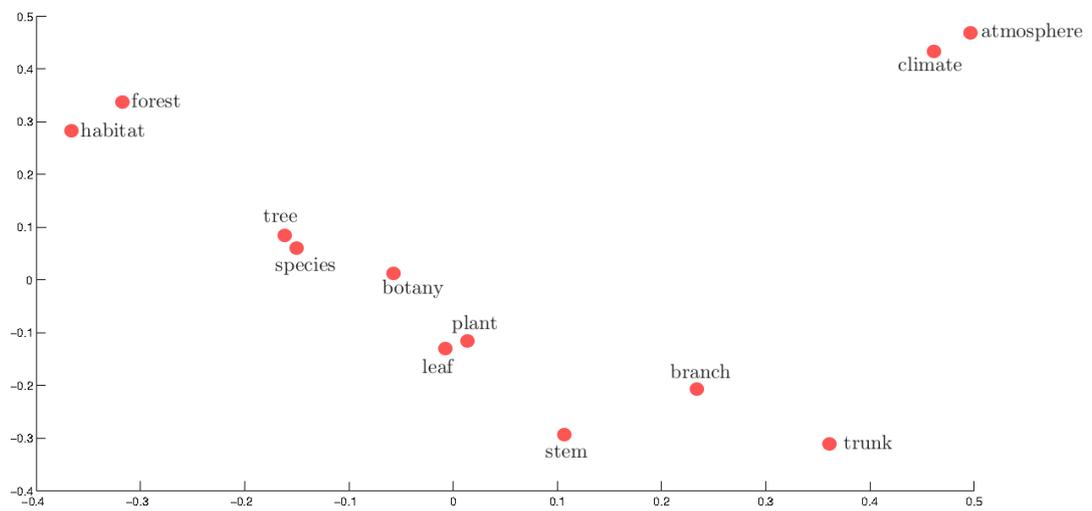

Figure A.3: MDS of a similarity matrix computed from the vector representation of the words extracted from a word-word matrix which has been computed using a sliding window.

# Appendix B

# A brief introduction to Singular Value Decomposition

Singular Value Decomposition (SVD) is one of the fundamental algorithms of linear algebra. It is largely used to factorize matrices in many field related to data analysis. In Computational Linguistics, SVD has been made popular through its use in distributional models that aim to derive (word) meaning representations through the analysis of (word) co-occurrence matrices, e.g. Latent Semantic Analysis (LSA) – refer to Section 2.3 for an introduction to these models.

The main motivation of SVD is to factorize matrices. This is interesting since several distributional models rely on highly dimensional but very sparse spaces. Therefore, their sizes tend to be very large which can be limiting for their processing in practice. SVD is therefore used as a way to reduce model sizes. In addition, since the process of factorization identifies and orders the dimensions along which data points exhibit the most variation, the reduction has the interesting property to reduce noise as well as sparsity. This contributes to improve the quality of the vector representations by focusing on informative dimensions. To this end, SVD consider that the optimal reduction, is the lower dimensional subspace onto which the original data have to be projected while preserving maximal data variability. This is done by grouping dimensions with highly correlated values – generally 300 dimensions are considered to build vector representation of words. It is however important to stress that the semantics of newly created dimensions can only be explained as a linear combination of the original dimensions  this why, when applied to distributional models, SVD are often considered as a method able to reveal latent concepts and high-order co-occurrences. Figure B.1 illustrates the process which is performed by SVD.

Nevertheless, one of the limits of this technique is that, as for any projection/factorization method, the reduction will necessarily imply some loss of information. Indeed, two points (e.g. words) that were originally differentiated into an highly dimensional space can both be projected into the same coordinates in the lower dimensional subspace that has been selected, i.e. they can be both reduced to the same vector representation. Another limit of SVD is their high algorithmic complexity. It can be limiting for processing large models - it must also be computed each time the model is updated.

Technical aspects of SVD are largely covered in the literature. We recommend referring to the seminal book of (Golub and Van Loan, 2012). The articles of (Deerwester et al., 1990; Landauer et al., 1998; Turney and Pantel, 2010) are also very good introductions to SVD applied to distributional models.

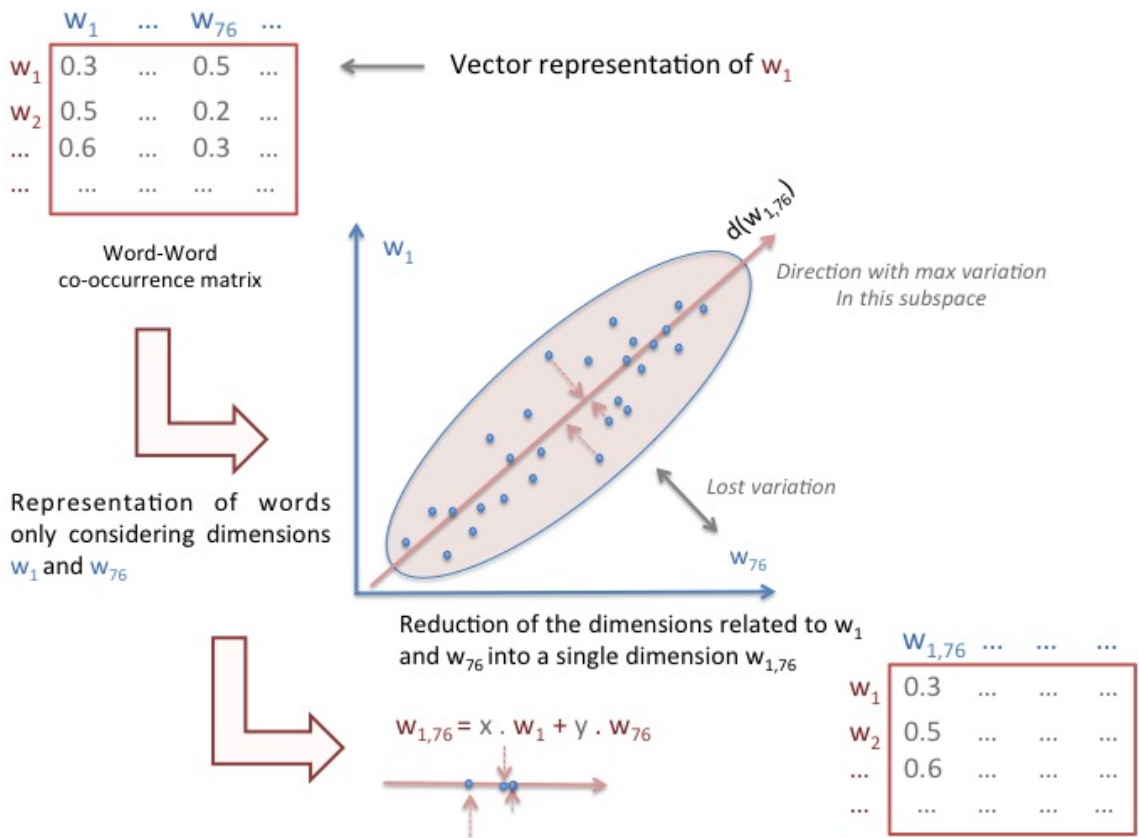

Figure B.1: Illustration of the process of singular value decomposition on a word-word matrix. The figure shows how the matrix space is reduced by detecting correlated dimensions, finding subspace directions with high variability of values, and factorizing this dimensions through linear composition.

# Appendix C

# A brief overview of other models for representing units of language

The definition of models from which the meaning of units of language can be derived is a vibrant and prolific topic in the literature since several decades. This problematic is indeed central for Computational Linguistics and plays an important role for the design of corpus-based semantic measures. As we saw, most corpus-based measures that can be used for comparing words only differ regarding the approach they adopt to represent word meaning, generally through vector-based representations. The way those vectors are next compared is most often unchanged, e.g. using the well-known cosine similarity measure. The large literature related to the definition and combination of models that can be used to manipulate the semantics of complex units of language is therefore central to deeply understand corpus-based semantic measures. Nevertheless, as the reader will understand, this book cannot provide an in-depth presentation of all the models that have been proposed so far. An introduction to several models that rely on the distributional hypothesis has already been proposed in Chapter 2, e.g., VSM, LDA, ESA. Understanding more sophisticated models most often requires technical knowledge about specific machine learning techniques (e.g. neural networks), statistics and linear algebra; this makes more complex their introduction to a large audience. In addition, these sophisticated models are most often not proposed in the aim of assessing semantic similarity/relatedness of units of language, or to generate explicit representations of their meaning. They rather envision a large goal, to model language. These models, denoted language models, can however generally be used to derive useful representations of units of language – if they do not explicitly integrate such representations in their modeling. Here, we propose a brief introduction to several popular language models. Our main goals are to (i) underline the intuitions that motivate their definitions and (ii) to stress how they can be used in the aim of defining semantic measures – most of the time to compare words. We will next briefly discuss some of the models that have been proposed to capture the semantics of larger units of language. Technical details of the contributions that we will present can be found on provided references.

## Language models

The distributional models we have seen so far can be used to derive word meaning representations based on the analysis of statistics on their usage – they rely on direct or indirect implementations of the distributional hypothesis (cf. Chapter 2). As from now, you can retain that both distributional and language models proposed in the literature, without exception, only differ on the way they capture and take advantage of this information. The aim of distributional models was generally to build word representations in a real space of $n$ dimensions, by representing words through real-value vectors. The language models we will now introduce propose to model language for answering specific tasks in Computational Linguistics or related fields, e.g. machine translation, speech recognition or handwriting recognition to mention a few. These approaches aim to model languages in order to enable predictions such as (i) what are the words that are more likely to appear after a given sequence of continuous words, or (ii) what is the more probable context (e.g. sequence of words) considering a specific word. Designing models that are efficient in answering those questions can for instance be used for question answering, spellchecking or disambiguation.

# N-gram models

A simple and popular language model that has been proposed to predict the next word according to given context is the n-gram model (Manning and Schütze, 1999) – since several variations have been proposed we can say that n-gram is a generic approach, a class of models. Depending on the design of the model, a n-gram can be a sequence of $n$ contiguous words, syllables, letters, etc. – here we only consider n-grams of words; $n$ is generally set between 1 and 5 in this case. N-gram models rely on the analysis of the statistical properties of n-grams in texts. These statistical properties are derived by scanning large corpora composed of billions of words using a sliding window of size $n$ with a specific step.[1] Therefore, using those statistics about word sequence observations, it is possible to predict the word $w_n$ based on the context of its $n-1$ preceding words $(w_1, \ldots, w_{n-1})$. More precisely, we can compute the probability that a specific word $w$ will occur after a given sequence $S$ of $n-1$ words; this is defined by $p(w|S)$ :

$$p(w|S) = \frac{o(S,w)}{o(S)} \tag{C.1}$$

With $o(S)$ the number of time the sequence $S$ has been observed and $o(S,w)$ the number of time $S$ was directly followed by the word $w$. Using basics of Information Theory, n-gram models have the ability to predict sequential data by estimating the probability $p(w_n|w_1, \ldots, w_{n-1})$. As any Markov model, these models assume the independence hypothesis, i.e. that $w_n$ only depends on $w1, \ldots, w_{n-1}$.[2] A n-gram model can therefore be used to compute the conditional probability distribution of words knowing a word sequence of size $n-1$. This can be used to derive the likelihood of a word knowing a specific $n-1$ sequence of words. More importantly for our concern, the conditional probabilities of a word for all the sequences that have been observed can be used to obtain a vector representation of a word. Once again, this representation encompasses information about the usage of the word, and is therefore assumed to give access to some of its meaning. Note that this representation is built considering a specific type of syntagmatic context (refer to Section 2.2). Nevertheless, this approach can be difficult to implement in practice because of the large number of sequences that are observed; this number corresponds to the number of dimensions of the space into which words will be represented. Indeed, due to the high variability of n-grams in language, the n-gram model suffers the curse of dimensionality. Otherwise stated, the number of parameters of n-gram models (i.e. number of $n-1$ sequences that have been observed) grows exponentially with $n$. Nevertheless, most of the dimensions will be set to zero and only very sparse vectors will have to be compared. In (Joubarne and Inkpen, 2011) the authors adopt two other approaches that study the distribution of words among n-grams. The first approach estimates the relatedness of two words by considering their PMI (Point-wise Mutual Information), i.e. the probability that the two words occur in the same n-gram divided by the probabilities they occur. The second approach relies on the Second Order Co-occurrence PMI (SOC-PMI) that have been defined in (Islam and Inkpen, 2006). N-gram statistics have also been used in the design of semantic measures dedicated to the comparison of sentences (Bär et al., 2012).

N-gram models provide good results when trained with large corpus. However, one of the main drawbacks of this probabilistic model is that, per definition, it poorly performs with n-grams that are not observed in the learning corpus. As an example, consider the following sequence of words: "*he was driven his bike*'. If the model is built using a 4-gram strategy, the likelihood associated to the word *bike* considering the context "*he was driving*" will be derived from the amount of time the association context+"*bike*" has been encountered in the corpus. Therefore, if the association never occurs, the likelihood of the word *bike* knowing aforementioned context will *de facto* be set to zero. Since the model suffers the curse of dimensionality, this problem increases the larger $n$ is set. This stresses the second limit of these models: their dependency to very large datasets. Several strategies have therefore been studied to overstep these limits. As an example, several forms of smoothing techniques have been studied to avoid and prevent problems induced by unseen n-grams. These techniques are essential to ensure good performance. Several other adaptations have been proposed to overcome the data sparsity issue, i.e. the fact that input data will only give access to a reduce set of the language diversity; skip-gram is one of these proposals (Guthrie et al., 2006). This model generalises n-gram by allowing words to be skipped in order to generate n-gram that would not have been seen while they could improve language modeling. As an example, considering the phrase "*France is a beautiful country...*", a 1-skip-3gram will generate the following 3-grams : "*France is a*", "*France a beautiful*", "*is a beautiful*", "*is beautiful country*", "*a

---

[1] Google provides interesting results on n-gram statistics over time at: https://books.google.com/ngrams
[2] The fact that this assumption is rough for language modeling is not questioned. This restriction is however required to ensure tractable algorithmic complexity.

*beautiful country*", "*a country*". Another interesting variation of n-gram models are cache models (Kuhn and De Mori, 1990). They rely on the simple idea that words recently observed are more likely to appear again in the short term. Consider for instance the case of rare words that will often occur in a specific text, e.g. the word *Paris* will often occur in a text talking about the French capital. Therefore, cache models propose to keep track of recent contexts when estimating the probability distributions of words. Others variants are class-based models (Brown et al., 1992); they propose to replace words by classes of words sharing specific features. By reducing the number of n-grams that are possible, they reduce the number of parameters of the model (i.e. fighting the curse of dimensionality). This approach can also be used to derive probability distributions for unseen contexts.

Contribution and results related to n-gram models are numerous; we have here introduced basics of this popular class of models. We have also discussed how they can be used to compute the relatedness of words. Despite their limits, n-gram models (including variants) are still very popular and competitive language models. They are used in numerous practical applications. This is especially true considering that improvements that are achieved using more complex models are often made at the cost of a significant increase of computational complexity – an increase that is often unacceptable considering the constraints of real world applications.

Even if adaptations have been proposed to solve drawbacks of n-gram models, the dependency of this approach to very large corpora is still considered as incompatible with several contexts – remind that we are talking about billions of words, constituting such a dataset can be problematic in specific domains. This has encouraged the study of models that are less greedy and that can be used considering significantly smaller input datasets. We can therefore say that by being very costly in terms of input datasets but very cheap in terms of computational complexity, n-gram models implement a kind of no-brain or brute force approach in modeling language. Alternative language models, by being cheap in terms of input dataset requirement but costly in terms of computational complexity, propose to adopt an opposite strategy using more refined learning approaches. Some of these models are now briefly introduced.

Some statistical learning models have been studied to overcome the intrinsic limitations of n-gram models. During the learning phase, these models directly learn and process representations of units of language – once again, for convenience, we will here focus on word representation. Based on word usage analysis, these approaches build (implicit) word representations by optimizing the performance of a statistical learning algorithm with respect to a specific goal. As an example, such a model can be a neural network that, similarly to n-gram, will be able to predict the most probable word considering a given context. Such language models implicitly or explicitly rely on word modeling. A representation of word, often in the form of a vector, can generally be derived from them. Therefore, compared to classical distributional models that will simply build word representations by considering a deterministic approach, learning methods will implicitly learn word representations using heuristics that will optimize models – that internally encompass or depend on these word representations. This is why these methods are generally said to learn word representations, or more explicitly learn models from which word representations can be extracted. It is important to stress that, as we said, the primary source of information available to all methods for learning word representations is still word usage in a corpus – they therefore also extensively rely on the distributional hypothesis. Several type of learning approaches can be used to derive language models. Among the most popular, we will discuss the language models that are based on neural networks. Other types of models will next be cited.

## Neural network models

The main rationale of Neural Network Language based Models (NNLM) is to take advantage of similarities between contexts for assessing the probability distribution of words considering a given context. Indeed, as we saw, when assessing the likelihood of words for a specific context, simplest n-gram models will only consider observations in which this exact context has been observed.[3] However, even if the sequence "*he was driven his bike*" has never been observed, it is intuitive to say that the context "*he was driven his*" is more likely to generate the words *car*, *bike* or *truck* than the words *mountain* or *house*. This can once again simply be explained by the distributional hypothesis: words that frequently occur with the word *drive* are more likely to be referring to things that can be driven (i.e. vehicle). It could therefore be interesting to provide a way to represent contexts in such a way that, by learning the probability distribution associated to a context, knowledge for similar contexts will be generalized – i.e. to design an approach that will be robust to slight variations of contexts. This is the main aim of NNLMs. Similar contexts are, in some sense, processed similarly in order to generalize knowledge that can be extracted from occurrences. As an example, the observation "*he was driven his bike*" will not only contributes to improve the likelihood to see *bike* after the context "*he was driven his*". It will also improve the likelihood to see the words *car, truck*, etc. after this specific context. Technically, the generalization is made possible by projecting observations onto a low-dimensional space, and by building a model that will generate word conditional distribution probabilities from this projection. Therefore, two observations that are similar will (hopefully) be projected the same way. Thus, similar sets of parameters will be learned for similar observations. Parameters of the model are in some sense shared among similar observations. This is an important point: NNLMs share parameters among similar observations in order to overcome the exponential increase of parameters when considering large contexts.

---

[3]We have seen that smoothing have next been proposed to correct this limit but put this aside for now.

A variety of NNLM architectures have been proposed. For language models that aim to compute conditional probabilities of words from a vocabulary $V$ knowing a specific context (i.e. sequence of continuous words), several NNLMs adopt the following setting. They consider that input words are represented by vectors – such representations can simply be built by considering that each word is associated to a unique identifier between 0 and $|V| - 1$, and that its corresponding vector is a vector of size $|V|$ filled with 0 values, expect for the dimension associated to its identifier which is set to 1. These representations are used to derive the representation of a n-word context (sequence observation) – generally setting $n$ between 5 and 10. This is done by building a $n \times |V|$ matrix that contains the vectors of the words of the context. Finally, considering numerous contexts and expected next word probability distribution provided by the input dataset, the NNLM learn optimal parameters of the model and optimal word representations. The aim is to set these variables such as the model will generate word probability distributions best fitting expectations. This setting has first been defined in (Bengio et al., 2003). In this case, word representations are simply learned by the model and can therefore be used in a straightforward manner to compare words using measures designed for comparing vectors.

Contributions on NNLMs have led to many results. Numerous contributions have for instance focused on designing recurrent NNLMs, e.g. to avoid the fact that in (Bengio et al., 2003) the context size had to be defined manually (Mikolov et al., 2013). Among them we invite the reader to refer to the numerous contributions of Mikolov. In conclusion, NNLMs are generally considered to be complex models with regard to their computational requirements. Nevertheless, they have the interesting property of achieving efficient results with reduced input datasets.

More generally we encourage the reader to refer to embedding-based methods that can be used to build elaborated representations of units of language from which semantic similarity and relatedness measures can be defined. In addition, other types of language models have been proposed and will not be introduced in this book. Among others we can cite: conditional restricted Boltzmann machine, (hierarchical) log-bilinear models, global log-bilinears (Pennington et al., 2014). A brief introduction of several models can be found in (Mikolov, 2012).

## Compositionality: representing complex units of language

As we saw, a substantial literature is dedicated to the definition and analysis of models that aim to represent word meaning. An important area of research is also dedicated to the study of how larger units of language such as compound words, noun-adjective pairs or sentences can be represented. This challenge aim to define compositional approaches that represent meaning of units of language by combining different models representing meaning of their constituents. Compositional approaches assume that, since complex units of language are built by compositions (of simpler units) – sentences are sequences of words -, their meaning is also formed by composition. It is here considered that "*the meaning of complex expression is a function of the meaning of its parts and of their syntactic mode of combination*".[4] Central questions faced by compositionality are therefore (i) how to finely capture the semantics of short units of language through representations and (ii) how to combine them to derive meaning of their combination. A large number of contributions related to the former question have already been introduced, we now expose how they can be combined to derive compositional models; once again the following introduction is far to cover the diversity of the literature on this topic.

Simple models of compositionality have for instance proposed to give access to the meaning of a sentence by aggregating vector representations of its words using linear models. As an example (Landauer and Dumais, 1997) propose to represent a sentence by averaging the vector representations of its word. More generally, (Mitchell and Lapata, 2010) present a framework to combine word representations in order to represent the meaning of phrase or sentence into a vector space. This framework can be used to generate models by defining composition through the aggregation of representations by vector functions such as addition and multiplication. Another approach that has been tested by several authors proposes to model composition in terms of matrix multiplication. In (Baroni and Zamparelli, 2010) the authors define a compositional model for pairs of adjective-words by multiplying vector representations of words by matrix representations of adjectives. Several operations that can be performed on vector model to derive compositional models are discussed in (Widdows, 2008). An original approach that defines compositionality in terms of similarity composition is also exposed in (Turney, 2012).

---

[4]http://plato.stanford.edu/entries/compositionality – we strongly encourage the reader to refer to this resource for a complete and historical introduction to the notion of compositionality.

Simplest approaches based on linear models fail to model the complexity of the semantics of large units of languages. This is mainly because word ordering is not taken into account. More generally, any model that only relies on commutative operations such as addition or multiplication will not take into account syntax, order of language constituents, and negation. More complex models have therefore been proposed; some of them are limiting as they generate highly dimensional representations that depend on a large number of parameters. In (Socher et al., 2011), an interesting model based on a nonlinear function is studied to overcome limits of simple compositional models. In (Socher et al., 2012) the authors propose to model compositionality by considering learning models in which words of a sentence, represented by vectors, are structured through a parse tree; matrices are also used to model the effect of the sequential ordering of words on the global semantics of the sentence.

Designing efficient models that are able to capture the meaning of large units of language is an important challenge faced by Computational linguistics. This is also central for the design of accurate semantic measures that will be able to compare complex units of language by finely analysing their semantics. Thanks to an increasing interest on compositional models, and to a better understanding of the perspectives they open, a large number of contributions have recently been made; they have lead to more efficient models but have also underlined the wideness of this complex field of study. We advise interested readers to consult (Kamp et al., 2014) and the website of the *Composes* project[5] that is dedicated to this vibrant topic (Baroni et al., 2014) – the website references important documentations and publications, as well as software, models and datasets used for evaluation.

---

[5]Compositional Operations in Semantic Space, European Research Council project (nr. 283554): `http://clic.cimec.unitn.it/composes`

# Appendix D

# Software tools and source code libraries

This appendix provides an overview of existing software solutions and source code libraries dedicated to semantic measure computation and analysis. These contributions are important to both usage and evaluations of semantic measures. First, they provide implementations of existing proposals to end-users and therefore contribute to the large adoption of semantic measures - this is not a little thing considering that implementations are most often very technical. Second, by providing common platforms for measure evaluation, these development projects are also essential to support research contributions in the field. They enable to reproduce experiments and provide easy-way for researchers to test and evaluate new proposals. This aspect is particularly important since evaluation of semantic measures extensively relies on empirical analyses (c.f. Chapter 4).

Throughout this book we have underlined that numerous groups of researchers are deeply involved in the study of semantic measures, e.g. Natural Language Processing, Artificial Intelligence, Semantic Web and Bioinformatics, to mention a few. Moreover, due to their popularity, numerous proposals exist for a wide range of applications - refer to the large collection of measures presented in chapters 2 and 3. Nevertheless, no extensive software tool today federates the various communities through a common development and application framework. Most of the contributions are supported by few developers and only exist thank to their will to develop, correct and support useful tools for the community.

A few popular and robust solutions exist for computing corpus-based measures. These tools can most often be used on any (english) corpora. Specific distributional models build from popular corpora are also generally provided. However, for a long time, most software solutions dedicated to knowledge-based measures were developed for particular applications and ontologies, e.g. in the biomedical domain.

The aim of this appendix is not to provide an extensive comparison of the tools that are presented. Ordering of tools is therefore made considering subjective criteria such as popularity of the tool in the community, functionalities provided, source code base, documentation available, support. We invite the reader to refer to the official websites for more information about these tools. The information presented hereafter has been obtained in January 2015, improvements could have been made and the developer base may have also change. In addition, the authors of this book are related to the Semantic Measures Library a project that will be presented hereafter. Nevertheless, even if we provide slightly more information about this project compared to the others, we will not provide detailed information about the project since our objective is to introduce the reader to the diversity of tools available to date. Since the large majority of projects focus on corpus-based or knowledge-based measures, but do not cover both types, we first present those dedicated to the former to next discuss those related to the latter. Finally, we discuss the tools that have been proposed to ease measure evaluation.

## Tools for corpus-based semantic measures

Table D.1 presents several free source code libraries and software solutions that can be used to compute the semantic similarity/relatedness between pairs of words and/or sentences. Most of the measures implemented in these tools evaluate semantic relatedness analysing syntagmatic relationships. Note also that the documentation of the tools sometimes refers to the notion of semantic similarity without considering the definition adopted in this book – and therefore semantic similarity most often refer to

| Name | Compare | Types | Language |
|---|---|---|---|
| DKPro Similarity (Bär et al., 2013) | Words/Sentences | LIB | Java |
| Semilar (Rus et al., 2013) | Words/Sentences | LIB, GUI | Java |
| Disco (Kolb, 2008) | Words | LIB, CLI | Java |
| NLTK (Bird, 2006) | Words | LIB | Python |
| GenSim (Rehurek and Sojka, 2010) | Words | LIB | Python |
| WikiBrain (Sen et al., 2014) | Words/Wikipedia Topics | LIB | Java |
| Mechaglot | Sentences | LIB | Java |
| Takelab sts (Sarić et al., 2012) | Sentences | LIB | Python |
| SemSim | Words | LIB | Java |

Table D.1: Examples of existing software solutions dedicated to corpus-based semantic measures. URLs are provided in the dedicated sections. LIB: source code library, CLI: Command Line Interface, GUI: Graphical User Interface.

semantic relatedness. For each tool we provide: its name, associated reference (and starting date), the type of objects that can be compared and the type of the tool: LIB refers to source code library, CLI to command-line tool[1], and GUI to Graphical User Interface. The programming language used for the development is also specified.

A brief description of each tool is also provided hereafter.

**DKPro Similarity** is a framework dedicated to the comparison of pairs of words and pairs of texts (Bär et al., 2013). It provides numerous implementations of state-of-the-art semantic relatedness measures in Java - some knowledge-based measures are also available to compare WordNet synsets. This project is part of the DKPro Core[2] project which develops a collection of software components dedicated to NLP. These components have the interesting properties to be based on the Apache UIMA framework. The last version of DKPro Similarity has been release in October 2013 (version 2.1.0), it is distributed under the open-source Apache Software Licence. The source code and extensive documentation are available from the dedicated website.
Website: `https://code.google.com/p/dkpro-similarity-asl`

**Semilar** proposes a software and development environment dedicated to corpus-based semantic measures (Rus et al., 2013). It provides both a Java library and a GUI-based interface. Semilar can be used to compare words and sentences. Numerous measures have been implemented, and several distributional models are made available. Semilar is still in alpha version (2014), consult the website for licence information.
Website: `http://www.semanticsimilarity.org`

**Disco** is an open-source Java library dedicated to the semantic similarity computation between words (Kolb, 2008, 2009). The tool is distributed under the Apache License, version 2.0. Several measures are implemented. Interestingly, numerous languages are also supported: Arabic, Czech, Dutch, English, French, German, Italian, Russian, and Spanish. Last version available to date is version 1.4 (released in 2013).
Website: `http://www.linguatools.de/disco/disco_en.html`

**NLTK** is a popular and elegant NLP platform developed in Python (Bird, 2006). It contains a module dedicated to WordNet which provides specific semantic measures for comparing two synsets (no extensive list of measures available). In addition, (Tomasik and Sutherland, 2008) present a NLTK module for large-scale computation of specific corpus-based measures. NLTK is distributed under the Apache 2.0

---

[1] Most of tools provide a way to execute source code using command lines. CLI refers to interfaces that are used to guide the user by providing documentation, multiple parameter tuning, batch computation capabilities, etc.

[2] `https://code.google.com/p/dkpro-core-asl`

license.
Website: `http://www.nltk.org`

**GenSim** is a Python platform dedicated to statistical semantics (Rehurek and Sojka, 2010). It is well documented and provides several efficient distributional models and measures implementations. Gensim is distributed under the LGPL Licence. Both open-source and business supports are provided. The last release has been developed in 2014.
Website: `https://radimrehurek.com/gensim`

**WikiBrain** is a well-documented Java project that proposes Wikipedia-based algorithms for semantic relatedness computation (Sen et al., 2014). It can be used to compare both sentences and Wikipedia pages (topics). According to the documentation, efforts have been made to ensure efficiency and shorten computational time. WikiBrain is distributed under the Apache 2.0 license. The project is today actively maintained. WikiBrain is also a nice project to consider for those interested in interacting with Wikipedia
Website: `http://shilad.github.io/wikibrain`

**TakeLab Semantic Text Similarity System** is a Python code that can be used to compare two sentences (Sarić et al., 2012). It is licensed under a derivative of a BSD-license that requires proper attribution (c.f. website). This source code corresponds to a submission proposed to the SemEval evaluation campaign (cf. to Chapter 4 for more information).
Website: `http://takelab.fer.hr/sts`

**SemSim** is a Java library that can be used to evaluate the semantic relatedness of words and to compute distributional models from texts. It can also be used to compute distributional models. The source code is distributed under licence LGPL v3. The last version has been released in 2013.
Website: `https://github.com/marekrei/semsim`

**Mechaglot** can be used to compare sentences. It is distributed under the Creative Commons Attribution-ShareAlike 4.0 International License. This project is still under development and to date only provides an alpha version for developers.
Website: `http://mechaglot.sourceforge.net`

Numerous web services and web interfaces can be used to compute semantic relatedness between words and texts. Since these links may change and/or the services may not be supported in the future, we orient the reader to the updated list of links provided at Wikipedia: `http://en.wikipedia.org/wiki/Semantic_similarity`.

## Tools for knowledge-based semantic measures

Most software solutions dedicated to knowledge-based semantic measures have been developed for a specific usage context and are dedicated to a specific ontology/ structured terminology, e.g., Wordnet (Pedersen et al., 2004), Unified Medical Language System (UMLS) (McInnes et al., 2009), the Gene Ontology (GO)[3], the Disease Ontology (DO) (Li et al., 2011), Medical Subject Headings (MeSH) (Pirró and Euzenat, 2010a). By supporting Web Ontology Language (OWL), Resource Description Framework (RDF) and Open Biomedical Ontologies (OBO) format, some tools have also been developed in a generic manner. They can be used with any ontology defined in these formats.

Table D.2 summarises some characteristics of existing libraries/tools. We present the name of the tool, the ontologies and formats that are supported, the type of tools, command-line tool (CLI) and/or source code library (LIB). We also specify the programming language and if the measure can be used to compare pairs of concepts (P) and/or pairs of groups of concepts (G).

In the following, for convenience, semantic similarity/relatedness refers to knowledge-based semantic similarity/relatedness. We also recall that DKPro Similarity and NLTK implement measures that can be used to compare two synsets.

---

[3]Note that half a dozen libraries/tools are dedicated to the Gene Ontology: `http://www.geneontology.org/GO.tools_by_type.semantic_similarity.shtml`

| Name | ontology | Types | Measures | Language |
|---|---|---|---|---|
| SML (Harispe et al., 2014a) | OWL, RDF, OBO, etc. | CLI, LIB | P, G | Java |
| FastSemSim (Guzzi et al., 2012) | OBO and others | CLI, LIB | P, G | Python |
| SimPack (Bernstein et al., 2005) | OWL, RDF | LIB | P | Java |
| SemMF (Oldakowski and Bizer, 2005) | OWL, RDF | LIB | P, G | Java |
| OntoSim (David and Euzenat, 2008) | OWL, RDF | LIB | P | Java |
| Ytex Semantic Similarity (Garla and Brandt, 2012) | SEE DOC | LIB | P, G | Java |
| Similarity Library (Pirró and Euzenat, 2010a) | *Wordnet, MeSH, GO* | CLI, LIB | P, G | Java |
| WordNet-Similarity (Pedersen et al., 2004) | *WordNet* | CLI, LIB | P, G | Perl |
| WS4J | *WordNet* | LIB | P | Java |
| UMLS-Similarity (McInnes et al., 2009) | *UMLS* | LIB | P | Perl |
| OWLSim (Washington et al., 2009) | OWL, RDF, OBO | LIB | P | Java |
| DOSim (Li et al., 2011) | *DO* | CLI, LIB | P, G | R |
| DOSE | *DO* | CLI, LIB | P, G | R |

Table D.2: Some characteristics of a selection of libraries/software tools that can be used to compute knowledge-based semantic measures. Types: Command Line Interface (CLI), Library (LIB). Measures: pair of concepts (Pairwise – P), pair of groups of concepts (Groupwise – G).

A brief description of each tool presented in Table D.2 is also provided:

**SML** stands for Semantic Measures Library (Harispe et al., 2014a). This is a Java library dedicated to semantic measure computation, development and analysis. It is developed by Sébastien Harispe, one of the author of this book. SML implements numerous measures for comparing concepts and groups of concepts defined into ontologies. Numerous formats (OBO, RDF, OWL) and domain-specific ontologies are supported (e.g., WordNet, MesH, SNOMED-CT). Fine tuning is made available to control measure parameters and data pre-processing. In addition to the library, a command-line toolkit is also developed for non-developers. Both the library and the toolkit are optimized to handle large dataset and to ensure fast computation[4]. They are distributed under a GPL compatible license. Last release: 2014.
Website: http://www.semantic-measures-library.org

**FastSemSim** is a Python library dedicated to semantic similarity computation (Guzzi et al., 2012). It can be used to compare pairs of concepts and pairs of groups of concepts. Any OBO ontology can be loaded. Other formats can also manually be loaded into specific data structures that can be considered as data sources by the library. FastSemSim is designed to provide efficient implementation of measures. It is distributed under the GPL Licence. Last version: 2014
Website: https://sites.google.com/site/fastsemsim/

**SimPack** is a Java library that can be used to compare pairs of concepts defined into ontologies (Bernstein et al., 2005). It implements a large variety of measures and can load RDF/OWL ontologies. SimPack is distributed under the LGPL license. SimPack has not been updated since 2008.
Website: https://files.ifi.uzh.ch/ddis/oldweb/ddis/research/simpack/

**SemMF** is a Java library that can be used to compare objects defined into RDF graphs. It is distributed under LGPL licence(Oldakowski and Bizer, 2005).
Website: http://semmf.ag-nbi.de/doc/index.html

---

[4]A comparison with other tools is provided at: https://github.com/sharispe/sm-tools-evaluation

**OntoSim** is an efficient Java library dedicated to the comparison of ontologies (David and Euzenat, 2008). Several measures provided in this library can also be used to compare objects defined into RDF graphs.
Website: `http://ontosim.gforge.inria.fr/`

**YTEX Semantic Similarity** is a Java library that can be used to compare two concepts or two groups of concepts defined into an ontology (Garla and Brandt, 2012). The library can be used with UMLS, SNOMED-CT and MeSH. It is also possible to load other ontologies through SQL queries.
Website: `https://code.google.com/p/ytex/wiki/SemanticSim_V06`

**The Similarity Library** is Java library that can be used to compare pairs of concepts defined into several ontologies (WordNet, MesH, GO) (Pirró and Euzenat, 2010a). The library is made available on request.
Website: `https://simlibrary.wordpress.com/`

**WordNet-Similarity** is a Perl module dedicated to semantic similarity and relatedness measures between WordNet synsets (Pedersen et al., 2004). This package has been largely used in the literature. Last version: 2008.
Website: `http://wn-similarity.sourceforge.net/`

**WS4J** is a Java library dedicated to the development of semantic measures for WordNet. It is distributed under the GPL licence. Last version: 2013
Website: `http://code.google.com/p/ws4j/`

**UMLS-Similarity** is a Perl module that can be used to compare concepts defined into UMLS (McInnes et al., 2009)
Website: `http://umls-similarity.sourceforge.net/`

**OWLSim** is a Java Library for the comparison of pairs of concepts defined in OBO or OWL format (Washington et al., 2009). Last version: 2013.
Website: `http://code.google.com/p/owltools/wiki/OwlSim`

**DOSim** is an R package dedicated to semantic similarity computation for the Disease Ontology (Li et al., 2011). DOSim is distributed under the GPL licence. Last version: 2010
Website: `http://210.46.85.150/platform/dosim/`

**DOSE** is another R package dedicated to semantic similarity computation for the Disease Ontology. DOSE is distributed under the Artistic-2.0 licence.
Website: `http://www.bioconductor.org/packages/release/bioc/html/DOSE.html`

# Tools for the analysis of semantic measures

Tools and source code libraries implementing numerous measures have been presented in the previous section. They represent important contributions for those interested in comparing new proposals, and can be used to select measures adapted to a specific domain. We now provide specific information relative to four tools that provide interesting functionalities for measure evaluation. Recall that protocols and dataset evaluation have been introduced in Chapter 4.

**DKPro Similarity** provides several functionalities that can ease the evaluation of semantic measures. It gives access to several evaluation datasets and experiments that are commonly used to compare corpus-based semantic measures, e.g. it contains several word pair similarity/relatedness datasets[5]. DKPro similarity can also be used to reproduce SemEval-2012 best system[6] – more information about this tool can be found in Section D.

---

[5] `https://code.google.com/p/dkpro-similarity-asl/wiki/WordPairSimilarity`
[6] `https://code.google.com/p/dkpro-similarity-asl/wiki/SemEval2013`

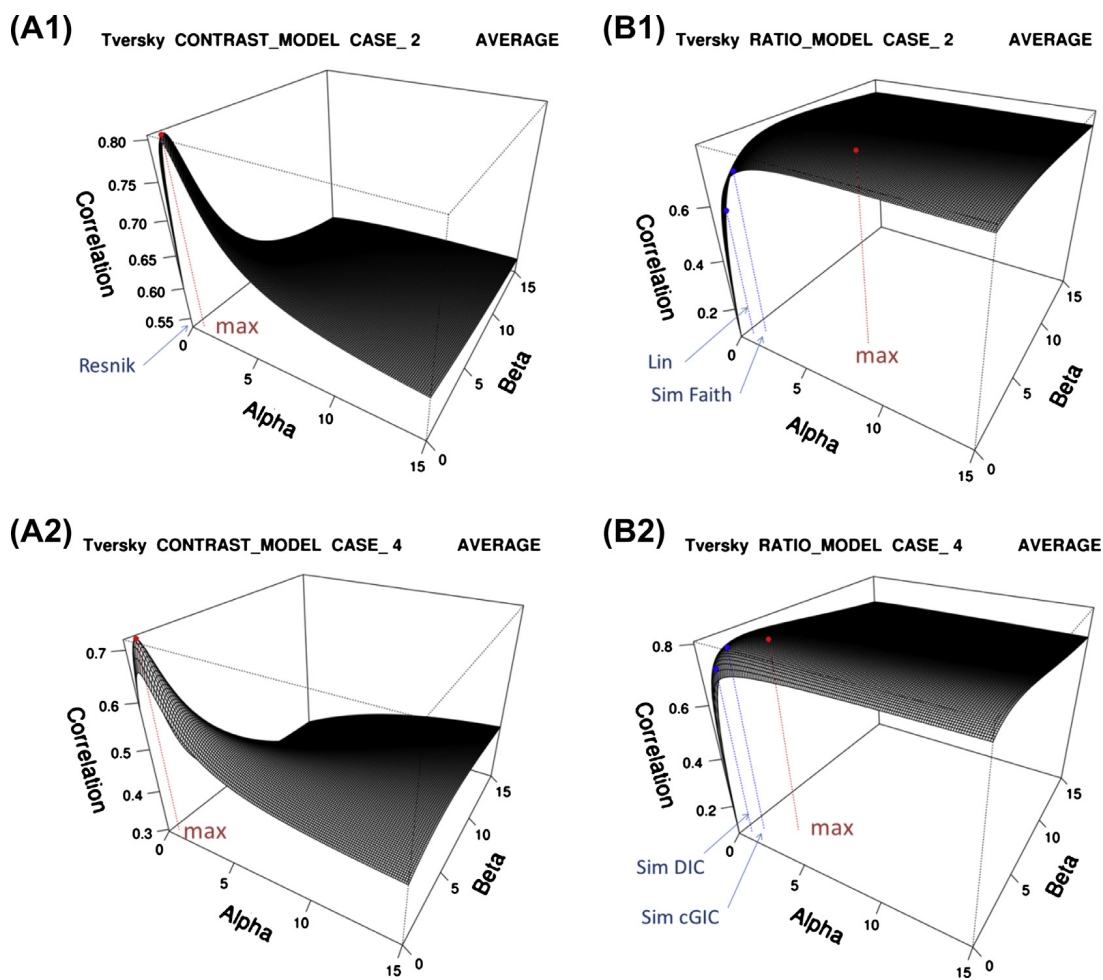

Figure D.1: Evaluation of measures that have been made using SML. From (Harispe et al., 2013c).

**SML** provides numerous knowledge-based measures implementation that can be used through parametric functions. These implementations and additional code included into the library provide an easy way to compare semantic measure, e.g. regarding their accuracy. As an example, in Figure D.1, the surfaces represent the accuracy of large configuration of measures – each point in the surface corresponds to the accuracy of a specific measure configuration. Tsversky ratio and contrast models abstract functions were used; these measures can be expressed through parametric functions depending on two parameters alpha and beta – here represented by the abscissa and the ordinate in the figures. Therefore, each figure corresponds to a specific expression of other parameters of the measures that are fixed. The evaluation is made by comparing estimated similarity with human expected values. This experiment underlines the benefits of efficient tools dedicated to semantic measure computation and to analysis semantic measures – details are provided in (Harispe et al., 2013c). More information about SML are also given in Section D.

**SemEval** is an evaluation campaign related to semantic analysis in which several tracks are related to semantic measures. We encourage the reader to consult the resources that are made available for these tasks. Tools and source codes are generally provided to ease measure evaluation. More information can be found on the wiki maintained by the IXA Group from the University of the Basque Country: http://ixa2.si.ehu.es/stswiki.

**CESSM** is a platform that can be used to evaluate semantic measures over the Gene Ontology (Pesquita et al., 2009b). Several evaluation strategies are implemented. New measures can thus be compared to existing results. Datasets used in the evaluations can also be downloaded. The source code is not available to download but can be requested to the platform maintainers.
Website: http://xldb.di.fc.ul.pt/tools/cessm

# Author's Biography

## Sébastien Harispe

**Sébastien Harispe** holds a Master and a PhD in Computer Science from the University of Montpellier II (France). His research focuses on Artificial Intelligence and more particularly on the diversity of methods which can be used to support decision making from text and knowledge base analysis, e.g. Information Extraction and Knowledge inference. Sébastien Harispe proposed several theoretical and practical contributions related to semantic measures. He dedicated his thesis to an in-depth analysis of knowledge-based semantic similarity measures from which he proposed a unifying theoretical framework for these measures. He is also the project leader and main developer of the Semantic Measures Library project, a project dedicated to the development of open-source software solutions for semantic measures computation and analysis.

## Sylvie Ranwez

**Sylvie Ranwez** is an Associate Professor in the LGI2P Research Center of the engineering school Ecole des Mines d'Alès, in France. Since 2000, she has been interested in the research endeavor of one part of the Artificial Intelligence, i.e. Knowledge engineering. Her research is dedicated to ontologies used as a guideline in conceptual annotation process and information retrieval systems, navigation over numerous resources and visualization. Since semantic measures underlie all of these processes, she also directs research in this domain. She holds a PhD (2000) and a habilitation (2013) in Computer Science (University of Montpellier 2, France).

## Stefan Janaqi

**Stefan Janaqi** is a research member of the LGI2P Research Center team of the Ecole des Mines d'Alès (France). He holds a PhD in Computer Science from University Joseph Fourier, Grenoble (France), dealing with geometric properties of graphs. His research focuses on mathematical models for optimization, image treatment, evolutionary algorithms and convexity in discrete structures such as graphs.

## Jacky Montmain

**Jacky Montmain** received the Master's degree from the Ecole Nationale Supérieure d'Ingénieurs Electriciens de Grenoble, Grenoble, France, in 1987 and the PhD degree from the National Polytechnic Institute, France, in 1992, both in control theory. He was a research engineer at the French Atomic Energy Commission from 1991 to 2005 where he was appointed as Senior Expert in the field "Mathematics, Computer Sciences, Software and System Technologies" in 2003. He is currently a Professor at the Ecole des Mines d'Alès, Nîmes, France. His research interests include the application of artificial intelligence techniques to model-based diagnosis and supervision, industrial performance improvement, multicriteria and fuzzy approaches to decision-making.